\documentclass{amsart}
\usepackage{graphicx} 

\usepackage{fullpage}
\usepackage{amsfonts}
\usepackage{amsmath}
\usepackage{amsxtra, amsfonts,amscd,amssymb,amsthm,mathtools}
\usepackage{algorithm,algorithmic}
\usepackage{color}
\usepackage{lipsum}
\usepackage{hyperref}
\usepackage[nameinlink]{cleveref}

\newtheorem{definition}{Definition}[section]
\newtheorem{theorem}{Theorem}[section]
\newtheorem{corollary}[theorem]{Corollary}
\newtheorem{lemma}[theorem]{Lemma}
\newtheorem{Observation}[theorem]{Observation}
\newtheorem{Remark}[theorem]{Remark}
\newtheorem{proposition}[theorem]{Proposition}

\usepackage[utf8]{inputenc}

\title{Stability of Parallel Transport Convolution under Riemannian Metric Perturbations}
\author{Hongnian Huang}
\address{
    Department of Mathematics and Statistics,
    University of New Mexico,
    Albuquerque, NM 87131, USA
}
\email{hnhuang@gmail.com}
\author{Andrew Stump}
\address{
    Department of Mathematics and Statistics,
    University of New Mexico,
    Albuquerque, NM 87131, USA
}
\email{stumpa@unm.edu}

\date{}

\begin{document}

\maketitle

\begin{abstract}
    Geometric convolutions on Riemannian manifolds depend on the geometry of the underlying domain, making stability under metric perturbations a fundamental question in geometric deep learning. This paper studies this question for parallel transport convolution through methods from differential geometry and analysis. On a fixed compact manifold, we establish that, for a continuous input signal and a continuous reference kernel supported in a sufficiently small normal neighborhood, the convolution output depends continuously in $L^2$ on the Riemannian metric in the $C^2$-topology, under the specified localization and transport-path conventions. The analysis develops perturbation estimates for geodesics, parallel transport, Jacobi fields, and exponential maps, and uses these estimates to compare the transported kernels. A central difficulty is the cut locus, which obstructs uniform comparison of the transport construction over the entire manifold. We address this difficulty by combining uniform estimates away from the cut locus with integral bounds on neighborhoods of arbitrarily small volume, while also controlling the change in the Riemannian volume measure. The resulting qualitative stability theorem connects perturbation theory for Riemannian geometric constructions with the analysis of spatial convolution operators in geometric deep learning.
\end{abstract}

\section{Introduction}
This paper studies a problem at the intersection of differential geometry, analysis, and geometric deep learning: the stability of a geometrically defined convolution under perturbations of the Riemannian metric. Geometric deep learning extends convolutional representation learning to data supported on graphs, surfaces, and manifolds \cite{bronstein2017geometric}. On a manifold, the metric determines distances and local neighborhoods, while its Levi-Civita connection determines how directions are transported between tangent spaces. When these structures enter the definition of a convolution, their dependence on the metric becomes a question about the stability of the resulting representation. We investigate this dependence on a fixed compact smooth manifold, allowing the Riemannian metric to vary while keeping the input signal and reference kernel fixed.

The problem has a complementary interpretation in each of the two fields. From the perspective of geometric deep learning, it asks whether a convolution produces similar outputs when the geometry supporting the data is slightly perturbed. From the perspective of differential geometry, it asks how a construction assembled from geodesics, parallel transport, exponential maps, and Riemannian integration varies with the metric. Answering the first question requires coordinated control of the geometric objects in the second. The analysis must also account for conjugate points and the cut locus, which constrain the invertibility of exponential maps and the choice of geodesics between prescribed endpoints. This interaction between local geometric differential equations and global geometric obstructions is central to the paper.

Stability has long provided a mathematical principle for the analysis of convolutional representations. In the Euclidean setting, Mallat \cite{mallat2012group} established deformation stability for wavelet scattering, relating small diffeomorphic changes of a signal to controlled changes in its representation. On graphs, Gama, Bruna, and Ribeiro \cite{gama2020stability} studied the response of graph neural networks to perturbations of the graph shift operator and identified filter conditions that balance stability and discriminability. These results highlight two complementary properties of geometric learning operators: predictable behavior under transformations that preserve the domain’s structure, and stability when that structure is perturbed. A precise stability statement must specify the admissible perturbations, the assumptions on the filters, and the norm used to compare the outputs.

Related questions have been investigated directly on manifolds. Perlmutter et al. \cite{perlmutter2020geometric} developed geometric wavelet scattering on compact Riemannian manifolds and established invariance and stability properties for suitable geometric transformations. Wang, Ruiz, and Ribeiro studied manifold neural networks under absolute and relative perturbations of the Laplace–Beltrami operator \cite{Wang2021Stability, Wang2021Relative}, and developed a broader analysis of manifold filters and networks under smooth deformations \cite{wang2024deformations}. Their filters are constructed through the Laplace–Beltrami operator and analyzed using spectral conditions. This literature provides substantial foundations for geometric stability, while leaving a distinct analytical problem for spatial filters whose dependence on the metric is expressed through geodesics, local inverse exponential maps, and parallel transport.

The present work addresses this problem for parallel transport convolution. Schonsheck, Dong, and Lai \cite{schonsheck2022parallel} introduced this construction to transport localized filters across a manifold while retaining directional information. Unlike Euclidean translation, the transport of a filter on a curved domain depends on the connection and the chosen transport paths. A perturbation of the Riemannian metric therefore affects several components simultaneously: the geodesics connecting points, the associated parallel transport maps, the exponential maps used to represent local neighborhoods, and the volume measure used in integration. Understanding the combined effect of these changes requires estimates that connect perturbations of the metric to perturbations of the convolution output.

We consider a compact Riemannian manifold \((M,g_1)\), a fixed reference point \(x_0\), a continuous input \(f:M\to\mathbb R\), and a continuous reference kernel \(K(x_0,\cdot)\) supported in a sufficiently small normal neighborhood of \(x_0\). For a nearby metric \(g\), the transported kernel is locally given by
\[
K_g(x,y)=K\!\left(x_0,
\exp_{x_0}^{g}\circ(P_{x_0,x}^{g})^{-1}
\circ(\exp_x^{g})^{-1}(y)\right),
\]
with the localization and transport-path conventions specified in Section~\ref{sec:app}. The corresponding convolution is
\[
T_gf(x)=\int_M f(y)K_g(x,y)\,d\operatorname{vol}_g(y).
\]
The main result can now be stated as follows, with the fixed data and the dependence on the prescribed tolerance made explicit.

\begin{theorem}[c.f. Theorem \ref{thrm:fin}] Let \((M,g_1)\) be a compact Riemannian manifold, fix \(x_0\in M\), and choose
\[
0<r<\tfrac12\operatorname{inj}(M,g_1).
\]
Let \(f\in C(M)\) and let \(K(x_0,\cdot)\in C(M)\) satisfy
\[
\operatorname{supp}K(x_0,\cdot)\subset B_{g_1}(x_0,r).
\]
For \(m=1,2\), let \(h_m=T_{g_m}f\) be the parallel transport convolution defined above, using the localization and transport-path conventions of Section~\ref{sec:app}. Then, for every \(\varepsilon>0\), there exists
\[
\delta=\delta(\varepsilon;M,g_1,x_0,r,f,K)>0
\]
such that every smooth Riemannian metric \(g_2\) on \(M\) satisfying
\[
\|g_2-g_1\|_{C^2(M,g_1)}<\delta
\]
also satisfies
\[
\|h_2-h_1\|_{L^2(M,g_1)}^2
=\int_M|h_2(x)-h_1(x)|^2\,d\operatorname{vol}_{g_1}(x)
<\varepsilon.
\]
\end{theorem}

\begin{Remark}
Here the \(C^2(M,g_1)\) norm measures the metric difference and its first two covariant derivatives with respect to the reference metric \(g_1\). In particular, the theorem asserts that, for fixed \(f\) and \(K\),
\[
\|g_2-g_1\|_{C^2(M,g_1)}\longrightarrow0
\quad\Longrightarrow\quad
\|T_{g_2}f-T_{g_1}f\|_{L^2(M,g_1)}\longrightarrow0.
\]
Thus, sufficiently small changes of the metric and its first two covariant derivatives produce arbitrarily small changes in the convolution output in the reference \(L^2\) norm. The result concerns the continuous geometric convolution with fixed data and kernel; it supplies a foundation for understanding the geometric sensitivity of architectures built from this operation.  
\end{Remark}

The paper is organized to make this passage from geometry to convolution explicit. Section~\ref{sec:pre} introduces the geometric norms and ordinary differential equation estimates used throughout the analysis. Section~\ref{sec:metricaprox} develops perturbation results for geodesics and parallel transport under \(C^1\) metric control. Section~\ref{sec:C2} uses \(C^2\) control to study curvature, Jacobi fields, and the exponential map. Section~\ref{sec:app} combines these results to establish the stability of parallel transport convolution in Theorem \ref{thrm:fin}. The proof overview below explains how the individual geometric results contribute to this conclusion.

\subsection{Proof strategy and the role of the auxiliary results}

\begin{enumerate}
    \item The proof of Theorem~\ref{thrm:fin} proceeds from local perturbation estimates for geometric differential equations to a global estimate for the convolution output. Its central step is Proposition~\ref{def:prop5}, which compares the geometric maps used to transport the kernel uniformly away from the reference cut locus. The preceding lemmas, propositions, and theorems establish the estimates and local invertibility needed for this comparison.

    \item{\bf From metric perturbations to differential equations.} Propositions~\ref{def:prop0} - \ref{def:prop1} and Lemma~\ref{lemma:ineq1} relate the intrinsic \(C^k\) norm to coordinate estimates, allowing metric perturbations to be inserted into the equations for geodesics, parallel transport, and Jacobi fields. Proposition~\ref{prop:neat} compares tangent-vector lengths and distances for the two metrics, so that bounds formulated using \(g_1\) remain useful for \(g_2\). Gr\"onwall’s inequality, Lemma~\ref{gron:int}, supplies the analytical foundation for the continuous-dependence estimates: Lemma~\ref{def:example1} treats the first-order linear systems arising in parallel transport, Lemma~\ref{def:example2} treats the geodesic equation, and Lemma~\ref{def:example3} treats the second-order linear systems governing Jacobi fields.

    \item{\bf Stability of geodesics and parallel transport.} Lemma~\ref{lemma:ineq0} gives a Christoffel-symbol estimate in suitable normal coordinates and supports Theorem~\ref{thrm:neat}, which isolates the effect of changing the metric while keeping the transport curve fixed. Proposition~\ref{prop:prop1} gives coordinate control of the difference between the two connections. Together with the differential equation estimates, this leads to Theorem~\ref{def:thrm3}: geodesics and their coordinate velocities remain close when the metrics and initial velocities are close. Corollary~\ref{def:cor1} specializes this conclusion to identical initial velocities and hence compares exponential maps. Theorem~\ref{def:thrm5} addresses the additional difficulty relevant to convolution: parallel transport remains close when both the connection and the geodesic change, including when the transported initial vectors also vary.

    \item{\bf Control of local inverses and connecting geodesics.} The use of \(C^2\) metric control enters through the curvature and Jacobi equations. Proposition~\ref{def:prop2} controls curvature perturbations, Theorem~\ref{thrm:neat1} establishes continuous dependence of Jacobi fields, and Corollary~\ref{cor:neat1} translates this dependence into control of the differential of the exponential map in local coordinates. Theorem~\ref{thrm:neat3} supplies a uniform short-time exclusion of conjugate points. Lemma~\ref{lemma:ineq2} controls the variation of Jacobi fields with the initial geodesic direction for the reference metric. These ingredients support Lemma~\ref{lemma:lemma3}, which provides uniform local invertibility and an image-containment statement for the perturbed exponential map on the specified sets away from the cut locus. Theorem~\ref{thrm2:neat} then produces a nearby \(g_2\)-geodesic joining the same endpoints as a reference minimizing \(g_1\)-geodesic. This endpoint-matching step is essential: closeness of geodesics with the same initial data alone would not ensure that both transport maps terminate at the same point \(x\).

    \item{\bf Comparison of the transported kernels.} Theorem~\ref{thrm:neat5} establishes a uniform lower bound on the injectivity radius for metrics sufficiently close to \(g_1\). Together with metric comparison, this ensures that the local logarithm maps required by the convolution are defined on the chosen reference balls. Corollary~\ref{cor:neat5} then gives uniform closeness of these inverse exponential maps. Proposition~\ref{def:prop5} combines this conclusion with the connecting-geodesic result, Theorem~\ref{thrm2:neat}, the parallel-transport estimate, Theorem~\ref{def:thrm5}, and the exponential-map estimates from Section~\ref{sec:metricaprox}. It therefore controls the entire composition
\[
\Phi_m(x,y)=\exp_{x_0}^{g_m}\circ(P_{x_0,x}^{g_m})^{-1}
\circ(\exp_x^{g_m})^{-1}(y),\qquad m=1,2,
\]
uniformly for \(x\) a fixed positive distance from \(\operatorname{Cut}_{g_1}(x_0)\) and \(y\) in the prescribed local ball. Uniform continuity of the reference kernel converts this geometric comparison into uniform closeness of \(K_1\) and \(K_2\) on that region, with the common zero-extension convention outside the local balls.

\item{\bf Passage to the global \(L^2\) estimate.} Write \(\mu_m=\operatorname{vol}_{g_m}\) and \(\rho=d\mu_2/d\mu_1\). The output difference separates as
\[
h_1(x)-h_2(x)
=\int_M f(y)\bigl(K_1(x,y)-K_2(x,y)\bigr)\,d\mu_1(y)
+\int_M f(y)K_2(x,y)\bigl(1-\rho(y)\bigr)\,d\mu_1(y).
\]
Lemma~\ref{lemma:fin}, which is used in the proof of Theorem~\ref{thrm:fin}, controls the second term uniformly in \(\ x \), using convergence of the volume density. For the first term, choose an open neighborhood \(E\) of the reference cut locus with arbitrarily small \(\mu_1\)-volume. On the compact complement \(A=M\setminus E\), Proposition~\ref{def:prop5} gives uniform kernel convergence. On \(E\), the bound \(|K_1-K_2|\leq 2K_0\), where \(K_0=\|K(x_0,\cdot)\|_\infty\), controls the contribution by the volume of \(E\). The decomposition is made in the output variable \(x\), where the cut-locus restriction occurs.

More precisely, let \(V=\mu_1(M)\), \(F=\|f\|_\infty\), suppose the volume-change term is bounded by \(\eta\), and let \(\omega\) bound \(|K_1-K_2|\) for \(x\in A\) and all \(y\). Cauchy–Schwarz and the preceding decomposition yield
\[
\|h_1-h_2\|_{L^2(M,g_1)}^2
\leq 2F^2V^3\omega^2
+8F^2V^2K_0^2\mu_1(E)
+2V\eta^2.
\]
Given \(\varepsilon>0\), first choose \(E\) so that the middle term is sufficiently small. With this neighborhood fixed, choose the metric perturbation small enough to control the first and last terms. This proves the asserted \(L^2\) continuity. The auxiliary results thus serve three complementary purposes: controlling geometric evolution, ensuring the local inverses and endpoint-matched transport paths are available, and converting uniform control away from the cut locus into a global integral estimate.

\end{enumerate}

\subsection*{Disclosure of AI use}
The main mathematical results of this paper were developed by the authors without the assistance of artificial intelligence. AI tools were used during manuscript preparation to improve readability, clarity, and presentation. The authors take full responsibility for the mathematical content and the final text.

\subsection*{Acknowledgments}
We thank Professor Rongjie Lai for introducing this problem to us and for insightful discussions at the outset of this work about possible approaches to its solution.

\section{Preliminary Setup}\label{sec:pre}

Everything that follows compares objects built from two metrics, namely geodesics, parallel transport and Jacobi fields, and we do this in coordinates. So we first need to turn “$g_2$ is close to $g_1$” into pointwise bounds. We define the $C^k$ distance between metrics and show that it bounds the coordinate components of $g_1 - g_2$ and its derivatives, and also the lengths of tangent vectors and the distances between points (proposition~\ref{prop:neat}). We then state Gr\"onwall’s inequality and use it to prove the three ODE estimates needed for parallel transport, geodesics and Jacobi fields. Sections~\ref{sec:metricaprox} and~\ref{sec:C2} apply these.

\subsection{The $C^k$-norm of a tensor}\label{sect:Cknorm}

Throughout this paper, $M$ is a compact manifold. To say that two Riemannian metrics $g_1$ and $g_2$ are  \lq\lq close", we need a way to measure the \lq\lq size" of the symmetric $(0,2)$-tensor $g_1 - g_2$ together with its covariant derivatives. Following~\cite{wittmann2019banach}, we use the maximum $C^k$-norm. Compactness of $M$ guarantees that the supremum in the following definition is finite for every smooth tensor. Only $k = 0, 1, 2$ are needed in this paper.

\begin{definition} \label{def:norm1}
    Let $g$ be a metric for the manifold $M$, and let $a \in \Gamma (Sym^2 (T^* M))$, then

    \begin{equation*}
        \parallel a \parallel_{C^k (M,g)}^2 = \sup_{p \in M} \bigg( |a|_g^2 + \sum_{l=1}^k |(\nabla^g)^l a|_g^2 \bigg)
    \end{equation*}

    Here, $\nabla^g$ is the Levi-Civita connection with respect to $g$. Then we have
    \begin{equation*}
        \begin{split}
            |a|_g^2 & = a_{ij}a_{\alpha \beta} g^{i \alpha} g^{j \beta}\\
            |(\nabla^g)^l a|_g^2 & = ((\nabla^g)^l a)_{j_1 j_2 \dots j_{l+2}} ((\nabla^g)^l a)_{\alpha_1 \alpha_2 \dots \alpha_{l+2}} g^{j_1 \alpha_1} \cdot g^{j_2 \alpha_2} \cdots g^{j_{l+2} \alpha_{l+2}}
        \end{split}
    \end{equation*}
    for $l = 1,2, \dots, k $. 
\end{definition}

The metric $g$ in this definition is called the \textbf{background metric}, because it is used to measure both the tensor and its derivatives, through $\nabla^g$ and the inverse metric $g^{ij}$. When we compare two metrics we take $a = g_1 - g_2$ and use $g_1$ as the background metric, which gives $\parallel g_1 - g_2 \parallel_{C^k (M,g_1)}$. The $C^1$ and $C^2$ cases are written out in Sections~\ref{sec:metricaprox} and~\ref{sec:C2}. This norm is intrinsic, but the estimates in the rest of the paper are carried out in coordinates, where geodesics, parallel transport and Jacobi fields satisfy differential equations. So in the next subsection we pass to a local coordinate system and compare this norm with a Euclidean one, in which $|a|$ is computed from the components of $a$ and their partial derivatives.

\subsubsection{Local coordinates and the Euclidean norm}

The norm in definition~\ref{def:norm1} is computed with the background metric $g$ and its covariant derivatives. The computation is much simpler when the background metric is Euclidean, $g_{ij} = \delta_{ij}$, because then covariant derivatives reduce to ordinary partial derivatives. A general metric is not Euclidean, but inside a coordinate chart we can compare it with the Euclidean metric that the chart induces. So we work locally: $U$ always denotes a fixed neighborhood whose closure lies in a coordinate chart, as recorded in Remark~\ref{def:rmk1} at the end of this subsection. We begin with the Euclidean metric induced by the chart.
\begin{definition}\label{def:eucl}
    Let $U$ be a fixed coordinate neighborhood, and let $\{ X_l \}$ be the coordinate frame associated with the neighborhood, i.e. $X_l := \frac{\partial}{\partial x^l}$ in these coordinates. Let $h$ be the Euclidean metric defined on $U$ such that
    \begin{equation*}
        h_{ij} = h(X_i,X_j) = \delta_i^j
    \end{equation*}
    Consequently, for all $p \in U$ and all $v \in T_pM$ with $v=v^i X_i$, we have $|v|_{h(p)} = |(v^1,v^2, \dots , v^n)|$, where the right-hand side is the standard Euclidean norm on $\mathbb{R}^n$.
\end{definition}
Next we localize the norm of definition~\ref{def:norm1} by dropping the supremum.
\begin{definition}\label{def:nbhdnorm}
    Let $U$ be a fixed coordinate neighborhood, let $g$ be a metric defined on $U$, and let $a \in \Gamma( Sym^2 (T^*M))$, then we define
    \begin{equation*}
        \parallel a \parallel_{C^k (U, g)}^2 = |a|_g^2 + \sum_{l=1}^k |(\nabla^g)^l a|_g^2
    \end{equation*}
\end{definition}
Since this quantity depends on the point $p \in U$ at which it's evaluated, any inequality below involving $\parallel \cdot \parallel_{C^k (U,g)}$ or $\parallel \cdot \parallel_{C^k (U,h)}$ is meant to hold for all $p \in U$, with constants independent of $p$. To be clear, definition~\ref{def:nbhdnorm} is not necessarily a norm; it's just shorthand. 

Taking $g=h$, and since the Levi-Civita connection of $h$ is given by coordinate derivatives, we obtain

\begin{equation}\label{def:eq1}
\begin{split}
 \parallel a \parallel_{C^k (U, h)}^2 &= |a|_h^2 + \sum_{l=1}^k |(\nabla^h)^l a|_h^2\\
 &= \sum_{ij} (a_{ij})^2 + \sum_{1 \leq|\alpha| \leq k} \sum_{ij} (\partial^\alpha a_{ij})^2
\end{split}
\end{equation}
In particular, $\parallel a \parallel_{C^k (U,h)}$ controls every component $a_{ij}$ and every partial derivative $\partial^\alpha a_{ij}$ with $|\alpha| \leq k$.

\begin{Remark}\label{def:rmk1}
Throughout, $U$ denotes a fixed neighborhood whose closure $\overline{U}$ is contained in an ambient coordinate chart $(V, \phi)$. Since $\overline{U}$ is a closed subset of the compact manifold $M$, it is compact. So, the following are well-defined for an arbitrary metric $g$:
 \begin{align*}
     \max |\Gamma^g| & := \max_{ijk} \big( \sup_{p \in \overline{U}} |\Gamma_{ij}^{k,g}| \big)\\
     \max |\partial \Gamma^g| & := \max_{ijkl} \big( \sup_{p \in \overline{U}} | \frac{\partial \Gamma_{ij}^{k,g}}{\partial x^l}| \big)\\
     \max |\partial^2 \Gamma^g| & := \max_{ijkls} \big( \sup_{p \in \overline{U}} | \frac{\partial^2 \Gamma_{ij}^{k,g}}{\partial x^l \partial x^s}| \big)\\
 \end{align*}
\end{Remark}

The next step is to compare $\parallel a \parallel_{C^k (U,h)}$ with the global norm $\parallel a \parallel_{C^k (M,g)}$ of definition~\ref{def:norm1}. This is what lets us turn smallness of $\parallel g_1 - g_2 \parallel_{C^k (M,g_1)}$ into bounds on the components of $g_1 - g_2$ and their partial derivatives in coordinates.

\subsubsection{Comparing local and global norms}

We now relate the local quantity $\parallel \cdot \parallel_{C^k (U,g)}$ to the global norm and to the Euclidean norm of the previous subsection. There are two comparisons. Restricting the supremum to $U$ can only decrease the norm (Proposition~\ref{def:prop0}), and on $U$ the norm computed with the background metric $g$ is equivalent to the Euclidean norm (Proposition~\ref{def:prop1}). Together they give Lemma~\ref{lemma:ineq1}, which bounds the Euclidean norm of $a$ on $U$ by its global $C^k$-norm.

\begin{proposition}\label{def:prop0}
    Let $(M,g)$ be a compact Riemannian manifold, and let $U$ be a fixed neighborhood. Then for all $a \in \Gamma (Sym^2 (T^*M))$
    \begin{equation*}
        \parallel a \parallel_{C^k (U,g)} \leq \parallel a \parallel_{C^k (M,g)}
    \end{equation*}
\end{proposition}

\begin{proof}
 Choose $a \in \Gamma( Sym^2 (T^*M))$, then since $U \subset M$, then 
\begin{equation*}
    \sup_{p \in \overline{U}} \bigg( |a|_g^2 + \sum_{l=1}^k |(\nabla^g)^l a|_g^2 \bigg) \leq \sup_{p \in M} \bigg( |a|_g^2 + \sum_{l=1}^k |(\nabla^g)^l a|_g^2 \bigg)
\end{equation*}
which means
\begin{equation*}
    \parallel a \parallel_{C^k (U, g)}^2 \leq \sup_{p \in \overline{U}} \bigg( |a|_g^2 + \sum_{l=1}^k |(\nabla^g)^l a|_g^2 \bigg) \leq \sup_{p \in M} \bigg( |a|_g^2 + \sum_{l=1}^k |(\nabla^g)^l a|_g^2 \bigg) = \parallel a \parallel_{C^k (M, g)}^2
\end{equation*}
and therefore, $\parallel a \parallel_{C^k (U, g)}^2 \leq  \parallel a \parallel_{C^k (M, g)}^2$, since both are nonnegative and $\sqrt{\cdot}$ is monotone, then $\parallel a \parallel_{C^k (U, g)} \leq  \parallel a \parallel_{C^k (M, g)}$, as desired.
\end{proof}

Proposition~\ref{def:prop0} shows that restricting to $U$ can only decrease the norm, but the covariant derivatives in it are still those of the background metric $g$, which are difficult to compute in coordinates. To work in coordinates we need to replace $g$ by the Euclidean metric $h$ on $U$. The next proposition shows that, on $U$, the two norms are equivalent, with constants depending on $M$, $g$, $U$ and $k$.

\begin{proposition}\label{def:prop1}

Let $(M,g)$ be a compact Riemannian manifold and $U$ be a fixed neighborhood and let $h$ be the Euclidean metric on $U$, as in definition~\ref{def:eucl}, then
    \begin{equation*}
        \parallel \cdot \parallel_{C^k (U, g)} \sim  \parallel \cdot \parallel_{C^k (U, h)}
    \end{equation*}
    where the bounding constants $C_1$ and $C_2$ will be based on the manifold $M$, the metric $g$, the neighborhood $U$, and $k$, for $k = 0,1,2$.    

\end{proposition}

The proof bounds the ratio of the two norms of a vector above and below, uniformly on the compact set $\overline U$, and then extends the comparison to tensors because their norms do not depend on the choice of orthonormal basis. The details are in Appendix~\ref{app:prop1}.

Combining these two propositions gives the estimate we use throughout: the Euclidean norm of $a$ on $U$ is controlled by its global $C^k$-norm.

\begin{lemma}\label{lemma:ineq1}
    For a fixed coordinate neighborhood $U$, there exists a constant $C(M,g,U,k)>0$ such that for any $a \in \Gamma( Sym^2 (T^*M))$, we have
    \begin{equation*}
        \parallel a \parallel_{C^k (U, h)} \leq C(M,g,U,k) \parallel a \parallel_{C^k (M, g)}
    \end{equation*}
    where the constant is based on the manifold $M$, the metric $g$, the coordinate neighborhood $U$, and $k$ for $k=0,1,2$.   
\end{lemma}
\begin{proof}
    By proposition~\ref{def:prop1} we know that on $U$ we have
    \begin{equation*}
    \parallel a \parallel_{C^k (U, h)} \leq C(M,g,U,k) \parallel a \parallel_{C^k (U, g)}        
    \end{equation*}
    Then we combine this with proposition~\ref{def:prop0}, and we have
    \begin{equation*}
    \parallel a \parallel_{C^k (U, h)} \leq C(M,g,U,k) \parallel a \parallel_{C^k (U, g)} \leq C(M,g,U,k) \parallel a \parallel_{C^k (M, g)}      
    \end{equation*} 
    which is our desired inequality.
\end{proof}

Lemma~\ref{lemma:ineq1} bounds the Euclidean norm of $a$ on $U$ by the global norm. In the rest of the paper, however, we use it one component at a time: the differential equation estimates involve the components of $g_1 - g_2$ and of its partial derivatives (for example, in the Christoffel symbols and their derivatives), not the norm as a whole. The next observation records that each such component is bounded by the Euclidean norm, and hence, by Lemma~\ref{lemma:ineq1}, by the global norm.

\begin{Observation}\label{def:obs1}
    The Frobenius norm for matrices $b = (b_{ij})$ is defined by $\parallel b \parallel_F^2 = \sum_{ij} (b_{ij})^2$. From equation~\ref{def:eq1}, for any $a \in \Gamma( Sym^2 (T^*M))$, we have $\parallel a \parallel_F^2 = \sum_{ij} (a_{ij})^2 = |a|_h^2 \leq \parallel a \parallel_{C^k (U, h)}^2$. Additionally, from equation~\ref{def:eq1}, for any $\alpha$ with $1 \leq |\alpha| \leq k$, we have
    \begin{equation*}
        \parallel \partial^\alpha a \parallel_F^2 = \sum_{ij} ( \partial^\alpha a_{ij})^2 \leq \sum_{|\beta|=|\alpha|} \sum_{ij} ( \partial^\beta a_{ij})^2 \leq |(\nabla^h)^{|\alpha|} a|_h^2 \leq \parallel a \parallel_{C^k (U, h)}^2
    \end{equation*}
\end{Observation}
Combining this observation with Lemma~\ref{lemma:ineq1}, we have $|a_{ij}| \leq C(M,g,U,k) \parallel a \parallel_{C^k (M,g)}$ and for $1 \leq |\alpha| \leq k$ we have $|\partial^\alpha a_{ij}| \leq C(M,g,U,k) \parallel a \parallel_{C^k (M,g)}$.

The next subsection uses the same norm differently, to compare the metrics themselves: lengths of vectors and distances.

\subsubsection{Closeness of metrics: lengths and distances}

Lemma~\ref{lemma:ineq1} and observation~\ref{def:obs1} control $g_1 - g_2$ one coordinate at a time. The proposition below compares the two metrics directly, without coordinates: if the norm is small, then lengths of tangent vectors with respect to $g_2$ are within a factor $\sqrt{1 \pm \epsilon}$ of their lengths with respect to $g_1$, and the two distance functions are comparable by the same factor. The proof uses only the value of $\parallel g_1 - g_2 \parallel_{C^k (M,g_1)}$ at a single point, so the proposition holds for any $k \geq 0$.

\begin{proposition}\label{prop:neat}
    Let $(M,g_1)$ be a compact Riemannian manifold, and let $g_2$ be another metric on $M$. For all $0 < \epsilon <1$, if $\parallel g_1 - g_2 \parallel_{C^k (M,g_1)} < \epsilon$ then:
    \begin{enumerate}
        \item For all $q \in M$ and $v \in T_qM$ we have
        \begin{equation*}
            \sqrt{1- \epsilon} \ |v|_{g_1 (q)} \leq |v|_{g_2 (q)} \leq \sqrt{1 + \epsilon} \ |v|_{g_1 (q)}
        \end{equation*}
        \item For all $p,q \in M$ we have
        \begin{equation*}
            \sqrt{1 - \epsilon} \ d_{g_1}(p,q) \leq d_{g_2}(p,q) \leq \sqrt{1 + \epsilon} \ d_{g_1}(p,q)
        \end{equation*}
    \end{enumerate}   
\end{proposition}
\begin{proof}
\begin{enumerate}
    \item Fix a point $q \in M$. Since $g_1 (q)$ is an inner-product on $T_qM$, we can choose, via the Gram-Schmidt process, an orthonormal basis $\{ e_1, \dots, e_n \} \subset T_qM$ with respect to $g_1 (q)$, that is, $\Tilde{g}_{1ij} :=g_1 (q)(e_i,e_j)= \delta_i^j$. Let $\Tilde{g}_{2ij} : =g_2 (q)(e_i,e_j)$. Since $g_2$ is an inner product at $q$, $(\Tilde{g}_{2ij})$ is positive definite, so by theorem 10.7, p.233,~\cite{roman2007advanced}, there exists $Q \in O(n)$ such that $Q^T (\Tilde{g}_{2ij})Q = (d_{ij})$ is a diagonal matrix. Then let $u_i = Qe_i$, and $(g_{mij}) := g_m (q) (u_i,u_j)$, and we'll have
    \begin{equation*}
        (g_{2ij}) = \sum_{kl} Q_{ki} \Tilde{g}_{2kl} Q_{lj} = Q^T (\Tilde{g}_{2ij})Q = (d_{ij})
    \end{equation*}
    and
    \begin{equation*}
        (g_{1ij}) = \sum_{kl} Q_{ki} \Tilde{g}_{1kl} Q_{lj} =  Q^T (\Tilde{g}_{1ij})Q = Q^T (\delta_i^j)Q = (\delta_i^j)
    \end{equation*}
    Since $|\cdot|_{g_1}^2$ is obtained by contracting a symmetric $(0,2)$-tensor against two copies of the inverse metric $(g_1^{ij})$, its values at a point are independent of the choice of basis, then we have
\begin{align*}
    |g_1 - g_2|_{g_1}^2 & = \sum_{ij \alpha \beta} (g_{1ij} - g_{2ij})(g_{1 \alpha \beta} - g_{2 \alpha \beta})g_1^{i \alpha} g_1^{j \beta}\\
    & = \sum_{ij \alpha \beta} (\delta_i^j - g_{2ij})(\delta_\alpha^\beta - g_{2 \alpha \beta}) \delta_i^\alpha \delta_j^\beta\\
    & = \sum_{ij}(\delta_i^j - g_{2ij})^2 = \sum_{i=1}^n (d_{ii}-1)^2
\end{align*}
Since $\parallel g_1 - g_2 \parallel_{C^k (M,g_1)}$ is a supremum over all of $M$, the value of $\parallel g_1 - g_2 \parallel_{C^k (M,g_1)}^2$ is at least the value of $|g_1 - g_2|_{g_1}^2$ at the point $q$. So, we have
\begin{equation*}
    \sum_{i=1}^n (d_{ii}-1)^2 = |g_1 - g_2|_{g_1(q)}^2 \leq \parallel g_1 - g_2 \parallel_{C^k (M,g_1)}^2 < \epsilon^2
\end{equation*}
This means that $|d_{ii}-1|< \epsilon$ and
\begin{equation*}
    1 - \epsilon < d_{ii} < 1 + \epsilon
\end{equation*}
Therefore,
\begin{equation*}
    (1 - \epsilon) \ \delta_i^j < (d_{ij}) < (1 + \epsilon) \ \delta_i^j
\end{equation*}
This means that for all $v \in T_qM$
\begin{equation*}
    (1- \epsilon)v^T(\delta_i^j)v \leq v^T(d_{ij})v \leq (1+\epsilon)v^T(\delta_i^j)v
\end{equation*}
Then we can say that
\begin{equation*}
    (1- \epsilon)(\delta_i^j) < (g_{2ij}) < (1+\epsilon)( \delta_i^j)
\end{equation*}
This means that, in the basis $\{u_i \}$, for all $v \in T_qM$ we have
\begin{equation*}
    (1- \epsilon)v^T(g_{1ij})v \leq v^T(g_{2ij})v \leq (1+ \epsilon)v^T(g_{1ij})v
\end{equation*}
in other words
\begin{equation*}
    \sqrt{1-\epsilon}|v|_{g_1 (q)} \leq |v|_{g_2 (q)} \leq \sqrt{1+\epsilon}|v|_{g_1 (q)}
\end{equation*}
and since $q$ was arbitrary, this holds on all of $M$.
\item Now let $\alpha : [0,1] \to M$ be a differentiable curve connecting $p$ and $q$, then we have
\begin{equation*}
    \sqrt{1 - \epsilon} \int_0^1 |\alpha^\prime (t)|_{g_1 (\alpha (t))} dt \leq \int_0^1 |\alpha^\prime (t)|_{g_2 (\alpha (t))} dt \leq \sqrt{1 + \epsilon} \int_0^1 |\alpha^\prime (t)|_{g_1 (\alpha (t))} dt
\end{equation*}
This means that
\begin{equation*}
    \sqrt{1 - \epsilon} \ L_{g_1} (\alpha) \leq L_{g_2} (\alpha) \leq \sqrt{1 + \epsilon} \ L_{g_1} (\alpha)
\end{equation*}
Since $L_{g_2}(\alpha) \geq \sqrt{1- \epsilon}L_{g_1}(\alpha)$ holds for all differentiable curves $\alpha$ connecting $p$ and $q$, the quantity $\sqrt{1 - \epsilon} \inf_\beta L_{g_1}(\beta)$ is a lower bound for $L_{g_2}(\alpha)$ over all $\alpha$, i.e. $L_{g_2}(\alpha) \geq  \sqrt{1- \epsilon}L_{g_1}(\alpha) \geq \sqrt{1 - \epsilon} \inf_\beta L_{g_1}(\beta)$. Since $\inf_\alpha L_{g_2}(\alpha)$ is the greatest lower bound over the curves $\alpha$, it's at least as large as this particular lower bound, thus $\inf_\alpha L_{g_2}(\alpha) \geq \sqrt{1 - \epsilon} \inf_\beta L_{g_1}(\beta)$. 
Similarly, since $L_{g_2} (\alpha) \leq \sqrt{1 + \epsilon} \ L_{g_1} (\alpha)$, for all such $\alpha$, we have $\inf_\alpha L_{g_2}(\alpha) \leq \sqrt{1+\epsilon} L_{g_1}(\alpha)$ for all such $\alpha$. So, $\inf_\alpha L_{g_2} (\alpha)$ is a lower bound for $\sqrt{1+\epsilon} L_{g_1}(\alpha)$ over all such $\alpha$, then the greatest lower bound must be at least as big, i.e. $\inf_\alpha L_{g_2}(\alpha) \leq \sqrt{1+\epsilon} \inf_\beta L_{g_1}(\beta)$. Then since $d_{g_m}(p,q) : = \inf_{\alpha} L_{g_m} (\alpha)$, we see that
\begin{align*}
    d_{g_2} (p,q) & = \inf_{\alpha} L_{g_2} (\alpha)\\
    & \geq \sqrt{1 - \epsilon} \ \inf_{\beta} L_{g_1} (\beta)\\
    & = \sqrt{1 - \epsilon} \ d_{g_1} (p,q)
\end{align*}
Then, similarly we have
\begin{align*}
    d_{g_2} (p,q) & = \inf_{\alpha} L_{g_2} (\alpha)\\
    & \leq \sqrt{1 + \epsilon} \ \inf_{\beta} L_{g_1} (\beta)\\
    & = \sqrt{1 + \epsilon} \ d_{g_1} (p,q)
\end{align*}
Which gives us
\begin{equation*}
     \sqrt{1 - \epsilon} \ d_{g_1} (p,q) \leq d_{g_2} (p,q) \leq \sqrt{1 + \epsilon} \ d_{g_1} (p,q)
\end{equation*}
\end{enumerate}   
\end{proof}

Part (1) says that $g_1$ and $g_2$ give nearly the same length to every tangent vector, and part (2) says that they give nearly the same distance between every pair of points. We use both when we compare the two metrics on a fixed neighborhood, for example if we choose to make $\parallel g_1 - g_2 \parallel_{C^2 (M,g_1)}$ sufficiently small, so that $|v|_{g_1}$ and $|v|_{g_2}$ are within a factor of 2 of each other. Together with the coordinate bounds above, this is all we need from the $C^k$-norm itself.

\subsection{ODE Theory}\label{sectionode}

The perturbation estimates in this paper reduce to comparing solutions of ordinary differential equations. When the metric changes, the coefficients of these systems (Christoffel symbols and curvature terms) change, and we need to know that the solutions change only slightly. The tool is Gr\"onwall’s inequality. We state it first, use it to derive a general estimate for the dependence of a solution on its initial condition and on a parameter, and then prove one estimate for each of the three systems that arise: parallel transport, the geodesic equation, and the equation for Jacobi fields.

\begin{lemma}\label{gron:int} (Gr\"onwall's inequality - integral form, p.712,~\cite{evans2010partial})
Let $u(t)$ be a real-valued, continuous, and non-negative function on the interval $[0,1]$. Suppose that there are constants $A, B \in \mathbb{R}$ with $A>0$ and $B \geq 0$, such that

\begin{equation*}
    u (t) \leq B + \int_0^t A u (s) ds
\end{equation*}
for all $t \in [0,1]$. Then $u (t)$ satisfies the inequality
\begin{equation*}
    u (t) \leq B e^{At} \leq B e^A
\end{equation*} 
for all $t \in [0,1]$.
\end{lemma}

Gr\"onwall’s inequality gives a general bound on how a solution of an ordinary differential equation depends on its initial value and on a time-dependent parameter. To derive it, let $y$ and $z$ be differentiable functions from $[0,1]$ to $\mathbb{R}^n$, driven by two parameter functions $\lambda$ and $\mu$.

\begin{align*}
        y(t) = \begin{bmatrix}
            y_1 (t)\\
            y_2 (t)\\
            \vdots\\
            y_n (t)
        \end{bmatrix} & & z(t) = \begin{bmatrix}
            z_1 (t)\\
            z_2 (t)\\
            \vdots\\
            z_n (t)
        \end{bmatrix}
\end{align*}
Then, suppose that we have the continuous system

\begin{equation*}
    F (t,x, \eta) = \begin{bmatrix}
        f_1 (t,x, \eta)\\
        f_2 (t,x, \eta)\\
        \vdots\\
        f_n (t,x, \eta)\\
    \end{bmatrix}
\end{equation*}
such that
\begin{align*}
    \frac{dy}{dt} & = F (t,y(t),\lambda (t)), \ y(0)=y_0\\
    \frac{dz}{dt} & = F (t,z(t),\mu (t)), \ z(0)=z_0
\end{align*}
where $\lambda (t)$ and $\mu (t)$ are continuous with respect to $t$ and are fixed throughout each system's solve. From here, we'll assume that $F$ is Lipschitz continuous in both the $x$ variable and the $\eta$ variable. That is, there exist $K_1, K_2 >0$ such that

\begin{align*}
    |F (t,x_1, \eta) - F(t,x_2, \eta)| & \leq K_1 |x_1 - x_2|\\
    |F (t,x, \eta_1) - F(t,x, \eta_2)| & \leq K_2 |\eta_1 - \eta_2|
\end{align*}
where $K_1$ and $K_2$ are constant with respect to $t$. Then, for $t \in [0,1]$, we see that

\begin{align*}
    y (t) & = y_0 + \int_0^t F (s, y(s), \lambda (s)) ds\\
    z (t) & = z_0 + \int_0^t F (s, z(s), \mu (s)) ds
\end{align*}
Then we have
\begin{equation*}
    y(t) - z(t) = y_0 - z_0 + \int_0^t (F(s,y(s), \lambda (s)) - F(s, z(s), \mu (s))) ds
\end{equation*}
Then, to set up our bound
\begin{align*}
    |y(t) - z(t)| & \leq |y_0 - z_0| + \int_0^t |F(s,y(s), \lambda (s)) - F(s, z(s), \mu (s))| ds\\
    & \leq |y_0 - z_0| + \int_0^t |F(s,y(s), \lambda (s)) - F(s, z(s), \lambda (s))| + |F(s,z(s), \lambda (s)) - F(s, z(s), \mu (s))|  ds\\
    & \leq |y_0 - z_0| + \int_0^t K_1 |y(s) - z(s)|ds + \int_0^t K_2 |\lambda (s) - \mu (s)| ds\\
    & \leq |y_0 - z_0| + \int_0^t K_1 |y(s) - z(s)|ds + \int_0^t K_2 \sup_{s \in [0,1]} |\lambda (s) - \mu (s)| ds \\
    & = |y_0 - z_0| + K_2 \sup_{s \in [0,1]}|\lambda (s) - \mu (s)|t + \int_0^t K_1 |y(s) - z(s)|ds\\
    & \leq (|y_0 - z_0| + K_2 \sup_{s \in [0,1]}|\lambda (s) - \mu (s)| ) + \int_0^t K_1 |y(s) - z(s)|ds
\end{align*}
So, for all $t \in [0,1]$ we have
\begin{equation*}
    |y(t) - z(t)| \leq (|y_0 - z_0| + K_2 \sup_{s \in [0,1]}|\lambda (s) - \mu (s)| ) + \int_0^t K_1 |y(s) - z(s)|ds
\end{equation*}
which is the condition that allows us to use Gr\"onwall's inequality. So, we'll have

\begin{equation*}
    \begin{split}
    |y(t) - z(t)| & \leq (|y_0 - z_0| + K_2 \sup_{s \in [0,1]}|\lambda (s) - \mu (s)| ) e^{t K_1}\\
    & \leq (|y_0 - z_0| + K_2 \sup_{s \in [0,1]} |\lambda (s) - \mu (s)| ) e^{ K_1}        
    \end{split}
\end{equation*}

This estimate is the template for the three lemmas below. In each of them, the Lipschitz constants are computed along the solutions from the bounds assumed in the statement, because the systems need not be globally Lipschitz (the geodesic equation, for example, is quadratic in $f^\prime$). The first system is the linear first-order equation satisfied by parallel transport.

\begin{lemma}\label{def:example1}

    Suppose that on the interval $[0,1]$, for $m=1,2$, $f_m (t) = (f_{m1}(t), \dots , f_{mn}(t))$ solves the linear homogeneous differential equation

    \begin{equation*}
        f_{mk}^\prime (t) + \sum_{j} f_{mj}(t) \lambda_{mj}^k (t) =0
    \end{equation*}
    
    Then we'll define
    \begin{equation*}
        F(t, f_m, \lambda_m) = \begin{bmatrix}
            - \sum_{j} f_{mj}(t) \lambda_{mj}^1 (t)\\
            \vdots\\
            - \sum_{j} f_{mj}(t) \lambda_{mj}^n (t)
        \end{bmatrix}
    \end{equation*}
    where $\lambda_{mj}^k$ are continuous.
    
    Now suppose that $\zeta \geq 0$ can be arbitrary, and suppose that there exist $C_0, C_1 \geq 0$ such that
    \begin{align*}
        \max_{1 \leq jk \leq n, t\in [0,1]}|\lambda_{1j}^k (t)|  &  \leq C_0\\
         |f_1 (t)| & \leq C_1\\
         |\lambda_{1j}^k (t) - \lambda_{2j}^k (t)| & \leq \zeta
    \end{align*}
    for all $1 \leq j,k \leq n$ and $t \in [0,1]$. Then by letting $A = n^2(C_0 + \zeta)$ and $B = |f_1(0) - f_2 (0)| + n^2C_1 \zeta$,
     \begin{equation*}
        \begin{split}
            |f_1 (t) - f_2 (t)| & \leq (|f_1 (0) - f_2 (0)| + n^2C_1 \zeta)e^{n^2(C_0 + \zeta)} 
        \end{split}
    \end{equation*}
\end{lemma}

\begin{proof}
    We'll see that
    \begin{align*}
        |F(t,f_1, \lambda_1) - F (t,f_1, \lambda_2)| & \leq \sum_{jk} |f_{1j} \lambda_{1j}^k (t) - f_{1j}(t) \lambda_{2j}^k (t)|\\
        & \leq \sum_{jk} |f_{1j}(t) | |\lambda_{1j}^k (t) - \lambda_{2j}^k (t)|\\
        & \leq \sum_{jk} C_1 |\lambda_{1j}^k (t) - \lambda_{2j}^k (t)|\\
        & \leq n^2 C_1 \zeta\\
        |F(t,f_1, \lambda_2) - F(t, f_2, \lambda_2)| & \leq \sum_{jk} |f_{1j}(t) \lambda_{2j}^k (t) - f_{2j}(t) \lambda_{2j}^k (t)|\\
        & \leq \sum_{jk} | \lambda_{2j}^k (t) | |f_{1j}(t)  - f_{2j}(t) |\\
        & \leq  \sum_{jk} (| \lambda_{1j}^k (t) | + |\lambda_{1j}^k (t) - \lambda_{2j}^k (t)|) |f_{1j}(t)  - f_{2j}(t) |\\
        & \leq n^2 (C_0 + \zeta) |f_{1}(t)  - f_{2}(t) |
    \end{align*}
    By combining what we've just shown, we have
    \begin{align*}
        |f_1 (t) - f_2(t)| & \leq |f_1 (0) - f_2(0)| + \int_0^t n^2 (C_0 + \zeta) |f_{1}(s)  - f_{2}(s) | + n^2 C_1 \zeta ds\\
        & \leq |f_1 (0) - f_2(0)| + n^2 C_1 \zeta + \int_0^t n^2 (C_0 + \zeta) |f_{1}(s)  - f_{2}(s) | \ ds
    \end{align*}
    Then by letting $A=n^2 (C_0 + \zeta)$, $B=|f_1 (0) - f_2 (0)| + n^2 C_1 \zeta$, we have the inequality
    \begin{equation*}
        |f_1(t) - f_2 (t)|  \leq B + \int_0^t A   |f_1(s) - f_2 (s)| \ ds
    \end{equation*}
    Then by the integral form of Gr\"onwall's inequality, lemma~\ref{gron:int}, we have
    \begin{equation*}
        \begin{split}
            |f_1 (t) - f_2 (t)| & \leq Be^{tA}\\
            & \leq Be^{A} 
        \end{split}
    \end{equation*} 
    as desired.
\end{proof}

Next we treat the geodesic equation. It is nonlinear, because the coefficients $\lambda$ depend on the solution itself, which is the reason for the extra term $I$ in the proof below.

\begin{lemma}\label{def:example2}
    Suppose that on the interval $[0,1]$, for $m=1,2$, $f_m (t) = (f_{m1}(t), \dots , f_{mn}(t))$ solves the homogeneous system
 \begin{equation*}
    f_{mk}^{\prime \prime}(t) + \sum_{ij} \lambda_{mij}^k (f_m(t)) f_{mi}^\prime (t) f_{mj}^\prime (t) =0
\end{equation*}  
where $\lambda_{mij}^k$ are continuous. Now suppose that $\zeta \geq 0$ is arbitrary, and suppose that there exist $C_0, C_1, C_2 \geq 0$, such that
\begin{align*}
    |\lambda_{1ij}^k (f_1 (t)) - \lambda_{1ij}^k (f_2 (t))| & \leq C_0 |f_1(t) - f_2 (t)|\\
     \max_{1 \leq ijk \leq n, x\in f_1([0,1]) \cup f_2([0,1])}|\lambda_{1ij}^k (x)| & \leq C_1\\
    |f_m^\prime (t)| & \leq C_2\\
    |\lambda_{1ij}^k (x) - \lambda_{2ij}^k (x)| & \leq \zeta
\end{align*}
for all $1 \leq i,j,k \leq n$, $t \in [0,1]$, and $x \in f_2 ([0,1])$. Then, by letting

\begin{align*}
          y_m (t) = \begin{bmatrix}
        f_m (t)\\
        f_m^\prime (t)
    \end{bmatrix}, & &
    A = 1+ n^3 C_2^2C_0 + 2n^3 C_1 C_2, & &
    B=|y_1 (0) - y_2 (0)| + n^3 C_2^2 \zeta,
\end{align*}
we'll have
\begin{equation*}
    \begin{split}
        |y_1 (t) - y_2 (t)| & \leq (|y_1 (0) - y_2 (0)| + n^3 C_2^2 \zeta)e^{A} 
    \end{split}
\end{equation*}

\end{lemma}

\begin{proof}
First, we let
\begin{equation*}
    F(t, y_m, \lambda_m) = \begin{bmatrix}
        f_{m1}^\prime (t)\\
        \vdots\\
        f_{mn}^\prime (t)\\
        - \sum_{ij} \lambda_{mij}^1 (f_m (t)) f_{mi}^\prime (t) f_{mj}^\prime (t)\\
        \vdots\\
        - \sum_{ij} \lambda_{mij}^n (f_m (t)) f_{mi}^\prime (t) f_{mj}^\prime (t)
    \end{bmatrix}
\end{equation*}
Then we see that
\begin{align*}
    |F(t, y_1, \lambda_1) - F(t, y_2, \lambda_1)|  & \leq |f_1^\prime (t) - f_2^\prime (t)| + \sum_{ijk} |\lambda_{1ij}^k (f_1 (t))f_{1i}^\prime (t) f_{1j}^\prime (t) - \lambda_{1ij}^k (f_2 (t)) f_{2i}^\prime (t) f_{2j}^\prime (t)|\\
    & \leq |f_1^\prime (t) - f_2^\prime (t)| + \sum_{ijk} |\lambda_{1ij}^k (f_1 (t)) f_{1i}^\prime (t) f_{1j}^\prime (t) - \lambda_{1ij}^k (f_2 (t)) f_{1i}^\prime (t) f_{1j}^\prime (t)|\\
    & + \sum_{ijk} |\lambda_{1ij}^k (f_2 (t)) f_{1i}^\prime (t) f_{1j}^\prime (t) - \lambda_{1ij}^k (f_2 (t)) f_{2i}^\prime (t) f_{2j}^\prime (t)|\\
    & = |f_1^\prime (t) - f_2^\prime (t)| + I + II = \star
\end{align*}
Then for $I$, we have
\begin{align*}
    I & = \sum_{ijk} |\lambda_{1ij}^k (f_1 (t)) f_{1i}^\prime (t) f_{1j}^\prime (t) - \lambda_{1ij}^k (f_2 (t)) f_{1i}^\prime (t) f_{1j}^\prime (t)|\\
    & = \sum_{ijk} |f_{1i}^\prime (t)| |f_{1j}^\prime (t)| |\lambda_{1ij}^k (f_1 (t))  - \lambda_{1ij}^k (f_2 (t))|\\
    & \leq n^3 C_2^2 C_0 |f_1 (t) - f_2 (t)|
\end{align*}
Next for $II$, we have
\begin{align*}
    II & = \sum_{ijk} |\lambda_{1ij}^k (f_2 (t)) f_{1i}^\prime (t) f_{1j}^\prime (t) - \lambda_{1ij}^k (f_2 (t)) f_{2i}^\prime (t) f_{2j}^\prime (t)|\\
    & \leq \sum_{ijk} C_1 (|f_{1i}^\prime (t)| |f_{1j}^\prime (t) - f_{2j}^\prime (t)|+ |f_{2j}^\prime (t)| |f_{1i}^\prime (t) - f_{2i}^\prime (t)|)\\
    & \leq \sum_{ijk}C_1 (2C_2 |f_1^\prime (t) - f_2^\prime (t)|)\\
    & \leq 2n^3 C_1 C_2 |f_1^\prime (t) - f_2^\prime (t)|
\end{align*}
Putting this together, we have
\begin{align*}
    \star & = |f_1^\prime (t) - f_2^\prime (t)| + I + II\\
    & \leq |f_1^\prime (t) - f_2^\prime (t)| + n^3 C_2^2 C_0 |f_1 (t) - f_2 (t)| + 2n^3 C_1 C_2 |f_1^\prime (t) - f_2^\prime (t)|\\
    & \leq (1+ n^3 C_2^2C_0 + 2n^3 C_1 C_2) |y_1 (t) - y_2 (t)|
\end{align*}
Then, we see that
\begin{align*}
      |F(t, y_2, \lambda_1) - F(t, y_2, \lambda_2)| & \leq |f_2^\prime (t) - f_2^\prime (t)| + \sum_{ijk} |\lambda_{1ij}^k (f_2 (t)) f_{2i}^\prime (t) f_{2j}^\prime (t) - \lambda_{2ij}^k (f_2 (t)) f_{2i}^\prime (t) f_{2j}^\prime (t) |\\
    & = \sum_{ijk} |f_{2i}^\prime (t)| |f_{2j}^\prime (t)| |\lambda_{1ij}^k (f_2 (t)) - \lambda_{2ij}^k (f_2 (t))|\\
    & \leq n^3 C_2^2 \zeta
\end{align*}

This means that we'll have

\begin{align*}
    |y_1 (t) - y_2 (t)| & \leq |y_1 (0) - y_2 (0)| + \int_0^t (1+ n^3 C_2^2C_0 + 2n^3 C_1 C_2) |y_1 (s) - y_2 (s)| + n^3 C_2^2 \zeta \ ds\\
    & \leq |y_1 (0) - y_2 (0)| + n^3 C_2^2 \zeta + \int_0^t (1+ n^3 C_2^2C_0 + 2n^3 C_1 C_2) |y_1 (s) - y_2 (s)| \ ds
\end{align*}

Then by letting $A = 1+ n^3 C_2^2C_0 + 2n^3 C_1 C_2$, and $B= |y_1 (0) - y_2 (0)| + n^3 C_2^2 \zeta$, we'll have the inequality
\begin{equation*}
    |y_1(t) - y_2 (t)| \leq B+ \int_0^t A |y_1 (s) - y_2 (s)| \ ds
\end{equation*}
Then by the integral form of Gr\"onwall's inequality, lemma~\ref{gron:int}, we have
\begin{equation*}
    \begin{split}
        |y_1 (t) - y_2 (t)| & \leq (|y_1 (0) - y_2 (0)| +  n^3 C_2^2 \zeta)e^{tA}\\
        & \leq (|y_1 (0) - y_2 (0)| +  n^3 C_2^2 \zeta)e^{A}
    \end{split}
\end{equation*}  
as desired.
\end{proof}

The third system is the linear second-order equation satisfied by Jacobi fields. We write it as a first-order system for $y = (f, f^\prime)$.

\begin{lemma}\label{def:example3}
    Suppose that on the interval $[0,1]$, for $m=1,2$, $f_m (t) = (f_{m1}(t), \dots, f_{mn}(t))$ solves the linear homogeneous system
    \begin{equation*}
    f_{mk}^{\prime \prime} (t) + \sum_{j=1}^n a_{kj}^m (t)f_{mj}^\prime (t) + \sum_{j=1}^n b_{kj}^m (t) f_{mj}(t) =0
    \end{equation*}   
    Then let $\lambda_m  = (A_m (t), B_m (t)) = (- (a_{kj}^m (t)), - (b_{kj}^m (t)))$. Now suppose that $\zeta \geq 0$ is arbitrary, and that there exist $C_0, C_1 \geq 0$, which are constant with respect to $t$, such that
    \begin{align*}
        |y_1 (t)|  & \leq C_0\\
        |M(t, \lambda_1)| & \leq C_1\\
        |M(t, \lambda_1) - M(t, \lambda_2)| =|A_1 (t) - A_2 (t)| + |B_1 (t) - B_2 (t)| & \leq \zeta
    \end{align*}
    for all $t \in [0,1]$, where 
    \begin{align*}
        y_m (t) = \begin{bmatrix}
            f_m (t)\\
            f_m^\prime (t)
        \end{bmatrix}, & &
         M(t, \lambda_m) = \begin{bmatrix}
            0 & I\\
            - (b_{kj}^m (t)) & - (a_{kj}^m (t))  
        \end{bmatrix}.
    \end{align*}
    Then, for $A = C_1 + \zeta$ and $B = |y_1(0) - y_2 (0)| + C_0 \zeta $, we'll have
    \begin{equation*}
        \begin{split}
            |y_1 (t) - y_2 (t)| & \leq B e^{ tA}\\
            & \leq B e^{ A}
        \end{split}
    \end{equation*}  
\end{lemma}

\begin{proof}
We see that

    \begin{align*}
        \frac{d y_m}{dt} & = \begin{bmatrix}
            f_m^\prime (t)\\
            f_m^{\prime \prime}(t)
        \end{bmatrix} 
        = \begin{bmatrix}
            f_m^\prime (t)\\
            B_m (t)f_m (t) + A_m (t)f_m^\prime (t) 
        \end{bmatrix}\\
        & = \begin{bmatrix}
            0 & I\\
            - (b_{kj}^m (t)) & - (a_{kj}^m (t))  
        \end{bmatrix}
        \begin{bmatrix}
            f_m (t)\\
            f_m^\prime (t)
        \end{bmatrix}
        = M(t, \lambda_m) y_m (t)
    \end{align*}
    Then let
    \begin{equation*}
        F(t, x, \lambda_m) = M(t, \lambda_m) x
    \end{equation*}
    Then we have
    \begin{align*}
        |F (t, y_1 (t), \lambda_1) - F (t, y_1 (t), \lambda_2)| & \leq (|A_1 (t) - A_2 (t)|+|B_1 (t) - B_2 (t)|) |y_1 (t)|\\
        & \leq C_0 \zeta\\
        |F (t, y_1 (t), \lambda_2) - F (t, y_2 (t), \lambda_2)| & \leq |M(t, \lambda_2) (y_1 (t) - y_2 (t))|\\
        & \leq (|M(t, \lambda_1)| + |M(t, \lambda_1) - M(t, \lambda_2) |)|y_1 (t) - y_2 (t)|\\
        & \leq  (C_1 + |A_1 (t) - A_2 (t)| + |B_1 (t) - B_2 (t)|  )|y_1 (t) - y_2 (t)|\\
        & \leq (C_1 + \zeta)|y_1 (t) - y_2 (t)|
    \end{align*}

    Then, putting this together we'll get
    \begin{align*}
        |y_1 (t) - y_2 (t)| & \leq |y_1(0) - y_2 (0)| + C_0 \zeta + \int_0^t( C_1 +\zeta)|y_1 (s) - y_2 (s)| ds
    \end{align*}
    Letting $A = C_1 + \zeta$ and $B = |y_1(0) - y_2 (0)| + C_0 \zeta $, by the integral form of Gr\"onwall's inequality, lemma~\ref{gron:int}, we have
    \begin{equation*}
        \begin{split}
            |y_1 (t) - y_2 (t)| & \leq (|y_1(0) - y_2 (0)| + C_0 \zeta) e^{t (C_1 + \zeta)}\\
            & \leq (|y_1(0) - y_2 (0)| + C_0 \zeta) e^{( C_1 + \zeta)}\\
        \end{split}
    \end{equation*}      
as desired.
\end{proof}

In the rest of the paper, we apply these estimates to the geodesic, parallel-transport, and Jacobi equations, where the coefficients are built from the Christoffel symbols of $g_1$ and $g_2$ and, for Jacobi fields, their derivatives.

\section{Geodesics and parallel transport under $C^1$ metric control}\label{sec:metricaprox}

The geometric norms and ordinary differential equation estimates of Section~\ref{sec:pre} are now applied to the objects that define parallel transport convolution: geodesics and parallel transport. Throughout this section, $(M,g_1)$ is an $n$-dimensional compact, connected Riemannian manifold, and $g_2$ is another Riemannian metric on $M$. Since $M$ is compact, both metrics are complete (see the Hopf–Rinow theorem, Chapter 6 of~\cite{lee2018introduction}). The question that drives the section is the following. Choose distinct points $p,q \in M$, and let $\gamma_1$ and $\gamma_2$ be geodesics from $p$ to $q$ with respect to $g_1$ and $g_2$. If $g_1$ and $g_2$ are sufficiently similar, are the parallel transports along $\gamma_1$ and $\gamma_2$ similar as well? To make \lq\lq sufficiently similar” precise, we apply the norm of Definition~\ref{def:norm1}, with $k = 1$ and background metric $g_1$, to the symmetric $(0,2)$-tensor $g_1 - g_2$.

\begin{definition}\label{def:normc1}
    Let $g_1$ and $g_2$ be metrics on the compact manifold $M$, then the $C^1$-norm between $g_1$ and $g_2$ is defined by
    \begin{equation*}
     \parallel g_1 - g_2 \parallel_{C^1 (M, g_1)}^2 = \sup_{p \in M} \big( |g_1 - g_2|_{g_1}^2+|\nabla^{g_1} (g_1 - g_2)|_{g_1}^2 \big)
    \end{equation*}
\end{definition}

Here $\nabla^{g_1}$ is the Levi-Civita connection of $g_1$, and $| \cdot |_{g_1}$ is the norm induced by $g_1$ on $\Gamma (Sym^2 (T^*M))$, as in definition~\ref{def:norm1}. Thus, definition~\ref{def:normc1} is the case $k = 1$ of definition~\ref{def:norm1}, with $a = g_1 - g_2$ and background metric $g_1$, so every result of section~\ref{sec:pre}, including proposition~\ref{prop:neat} and lemma~\ref{lemma:ineq1}, applies to it. This norm involves $g_2$ and its first derivatives, which are exactly the data that determine the Christoffel symbols of $g_2$, and therefore its geodesics and parallel transport. This is why $C^1$ control is the natural place to begin. 

We answer the question in three steps. Section~\ref{sec:fixcurve} fixes a curve, so that only the connection changes, and compares parallel transport along it. Section~\ref{sec:stabgeo} compares geodesics, using the ordinary differential equation estimates of Section~\ref{sectionode}. Section~\ref{sec:paratrans} combines the two preceding results, closeness of parallel transport along a fixed curve and closeness of the geodesics, to compare parallel transport along $\gamma_1$ and $\gamma_2$. The answer obtained there is incomplete, since it requires the initial velocities of $\gamma_1$ and $\gamma_2$ to be close. Closing that gap requires $C^2$ control and is the subject of the next section.

\subsection{Parallel transport along a fixed curve}\label{sec:fixcurve}

We begin with the simplest comparison, in which only the connection changes. Fix a curve $\gamma$ and transport a vector along it with respect to $g_1$ and with respect to $g_2$. The curve is the same in both cases, so any difference in the results comes from the two metrics alone, through their Levi-Civita connections. The hypotheses of the theorem below are phrased in terms of lengths, so we recall that the length of a differentiable curve $\alpha : [0,1] \to M$ with respect to $g_m$ $(m=1,2)$ is
\begin{equation*}
 L_{g_m} (\alpha) = \int_0^1 |\alpha^\prime (t)|_{g_m (\alpha (t))} dt   
\end{equation*}
(see~\cite[p.~43]{do1992riemannian}).

The theorem below states that the $C^1$-norm controls parallel transport along a fixed curve. Let $\gamma$ be a geodesic of $g_1$, and transport a vector of bounded length along $\gamma$ with respect to each metric. The two results are close when $g_1$ and $g_2$ are close in the $C^1$-norm.

\begin{theorem} \label{thrm:neat}
    Choose distinct $p,q \in M$, $L>0$. Let $\gamma : [0,1] \to M$ be a geodesic with respect to $g_1$ connecting $p$ to $q$, such that $L_{g_m} (\gamma)< L$ for $m=1,2$. Then choose $v \in T_pM$ such that $|v|_{g_m(p)} \leq L$, and let $P_{g_1}^\gamma, P_{g_2}^\gamma : T_pM \to T_qM$ be the parallel transports along $\gamma$ with respect to $g_1$ and $g_2$, respectively. Let $P_{g_1}^\gamma (v)=u_1$ and $P_{g_2}^\gamma (v)=u_2$. For every $\epsilon >0$, there exists $\delta (L, \epsilon) > 0$ such that $\parallel g_1 - g_2 \parallel_{C^1 (M,g_1)} < \delta (L, \epsilon)$ implies
    \begin{equation*}
        |u_1 - u_2|_{g_1 (q)} < \epsilon
    \end{equation*}
\end{theorem}

The proof compares the two transports through a $g_1$-parallel frame along $\gamma$. Let $V_1 (t)$ be the $g_1$-parallel field along $\gamma$ with $V_1 (0) = v$, and let $V_2 (t)$ be the $g_2$-parallel field with $V_2 (0) = v$. Since $V_1 (t)$ has no $g_1$-covariant derivative, the difference $V_1 (t) - V_2 (t)$ changes at exactly the rate of the $g_1$-covariant derivative of $V_2 (t)$, and this rate is expressed through the Christoffel symbols of $g_2$. The following lemma shows that the $C^1$-norm controls these symbols at a point, in suitable normal coordinates. We prove the lemma here.

\begin{lemma}\label{lemma:ineq0}
    Suppose $0< \epsilon < \frac{1}{2}$ and $q \in M$, if $\parallel g_1 - g_2 \parallel_{C^1 (M, g_1)}< \epsilon$, then there exists a coordinate chart about $q$ such that $|\Gamma_{ij}^{k,g_2} (q)|< 3n \epsilon$.
\end{lemma}
\begin{proof}
    Choose $q \in M$. By the Simultaneous diagonalization theorem (Exercise 21, p. 202,~\cite{roman2007advanced}), there exists a $g_1$-orthonormal basis of $T_qM$ such that $g_2 (q) =(g_{2ij}) = D = (d_{ij})$ where $d_{ij}=0$ for $i \neq j$; since any $g_1 (q)$-orthonormal basis defines a valid normal coordinate system about $q$ with respect to $g_1$, we choose our normal coordinate system to be this one. From proposition 5.24, p. 132,~\cite{lee2018introduction}, we have $g_1 (q) = \delta_i^j$. Then, we have
\begin{align*}
    |g_1 - g_2|_{g_1}^2 & = \sum_{ij \alpha \beta} (g_{1ij} - g_{2ij})(g_{1 \alpha \beta} - g_{2 \alpha \beta})g_1^{i \alpha} g_1^{j \beta}\\
    & = \sum_{ij \alpha \beta} (\delta_i^j - g_{2ij})(\delta_\alpha^\beta - g_{2 \alpha \beta}) \delta_i^\alpha \delta_j^\beta\\
    & = \sum_{ij}(\delta_i^j - g_{2ij})^2 = \sum_{i=1}^n (d_{ii}-1)^2
\end{align*}
Since $\parallel g_1 - g_2 \parallel_{C^1 (M,g_1)}$ is a supremum over all of $M$, the value of $\parallel g_1 - g_2 \parallel_{C^1 (M,g_1)}^2$ is at least the value of $|g_1 - g_2|_{g_1}^2$ at the point $q$. So, we have
\begin{equation*}
    \sum_{i=1}^n (d_{ii}-1)^2 = |g_1 - g_2|_{g_1(q)}^2 \leq \parallel g_1 - g_2 \parallel_{C^1 (M,g_1)}^2 < \epsilon^2
\end{equation*}

This means that $|d_{ii}-1|< \epsilon$, and we have
\begin{equation*}
    1 - \epsilon < d_{ii} < 1 + \epsilon
\end{equation*}

Since $0 < \epsilon < \frac{1}{2}$, we'll have
\begin{equation*}
    d_{ii}^{-1} < \frac{1}{1-\epsilon} < 2
\end{equation*}
 
    Also, from proposition 5.24, p.132, ~\cite{lee2018introduction}, we have 
    \begin{equation*}
        \Gamma_{ij}^{k,g_1} (q)= 0 = \frac{\partial g_{1ij}}{\partial x_k} (q)
    \end{equation*}
    which means that
    \begin{align*}
     \nabla^{g_1} g_1 (q) & = \big( \frac{\partial g_{1ij}}{\partial x_k} (q) - \big( \sum_{l=1}^n ( g_{1lj} (q) \Gamma_{ki}^{l,g_1} (q) + g_{1il}(q) \Gamma_{kj}^{l,g_1} (q) ) \big) \big) dx^k \otimes dx^i \otimes dx^j\\
    & =  0
    \end{align*}
    Then since we're taking the Levi-Civita connection with respect to $g_1$, we also have
    \begin{align*}
        \nabla^{g_1} g_2 (q) & = \big( \frac{\partial g_{2ij}}{\partial x_k} (q) -  \big( \sum_{l=1}^n ( g_{2lj}(q) \Gamma_{ki}^{l,g_1} (q) + g_{2il}(q) \Gamma_{kj}^{l,g_1} (q) ) \big) \big) dx^k \otimes dx^i \otimes dx^j\\
        & =\frac{\partial g_{2ij}}{\partial x_k} (q) dx^k \otimes dx^i \otimes dx^j\\
    \end{align*}
    With the difference between the two tensors, at our point $q$, being
    \begin{equation*}
        \nabla^{g_1} g_1 - \nabla^{g_1} g_2 = - \nabla^{g_1} g_2 =  -\frac{\partial g_{2ij}}{\partial x_k} dx^k \otimes dx^i \otimes dx^j
    \end{equation*}
    Then, putting this together with the $C^1$-norm, gives us
    \begin{align*}
     |\nabla^{g_1} g_1 - \nabla^{g_1} g_2|_{g_1}^2 = |\nabla^{g_1} g_2|_{g_1}^2 & = 
     \big( \frac{\partial g_{2ij}}{\partial x_k} \big) \big( \frac{\partial g_{2\alpha \beta}}{\partial x_l} \big) g_1^{kl} g_1^{i \alpha} g_1^{j \beta}\\
    & = \big( \frac{\partial g_{2ij}}{\partial x_k} \big) \big( \frac{\partial g_{2 \alpha \beta}}{\partial x_l} \big) \delta_k^l \delta_i^\alpha \delta_j^\beta\\
    & = \big( \frac{\partial g_{2ij}}{\partial x_k} \big)^2
    \end{align*}
    Then, combining this with what we know about the metrics in our neighborhood and the $C^1$-norm, we know that
    \begin{align*}
        |\frac{\partial g_{2ij}}{\partial x_k} | & = \sqrt{\big( \frac{\partial g_{2ij}}{\partial x_k} \big)^2}\\
        & \leq \sqrt{ \sum_{i,j =1}^n \big( \frac{\partial g_{2ij}}{\partial x_k} \big)^2 }\\
        & \leq |\nabla^{g_1} g_2|_{g_1} = \sqrt{ |\nabla^{g_1} g_1 - \nabla^{g_1} g_2|_{g_1}^2 }\\
        & \leq \parallel g_1 - g_2 \parallel_{C^1 (M, g_1)} < \epsilon
    \end{align*}
    Then we can use this to put a bound on $\Gamma_{ij}^{k,g_2}(q)$. We have
    \begin{align*}
        \bigg| \sum_l \Gamma_{ij}^{l,g_2}g_{2 \alpha l} \bigg| & = \bigg| \frac{1}{2} \big( \frac{\partial g_{2\alpha i}}{\partial x_j} + \frac{\partial g_{2 \alpha j}}{\partial x_i} - \frac{\partial g_{2ij}}{\partial x_\alpha} \big) \bigg|\\
        & \leq \frac{1}{2} \bigg( \bigg| \frac{\partial g_{2\alpha i}}{\partial x_j} \bigg| + \bigg| \frac{\partial g_{2 \alpha j}}{\partial x_i} \bigg| + \bigg| \frac{\partial g_{2ij}}{\partial x_\alpha} \bigg| \bigg) \\
        & < \frac{3 \epsilon}{2}
        \end{align*}
    Which means that at $q$, we have
    \begin{equation*}
       \big| \sum_l \Gamma_{ij}^{l,g_2}g_{2 \alpha l} \big| < \frac{3 \epsilon}{2}
    \end{equation*}
    Then at our point $q$, we have
        \begin{equation*}
          \Gamma_{ij}^{k,g_2} = \sum_{l\alpha} \Gamma_{ij}^{l,g_2} g_{2 \alpha l} g_2^{\alpha k}  
        \end{equation*}

    Since we know that $(d_{ii})^{-1} < \frac{1}{1- \epsilon} < 2$, we make the observation
    \begin{equation*}
        (g_2^{ij}) = (d_{ij}^{-1})
    \end{equation*}
    where $d_{ij}^{-1}=0$ when $i \neq j$. So, we have $|g_2^{ij}| =|d_{ij}^{-1}| < 2$. Then
    \begin{align*}
       | \Gamma_{ij}^{k,g_2} | &= \big| \sum_{l \alpha} \Gamma_{ij}^{l,g_2} g_{2 \alpha l}g_2^{\alpha k} \big|\\
       & = \big| \sum_\alpha g_2^{\alpha k} \big( \sum_l \Gamma_{ij}^{l,g_2} g_{2 \alpha l} \big) \big|\\
       & \leq \big| \sum_\alpha |g_2^{\alpha k}| \big| \sum_l \Gamma_{ij}^{l,g_2} g_{2 \alpha l}  \big|\\
       & < \sum_\alpha 2 \big| \sum_l \Gamma_{ij}^{l,g_2} g_{2 \alpha l} \big|\\
       & = \big| \sum_l \Gamma_{ij}^{l,g_2} g_{2 \alpha l} \big| 2n\\
       & < \frac{3 \epsilon}{2} \cdot 2n = 3n \epsilon
    \end{align*}
    Which gives us our desired bound.
\end{proof}

With this lemma, the proof of theorem~\ref{thrm:neat} compares the two transported fields in orthonormal frames. Lemma~\ref{lemma:ineq0} makes the $g_1$-covariant derivative of the $g_2$-transported field small along $\gamma$, and integrating this bound from $p$, where the two fields agree, controls their difference at $q$. The details are in Appendix~\ref{app:neat}. This settles the case of a fixed curve. What remains of the question posed at the start of this section is that the curve itself changes with the metric. 

\subsection{Stability of geodesics}\label{sec:stabgeo}

Now we compare geodesics themselves. A geodesic solves a system of ordinary differential equations whose coefficients are the Christoffel symbols of the metric, so we first compare the coefficients. Lemma~\ref{lemma:ineq0} bounds $\Gamma^{g_2}$ at a single point in special coordinates, but here we need the difference $\Gamma^{g_1} - \Gamma^{g_2}$ on a whole coordinate neighborhood. Because the Christoffel symbols are built from the metric and its first derivatives, the $C^1$-norm controls this difference.

\begin{proposition}\label{prop:prop1}
    Suppose we fix a local coordinate system $U$. For $k = 1,2$, and for all $\epsilon >0$ there exists $\delta (M,g_1,U,k, \epsilon)>0 $ such that $\parallel g_1 - g_2 \parallel_{C^k (M, g_1)} < \delta (M,g_1,U,k, \epsilon)$ implies that
    \begin{equation*}
        |\Gamma_{ij}^{l,g_1} - \Gamma_{ij}^{l,g_2}| < \epsilon
    \end{equation*}
    on the coordinate system $U$.
\end{proposition}

The proof first bounds the inverse metric and its difference from $g_1^{-1}$ by a Neumann series. The Christoffel symbols, which involve the inverse metric and first derivatives, then differ by a constant times the $C^1$ distance. The details are in Appendix~\ref{app:prop-prop1}.

Proposition~\ref{prop:prop1} puts the coefficients of the geodesic equations for $g_1$ and $g_2$ within $\epsilon$ of each other on $U$. By lemma~\ref{def:example2}, solutions of such equations depend continuously on their coefficients and on their initial data. So geodesics of $g_1$ and $g_2$ whose initial velocities are close should remain close. The next theorem makes this precise, measuring closeness by the distance $d_{g_1}$ and by the Euclidean norm of the coordinate velocities.

\begin{theorem}\label{def:thrm3}
For a fixed coordinate neighborhood $U$, for $m=1,2$, suppose $\gamma_m  : [0,1] \to M$ are geodesics with respect to $g_m$, emanating from $p \in M$, such that $\gamma_1 ([0,1]) \cup \gamma_2 ([0,1]) \subset U$, with $L_{g_m}(\gamma_m) < L$. Then, for all $ \epsilon > 0$, there exist $\delta_1 (M, g_1, U, k , L, \epsilon) >0$ and $\delta_2 (M, g_1, U, L, \epsilon)>0$ (for $k=1,2$), such that $\parallel g_1 - g_2 \parallel_{C^k(M, g_1)} < \delta_1 (M, g_1, U, k , L, \epsilon)$ and $|\gamma_1^\prime (0) - \gamma_2^\prime (0)|_{g_1 (p)} < \delta_2 (M, g_1, U, L, \epsilon)$ imply that for all $t \in [0,1]$, we have
\begin{equation*}
    d_{g_1} (\gamma_1 (t), \gamma_2 (t)) < \epsilon
\end{equation*}
Moreover, for all $t \in [0,1]$, we have
\begin{equation*}
    |\gamma_1^\prime (t) - \gamma_2^\prime (t)| < \epsilon
\end{equation*}
where $|\cdot|$ is the Euclidean norm from definition~\ref{def:eucl}.
\end{theorem}

The proof applies lemma~\ref{def:example2} to the geodesic equation, with the Christoffel symbols of $g_1$ and $g_2$ as coefficients. Proposition~\ref{prop:prop1} supplies the closeness of the coefficients, and proposition~\ref{prop:neat} together with the equivalence of $g_1$ and the Euclidean norm on $U$ bounds the remaining constants. The details are in Appendix~\ref{app:thrm3}.

Theorem~\ref{def:thrm3} controls the geodesics through two quantities: the norm $\parallel g_1 - g_2 \parallel_{C^k (M,g_1)}$ and the difference of the initial velocities. It is natural to ask whether the second can be removed, that is, whether there is a $g_2$-geodesic starting at $p = \gamma_1 (0)$ for which $\parallel g_1 - g_2 \parallel_{C^k (M,g_1)}$ alone suffices. The geodesic with the same initial velocity is such a choice, and it gives the following corollary, which compares the exponential maps of $g_1$ and $g_2$.

\begin{corollary}\label{def:cor1}
    Fix a coordinate neighborhood $U$, then choose $p \in U$, and $v \in T_pM$ such that $|v|_{g_m (p)}<L$, for $m=1,2$, and $\gamma_1 ([0,1]) \subset U$ for $\gamma_1 (t) = exp_p^{g_1} (vt)$ and suppose that $\gamma_2 ([0,1]) \subset U$ (where $\gamma_2 (t) = exp_p^{g_2} (vt)$). Then for all $\epsilon > 0$, there exists $\delta (M, g_1, U, k, L, \epsilon)>0$ such that $\parallel g_1 - g_2 \parallel_{C^k (M, g_1)} < \delta (M, g_1, U, k, L, \epsilon)$ implies that for all $t \in [0,1]$, we have
    \begin{equation*}
        d_{g_1} (\gamma_1 (t), \gamma_2 (t)) < \epsilon
    \end{equation*}
\end{corollary}

\begin{proof}
    First, choose $\epsilon>0$, from theorem~\ref{def:thrm3} we know that there exist $\delta_1 (M,g_1,U,k,L, \epsilon)>0$ and $\delta_2 (M,g_1, U, L, \epsilon)>0$, such that $\parallel g_1 - g_2 \parallel_{C^k (M,g_1)} < \delta_1 (M,g_1,U,k,L, \epsilon)$ and $|\gamma_1^\prime (0) - \gamma_2^\prime (0)|_{g_1 (p)} < \delta_2 (M,g_1, U, L, \epsilon)$, then for all $t \in [0,1]$, we have
    \begin{equation*}
        d_{g_1}(\gamma_1 (t), \gamma_2 (t)) < \epsilon
    \end{equation*}

    By how we've defined these geodesics, we see that $\gamma_1^\prime (0) = v = \gamma_2^\prime (0)$, and $|\gamma_1^\prime (0) - \gamma_2^\prime (0)| = 0 = |\gamma_1^\prime (0) - \gamma_2^\prime (0)|_{g_1 (p)}$. So, we have $|\gamma_1^\prime (0) - \gamma_2^\prime (0)|_{g_1 (p)} < \delta_2 (M, g_1, U, L, \epsilon)$ for any $\delta_2 (M,g_1, U, L, \epsilon)>0$. So let $\delta (M,g_1,U,k,L, \epsilon) = \delta_1 (M,g_1,U,k,L, \epsilon)$. Then, $\parallel g_1 - g_2 \parallel_{C^k (M,g_1)} < \delta (M,g_1,U,k,L, \epsilon)$ implies that
    \begin{equation*}
        d_{g_1}(\gamma_1 (t), \gamma_2 (t)) < \epsilon
    \end{equation*}
    as desired.   
\end{proof}

In particular, the exponential maps of $g_1$ and $g_2$ at $p$ are close, on vectors whose geodesics stay in $U$, when the metrics are close in the $C^k$-norm for $k = 1,2$.

\subsection{Parallel transport along different geodesics}\label{sec:paratrans}

We now return to the question posed at the beginning of this section, in which both the connection and the curve change. Let $\gamma_1$ and $\gamma_2$ be geodesics from $p$ to $q$ with respect to $g_1$ and $g_2$. Parallel transport along $\gamma_m$ solves a linear first-order system of ordinary differential equations, whose coefficients are the Christoffel symbols of $g_m$ along $\gamma_m$, contracted with its velocity. Proposition~\ref{prop:prop1} and theorem~\ref{def:thrm3} show that these coefficients, for $m = 1, 2$, are close when the metrics and the initial velocities of the geodesics are close. Lemma~\ref{def:example1} converts this into closeness of the transported vectors. The following theorem combines these results.

\begin{theorem}\label{def:thrm5}
    Fix a coordinate neighborhood $U$, then choose distinct $p,q \in U$. For $m =1,2$, let $\gamma_m : [0,1] \to M$ be geodesics for $g_m$, connecting $p$ and $q$, with $L_{g_m}(\gamma_m) <L$, such that $\gamma_1 ([0,1]) \subset U$ and suppose that $\gamma_2 ([0,1]) \subset U$. Then for all $ \epsilon > 0$, there exist $\delta_1 (M, g_1, U, k, L, \epsilon)>0$ and $\delta_2 (M, g_1, U, L, \epsilon)>0$, such that

    \begin{equation*}
    \begin{split}
      \parallel g_1 - g_2 \parallel_{C^k (M, g_1)} & < \delta_1 (M, g_1, U, k, L, \epsilon)\\
      |\gamma_1^\prime (0) - \gamma_2^\prime (0)|_{g_1 (p)} & < \delta_2 (M, g_1, U, L, \epsilon)
    \end{split}
    \end{equation*}
    implies that for all $v \in T_pM$ with $|v|_{g_1 (p)} <L$, we have
    \begin{equation*}
        |P_{g_1}^{\gamma_1} (v) - P_{g_2}^{\gamma_2} (v) |_{g_1 (q)} < \epsilon
    \end{equation*}

    Moreover, for all $\epsilon >0$, for $v_1, v_2 \in T_pM$ with $|v_1|_{g_1(p)},|v_2|_{g_1(p)} < L$, there exist $\delta_1 (M, g_1, U, k, L, \epsilon)>0$, $\delta_2 (M, g_1, U, L, \epsilon)>0$, and $\delta_3 (\epsilon)>0$ such that
        \begin{equation*}
    \begin{split}
      \parallel g_1 - g_2 \parallel_{C^k (M, g_1)} & < \delta_1 (M, g_1, U, k, L, \epsilon)\\
      |\gamma_1^\prime (0) - \gamma_2^\prime (0)|_{g_1 (p)} & < \delta_2 (M, g_1, U, L, \epsilon)\\
      |v_1-v_2|_{g_1 (p)} & < \delta_3 (\epsilon)
    \end{split}
    \end{equation*}
    implies that
    \begin{equation*}
    |P_{g_1}^{\gamma_1} (v_1) - P_{g_2}^{\gamma_2} (v_2) |_{g_1 (q)} < \epsilon
    \end{equation*}
       
\end{theorem}

The proof writes parallel transport along each curve as the solution of a linear system of ordinary differential equations whose coefficients are the Christoffel symbols along the curve. It compares the two coefficients by separating the effect of the curves, of their velocities, and of the connections, and then applies Gr\"onwall’s inequality. The details are in Appendix~\ref{app:thrm5}.

Theorem~\ref{def:thrm5} answers the question posed at the start of this section only in part. As in theorem~\ref{def:thrm3}, it assumes that the initial velocities of $\gamma_1$ and $\gamma_2$ are close, in addition to the closeness of the metrics. In corollary~\ref{def:cor1} this assumption was not needed, because the geodesics had the same initial velocity. When the geodesics are specified only by their endpoints $p$ and $q$, however, nothing guarantees that their initial velocities are close. The next section shows that, for $g_2$ close to $g_1$, a $g_2$-geodesic from $p$ to $q$ can be chosen with initial velocity close to that of $\gamma_1$.

\section{Jacobi fields and conjugate points under $C^2$-metric control}\label{sec:C2}

The previous section left us with one missing step: producing, for points $p$ and $q$, a $g_2$-geodesic from $p$ to $q$ whose initial velocity is close to that of $\gamma_1$. This amounts to inverting the exponential map of $g_2$ near that of $g_1$, which is controlled through its differential. The differential is described by Jacobi fields, and the Jacobi equation involves the curvature, and hence two derivatives of the metric. Throughout this section, $(M,g_1)$ is a compact, connected Riemannian manifold and $g_2$ is another Riemannian metric on $M$, as in section~\ref{sec:metricaprox}.

We proceed in four steps. Section~\ref{sec:C2control} shows that the curvature of $g_2$ is controlled by the $C^2$-norm. Section~\ref{sec:jac} uses this to compare Jacobi fields, and from them the differentials of the exponential maps. Section~\ref{sec:conj} shows that, under the same control, $g_2$ has no conjugate points, where these differentials would degenerate, along geodesics of a fixed length. Section~\ref{sec:cut} combines these results with those of section~\ref{sec:metricaprox} to construct the connecting geodesic. We begin by making $C^2$ control precise.

\begin{definition}\label{def:norm12}
    Let $(M, g_1)$ be an $n$-dimensional complete, connected, compact Riemannian manifold. Let $g_2$ be another Riemannian metric on $M$, which makes $(M, g_2)$ complete. We define the $C^2$-norm between these two metrics as
    \begin{equation*}
     \parallel g_1 - g_2 \parallel_{C^2 (M,g_1)}^2 = \sup_{p \in M} \big( |g_1 - g_2|_{g_1}^2 + |\nabla^{g_1} (g_1 - g_2)|_{g_1}^2 + | (\nabla^{g_1})^2 (g_1 -  g_2)|_{g_1}^2 \big)   
    \end{equation*}
\end{definition}

Here $\nabla^{g_1}$ is the Levi-Civita connection of $g_1$, and $|\cdot|_{g_1}$ is the norm induced by $g_1$ on tensors of the appropriate rank, as in definition~\ref{def:norm1}. Thus, definition~\ref{def:norm12} is the case $k = 2$ of definition~\ref{def:norm1}, with $a= g_1 - g_2$ and background metric $g_1$, so every result of Section~\ref{sec:pre} applies to it, as for the $C^1$-norm. Since it also contains the terms of the $C^1$-norm, it dominates the latter, and everything proved in section~\ref{sec:metricaprox} remains available.

\subsection{Curvature under $C^2$ control}\label{sec:C2control}

The Jacobi equation involves the curvature of the metric, so we begin by showing that the $C^2$-norm controls it. The $C^1$-norm already controlled the Christoffel symbols of $g_2$ (proposition~\ref{prop:prop1}), and these determine its geodesics and parallel transport. The curvature, however, is built from the Christoffel symbols together with their first derivatives, which involve two derivatives of the metric. We work in a coordinate neighborhood $U$, as in remark~\ref{def:rmk1}.

Let $\{ X_l \}$ be the coordinate frame for $U$. Then the curvature for $g_m$ is 
\begin{equation*}
 R^{g_m} = \sum_{ijkl} R_{ijk}^{l,g_m} dx^i \otimes dx^j \otimes dx^k \otimes \partial_l   
\end{equation*}
where 
\begin{equation*}
    R^{g_m} (X_i, X_j)X_k = \sum_l R_{ijk}^{l, g_m} X_l
\end{equation*}
In any coordinate chart, the components of the curvature are given in terms of the Christoffel symbols by the following formula (proposition 7.4 of~\cite[p. 197]{lee2018introduction}):
\begin{equation*}
 R_{ijk}^{l, g_m} = \partial_i \Gamma_{jk}^{l, g_m}  -  \partial_j \Gamma_{ik}^{l, g_m}  + \Gamma_{jk}^{s, g_m} \Gamma_{is}^{l, g_m} - \Gamma_{ik}^{s, g_m} \Gamma_{js}^{l, g_m} 
\end{equation*}

The formula above shows that the components $R_{ijk}^{l,g_m}$ are determined by the Christoffel symbols and their first derivatives. The $C^2$-norm therefore gives the following local control.

\begin{proposition}\label{def:prop2}
For a fixed coordinate neighborhood $U$, for all $\epsilon >0$ there exists $\delta (M, g_1, U, \epsilon) >0$ such that $\parallel g_1 - g_2 \parallel_{C^2 (M, g_1)} < \delta (M, g_1, U, \epsilon)$ implies that
\begin{equation*}
    |R_{ijk}^{l, g_1} - R_{ijk}^{l, g_2}| < \epsilon
\end{equation*}
on $U$.
\end{proposition}

The proof differentiates the inverse metric and the Christoffel symbols once more, so the curvature differs by a constant times the $C^2$ distance. The details are in Appendix~\ref{app:prop2}.

For later use, we also record notation for the sizes of the curvature and of its derivative on $U$.

\begin{Remark}\label{def:rmk2}
    From this point on, since $\overline{U}$ is compact (remark~\ref{def:rmk1}), for $m=1,2$, we define:
    \begin{align*}
        \max |R^{g_m}| & := \max_{ijkl} \big( \sup_{p \in \overline{U}} |R_{ijk}^{l, g_m}| \big)\\
        \max | \partial R^{g_m}| & := \max_{ijkls} \big( \sup_{p \in \overline{U}}  \big| \frac{\partial R_{ijk}^{l, g_m}}{\partial x^s} \big| \big)
    \end{align*} 
\end{Remark}

Proposition~\ref{def:prop2} is the curvature analogue of proposition~\ref{prop:prop1}. It supplies the coefficients of the Jacobi equation, which we derive in the next subsection.

\subsection{Jacobi fields and the differential of the exponential map}\label{sec:jac}

Jacobi fields give a concrete description of the differential of the exponential map. Fix $p \in M$. For $m=1,2$ and $w,u \in T_pM$, let $J_{mwu}$ be the Jacobi field of $g_m$ along the geodesic $t \mapsto exp_p^{g_m}(tw)$ with $J_{mwu}(0)=0$ and $J_{mwu}'(0)=u$. Then $d(exp_p^{g_m})_w(u)=J_{mwu}(1)$. So to compare the differentials of $exp_p^{g_1}$ and $exp_p^{g_2}$, it is enough to compare the corresponding Jacobi fields. We do this in a fixed coordinate neighborhood $U$, where the Jacobi equation becomes a linear second-order system for the components of the field, of the kind treated in lemma~\ref{def:example3}. We set up this system next.

Fix a coordinate neighborhood $U$, then for $p \in U$, choose $v \in T_pM$, such that $\gamma_1 ([0,1]) \subset U$ for $\gamma_1 (t) = exp_p^{g_1}(tv)$, and suppose that $\gamma_2 (t)=exp_p^{g_2}(tv)$ satisfies $\gamma_2 ([0,1]) \subset U$. Then, let $\{ X_i \}$ be the coordinate frame associated with $U$ and let $(\gamma_m^1, \gamma_m^2, \dots , \gamma_m^n)$ be the representation of $\gamma_m (t)$ in the neighborhood $U$. The derivative of $\gamma_m$ gives us a vector field along the curve $\gamma_m$ which, by the chain rule, is given in coordinates by,

\begin{equation*}
    \gamma_m^\prime (t) = \sum_i \dot{\gamma}_m^i X_i (\gamma_m (t))
\end{equation*}
Let $J_m$ be a Jacobi field along $\gamma_m$, with components $f_{mj}$ in the frame $\{ X_l \}$,
\begin{equation*}
    J_m (t) = \sum_j f_{mj}(t)X_j (\gamma_m (t))
\end{equation*}
where $J_m (t)$ solves the Jacobi equation
\begin{equation*}
    (\nabla_{\gamma_m^\prime(t)}^{g_m})^2J_m(t)+ R^{g_m}(J_m (t), \gamma_m^\prime (t)) \gamma_m^\prime (t)=0
\end{equation*}
A natural way to write the Jacobi equation would be to transport an orthonormal frame along $\gamma_m$, in which the only coefficient is the curvature. But $J_1$ and $J_2$ lie along different curves, and the parallel orthonormal frames of $g_1$ and $g_2$ are different frames, so their components cannot be compared until the frames themselves are compared. That would be a further comparison of parallel transports, and the conclusion we want, closeness of the differentials of the exponential maps, is measured in coordinates in any case. We therefore keep the fixed coordinate frame $\{X_l\}$ of $U$, which is the same for both metrics and both curves. The price is a longer equation.

Expanding the covariant derivatives and the curvature term in the frame $\{X_l\}$, and using the geodesic equation to eliminate $\ddot\gamma_m$, turns the Jacobi equation into a linear homogeneous system for the components. The derivation is a direct computation, given in Appendix~\ref{app:jaceq}. The components satisfy

\begin{equation}\label{def:jaceq}
    f_{mk}^{\prime \prime}(t) + a_{kj}^{g_m}(t) f_{mj}^\prime (t) + b_{k j}^{g_m}(t) f_{m j}(t) =0
\end{equation}
where, in Einstein's summation convention, $a_{kj}^{g_m}(t) = 2  \Gamma_{ij}^{k, g_m}(\gamma_m (t)) \dot{\gamma}_m^i$ and $b_{k j}^{g_m}(t) =   \dot{\gamma}_m^\mu \dot{\gamma}_m^s \frac{\partial \Gamma_{\mu j}^{k, g_m}}{\partial x^s} (\gamma_m (t)) - \dot{\gamma}_m^\nu \dot{\gamma}_m^c \Gamma_{c \nu}^{\mu, g_m} (\gamma_m (t)) \Gamma_{\mu j}^{k, g_m} (\gamma_m (t)) + \dot{\gamma}_m^\mu \dot{\gamma}_m^\nu \Gamma_{\mu j}^{\beta, g_m} (\gamma_m (t)) \Gamma_{\nu \beta}^{k, g_m} (\gamma_m (t)) + \dot{\gamma}_m^\mu \dot{\gamma}_m^\nu R_{j \nu \mu }^{k, g_m} (\gamma_m (t)) $, which are smooth along $\gamma_m (t)$.

The coefficients $a$ and $b$ are built from the Christoffel symbols along $\gamma_m$, their first derivatives and the curvature. Propositions~\ref{prop:prop1} and~\ref{def:prop2} compare these quantities for $g_1$ and $g_2$, and theorem~\ref{def:thrm3} compares the curves $\gamma_1$ and $\gamma_2$ along which they are evaluated. The following theorem uses this to compare the solutions.

\begin{theorem}\label{thrm:neat1}
Fix a coordinate neighborhood $U$, then choose $p \in U$ and $v,w \in T_pM$ such that $|v|_{g_m (p)}<L$ and $|w|_{g_m (p)}<L$, for $m=1,2$ and $\gamma_1 ([0,1]) \subset U$ for $\gamma_1 (t) = exp_p^{g_1}(tv)$ and suppose that $\gamma_2 ([0,1]) \subset U$ (where $\gamma_2 (t) = exp_p^{g_2}(tv)$). Let $J_m (t)$ be a Jacobi field along $\gamma_m$ with $J_1 (0)= J_2 (0)=0$ and $J_1^\prime (0) = J_2^\prime (0)=w$. Now let
\begin{equation*}
    f_m (t) = (f_{m1}(t), f_{m2}(t), \dots , f_{mn}(t))
\end{equation*}
Then for all $\epsilon >0$ there exists $\delta (M, g_1, U, L, \epsilon) >0$ such that $\parallel g_1 - g_2 \parallel_{C^2 (M, g_1)} < \delta (M, g_1, U, L, \epsilon)$ implies that for all $t \in [0,1]$, we have
\begin{equation*}
    |f_1 (t) - f_2 (t)|< \epsilon
\end{equation*}
where $| \cdot |$ is the Euclidean norm from definition~\ref{def:eucl}.

\end{theorem} 

The proof writes the Jacobi equation as a first-order system for $(f_m, f_m')$, as in lemma~\ref{def:example3}, and bounds the difference of its solutions by Gr\"onwall’s inequality. It then estimates the difference of the coefficients term by term. The details are in Appendix~\ref{app:neat1}.

From theorem~\ref{thrm:neat1}, we can observe that for the Jacobi field $J_{mwu}(t)$, along the geodesic $\gamma_{mw} (t) = exp_p^{g_m}(tw)$, with initial position $J_{mwu} (0)=0$ and initial velocity $J_{mwu}^\prime (0)=u$, we'll have the unique representation
\begin{equation*}
    f_{mwu}(t) = (f_{mwu1}(t), \dots, f_{mwun}(t))
\end{equation*}
where $d(exp_p^{g_m})_w (u) = \sum_{i=1}^n f_{mwui}(1) X_{exp_p^{g_m}(w)}^i$. Then we can define the norm for $d(exp_p^{g_m})_w (u)$,
\begin{equation*}
    \parallel d(exp_p^{g_m})_w (u) \parallel_E^2 = |f_{mwu}(1)|^2 = \sum_{i=1}^n ( f_{mwui}(1))^2
\end{equation*}
Since the differential $d(exp_p^{g_m})_w$ is a linear map on $T_pM$, we measure it with an operator norm. Theorem~\ref{thrm:neat1} controls $d(exp_p^{g_m})_w(u)$ one vector at a time, and the following norm and corollary make this control uniform in $u$.

\begin{definition}
For $p \in M$, and $w \in T_pM$, we will say
    \begin{equation*}
        \parallel d(exp_p^{g_m})_w \parallel = \sup_{u \in  T_pM, u \neq 0} \frac{\parallel d(exp_p^{g_m})_w (u) \parallel_E}{|u|_{g_1 (p)}}
    \end{equation*}
\end{definition}

Combining this norm with theorem~\ref{thrm:neat1} gives the following corollary.

\begin{corollary}\label{cor:neat1}
    Fix a coordinate neighborhood $U$, then choose $p \in U$ and $w \in T_pM$ such that $|w|_{g_m (p)} < L$ for $m=1,2$ and $\gamma_1 ([0,1]) \subset U$ for $\gamma_1 (t) = exp_p^{g_1} (tw)$ and suppose that $\gamma_2 ([0,1]) \subset U$ (for $\gamma_2 (t) = exp_p^{g_2} (tw)$). Then, for all $\epsilon >0$ there exists $\delta (M, g_1, U, L, \epsilon) >0$ such that $\parallel g_1 - g_2 \parallel_{C^2 (M, g_1)}< \delta (M, g_1, U, L, \epsilon)$ implies that
    \begin{equation*}
        \parallel d (exp_p^{g_1})_w - d (exp_p^{g_2})_w \parallel < \epsilon
    \end{equation*}
\end{corollary}

\begin{proof}
    First we observe that for all $u \in T_pM$ with $u \neq 0$, we have $| \frac{u}{|u|_{g_1 (p)}} |_{g_1 (p)} =1$. Then by the linearity of the maps $d(exp_p^{g_m})_w$, we have that
    \begin{equation*}
        d (exp_p^{g_m})_w (\frac{u}{|u|_{g_1 (p)}} ) = \frac{1}{|u|_{g_1 (p)}}  d (exp_p^{g_m})_w (u)
    \end{equation*}
    Then
    \begin{equation*}
        \parallel d (exp_p^{g_1})_w (\frac{u}{|u|_{g_1 (p)}} ) - d (exp_p^{g_2})_w (\frac{u}{|u|_{g_1 (p)}} ) \parallel_E = \frac{1}{|u|_{g_1 (p)}} \parallel d (exp_p^{g_1})_w ( u ) - d (exp_p^{g_2})_w ( u ) \parallel_E
    \end{equation*}
    From theorem~\ref{thrm:neat1} there exists $\delta (M, g_1, U, \max \{ L, \frac{3}{2} \}, \epsilon) >0$ such that $\parallel g_1 - g_2 \parallel_{C^2 (M, g_1)}< \delta (M, g_1, U, \max \{ L, \frac{3}{2} \}, \epsilon)$ implies that we have that
    \begin{equation*}
        \parallel d (exp_p^{g_1})_w (\frac{u}{|u|_{g_1 (p)}} ) - d (exp_p^{g_2})_w (\frac{u}{|u|_{g_1 (p)}} ) \parallel_E = |f_{1w \frac{u}{|u|_{g_1 (p)}}} (1) - f_{2w \frac{u}{|u|_{g_1 (p)}}} (1)| < \epsilon
    \end{equation*}
    By combining these two facts, we have that for all $u \in T_pM$ with $u \neq 0$,
    \begin{equation*}
        \frac{\parallel d (exp_p^{g_1})_w ( u ) - d (exp_p^{g_2})_w ( u ) \parallel_E }{|u|_{g_1(p)}} < \epsilon
    \end{equation*}
    Since $\max\{L, \frac{3}{2} \}$ is still based on $L$, then let $\delta (M, g_1, U, L, \epsilon) = \delta (M, g_1, U, \max \{L, \frac{3}{2} \}, \epsilon)>0$. Then $\delta (M, g_1, U, L, \epsilon) >0$ is such that $\parallel g_1 - g_2 \parallel_{C^2 (M, g_1)}< \delta (M, g_1, U, L, \epsilon)$ implies that
    \begin{align*}
        \parallel d (exp_p^{g_1})_w - d (exp_p^{g_2})_w \parallel & = \sup_{u \in T_pM,u \neq 0} \frac{\parallel d (exp_p^{g_1})_w (u) - d (exp_p^{g_2})_w (u) \parallel_E}{ |u|_{g_1 (p)}}\\
        & < \sup_{u \in T_pM,u \neq 0} \epsilon = \epsilon
    \end{align*}
\end{proof}

\subsection{Conjugate points}\label{sec:conj}

Corollary~\ref{cor:neat1} shows that the differentials of the exponential maps of $g_1$ and $g_2$ are close. To invert the exponential map of $g_2$ near that of $g_1$, as described at the start of this section, closeness is not enough; we also need the differential of $exp_p^{g_2}$ to be invertible. This fails exactly at conjugate points: a point $\gamma(t_0)$ is conjugate to $p$ along a geodesic $\gamma$ if there is a nonzero Jacobi field along $\gamma$ that vanishes at $0$ and at $t_0$, and this happens precisely when the differential of the exponential map is singular at $t_0\gamma'(0)$. We therefore ask how far along a $g_2$-geodesic we can travel before meeting a conjugate point. An upper bound on the sectional curvature excludes conjugate points up to a length controlled by the bound, and under $C^2$ control the sectional curvature of $g_2$ on $U$ is bounded by a constant depending only on $g_1$. We first recall the Riemann curvature tensor and the sectional curvature.

The Riemann curvature tensor with respect to $g_m$ is defined by

\begin{equation*}
    Rm^{g_m} = R_{ijks}^{g_m} dx^i \otimes dx^j \otimes dx^k \otimes dx^s
\end{equation*}
where
\begin{equation*}
    R_{ijks}^{g_m} = g_m (R^{g_m}(X_i, X_j)X_k, X_s) = \sum_l R_{ijk}^{l, g_m} g_{mls}
\end{equation*}

The sectional curvature is then defined by
\begin{equation*}
    K^{g_m}(Y_1, Y_2)= \frac{Rm^{g_m}(Y_1,Y_2,Y_2,Y_1)}{|Y_1 \wedge Y_2|_{g_m}^2} =  \frac{Rm^{g_m}(Y_1,Y_2,Y_2,Y_1)}{|Y_1|_{g_m}^2|Y_2|_{g_m}^2 - g_m (Y_1,Y_2)^2}
\end{equation*}
Since $M$ is compact, the Riemann curvature tensor for $g_1$ is bounded, i.e. for all $i,j, k, s = 1, \dots , n$
\begin{equation*}
    |R_{ijks}^{g_1}| \leq K_{g_1}
\end{equation*}

The following theorem shows how far along a $g_2$-geodesic we can travel before meeting a conjugate point.

\begin{theorem}\label{thrm:neat3}
    For a fixed coordinate neighborhood $U$, choose $p \in U$ and $v \in T_pM$ with $|v|_{g_1 (p)} < L$, and let $\gamma_2: [0,1] \to U$ be a geodesic with respect to $g_2$ that stays in $U$, such that $\gamma_2^\prime (0)= v$. There exists a $\delta (M, g_1, U, L)>0$ such that  $\parallel g_1 - g_2 \parallel_{C^2 (M, g_1)} < \delta (M, g_1, U)$ implies that there exists a constant based on $g_1$, $C(g_1)>0$, such that for $t \leq \frac{\pi}{2L \sqrt{C(g_1) +1}}$, $\gamma_2 (t)$ has no conjugate points.
\end{theorem}

\begin{proof}
The proof has three steps. We first show that the curvature components of $g_2$ are close to those of $g_1$, and hence bounded, on $U$. We then use this to bound the sectional curvature of $g_2$ on $U$ by a constant $C(g_1)$. Finally, we rescale the geodesic to unit speed and apply a comparison theorem for sectional curvature bounded above.

First, we want to show that on $U$, for all $\epsilon >0$, there exists a $\delta (M,g_1,U) > 0$ such that $\parallel g_1 - g_2 \parallel_{C^2 (M,g_1)} < \delta (M,g_1,U)$ implies that for all $1 \leq i,j,k,s \leq n$, we have
\begin{equation*}
    | R_{ijks}^{g_1} - R_{ijks}^{g_2}| < \epsilon
\end{equation*}

So, choose $\epsilon >0$. From propositions~\ref{def:prop2} and~\ref{def:prop1}, we know that there exists $\delta (M, g_1, U)>0$ such that $\parallel g_1 - g_2 \parallel_{C^2 (M,g_1)} < \delta (M, g_1, U)$ implies that 
\begin{equation*}
    \begin{split}
        |R_{ijk}^{l, g_1} - R_{ijk}^{l, g_2}| & < \zeta_1\\
        |g_{1ij} - g_{2ij}| & < \zeta_2
    \end{split}
\end{equation*}
on $U$. Then we observe that on $U$

\begin{align*}
    |  R_{ijks}^{g_1} - R_{ijks}^{g_2}| & = |  R_{ijk}^{l, g_1}g_{1ls} - R_{ijk}^{l, g_2}g_{2ls} |\\
    & \leq |R_{ijk}^{l, g_1}g_{1ls} - R_{ijk}^{l, g_1}g_{2ls}| + |R_{ijk}^{l, g_1}g_{2ls} - R_{ijk}^{l, g_2}g_{2ls}|\\
    & =  |R_{ijk}^{l, g_1}| |g_{1ls}-g_{2ls}| + |g_{2ls}| |R_{ijk}^{l,g_1} - R_{ijk}^{l, g_2}|\\
    & \leq n (\max|R^{g_1}| \zeta_2 + (|g_{1ls}| + \zeta_2)\zeta_1)\\
    & \leq n \max|R^{g_1}| \zeta_2 + n (\max|g_1|+ \zeta_2)\zeta_1
\end{align*}
where $\max|R^{g_1}|$ was defined in remark~\ref{def:rmk2} and $\max|g_1| = \max_{ij} \big( \sup_{p \in \overline{U}} |g_{1ij}(p)| \big)$. Then for $\zeta_2 = \min \{ 1, \frac{\epsilon}{2n \max |R^{g_1}|} \}$ and $\zeta_1 = \frac{\epsilon}{2n (\max |g_1|+1)}$, we see that $\parallel g_1 - g_2 \parallel_{C^2 (M, g_1)} < \delta(M, g_1, U)$ implies that
\begin{equation*}
    |R_{ijks}^{g_1} - R_{ijks}^{g_2}| < \epsilon
\end{equation*}
on $U$.

Now we'll let $\delta(M, g_1, U, L) >0$ be such that $\parallel g_1 - g_2 \parallel_{C^2 (M, g_1)} < \delta(M, g_1, U, L)$ implies that
\begin{equation*}
    |R_{ijks}^{g_1} - R_{ijks}^{g_2}| < 1
\end{equation*}
Then we have
\begin{equation*}
    |R_{ijks}^{g_2}| < |R_{ijks}^{g_1}|+1 < K_{g_1}+1
\end{equation*}

Then any arbitrary 2-plane $\sigma$ is spanned by 2 $g_2$-orthonormal vectors $Y_m$ for $m=1,2$,
\begin{equation*}
    Y_m = \sum_{i=1}^n y_m^i X_i
\end{equation*}
where $|Y_m|_{g_2} =1$ and $g_2 (Y_1, Y_2)=0$. Then we see that
\begin{equation*}
    |y_m^i| \leq |Y_m| \leq \frac{2}{A_{g_1}} |Y_m|_{g_2} = \frac{2}{A_{g_1}}
\end{equation*}
Then on $U$, we see that
\begin{align*}
    |K_{g_2} (\sigma)| & = |K^{g_2}(Y_1, Y_2)|\\
    & = \bigg| \frac{Rm^{g_2}(Y_1,Y_2,Y_2,Y_1)}{|Y_1 \wedge Y_2|_{g_2}^2} \bigg|\\
    & = \bigg| \frac{Rm^{g_2}(Y_1,Y_2,Y_2,Y_1)}{|Y_1|_{g_2}^2|Y_2|_{g_2}^2 - g_2 (Y_1,Y_2)^2} \bigg|\\
    & = |Rm^{g_2} (Y_1,Y_2, Y_2, Y_1)|\\
    & = \bigg| \sum_{ijks} y_1^i y_2^j y_2^k y_1^s Rm^{g_2}(X_i, X_j, X_k, X_s) \bigg|\\
    & = \bigg| \sum_{ijks} y_1^i y_2^j y_2^k y_1^s R_{ijks}^{g_2}  \bigg|\\
    & \leq \sum_{ijks} |y_1^i| |y_2^j| |y_2^k| |y_1^s| |R_{ijks}^{g_2}|\\
    & < n^4 \big( \frac{2}{A_{g_1}} \big)^4 (K_{g_1}+1)\\
    & = \frac{16n^4}{A_{g_1}^4} (K_{g_1}+1) = C(g_1)
\end{align*}

Now from our assumption, we have that for $\gamma_2 : [0,1] \to U$ is a geodesic with respect to $g_2$ that stays in $U$ and $|\gamma_2^\prime (0)|_{g_1} = |v |_{g_1} <L$, then for $\delta (M, g_1, U) \leq 1$, we have
\begin{equation*}
    |\gamma_2^\prime (0)|_{g_2} = |v |_{g_2} <  2|v |_{g_1} <2 L
\end{equation*}
Now for $v \neq 0$, let $c = |v|_{g_2}$ and define a new geodesic
\begin{equation*}
    \Tilde{\gamma}_2 (s) = \gamma_2 (\frac{s}{c})
\end{equation*}
and we see that
\begin{equation*}
    \Tilde{\gamma}_2^\prime (0) = \frac{1}{c} \gamma_2^\prime (0) = \frac{v}{|v|_{g_2}}
\end{equation*}
which is a unit vector. Here, we're saying that $s=ct$. So, $U$ is an open subset of $M$. Which means that it's an $n$-dimensional manifold, and we can use proposition 2.4 (p.218, ~\cite{do1992riemannian}). Which says that $\Tilde{\gamma}_2 (s)$ has no conjugate points for $s < \frac{\pi}{\sqrt{C(g_1)}}$, then this becomes $tc < \frac{\pi}{\sqrt{C(g_1)}}$ and

\begin{equation}
    t < \frac{\pi}{c \sqrt{C(g_1)}} = \frac{\pi}{ |v|_{g_2} \sqrt{C(g_1)}}
\end{equation}
But, since $|v|_{g_2} < 2L$, we have $\frac{\pi}{ 2L \sqrt{C(g_1)}} < \frac{\pi}{ |v|_{g_2} \sqrt{C(g_1)}}$. Hence, $\parallel g_1 - g_2 \parallel_{C^2 (M, g_1)} < \delta (M, g_1, U)$ implies that $\gamma_2 (t)$ has no conjugate points when
\begin{equation*}
    t \leq \frac{\pi}{ 2L \sqrt{C(g_1)+1}} < \frac{\pi}{ 2L \sqrt{C(g_1)}}
\end{equation*} 
Then $C(g_1)$ is our desired constant.
\end{proof}

Theorem~\ref{thrm:neat3} handles the start of each geodesic, and Corollary~\ref{cor:neat1} the rest. Together they control the differentials of the exponential map of $g_2$ well enough to invert it. We combine these results in the next subsection. 

\subsection{Connecting geodesics away from the cut locus}\label{sec:cut}

We can now answer the question left open at the end of section~\ref{sec:metricaprox}. Given points $p$ and $q$ with $q$ away from the cut locus of $g_1$, we produce a $g_2$-geodesic joining them whose initial velocity and path are close to those of the minimizing $g_1$-geodesic. The argument has three parts. First, since $q$ lies at distance at least $l$ from the cut locus, the vector $v = exp_p^{g_1,-1}(q)$ lies in a compact star-shaped subset of the injectivity domain of $g_1$. Second, the control of the differential of the exponential map and of conjugate points from sections~\ref{sec:jac} and~\ref{sec:conj} shows that $exp_p^{g_2}$ is a diffeomorphism near $v$ onto a neighborhood containing $q$, which yields a vector $u$ near $v$ with $exp_p^{g_2}(u)=q$. Third, because $u$ and $v$ are close, theorem~\ref{def:thrm3} shows that the two geodesics are close.

We begin with the terminology needed to state the result. Let $Cut_{g_m}(p)$ denote the cut locus of $p$ with respect to $g_m$, 
and let $ID^{g_m}(p)$ denote the injectivity domain of $p$ with respect to $g_m$, which is the open star-shaped subset of $T_pM$ that $exp_p^{g_m}$ maps diffeomorphically onto $M \setminus Cut_{g_m}(p)$ (\cite[p. 310]{lee2018introduction}).

\begin{theorem}\label{thrm2:neat}

    For a fixed $p \in M$ and $l >0$ with $0 < l < d_{g_1}(p, Cut_{g_1}(p))$, let $q \in M \setminus Cut_{g_1}(p)$ such that $d_{g_1}(q, Cut_{g_1}(p) ) \geq l$, and let $\gamma_1 : [0,1] \to M$ be the unique minimizing geodesic connecting $p$ and $q$. Then for all $\epsilon_1, \epsilon_2 >0$ there exists $\delta (M, g_1, l,\epsilon_1, \epsilon_2 )>0$ such that $\parallel g_1 - g_2 \parallel_{C^2 (M, g_1)} < \delta (M, g_1, l,\epsilon_1, \epsilon_2 )$ implies that there exists a geodesic with respect to $g_2$ connecting $p$ and $q$, $\gamma_2$, such that
    \begin{equation*}
    \begin{split}
            | \gamma_1^\prime (0) - \gamma_2^\prime (0)|_{g_1 (p)} & < \epsilon_1\\
           d_{g_1}(\gamma_1 (t), \gamma_2 (t)) & < \epsilon_2
    \end{split}
    \end{equation*}

\end{theorem}

The proof uses two further ingredients: a notion of $g_m$-star-shaped set, which describes the region around the $g_1$-geodesic in $T_pM$, and a lemma which will be needed to prove theorem~\ref{thrm2:neat}.

\begin{definition}\label{def:star}
    Let $p \in M$, then let $U$ be a set that contains $p$. We say that $U$ is $g_m$-geodesically star-shaped with respect to $p$, if for all $q \in U$, there exists at least one $v \in T_pM$ such that $exp_p^{g_m}(v)=q$ and the geodesic $\gamma_{mv}(t) =exp_p^{g_m}(tv)$ for $t \in [0,1]$ is contained in $U$, i.e. $\gamma_{mv}([0,1]) \subset U$.
\end{definition}

In all applications of definition~\ref{def:star}, the sets we consider all lie away from the cut locus, where $exp_p^{g_m}$ is a diffeomorphism on the injectivity domain. On such sets the condition above can be checked on the level of $T_pM$, where it is a statement about the star-shapedness of the preimage with respect to $0$. The following remark makes this precise, and it is the form in which we will use the definition.

\begin{Remark}\label{rmk:star}
    Choose $p \in M$ and set $U$ such that $p \in U \subset M \setminus Cut_{g_m}(p)$. Then since $U$ is away from the cut locus, then $U \subset exp_p^{g_m}( ID^{g_m}(p) )$ where $exp_p^{g_m}: ID^{g_m}(p) \to M \setminus Cut_{g_m}(p)$ is a diffeomorphism. Then for all $q \in U$ there's a unique $v_0 \in ID^{g_m}(p)$ such that $exp_p^{g_m}(v_0)=q$. Let $V := exp_p^{g_m,-1}(U)$, and we see that $V \subset ID^{g_m}(p)$.
    
    If $U$ is $g_m$-geodesically star-shaped with respect to $p$, then for all $q \in U$, there exists $v_1 \in T_pM$ such that $exp_p^{g_m}(v_1)=q$ and the geodesic $\gamma_{mv_1}(t)= exp_p^{g_m}(tv_1)$ is such that $\gamma_{mv_1}([0,1]) \subset U$. Let $v_0 \in ID^{g_m}(p)$ be the unique vector from above such that $exp_p^{g_m}(v_0)=q$. Now let $w (t):= exp_p^{g_m,-1}(\gamma_{mv_1}(t))$. We see that since $\gamma_{mv_1}([0,1]) \subset U$ then $w ([0,1]) \subset V \subset ID^{g_m}(p)$. We want to show that $w(t)=tv_1$. We see that $w(0)=0$. Let $f(t) = w(t)-tv_1$, then let $S = f^{-1}(0)$, since $0 \in S$, then $S$ is nonempty. Then by how we defined $w(t)$, it's continuous, then $S$ is the pre-image of point through the continuous function, $f$, which means that $S$ is closed. Now suppose that there exists $t_0 \in S$ then $f(t_0)=0$, this means that $t_0 v_1 = w (t_0) \in ID^{g_m}(p)$, then by the openness of $ID^{g_m}(p)$, and by the continuity of $t \to tv_1$, then $tv_1 \in ID^{g_m}(p)$. Then by the injectivity of $exp_p^{g_m}$ in $ID^{g_m}(p)$, for $t$ near $t_0$, we have $w(t)=tv_1$. Then all points in $S$ are contained in an open neighborhood that's also contained in $S$, which means that $S$ is open. Then $S$ is closed and open on a connected set $[0,1]$. This means that $S=[0,1]$ and on $ID^{g_m}(p)$, $w(t)=tv_1$, and since $\gamma_{mv_1}([0,1]) \subset U$, then $v_1 \cdot [0,1] = w([0,1]) \subset V$. Then for $t=1$, we have $v_1 = w(1) = exp_p^{g_m,-1}(q)=v_0$. Then since $q$ was an arbitrary point in $U$, by how we defined $V$, then we can see that for all $v_0 \in V$, we'll have $v_0 \cdot [0,1] \subset V$, and $V$ is star-shaped with respect to $0$.
    
    For the other direction, if $V$ is star-shaped with respect to $0$, then for all $q \in U$, there exists $v_0 \in V$ such that $exp_p^{g_m}(v_0)=q$, then $tv_0 \in V$ for $t \in [0,1]$. Then by the definition of $exp_p^{g_m}$ and by how we've defined $V$, the geodesic $\gamma_{mv_0}(t)= exp_p^{g_m}(tv_0)$ is contained in $U$. Thus, $U$ is $g_m$-geodesically star-shaped with respect to $p$.

    Then $U$ being $g_m$-geodesically star-shaped with respect to $p$ is equivalent to $V$ being star-shaped with respect to $0$.
\end{Remark}

The following lemma is where the results of sections~\ref{sec:jac} and~\ref{sec:conj} are used. Its first part gives a uniform radius on which the exponential maps are diffeomorphisms onto their images, and its second part shows that the image of $exp_p^{g_2}$ on such a ball contains the point $exp_p^{g_1}(v)$.

\begin{lemma}\label{lemma:lemma3}
We have the following facts:
\begin{enumerate}

    \item For $p \in M$, suppose that we have fixed open neighborhoods $U_1$ and $U_2$, such that $p \in U_1 \subset \overline{U}_1 \subset U_2 \subset \overline{U}_2 \subset M \setminus Cut_{g_1}(p)$, $\overline{U}_1$ is $g_1$-geodesically star-shaped with respect to $p$ (definition~\ref{def:star}), and $\overline{U}_2$ is contained in a coordinate chart $(V, \phi)$ (as in remark~\ref{def:rmk1}). Then there exists $\delta (M,g_1,U_1,U_2)>0$ such that $\parallel g_1 - g_2 \parallel_{C^2(M,g_1)} < \delta (M,g_1,U_1,U_2)$ implies that there exists $R(U_1,U_2)>0$ such that for all $q \in \overline{U}_1$, with $v = exp_p^{g_1, -1}(q)$, $exp_p^{g_m} \big|_{B_{R(U_1,U_2)}^{g_1}(v)}$ is diffeomorphic to its image.

    \item Let $U$ be a fixed coordinate neighborhood with $p \in U$, and let $W \subset T_pM$ be compact. If there exists a fixed $R>0$ such that for all $v \in W$, $exp_p^{g_m} \big|_{B_R^{g_1}(v)}$ is diffeomorphic to its image, and $exp_p^{g_m} (B_R^{g_1}(v)) \subset U$, then there exists $\delta >0$ such that $\parallel g_1 - g_2 \parallel_{C^2(M,g_1)} < \delta$ implies that for all $v \in W$, $exp_p^{g_1}(v) \in exp_p^{g_2}(B_R^{g_1}(v))$.   
\end{enumerate}
\end{lemma}

The proof of lemma~\ref{lemma:lemma3} first shows, by a continuity argument, that the $g_2$-geodesics from $p$ stay in $U_2$. The Jacobi-field estimates then show that $\exp_p^{g_2}$ remains a local diffeomorphism on a slightly enlarged star-shaped set, and the inverse function theorem finishes. The details are in Appendix~\ref{app:lemma3}.

\begin{proof}[Proof of Theorem \ref{thrm2:neat}]

Let $\epsilon_1, \epsilon_2>0$.

We work in $T_pM$. For $u \in S^{g_1}(0)$, let $t_{cut}^{g_1} (u) = \sup \{ b > 0 \ : \ the \ restriction \ of \ \gamma_{1u}(t) = exp_p^{g_1}(tu) \ to \ [0,b] \ is \ minimizing \}$ (from~\cite{lee2018introduction}, p. 307). Then define
\begin{equation*}
    V_1 = \{ tu \ : \ u \in S^{g_1}(0) \ and \ 0 \leq t < t_{cut}^{g_1} (u) - l \} 
\end{equation*}
which is star-shaped with respect to $0$ and so is $\overline{V}_1$. Then, from remark~\ref{rmk:star}, for $U_1 = exp_p^{g_1} (V_1)$ and $\overline{U}_1 = exp_p^{g_1}(\overline{V}_1)$, both are $g_1$-geodesically star-shaped with respect to $p$.

Now suppose that for $q \in M \setminus Cut_{g_1}(p)$, $d_{g_1}(q, Cut_{g_1}(p)) \geq l$, let $v = exp_p^{g_1,-1}(q)$, then $u_v = \frac{v}{|v|_{g_1 (p)}}$, and we have
\begin{equation*}
    t_q = |v|_{g_1 (p)} = d_{g_1}(p,q)
\end{equation*}
Let $c_v = exp_p^{g_1} (t_{cut}^{g_1}(u_v)u_v)$ be the cut point along the same ray. Then we have $d_{g_1}(q, c_v)= t_{cut}^{g_1} (u_v) - t_q$. Since $c_v \in Cut_{g_1}(p)$ and $d_{g_1}(q, Cut_{g_1}(p))$ is an infimum over all of $Cut_{g_1}(p)$, we have
\begin{equation}\label{eq:eq0}
    l \leq d_{g_1}(q, Cut_{g_1}(p)) \leq d_{g_1}(q,c_v)= t_{cut}^{g_1} (u_v) - t_q
\end{equation}
Then it follows that
\begin{equation*}
    t_q \leq t_{cut}^{g_1} (u_v) -l
\end{equation*}
So, $v \in \overline{V}_1$, and $q \in \overline{U}_1$. Let $U_2 = exp_p^{g_1} (\{ tu \ : \ u \in S^{g_1}(0) \ and \ 0 \leq t < t_{cut}^{g_1} (u) - \frac{l}{2} \})$ and we have $p \in U_1 \subset \overline{U}_1 \subset U_2 \subset \overline{U}_2 \subset M \setminus Cut_{g_1}(p)$. Since $\overline{U}_2 \subset M \setminus Cut_{g_1}(p)$, it lies in the chart given by the normal coordinates $exp_p^{g_1,-1} : M \setminus Cut_{g_1}(p) \to ID^{g_1}(p)$.

By lemma~\ref{lemma:lemma3}, using part (1), there exists $\delta_1 (M,g_1, U_1,U_2)>0$ such that $\parallel g_1 - g_2 \parallel_{C^2 (M,g_1)} < \delta_1 (M,g_1, U_1,U_2)$ implies that for all $y \in \overline{U}_1$, where $exp_p^{g_1,-1}(y)=v$, there exists $R>0$ such that $exp_p^{g_m} \bigg|_{B_R^{g_1}(v)}$ is diffeomorphic to its image. Then let $\epsilon_3 = \min \{ \sigma_1, \epsilon_1, R \}$ (we'll define $\sigma_1$ later in the proof), and we know that for all $y \in \overline{U}_1$, $exp_p^{g_m} \bigg|_{B_{\epsilon_3}^{g_1}(v)}$ is diffeomorphic to its image. Then from lemma~\ref{lemma:lemma3}, using part (2), there exists $\delta_2 (M,g_1, U_1,U_2, \epsilon_1)>0$ such that $\parallel g_1 - g_2 \parallel_{C^2 (M,g_1)} < \delta_2 (M,g_1, U_1,U_2, \epsilon_1)$ implies that there exists $u \in B_{\epsilon_3}^{g_1}(v)$, such that $exp_p^{g_2}(u)=y$ and the geodesic $\eta_{2u}(t) = exp_p^{g_2}(tu) \in U_2$ for all $t \in [0,1]$. This means $exp_p^{g_2}(u) = y = exp_p^{g_1}(v)$.

Then $ \parallel g_1 - g_2 \parallel_{C^2 (M,g_1)} < \min \{ \delta_1, \delta_2 \}$ implies that for $\gamma_1 (t) = exp_p^{g_1}(tv)$ and $\gamma_2 (t) = exp_p^{g_2}(tu)$, we have $\gamma_m ([0,1]) \subset U_2$ for $m=1,2$, and
\begin{equation*}
    |\gamma_1^\prime (0) - \gamma_2^\prime (0)|_{g_1 (p)} < \epsilon_3 \leq \epsilon_1
\end{equation*}
Let $L_0 > 0$ be a bound on the $g_1$- and $g_2$-lengths of $\gamma_1$ and $\gamma_2$. It exists because $\overline{V}_1$ is compact, and by proposition~\ref{prop:neat}, $|u|_{g_2(p)} < 2|u|_{g_1(p)}$ once $\|g_1 - g_2\|_{C^2(M,g_1)}$ is small.

Then from theorem~\ref{def:thrm3}, there exists $\delta_3 (M,g_1,U_2,L_0, \epsilon_2)>0$ and $\sigma_1 (M,g_1,U_2,L_0, \epsilon_2)>0$ such that $\parallel g_1 - g_2 \parallel_{C^2 (M,g_1)} < \delta_3 (M,g_1, U_2,L_0, \epsilon_2)$ implies that if $|\gamma_1^\prime (0) - \gamma_2^\prime (0)|_{g_1 (p)} < \sigma_1$, then for all $t \in [0,1]$, we have
\begin{equation*}
    d_{g_1}(\gamma_1 (t), \gamma_2 (t)) < \epsilon_2
\end{equation*}

Then $\delta (M,g_1, l,\epsilon_1, \epsilon_2 ) = \min \{ \delta_1 (M,g_1,U_1,U_2), \delta_2 (M,g_1,U_1,U_2, \epsilon_1), \delta_3 (M,g_1, U_2,L_0, \epsilon_2) \} >0$ and $\parallel g_1 - g_2 \parallel_{C^2 (M,g_1)} < \delta (M,g_1, l,\epsilon_1, \epsilon_2)$ implies that there exists $u \in T_pM$ such that $exp_p^{g_2}(u)=q$ and the geodesic $\gamma_2 (t) = exp_p^{g_2}(tu)$ is our desired geodesic.

\end{proof}

Theorem~\ref{thrm2:neat} supplies what section~\ref{sec:metricaprox} lacked: a $g_2$-geodesic from $p$ to $q$ whose initial velocity is close to that of $\gamma_1$. Together with theorem~\ref{def:thrm5}, it controls parallel transport along geodesics with the same endpoints, which is the quantity that enters the convolution. We use this in the next section to prove the stability of the convolution.

\section{Stability of the parallel transport convolution under $C^2$-metric control}\label{sec:app}

The preceding sections control geodesics, parallel transport and the differential of the exponential map under small perturbations of the metric. We now use these results to prove the main theorem of the paper, theorem~\ref{thrm:fin}: if $g_2$ is sufficiently close to $g_1$ in $C^2(M,g_1)$, then the parallel transport convolutions $h_1$ and $h_2$ of a continuous function $f$ with a filter $K$ are close in $L^2(M,g_1)$. The proof is organized into four steps. In section~\ref{sec:setup} we recall how a filter supported near a fixed point $x_0$ is transported over $M$, which defines the kernels $K_m$ and the convolutions $h_m$. In section~\ref{sec:invexp} we show that the inverse exponential maps entering $K_2$ are defined on the same balls as those entering $K_1$, and that they are uniformly close. In section~\ref{sec:kernels} we combine this with the results of sections~\ref{sec:metricaprox} and~\ref{sec:C2} to compare $K_1$ and $K_2$ uniformly at points a fixed distance from the cut locus of $x_0$. Finally, in section~\ref{sec:L2} we control the contribution of a small neighborhood of the cut locus and the change in the volume form, which gives the $L^2$ estimate.

\subsection{The transported kernel and the convolution}\label{sec:setup}

We begin with the data that are fixed throughout this section. Fix $x_0\in M$, let $\operatorname{inj}(M,g_1)$ denote the injectivity radius of $M$ with respect to $g_1$, and choose $0<r<\tfrac{\operatorname{inj}(M,g_1)}{2}$. We assume that $K(x_0,\cdot)$ is continuous and localized near $x_0$, that is, $K(x_0,y)=0$ for $y\notin B_{g_1}(x_0,r)$, or equivalently $\operatorname{supp}K(x_0,\cdot)\subset B_{g_1}(x_0,r)$.

Parallel transport convolution moves this localized filter from $x_0$ to an arbitrary point $x\in M$ while retaining its orientation. For $m=1,2$, the transported kernel is

\begin{equation*}
    \begin{split}
        K_m(x, \cdot): & B_{g_1} (x,r) \to \mathbb{R}\\
        & y \mapsto K (x_0, \exp_{x_0}^{g_m} \circ (P_{x_0,x}^{g_m})^{-1} \circ \exp_x^{g_m,-1}(y)).
    \end{split}
\end{equation*}
Here $P^{g_m}_{x_0,x}$ denotes parallel transport along the $g_m$-geodesic from $x_0$ to $x$, that is, along $\gamma_m(t)=\exp_{x_0}^{g_m}(tu)$, where $\gamma_m(1)=\exp_{x_0}^{g_m}(u)=x$. Reading the composition from the right, $\exp_x^{g_m,-1}$ pulls $y$ back to a vector in $T_xM$, $(P^{g_m}_{x_0,x})^{-1}$ carries this vector to $T_{x_0}M$, and $\exp_{x_0}^{g_m}$ places it on $M$, where the filter $K(x_0,\cdot)$ is evaluated. For a continuous function $f:M\to\mathbb{R}$, the parallel transport convolution of $f$ with the filter $K$, centered at $x_0$, is then
\begin{equation*}
    \begin{split}
        h_m (x) = \big(f *_{M_{g_m}} K \big) (x) : &= \int_M f(y)K_m(x,y) \ d \mu_m (y)\\
        & = \int_M f(y) K (x_0, \exp_{x_0}^{g_m} \circ (P_{x_0,x}^{g_m})^{-1} \circ \exp_x^{g_m,-1}(y)) \  d \mu_m (y),
    \end{split}
\end{equation*}
where $\mu_m$ denotes the Riemannian volume measure of $g_m$ ($d \mu_m = d M_{(g_m, y)}$).

The kernel $K_m$ and the output $h_m$ depend on the metric through the following ingredients: the inverse exponential map $\exp_x^{g_m,-1}$, the transport path and the parallel transport along it, the exponential map $\exp_{x_0}^{g_m}$, and the volume form of $g_m$. Comparing $h_1$ and $h_2$ therefore requires controlling each of them, and the remainder of the section does this in turn. The first difficulty is that $K_2(x,\cdot)$ is defined on $B_{g_1}(x,r)$ only if $\exp_x^{g_2,-1}$ is defined there, which is not clear a priori; this is the subject of the next subsection. The second is that the $g_2$-geodesic from $x_0$ to $x$ need not be unique, so for $g_2$ the transport path must be specified; we fix this convention in section~\ref{sec:kernels}.

\subsection{Inverse exponential maps under metric perturbation}\label{sec:invexp}

On $B_{g_1}(x,r)$ the map $\exp_x^{g_1,-1}$ is defined, because $0 < r < \frac{\operatorname{inj}(M,g_1)}{2}$, but for $g_2$ nothing of the kind is known a priori. We settle this in two steps, and then show that the resulting inverse exponential maps are uniformly close. The first step is a uniform lower bound for the injectivity radius of $g_2$.

\begin{theorem}\label{thrm:neat5}
    There exists $\delta_1 >0$ such that $\parallel g_1 - g_2 \parallel_{C^2 (M,g_1)} < \delta_1$ implies that $\frac{\operatorname{inj}(M,g_1)}{2} \leq \operatorname{inj}(M,g_2)$.
\end{theorem}

The proof is by contradiction. Compactness and the uniform convergence of the distance functions give limit points, and the two possibilities for the nearest cut point, a conjugate point or a geodesic loop, each lead to a contradiction. The details are in Appendix~\ref{app:neat5}.

The second step is to place the ball $B_{g_1}(x,r)$ inside a $g_2$-ball on which $\exp_x^{g_2}$ is invertible. By theorem~\ref{thrm:neat5}, $\exp_x^{g_2}$ is a diffeomorphism onto its image on $B_{g_2}(x,\frac{\operatorname{inj}(M,g_1)}{2})$, so it suffices to find $\delta_0(r)>0$ such that $\parallel g_1-g_2 \parallel_{C^2(M,g_1)}<\delta_0(r)$ implies $B_{g_1}(x,r)\subset B_{g_2}(x,\frac{\operatorname{inj}(M,g_1)}{2})$. By proposition~\ref{prop:neat}, for $y\in B_{g_1}(x,r)$ we have
\begin{align*}
    d_{g_2}(x,y) & \leq \sqrt{1 + \delta_0} \ d_{g_1}(x,y)\\
    & < \sqrt{1 + \delta_0} \ r
\end{align*}
This is at most $\frac{\operatorname{inj}(M,g_1)}{2}$ as soon as $\sqrt{1+\delta_0}\leq \frac{\operatorname{inj}(M,g_1)}{2r}$, which we guarantee by setting
\begin{equation}\label{delta:eq}
    \delta_0 (r):= \min \{  \big( \frac{\operatorname{inj}(M,g_1)}{2r} \big)^2 -1, \frac{1}{2} \}.
\end{equation}
With this choice of $\delta_0$, together with theorem~\ref{thrm:neat5}, the inverse exponential map $\exp_x^{g_m,-1}$ is defined on $B_{g_1}(x,r)$ for $m=1,2$. The following corollary shows, in addition, that for any prescribed tolerance these maps are uniformly close when $g_2$ is sufficiently close to $g_1$.

\begin{corollary}\label{cor:neat5}
    For all $\epsilon >0$, there exists $\delta(r, \epsilon) \leq \min \{ \delta_1,\delta_0 \}$ (where $\delta_1$ is from theorem~\ref{thrm:neat5} and $\delta_0$ is from equation~\ref{delta:eq} ), such that $\parallel g_1 - g_2\parallel_{C^2(M,g_1)} < \delta(r, \epsilon)$ implies that for all $x \in M$ and all $y \in B_{g_1} (x,r)$, we have
    \begin{equation*}
        |exp_x^{g_1,-1}(y) - exp_x^{g_2,-1}(y)|_{g_1 (x)} < \epsilon
    \end{equation*}
\end{corollary}

\begin{Remark}
    The hypotheses on $\delta_1$ and $\delta_0$ and the compactness of $M$ are used only to ensure that, for $m = 1,2$, $exp_x^{g_m,-1}$ is defined on $B_{g_1}(x,r)$, that $exp_x^{g_m}$ is injective on the ball of radius $r$, and that the estimate is uniform in $x$. The estimate itself is a local consequence of continuous dependence of the geodesics on the metric.
\end{Remark}

The proof of corollary~\ref{cor:neat5} is by contradiction. Compactness gives limit points, and the initial velocities of the geodesics stay bounded, so continuous dependence on initial data passes to a limit that contradicts the assumption. The details are in Appendix~\ref{app:cor-neat5}.

With the inverse exponential maps under control, the remaining ingredients of $K_m$ are the transport path and the exponential map based at $x_0$, which we treat in the next subsection.

\subsection{Comparison of the transported kernels}\label{sec:kernels}

We now compare $K_1$ and $K_2$. Fix $x\in M\setminus\operatorname{Cut}_{g_1}(x_0)$, and let $\gamma_1$ be the minimizing $g_1$-geodesic from $x_0$ to $x$. The $g_2$-geodesic joining $x_0$ to $x$ need not be unique, so we must say which one defines $P^{g_2}_{x_0,x}$. When $\|g_1-g_2\|_{C^2(M,g_1)}$ is sufficiently small, theorem~\ref{thrm2:neat} provides a $g_2$-geodesic $\gamma_2$ from $x_0$ to $x$ that stays uniformly close to $\gamma_1$ and whose initial velocity is close to that of $\gamma_1$. We take $P^{g_2}_{x_0,x}$ to be parallel transport along $\gamma_2$. The inverse $(P^{g_m}_{x_0,x})^{-1}=P^{g_m}_{x,x_0}$ is then parallel transport along the same geodesic traversed backwards, namely the $g_m$-geodesic $\eta_m(t)=\exp_x^{g_m}\bigl(-t\,P^{g_m}_{x_0,x}(\gamma_m'(0))\bigr)$ from $x$ to $x_0$. The next proposition shows that the full composition defining $K_m$ is stable under this convention, uniformly for $x$ at distance at least $l$ away from $\operatorname{Cut}_{g_1}(x_0)$ and for $y\in B_{g_1}(x,r)$. The neighborhood of the cut locus is treated in section~\ref{sec:L2}.

\begin{proposition}\label{def:prop5}
For a fixed $x_0 \in M$, let $0 < l < d_{g_1}(x_0, Cut_{g_1}(x_0))$, and let $0 < r < \frac{inj(M,g_1)}{2}$. For all $\epsilon >0$, there exists $\delta (M,g_1,r,l, \epsilon) > 0$ such that $\parallel g_1 - g_2 \parallel_{C^2 (M,g_1)} < \delta (M,g_1,r,l, \epsilon)$ implies that for all $x \in M \setminus Cut_{g_1}(x_0)$ with $d_{g_1}(x, Cut_{g_1}(x_0)) \geq l$, and for all $y \in B_{g_1}(x,r)$ we have
\begin{equation*}
   d_{g_1} (exp_{x_0}^{g_1} \circ (P_{x_0, x}^{g_1})^{-1} \circ exp_x^{g_1, -1} (y), exp_{x_0}^{g_2} \circ (P_{x_0, x}^{g_2})^{-1} \circ exp_x^{g_2, -1} (y)) < \epsilon 
\end{equation*}

\end{proposition}

\begin{proof}

The proof fixes the sets on which the relevant geodesics lie, together with a bound $L_1(l)$ on the initial velocities, so that theorems~\ref{def:thrm3},~\ref{thrm2:neat}, and~\ref{def:thrm5} apply on them. Theorem~\ref{thrm2:neat} then gives a $g_2$-geodesic whose initial velocity is close to $u_1$, theorem~\ref{def:thrm5} makes the corresponding parallel transports close, theorem~\ref{def:thrm3} converts closeness of initial velocities into closeness of endpoints, and corollary~\ref{cor:neat5} supplies closeness of the inverse exponential maps at $x$.

Let $l_0 := \frac{inj(M,g_1)}{2}$. We will be working in $T_{x_0}M$. For $u \in S^{g_1}(0)$, let $t_{cut}^{g_1} (u) = \sup \{ b>0 \ : \ the \ restriction \ of \ \gamma_{1u}(t)= exp_{x_0}^{g_1}(tu) \ to \ [0,b] \ is \ minimizing \}$ (\cite{lee2018introduction}, p. 307). For $l \leq l_0$, define:
\begin{align*}
    V_{1,l} & = \{ tu \ : \ u \in S^{g_1}(0) \ and \ 0 \leq t < t_{cut}^{g_1}(u) - l \}\\
    V_{2,l} & = \{ tu \ : \ u \in S^{g_1}(0) \ and \ 0 \leq t < t_{cut}^{g_1}(u) - \frac{l}{2} \}
\end{align*}
and let $U_{1,l}= exp_{x_0}^{g_1} (V_{1,l})$ and $U_{2,l}= exp_{x_0}^{g_1} (V_{2,l})$. Since $exp_{x_0}^{g_1}$ is a diffeomorphism on $ID^{g_1}(x_0)$, for $i=1,2$ we have $\overline{U}_{i,l}= exp_{x_0}^{g_1} (\overline{V}_{i,l})$. For $l > l_0$, let $U_{1,l} = U_{1, l_0}$ and $U_{2,l} = U_{2, l_0}$. 

As in the proof of theorem~\ref{thrm2:neat}, $V_{1,l}$ and $\overline{V}_{1,l}$ are star-shaped with respect to $0$, so from remark~\ref{rmk:star}, $U_{1,l}$ and $\overline{U}_{1,l}$ are $g_1$-geodesically star-shaped with respect to $x_0$ and $x_0 \in U_{1,l} \subset \overline{U}_{1,l} \subset U_{2,l} \subset \overline{U}_{2,l} \subset M \setminus Cut_{g_1}(x_0)$.

Suppose $l \leq l_0$. Since $t_{cut}^{g_1}(u) \geq inj(M,g_1)$ for all $u \in S^{g_1}(0)$, the following holds for every $y \in B_{g_1} (x_0, \frac{inj(M,g_1)}{2})$. Let $u_y = \frac{exp_{x_0}^{g_1,-1}(y)}{|exp_{x_0}^{g_1,-1}(y)|_{g_1 (x_0)}}$, so that $y = exp_{x_0}^{g_1}(t_y u_y)$ and $t_y = d_{g_1}(x_0, y) < \frac{inj(M,g_1)}{2}$. Then we have
\begin{equation*}
    t_{cut}^{g_1}(u_y)-t_y > inj(M,g_1) - \frac{inj(M,g_1)}{2} = \frac{inj(M,g_1)}{2} \geq l
\end{equation*}
So, $t_y < t_{cut}^{g_1}(u_y) - l$, i.e. $exp_{x_0}^{g_1,-1}(y) \in V_{1,l}$ and $y \in U_{1,l}$, which implies that $B_{g_1}(x_0, \frac{inj(M,g_1)}{2}) \subset U_{1,l}$. For $l >l_0$, by how we defined $U_{1,l}$, it's trivially true that $B_{g_1}(x_0, \frac{inj(M,g_1)}{2}) \subset U_{1,l}$. So, we have this containment for all $l>0$.

Then for $x \in M \setminus Cut_{g_1}(x_0)$, with $d_{g_1}(x, Cut_{g_1}(x_0)) \geq l$, the same inequality chain from~\ref{eq:eq0} in the proof of theorem~\ref{thrm2:neat}, gives us $x \in \overline{U}_{1,l}$.

Next, suppose that our $\delta$'s are all sufficiently small so that for all $x \in U_{2,l}$ and all $v \in T_xM$, we have
\begin{equation*}
    \frac{1}{2}|v|_{g_2 (x)} \leq |v|_{g_1 (x)} \leq 2 |v|_{g_2 (x)}
\end{equation*}
From proposition~\ref{prop:neat}, this is possible.

Since $\overline{U}_{2,l} \subset M \setminus Cut_{g_1}(x_0)$, there exists $L_0 (l)>0$ such that for all $v \in exp_{x_0}^{g_1,-1}(\overline{U}_{2,l})$, we have $|v|_{g_1 (x_0)} < L_0(l)$. Let $L_1 (l) = \max \{ 2L_0(l)+1, 3 \cdot (inj(M,g_1))+1 \}$ and we see that
\begin{equation*}
    |v|_{g_2 (x_0)} < 2 |v|_{g_1 (x_0)} < 2 L_0 (l) < L_1 (l)
\end{equation*}
and hence $|v|_{g_m (x_0)} < L_1 (l)$ for $m = 1,2$ and all $v \in exp_{x_0}^{g_1,-1}(\overline{U}_{2,l})$.

Then let $\epsilon >0$, and from theorem~\ref{def:thrm3}, there exists $\delta_1 (M,g_1, U_{2,l}, L_1 (l))>0$ such that $\parallel g_1 - g_2 \parallel_{C^2 (M,g_1)} < \delta_1 (M,g_1, U_{2,l}, L_1 (l))$ implies that there exists $\sigma_1 >0$ such that for all $w_1, w_2 \in B_r^{g_1}(0) \subset T_{x_0}M$, we have
\begin{equation*}
    |w_1 - w_2|_{g_1 (x_0)} < \sigma_1
\end{equation*}
this implies that
\begin{equation*}
    d_{g_1} (exp_{x_0}^{g_1}(w_1), exp_{x_0}^{g_2}(w_2)) < \epsilon
\end{equation*}
Let $x \in M \setminus Cut_{g_1}(x_0)$ with $d_{g_1}(x,Cut_{g_1}(x_0)) \geq l$. Then $x \in \overline{U}_{1,l}$ (by~\ref{eq:eq0}), and since $\overline{U}_{1,l}$ is $g_1$-geodesically star-shaped with respect to $x_0$, the minimizing $g_1$-geodesic $\gamma_{1,x_0,x}$ from $x_0$ to $x$ satisfies $\gamma_{1,x_0,x}([0,1]) \subset \overline{U}_{1,l} \subset U_{2,l}$. Let $u_1 = exp_{x_0}^{g_1,-1}(x)$ be the initial velocity of $\gamma_{1,x_0,x}$. By theorem~\ref{thrm2:neat}, there exists $\delta_2 (M,g_1, U_{2,l}, U_{1,l},L_1 (l)) > 0$ such that $\parallel g_1 - g_2 \parallel_{C^2 (M,g_1)} < \delta_2 (M,g_1, U_{2,l}, U_{1,l},L_1 (l))$ implies that there exists $u_2 \in T_{x_0}M$ such that $exp_{x_0}^{g_2}(u_2)=x$ and 

\begin{equation*}
    |u_1 - u_2|_{g_1(x_0)} < \zeta_1
\end{equation*}

Then from theorem~\ref{def:thrm5}, there exists $\delta_3 (M,g_1,U_{2,l},L_1 (l)) > 0$ such that $\parallel g_1 - g_2 \parallel_{C^2 (M,g_1)} < \delta_3 (M,g_1,U_{2,l},L_1 (l))$ implies that for $u_1,u_2 \in T_{x_0}M$ with $|u_1|_{g_m (x_0)}, |u_2|_{g_m (x_0)} < L_1 (l)$ ($L_1(l) >0$ bounds $\overline{V}_{2,l} = exp_{x_0}^{g_1,-1} (\overline{U}_{2,l})$), we have
\begin{equation*}
    |u_1 - u_2|_{g_1 (x_0)} < \sigma_2
\end{equation*}
which implies that
\begin{equation*}
    |P_{x_0,x}^{g_1}(u_1) - P_{x_0,x}^{g_2}(u_2)|_{g_1 (x)} < \zeta_2
\end{equation*}
Let $\eta_{m,x_0,x}(t) = exp_{x_0}^{g_m}(tu_m)$ and $\eta_{m,x,x_0}(t) = exp_x^{g_m}(-t P_{x_0,x}^{g_m}(u_m))$ be the $g_m$-geodesics connecting $x_0$ to $x$ and $x$ to $x_0$, respectively. Let $\zeta_1 = \sigma_2$. Since $\eta_{m,x_0,x}([0,1]) \subset U_{2,l}$, let $\zeta_2 = \min \{ \sigma_1, \sigma_2 \}$. We see that $\eta_{m,x,x_0}([0,1]) \subset U_{2,l}$, and 
\begin{equation*}
    |\eta_{1,x,x_0}^\prime (0) - \eta_{2,x,x_0}^\prime (0)|_{g_1 (x)} = |P_{x_0,x}^{g_1}(u_1) - P_{x_0,x}^{g_2}(u_2)|_{g_1 (x)} < \zeta_2 \leq \sigma_2
\end{equation*}
then for $v_1,v_2 \in T_xM$ with $|v_1|_{g_m (x)}, |v_2|_{g_m (x)} < L_1 (l)$ we have that
\begin{equation*}
    |v_1 - v_2|_{g_1 (x)} < \sigma_2
\end{equation*}
implies that
\begin{equation*}
        |(P_{x_0,x}^{g_1})^{-1}(v_1) - (P_{x_0,x}^{g_2})^{-1}(v_2)|_{g_1 (x_0)} =    |P_{x,x_0}^{g_1}(v_1) - P_{x,x_0}^{g_2}(v_2)|_{g_1 (x_0)} < \sigma_1
\end{equation*}
Then we see that $\delta_a (M,g_1, U_{1,l}, U_{2,l}, L_1 (l)) = \min \{ \delta_1 (M,g_1,U_{2,l}, L_1 (l)), \delta_2 (M,g_1, U_{2,l},U_{1,l}, L_1 (l)),\\ \delta_3 (M,g_1,U_{2,l}, L_1(l)) \} >0$ gives us the property that $\parallel g_1 - g_2 \parallel_{C^2 (M,g_1)} < \delta_a (M,g_1, U_{1,l}, U_{2,l}, L_1 (l))$ implies that for all $x \in M \setminus Cut_{g_1}(x_0)$ with $d_{g_1}(x, Cut_{g_1}(x_0)) \geq l$, and $v_1,v_2 \in T_xM$ with $|v_1|_{g_m (x)}, |v_2|_{g_m (x)} < L_1 (l)$, from theorem~\ref{def:thrm3},
\begin{equation*}
    |v_1 - v_2|_{g_1 (x)} < \sigma_2
\end{equation*}
implies that
\begin{equation*}
    |(P_{x_0,x}^{g_1})^{-1} (v_1) - (P_{x_0,x}^{g_2})^{-1} (v_2)|_{g_1 (x_0)} < \sigma_1
\end{equation*}
which then implies that
\begin{equation*}
    d_{g_1} (exp_{x_0}^{g_1} (P_{x_0,x}^{g_1})^{-1}(v_1), exp_{x_0}^{g_2} (P_{x_0,x}^{g_2})^{-1}(v_2)) < \epsilon
\end{equation*}

Then from corollary~\ref{cor:neat5} there exists $\delta_b (r, \zeta_3)>0$ such that $\parallel g_1 - g_2 \parallel_{C^2 (M,g_1)} < \delta_b (r, \zeta_3)$ implies that for all $z \in M$, $exp_z^{g_2,-1}$ is defined on $B_{g_1}(z,r)$ and for all $y \in B_{g_1}(z,r)$ we have
\begin{equation*}
    |exp_z^{g_1,-1}(y) - exp_z^{g_2,-1}(y)|_{g_1 (z)} < \zeta_3
\end{equation*}
So, let $\zeta_3 = \min \{ \sigma_2, \frac{r}{2} \}$ and for all $y \in B_{g_1}(z,r)$ we have

\begin{align*}
    |exp_z^{g_2,-1}(y)|_{g_1 (z)} & \leq |exp_z^{g_1,-1}(y) - exp_z^{g_2,-1}(y)|_{g_1 (z)} +  |exp_z^{g_1,-1}(y)|_{g_1 (z)}\\
    &  < \frac{r}{2} + r = \frac{3r}{2}
\end{align*}
Then we have
\begin{equation*}
     |exp_z^{g_1,-1}(y)|_{g_2 (z)} < 2 |exp_z^{g_1,-1}(y)|_{g_1 (z)} < 2r < L_1 (l)
\end{equation*}
and
\begin{equation*}
     |exp_z^{g_2,-1}(y)|_{g_2 (z)} < 2 |exp_z^{g_2,-1}(y)|_{g_1 (z)} < 3r < L_1 (l)
\end{equation*}
this means that $|exp_z^{g_1,-1}(y)|_{g_m (z)}, |exp_z^{g_m,-1}(y)|_{g_1 (z)} < L_1 (l)$.
Then the number $\delta (M,g_1,r,l, \epsilon) =\\ \min \{ \delta_a (M,g_1, U_{1,l}, U_{2,l}, L_1 (l)), \delta_b (r, \zeta_3) \} >0$ has the property that $\parallel g_1 - g_2 \parallel_{C^2 (M,g_1)} < \delta (M,g_1,r,l, \epsilon)$ implies that for all $x \in M \setminus Cut_{g_1}(x_0)$ with $d_{g_1}(x, Cut_{g_1}(x_0)) \geq l$ and all $y \in B_{g_1}(x,r)$, we have
\begin{equation*}
    d_{g_1}(exp_{x_0}^{g_1} \circ (P_{x_0,x}^{g_1})^{-1} \circ exp_x^{g_1,-1}(y), exp_{x_0}^{g_2} \circ (P_{x_0,x}^{g_2})^{-1} \circ exp_x^{g_2,-1}(y)) < \epsilon
\end{equation*}
as desired.
\end{proof}

Proposition~\ref{def:prop5} controls the geometric part of the transported kernel: the point $\exp_{x_0}^{g_m}\circ(P^{g_m}_{x_0,x})^{-1}\circ\exp_x^{g_m,-1}(y)$ at which the filter is evaluated moves by less than $\epsilon$, uniformly for $x$ at distance at least $l$ from $\operatorname{Cut}_{g_1}(x_0)$. Since $K(x_0,\cdot)$ is continuous on the compact manifold $M$, it is uniformly continuous, and this converts the geometric comparison into uniform closeness of $K_1$ and $K_2$ on that region. It remains to pass from this to the $L^2$ estimate, which requires handling the neighborhood of the cut locus and the change in the volume form. This is the subject of the next subsection.

\subsection{From kernel closeness to the $L^2$ estimate}\label{sec:L2}

Two obstacles remain before the kernel comparison of the previous subsection yields closeness of $h_1$ and $h_2$ in $L^2(M,g_1)$. The integrals defining $h_1$ and $h_2$ are taken with respect to different volume forms, and proposition~\ref{def:prop5} gives no control for $x$ near the cut locus. We resolve the first with a lemma, and the second by using that the cut locus has measure zero. This gives the main theorem of the paper.

\begin{theorem}\label{thrm:fin}

For all $\epsilon >0$, there exists $\delta (M, g_1, f, K, \epsilon)>0$, such that $\parallel g_1 - g_2 \parallel_{C^2 (M,g_1)} < \delta (M, g_1, f, K, \epsilon)$ implies that
\begin{equation*}
    \parallel  h_1  - h_2  \parallel_{L^2 (M, g_1)}^2 < \epsilon
\end{equation*}

\end{theorem}

The proof combines three ingredients. Lemma~\ref{lemma:fin}, stated below, shows that replacing the volume form of $g_2$ by that of $g_1$ changes the output $h_2$ by an arbitrarily small amount, uniformly in $x$. Proposition~\ref{def:prop5}, together with the continuity of $K(x_0,\cdot)$, shows that $K_1$ and $K_2$ are uniformly close for $x$ outside a neighborhood of the cut locus. That neighborhood can be chosen with arbitrarily small volume, since the cut locus has measure zero, and on it we use only the bound $|K_1-K_2| \leq 2K_0$. Given $\epsilon > 0$, we first fix the neighborhood, and then the size of the metric perturbation.

\begin{lemma}\label{lemma:fin}

For all $\epsilon >0$, there exists $\delta (M,g_1, f,K, \epsilon) >0$ such that $\parallel g_1 - g_2 \parallel_{C^2 (M, g_1)} < \delta (M,g_1, f,K, \epsilon)$ implies that for all $x \in M$, we have
\begin{equation*}
    \big| \int_M f(y)K_2 (x,y) \ d \mu_1 (y)  - \int_M f(y)K_2(x,y) \ d \mu_2 (y) \big| < \epsilon
\end{equation*}   
\end{lemma}

The proof diagonalizes $g_2$ relative to $g_1$ at each point, so that in these coordinates the volume densities differ by a product of eigenvalues. The pointwise distance between the metrics therefore bounds how far the ratio of volume densities is from 1. The details are in Appendix~\ref{app:lemma-fin}.

\begin{proof}[Proof of Theorem \ref{thrm:fin}]

First, let $F= \parallel f \parallel_\infty$ and $K_0= \parallel K(x_0, \cdot) \parallel_\infty$ and let $V = \mu_1 (M)$; all three are finite because $f$ and $K(x_0,\cdot)$ are continuous on the compact manifold $M$.

Next, we observe that from lemma~\ref{lemma:fin}, for an arbitrary $0 < \eta$, there is a $\delta_1 (M, g_1, f, K, \eta)> 0$ such that $\parallel g_1 - g_2 \parallel_{C^2 (M, g_1)} < \delta_1 (M, g_1, f, K, \eta)$ implies that
\begin{equation*}
   \big| \int_M f (y) K_2 (x,y) \ d \mu_1 (y)  - \int_M f(y) K_2 (x,y) \ d \mu_2 (y) \big| < \eta
\end{equation*}

then we have

\begin{equation}
    \begin{split}
   |h_1 (x) - h_2 (x)|^2 & = \bigg| \int_M f(y) K_1(x,y) d \mu_1 (y)  -  \int_M f(y) K_2 (x,y) d \mu_2 (y) \bigg|^2\\ 
   & = \bigg| \int_M f(y) K_1(x,y) d \mu_1 (y) -  \int_M f(y) K_2 (x,y) d \mu_1 (y)\\
   & + \int_M f(y) K_2 (x,y) d \mu_1 (y) - \int_M f(y) K_2 (x,y) d \mu_2 (y) \bigg|^2\\
   & \leq 2 \bigg| \int_M f(y) K_1(x,y) d \mu_1 (y) -  \int_M f(y) K_2 (x,y) d \mu_1 (y) \bigg|^2\\
   & + 2 \bigg| \int_M f(y) K_2 (x,y) d \mu_1 (y) - \int_M f(y) K_2 (x,y) d \mu_2 (y) \bigg|^2\\
   & \leq 2 \bigg| \int_M f(y) K_1(x,y) d \mu_1 (y) -  \int_M f(y) K_2 (x,y) d \mu_1 (y) \bigg|^2 + 2 \eta^2 \\
   & = 2 \bigg| \int_M f(y) (K_1 (x,y) - K_2 (x,y)) d \mu_1 (y) \bigg|^2 + 2 \eta^2 \\
   & \leq 2 V \int_M |f(y)|^2 |K_1 (x,y) - K_2 (x,y)|^2 d \mu_1 (y) + 2 \eta^2\\
   & \leq 2F^2 V  \int_M   | K_1 (x,y) - K_2 (x,y)|^2 d \mu_1 (y) + 2 \eta^2\\
    \end{split}
\end{equation}

Since $Cut_{g_1}(x_0)$ is a compact set of measure zero (theorem 10.34, p.311~\cite{lee2018introduction}), for every $\sigma >0$ it can be covered by finitely many open balls whose $\mu_1$-measure is less than $\sigma$. Taking $\sigma = \zeta_1$, let $M_{g_1, \zeta_1}$ be the union of such balls, and write $E = M_{g_1, \zeta_1}$ and $A = M \setminus E$. Then $E$ is an open subset of $M$ such that $Cut_{g_1}(x_0) \subset E$ and $\mu_1 (E) < \zeta_1$, and $A \subset M \setminus Cut_{g_1}(x_0)$ is compact. Hence, there exists $0 < l_{\zeta_1}$ such that for all $x \in A$ we have $l_{\zeta_1} < d_{g_1}(x, Cut_{g_1}(x_0))$. Then we see that
\begin{equation*}
    |K_1 (x,y) -K_2 (x,y)| \leq |K_1 (x,y)| + |K_2 (x,y)| \leq 2 K_0
\end{equation*}

Next let $\epsilon >0$ and we observe
\begin{equation*}
\begin{split}
 \parallel  h_1  - h_2  \parallel_{L^2 (M, g_1)}^2 & = \int_M |h_1 (x) - h_2 (x)|^2 d \mu_1 (x) \\
 & \leq \int_M \bigg( 2F^2 V   \int_M   | K_1 (x,y) - K_2 (x,y)|^2 d \mu_1 (y)  + 2 \eta^2 \bigg) d \mu_1 (x)\\
 & = 2F^2 V \int_M \bigg(  \int_M   | K_1 (x,y) - K_2 (x,y)|^2 d \mu_1 (y)  \bigg) d \mu_1 (x) + 2 V \eta^2\\
 & = 2F^2 V \int_A  \big(   \int_{M }   | K_1 (x,y) - K_2 (x,y)|^2 d \mu_1 (y) \big) d \mu_1 (x)\\
 &+ 2F^2 V  \int_E  \big( \int_{M}   | K_1 (x,y) - K_2 (x,y)|^2 d \mu_1 (y) \big) d \mu_1 (x) + 2 V \eta^2\\
 & \leq 2F^2 V \int_A  \bigg(  \int_{M }   | K_1 (x,y) - K_2 (x,y)|^2 d \mu_1 (y) \bigg) d \mu_1 (x)\\
 & + 2F^2 V \int_E  \bigg( \int_{M}   4K_0^2 d \mu_1 (y) \bigg) d \mu_1 (x) + 2 V \eta^2\\
 & = 2F^2 V \int_A  \bigg(  \int_{M }   | K_1 (x,y) - K_2 (x,y)|^2 d \mu_1 (y) \bigg) d \mu_1 (x)\\
 & + 2F^2 V^2 \mu_1 (E)  4K_0^2 + 2 V \eta^2\\
 & \leq 2F^2 V \int_A  \bigg(  \int_{M}   | K_1 (x,y) - K_2 (x,y)|^2 d \mu_1 (y) \bigg) d \mu_1 (x)\\
 & + 8F^2 V^2 \zeta_1  K_0^2 + 2 V \eta^2 = \star 
\end{split}
\end{equation*}
Now let $\eta = \sqrt{\frac{\epsilon}{6 V}}$ and  $\zeta_1 = \frac{\epsilon}{24 K_0^2 F^2 V^2}$, and recall that $d_{g_1} (x, Cut_{g_1}(x_0)) > l_{\zeta_1}$ for all $x \in A$. Then for every $\zeta_2 > 0$, proposition~\ref{def:prop5} gives $\delta_2 (M,g_1, r, l_{\zeta_1}, \zeta_2) > 0$ such that $\parallel g_1 - g_2 \parallel_{C^2 (M,g_1)} < \delta_2 (M,g_1, r, l_{\zeta_1}, \zeta_2)$ implies that for all $x \in A$ and $y \in B_{g_1}(x,r)$, we have
\begin{equation*}
    d_{g_1}(exp_{x_0}^{g_1} \circ (P_{x_0,x}^{g_1})^{-1} \circ exp_x^{g_1,-1}(y), exp_{x_0}^{g_2} \circ (P_{x_0,x}^{g_2})^{-1} \circ exp_x^{g_2,-1}(y)) < \zeta_2
\end{equation*}
Then by the continuity of $K(x_0, \cdot)$, $\zeta_2$ sufficiently small implies that
\begin{equation*}
\begin{split}
     |K_1(x,y) - K_2 (x,y)| & = |K (x_0, \exp_{x_0}^{g_1} \circ (P_{x_0,x}^{g_1})^{-1} \circ \exp_x^{g_1,-1}(y))\\
     &- K (x_0, \exp_{x_0}^{g_2} \circ (P_{x_0,x}^{g_2})^{-1} \circ \exp_x^{g_2,-1}(y))|\\
     & < \omega
\end{split}
\end{equation*}
Since $K_1 (x,y) = K_2 (x,y)=0$ for $y \notin B_{g_1}(x,r)$ (since $K$ has compact support in $B_{g_1}(x_0,r)$), the bound $|K_1(x,y) - K_2 (x,y)| < \omega$ extends trivially from $B_{g_1}(x,r)$ to all $y \in M$, for every $x \in A$. Then we see that
\begin{align*}
  \int_A  \bigg(  \int_{M }   | K_1 (x,y) - K_2 (x,y)|^2 d \mu_1 (y) \bigg) d \mu_1 (x) & < \int_A \bigg(  \int_{M }   \omega^2 d \mu_1 (y) \bigg) d \mu_1 (x)\\
  & = V \mu_1  (A) \omega^2\\
  & \leq V^2 \omega^2
\end{align*}
since $A  \subset M$, we have $\mu_1 (A) \leq V$. Then let $\omega = \sqrt{\frac{\epsilon}{6 F^2 V^3}}$. Then for $\delta (M, g_1, f, K, \epsilon) = \min \{ \delta_1, \delta_2 \}$, we know that $\parallel g_1 - g_2 \parallel_{C^2 (M, g_1)} < \delta (M, g_1, f, K, \epsilon)$ implies that
\begin{equation*}
    \begin{split}
    \star & < 2F^2 V^3 \omega^2
    + 8F^2 V^2 \zeta_1  K_0^2 + 2 V \eta^2\\
    & = \frac{\epsilon}{3} + \frac{\epsilon}{3} + \frac{\epsilon}{3} = \epsilon
    \end{split}
\end{equation*}
as desired.
\end{proof}

This completes the proof of theorem~\ref{thrm:fin}: for fixed $f$ and $K$, the parallel transport convolution depends continuously, in $L^2(M,g_1)$, on the metric in the $C^2$ topology.

\section{Conclusion}

In this paper, we established the stability of parallel transport convolution under perturbations of the Riemannian metric on a fixed compact manifold. Specifically, we proved that, for a continuous input function and a continuous reference kernel with sufficiently small support, the convolution output depends continuously in \(L^2\) on the underlying metric with respect to the \(C^2\)-topology, under the prescribed localization and transport-path conventions. This result provides a rigorous stability guarantee for a spatial convolution operator whose construction depends explicitly on the geometry of the underlying manifold.

The main challenge arises from the dependence of the convolution on several geometric structures, including geodesics, parallel transport, exponential maps, and the Riemannian volume measure. We developed perturbation estimates for these structures and combined them to establish uniform convergence of the transported kernels away from the cut locus. Since the cut locus has measure zero, its contribution can be controlled through integration over neighborhoods of arbitrarily small volume. This allows the local geometric estimates to yield a global \(L^2\)-stability result.

Our analysis provides a mathematical foundation for understanding the robustness of parallel transport convolution to perturbations of the underlying geometry. The present result establishes qualitative continuity for a single convolution operator with fixed input and kernel. Natural directions for future research include obtaining quantitative stability estimates in terms of the metric perturbation, extending the analysis to compositions of convolutional layers and nonlinear architectures, and investigating stability under weaker regularity assumptions on the metric and input data. Such developments would further clarify the relationship between geometric perturbations and the stability of manifold-based learning methods.

\bibliography{main.bib}
\bibliographystyle{alpha}

\appendix

\section{Proofs for Section~\ref{sec:pre}}
In this section, we provide the proofs of the statements in Section~\ref{sec:pre}.

\subsection{Proof of Proposition \ref{def:prop1}}

\phantomsection\label{app:prop1}
\begin{proof}[Proof of Proposition \ref{def:prop1}]

    For $v \in T_pM$, let $|v|_{h(p)}^2 = h_p (v,v)$, and we'll use the quotient on $U$,
    \begin{equation*}
        \frac{g_p (v,v)}{|v|_{h(p)}^2}
    \end{equation*}
    where $p \in U$, and $v \neq 0$. Then on $\overline{U}$, we define
    \begin{equation*}
        \begin{split}
            \lambda_{\max} (p) & = \sup_{|v|_{h(p)}=1} \sqrt{ g_p (v,v)}\\
            \lambda_{\min} (p) & = \inf_{|v|_{h(p)}=1} \sqrt{ g_p (v,v) }\\
        \end{split}
    \end{equation*}
    Since we know that for the coordinate chart $(V, \phi)$ we have $\overline{U} \subset V$, then $\lambda_{\max} (p)$ and $\lambda_{\min} (p)$ are defined on all of $\overline{U}$.
    Then because $g$ and $h$ are continuous on $\overline{U}$ then $\lambda_{\max} (p)$ and $\lambda_{\min} (p)$ are continuous on $\overline{U}$ as well. Then choose $u \in T_pM$, then
\begin{equation*}
    \lambda_{\min}^2 (p) \leq \frac{g_p (u,u)}{|u|_{h(p)}^2} \leq \lambda_{\max}^2 (p)
\end{equation*}
and since $g$ is a positive definite matrix, $\lambda_{\min} (p) >0$ on $\overline{U}$. Now by the compactness of $\overline{U}$, we can define,
\begin{equation*}
    \begin{split}
        \lambda_{\max} & = \max_{p \in \overline{U}} \lambda_{\max} (p)\\
        \lambda_{\min} & = \min_{p \in \overline{U}} \lambda_{\min} (p)
    \end{split}
\end{equation*}
Then since $\overline{U}$ is compact, there are constants
\begin{equation*}
    0 < \lambda_{\min} \leq \lambda_{\max} < \infty
\end{equation*}
such that $0 < \lambda_{\min} \leq \lambda_{\min} (p) \leq \lambda_{\max}(p) \leq \lambda_{\max} < \infty$, where $\lambda_{\min}$ and $\lambda_{\max}$ are based on the choice of $U$. Then for $u \in T_pM$, then
\begin{equation*}
    0 < \lambda_{\min}^2 \leq g_p (\frac{u}{|u|_{h(p)}}, \frac{u}{|u|_{h(p)}}) \leq \lambda_{\max}^2 < \infty
\end{equation*}
which means that
\begin{equation}\label{def:eq2}
    \lambda_{\min}^2 |u|_{h(p)}^2 \leq g_p (u,u) \leq \lambda_{\max}^2 |u|_{h(p)}^2
\end{equation}
on all of $U$.

From here, choose $a \in \Gamma( Sym^2 (T^*M))$, fix $p \in U$, and let $\{ X_i \}$ be the coordinate frame on $U$. Since $h$ is the pullback of the Euclidean metric to $U$, $\{ X_i \}$ is $h$-orthonormal. Since $(g_{ij})$ is a real symmetric matrix, by theorem 10.7, p. 233,~\cite{roman2007advanced}, there exists $O \in O(n)$, such that $O^T (g_{ij})O=D$, where $D$ is a diagonal matrix and since $(g_{ij})$ is also positive definite, $d_i >0$ for $i=1, \dots, n$. Additionally, by what we've just shown, $d_i \in [\lambda_{\min}^2, \lambda_{\max}^2]$ for $i = 1, \dots , n$.
Now let $e_i = O_{ji}X_j$, and we'll  show that $|a|_g^2$ is invariant under this change of basis. Let $a_{ij} = a(X_i, X_j)$, $\Tilde{a}_{ij} = a(e_i, e_j)$, $g_{ij}= g (X_i, X_j)$, $\Tilde{g}_{ij}= g(e_i, e_j)$, $g^{ij}= (g(X_i, X_j))^{-1}$, and $\Tilde{g}^{ij} = (g(e_i, e_j))^{-1}$. This gives us the relationships:
\begin{align*}
    \Tilde{a}_{ij} &= \sum_{kl} O_{ki} a_{kl} O_{lj}\\
    \Tilde{g}^{i \alpha} & = \sum_{tq} O_{ti}g^{tq}O_{q \alpha }
\end{align*}
Then we have
\begin{align*}
    |\Tilde{a}|_{\Tilde{g}}^2 & = \sum_{ij \alpha \beta} \Tilde{a}_{ij} \Tilde{a}_{\alpha \beta} \Tilde{g}^{i \alpha} \Tilde{g}^{j \beta}\\
    & = \sum_{ij \alpha \beta} \sum_{klmntqrs} (O_{ki} a_{kl}O_{lj}) (O_{m \alpha } a_{mn}O_{n \beta }) (O_{ti}g^{tq}O_{q \alpha })(O_{rj}g^{rs}O_{s \beta })\\
    & = \sum_{ij \alpha \beta} \sum_{klmntqrs} (O_{ki}O_{ti})(O_{m \alpha }O_{q \alpha })(O_{rj}O_{lj})(O_{n \beta }O_{s \beta })a_{kl}a_{mn}g^{tq}g^{rs}\\
    & = \sum_{ij \alpha \beta} \sum_{klmntqrs} \delta_{kt} \delta_{mq} \delta_{rl} \delta_{ns} a_{kl} a_{mn} g^{tq} g^{rs}\\
    & = \sum_{klqn} a_{kl} a_{qn}g^{kq} g^{ln} = |a|_g^2
\end{align*}

So, $|a|_g^2$ is independent of this change of basis, as will $|a|_h^2$. Then for the $(0,3)$ and $(0,4)$-tensor cases, let $T_{i_1 \dots i_m} = T(X_{i_1}, \dots , X_{i_m})$ and $\Tilde{T}_{i_1 \dots i_m} = T(e_{i_1}, \dots , e_{i_m})$ then we have
\begin{equation*}
    \Tilde{T}_{i_1 \dots i_m} = \sum_{j_1 \dots j_m} O_{j_1i_1} \cdots O_{j_mi_m} T_{j_1 \dots j_m}
\end{equation*}
Then we see that
\begin{align*}
    |\Tilde{T}|_{\Tilde{g}}^2 & = \sum_{i_1 \dots i_mk_1 \dots k_m}  \Tilde{T}_{i_1 \dots i_m} \Tilde{T}_{k_1 \dots k_m} \Tilde{g}^{i_1k_1} \cdots \Tilde{g}^{i_mk_m}\\
    & = \sum_{i_1k_1 \dots i_mk_m j_1 \dots j_m l_1 \dots l_m t_1q_1 \dots t_mq_m} \big( \prod_{r=1}^m O_{j_ri_r} \big) \big( \prod_{r=1}^m O_{l_rk_r} \big) \big( \prod_{r=1}^m O_{t_ri_r} g^{t_rq_r}O_{q_rk_r} \big)  T_{j_1 \dots j_m} T_{l_1 \dots l_m}\\
    & =\sum_{i_1k_1 \dots i_mk_m j_1 \dots j_m l_1 \dots l_m t_1q_1 \dots t_mq_m} \bigg( \prod_{r=1}^m \big( O_{j_ri_r}O_{t_ri_r} \big) \big( O_{l_rk_r}O_{q_rk_r} \big) g^{t_rq_r} \bigg) T_{j_1 \dots j_m} T_{l_1 \dots l_m}\\
    & = \sum_{ j_1 \dots j_m l_1 \dots l_m t_1q_1 \dots t_mq_m} \big( \prod_{r=1}^m \delta_{j_r t_r} \delta_{l_rq_r}g^{t_rq_r} \big)  T_{j_1 \dots j_m} T_{l_1 \dots l_m}\\
    & = \sum_{j_1 \dots j_m l_1 \dots l_m} T_{j_1 \dots j_m} T_{l_1 \dots l_m} g^{j_1l_1} \cdots g^{j_ml_m} = |T|_g^2
\end{align*}

So, since $|\nabla^g a|_g^2$, $|(\nabla^g)^2 a|_g^2$, $|\nabla^h a|_h^2$, and $|(\nabla^h)^2 a|_h^2$ are $(0,3)$ and $(0,4)$-tensors, they are independent of an orthogonal change of basis. Then we have

\begin{align*}
    |a|_g^2 = |\Tilde{a}|_{\Tilde{g}}^2 & = \sum_{ij \alpha \beta} \Tilde{a}_{ij} \Tilde{a}_{\alpha \beta} (\sum_{tq} O_{ti} g^{tq}O_{q \alpha }) (\sum_{rs} O_{rj} g^{rs} O_{s \beta })\\
    & = \sum_{ij \alpha \beta} \Tilde{a}_{ij} \Tilde{a}_{\alpha \beta} (O^T (g^{ij})O)_{i \alpha} (O^T (g^{ij})O)_{j \beta}\\
    & = \sum_{ij \alpha \beta} \Tilde{a}_{ij} \Tilde{a}_{\alpha \beta} \frac{\delta_{i \alpha}}{d_i} \frac{\delta_{j \beta}}{d_j}\\
    & = \sum_{ij} \frac{(\Tilde{a}_{ij})^2}{d_i d_j}\\
    & \leq \sum_{ij} \frac{(\Tilde{a}_{ij})^2}{(\lambda_{\min})^4} = \frac{|\Tilde{a}|_{\Tilde{h}}^2 }{(\lambda_{\min})^4} =  \frac{|a|_h^2 }{(\lambda_{\min})^4}
\end{align*}
This also holds for
\begin{align*}
    |a|_g^2 = |\Tilde{a}|_{\Tilde{g}}^2 \geq \frac{|\Tilde{a}|_{\Tilde{h}}^2}{(\lambda_{\max})^4} = \frac{|a|_h^2}{(\lambda_{\max})^4}
\end{align*}

Then for the $(0,3)$ and the $(0,4)$-tensor cases we have
\begin{align*}
    |T|_g^2 = |\Tilde{T}|_{\Tilde{g}}^2 & = \sum_{i_1 \dots i_m k_1 \dots k_m} \Tilde{T}_{i_1 \dots i_m} \Tilde{T}_{k_1 \dots k_m} \prod_{r=1}^m \frac{\delta_{i_rk_r}}{d_{i_r}}\\
    & = \sum_{i_1 \dots i_m} \frac{( \Tilde{T}_{i_1 \dots i_m})^2 }{d_{i_1} \cdots d_{i_m}}\\
    & \leq \sum_{i_1 \dots i_m} \frac{( \Tilde{T}_{i_1 \dots i_m})^2 }{(\lambda_{\min})^{2m}} = \frac{|\Tilde{T}|_h^2}{(\lambda_{\min})^{2m}} = \frac{|T|_h^2}{(\lambda_{\min})^{2m}}
\end{align*}
Then in the other direction, we have
\begin{equation*}
     |T|_g^2 = |\Tilde{T}|_{\Tilde{g}}^2 \geq \frac{|\Tilde{T}|_{\Tilde{h}}^2}{(\lambda_{\max})^{2m}} = \frac{|T|_h^2}{(\lambda_{\max})^{2m}}   
\end{equation*}

Now, choose $a \in \Gamma( Sym^2 (T^*M))$ and we'll make the background metric swap for $k=0, 1, 2$:

\begin{enumerate}
    \item[k=0] 
    We already showed above that
    \begin{equation*}
    \frac{1}{ ( \lambda_{\max})^4} |a|_h^2 \leq |a|_g^2 \leq \frac{1}{ ( \lambda_{\min})^4} |a|_h^2
    \end{equation*}
    or
    \begin{equation*}
       (\lambda_{\min})^4 |a|_g^2  \leq |a|_h^2 \leq (\lambda_{\max})^4 |a|_g^2
    \end{equation*}
    
    and therefore,
    \begin{equation*}
        \frac{1}{( \lambda_{\max})^2} \parallel a \parallel_{C^0 (U,h)} \leq \parallel a \parallel_{C^0 (U,g)} \leq \frac{1}{( \lambda_{\min} )^2} \parallel a \parallel_{C^0 (U,h)}
    \end{equation*}
    where $\lambda_{\max}$ and $\lambda_{\min}$ are constants based on the chosen neighborhood $U$, and we have
    \begin{equation*}
        \parallel \cdot \parallel_{C^0 (U,g)} \sim \parallel \cdot \parallel_{C^0 (U,h)}
    \end{equation*}

    \item[k=1] 
    First, we'll recall that from above, for an $(0,3)$-tensor, $T_3$, we have
    \begin{equation*}
        \frac{|T_3|_h^2}{(\lambda_{\max})^6}  \leq |T_3|_g^2 \leq \frac{|T_3 |_h^2}{(\lambda_{\min})^6}
    \end{equation*}
    
    Next, we'll compute $\nabla^g a$, then we have
    \begin{equation*}
        \nabla^g a = dx^k \otimes \nabla^g_k a
    \end{equation*}
    where $\nabla_k^g = \nabla_{\frac{\partial}{\partial x_k}}^g$

    Then we observe that
    \begin{align*}
        dx_i (\frac{\partial}{\partial x_\alpha}) & = \delta_{i \alpha}\\
        (\nabla_k^g dx_i)(\frac{\partial}{\partial x_\alpha} ) + dx_i (\nabla_k^g (\frac{\partial}{\partial x_\alpha} ) ) & =0\\
        \nabla_k^g dx_i (\frac{\partial}{\partial x_\alpha} ) & = - \Gamma_{k \alpha}^{i, g}\\
        & = - \sum_l \Gamma_{k l}^{i, g} dx_l (\frac{\partial}{\partial x_\alpha}  )
    \end{align*}
    Then, $\nabla_k^g dx_i = - \sum_l \Gamma_{kl}^{i,g} dx_l$, which gives us
    \begin{align*}
        \nabla_k^g a & = \frac{\partial a_{ij}}{\partial x_k} dx^i \otimes dx^j + a_{ij} (\nabla_k^g dx^i) \otimes dx^j + a_{ij} dx^i \otimes (\nabla_k^g dx^j)\\
        &= \frac{\partial a_{ij}}{\partial x_k} dx^i \otimes dx^j - a_{ij} \sum_l \Gamma_{kl}^{i,g} dx_l \otimes dx^j - a_{ij} \sum_l \Gamma_{kl}^{j,g} dx^i \otimes dx_l\\
    \end{align*}
    It follows that
    \begin{align*}
        \nabla^g a & = dx^k \otimes \bigg( \frac{\partial a_{ij}}{\partial x_k} dx^i \otimes dx^j - a_{ij} \big( \sum_l \Gamma_{kl}^{i,g} dx_l \big) \otimes dx^j - a_{ij}  dx^i \otimes \big( \sum_l \Gamma_{kl}^{j,g} dx_l \big) \bigg)\\
        & = \sum_{k i j} \frac{\partial a_{ij}}{\partial x_k} dx_k \otimes dx_i \otimes dx_j - \sum_{k i j l} a_{ij} \Gamma_{kl}^{i,g} dx_k \otimes dx_l \otimes dx_j - \sum_{k i j l} a_{ij} \Gamma_{kl}^{j,g} dx_k \otimes dx_i \otimes dx_l\\
        & = \star
    \end{align*}
    Then under the first sum $\sum_{k i j l} a_{ij} \Gamma_{kl}^{i,g} dx_k \otimes dx_l \otimes dx_j$, we'll switch the indices $i$ and $l$, and get
    \begin{equation*}
        \sum_{k i j l} a_{lj} \Gamma_{ki}^{l,g} dx_k \otimes dx_i \otimes dx_j
    \end{equation*}
    Next, for the second sum $\sum_{k i j l} a_{ij} \Gamma_{kl}^{j,g} dx_k \otimes dx_i \otimes dx_l$ we'll switch the indices $j$ and $l$, and get
    \begin{equation*}
        \sum_{k i j l} a_{il} \Gamma_{kj}^{l,g} dx_k \otimes dx_i \otimes dx_j
    \end{equation*}
    In this way, we can reorganize $\star$ to give us
    \begin{align*}
        \star & = \sum_{k i j} \frac{\partial a_{ij}}{\partial x_k} dx_k \otimes dx_i \otimes dx_j - \sum_{k i j l} a_{lj} \Gamma_{ki}^{l,g} dx_k \otimes dx_i \otimes dx_j - \sum_{k i j l} a_{il} \Gamma_{kj}^{l,g} dx_k \otimes dx_i \otimes dx_j\\
        & =  \bigg( \frac{\partial a_{ij}}{\partial x_k} - \sum_l a_{lj} \Gamma_{ki}^{l,g} - \sum_l a_{il} \Gamma_{kj}^{l,g}  \bigg) dx^k \otimes dx^i \otimes dx^j\\
        & =  \bigg( \frac{\partial a_{ij}}{\partial x_k} - \sum_l ( a_{lj} \Gamma_{ki}^{l,g} + a_{il} \Gamma_{kj}^{l,g} )  \bigg) dx^k \otimes dx^i \otimes dx^j\\
    \end{align*}
    Then we have that
    \begin{equation*}
        \nabla^g a = (\nabla_k^g a)_{ij} dx^k \otimes dx^i \otimes dx^j
    \end{equation*}

    where 
    \begin{equation*}
        (\nabla_k^g a)_{ij} = \frac{\partial a_{ij}}{\partial x_k} - \sum_l ( a_{lj} \Gamma_{ki}^{l,g} + a_{il} \Gamma_{kj}^{l,g} ) 
    \end{equation*}

    or in schematic tensor notation (or Hamilton $\ast$ notation, p. 294,~\cite{hamilton1982three}), we have
    \begin{equation*}
        \nabla^g a = \partial a - \Gamma^g * a
    \end{equation*}
    Then since $h$ is the Euclidean metric, $\Gamma_{ij}^{k,h} \equiv 0$, which means
    \begin{equation*}
        \nabla^h a = \partial a = \nabla^g a + \Gamma *a
    \end{equation*}
    Then 
    \begin{equation*}
        |\Gamma *a|_h \leq C(n) \max |\Gamma^g||a|_h
    \end{equation*}
    where $C(n)$ is a constant depending on the number of factors and $n$ ($\max |\Gamma^g|$ and $\max |\partial \Gamma^g|$ are defined in remark~\ref{def:rmk1}). Then by the triangle inequality, we have
    \begin{equation*}
        |\nabla^h a|_h = |\partial a|_h \leq |\nabla^g a|_h +  C(n) \max |\Gamma^g||a|_h
    \end{equation*}
    Then using the Cauchy-Schwartz inequality for squares, we have
    \begin{align*}
        |\nabla^h a|_h^2 & = |\partial a|_h^2 \leq 2|\nabla^g a|_h^2 +  C(n) \max |\Gamma^g|^2|a|_h^2\\
        & \leq 2 (\lambda_{\max})^6 |\nabla^g a|_g^2 +   C(n) \max |\Gamma^g|^2 (\lambda_{\max})^4 |a|_g^2\\
    \end{align*}
    Then by letting $C_1 (M,g,U,1) = \max \{ 2 (\lambda_{\max})^6, (1+C(n) \max |\Gamma^g|^2 )(\lambda_{\max})^4  \}$, we have
    \begin{align*}
        |\nabla^h a|_h^2 + |a|_h^2 & \leq 2 (\lambda_{\max})^6 |\nabla^g a|_g^2 + (1+C(n) \max |\Gamma^g|^2 )(\lambda_{\max})^4 |a|_g^2\\
        & \leq C_1 (M,g,U,1) ( |\nabla^g a|_g^2 + |a|_g^2 )
    \end{align*}
    Then, in the other direction, we have
    \begin{equation*}
        \nabla^g a = \partial a - \Gamma*a
    \end{equation*}
    which gives us
    \begin{align*}
        (\lambda_{\min})^6 |\nabla^ga|_g^2 \leq |\nabla^g a|_h^2 & \leq 2|\partial a|_h^2 + C(n) \max |\Gamma^g|^2 |a|_h^2\\
        & \leq 2|\nabla^h a|_h^2 + C(n) \max |\Gamma^g|^2 |a|_h^2\\
        |\nabla^g a|_g^2 & \leq \frac{2|\nabla^h a|_h^2 + C(n) \max |\Gamma^g|^2 |a|_h^2 }{(\lambda_{\min})^6} 
    \end{align*}
    Then by letting $C = \max \{ \frac{2}{(\lambda_{\min})^6}, \frac{C(n) \max| \Gamma^g|^2}{(\lambda_{\min})^6} + \frac{1}{(\lambda_{\min})^4} \}$
    \begin{align*}
        |a|_g^2 + |\nabla^g a|_g^2 & \leq \frac{2}{(\lambda_{\min})^6} |\nabla^h a|_h^2 + \big( \frac{C(n) \max| \Gamma^g|^2}{(\lambda_{\min})^6} + \frac{1}{(\lambda_{\min})^4} \big) |a|_h^2\\
        & \leq C (|\nabla^h a|_h^2 + |a|_h^2)
    \end{align*}
    Then by letting $C_2 (M,g,U,1) = \frac{1}{C}$ we have

    \begin{equation*}
        C_2 ( |\nabla^g a|_g^2 + |a|_g^2 ) \leq |\nabla^h a|_h^2 + |a|_h^2 \leq C_1 ( |\nabla^g a|_g^2 + |a|_g^2 )
    \end{equation*}

    Then on our chosen fixed neighborhood $U$, we have
    \begin{equation*}
       \sqrt{ C_2 (M,g, U,1)} \parallel a \parallel_{C^1 (U,g)} \leq \parallel a \parallel_{C^1 (U,h)} \leq \sqrt{ C_1 (M,g, U,1)} \parallel a \parallel_{C^1 (U,g)}
    \end{equation*}
    
    \item[k=2] 
    First, we recall from above that, For an $(0,4)$-tensor, $T_4$, we have
    \begin{equation*}
        \frac{ |T_4 |_h^2}{(\lambda_{\max})^8}  \leq |T_4 |_g^2 \leq \frac{|T_4 |_h^2}{(\lambda_{\min})^8}
    \end{equation*}
    
    Next, in schematic tensor notation, we have,
    \begin{align*}
        (\nabla^g)^2a & = \nabla^g (\nabla^g a)\\
        & = \partial (\nabla^g a) + \Gamma^g * \nabla^g a\\
        & = \partial^2 a + \partial \Gamma^g *a + \Gamma^g * \partial a + \Gamma^g * \nabla^g a\\
        & = \partial^2 a + \partial \Gamma^g *a + \Gamma^g * (\nabla^g a  - \Gamma^g *a) + \Gamma^g * \nabla^g a\\
        & = \partial^2 a + \Gamma^g * \nabla^g a + (\partial \Gamma^g + \Gamma^g * \Gamma^g)*a\\
        & = (\nabla^h)^2 a + \Gamma^g * \nabla^g a + (\partial \Gamma^g + \Gamma^g * \Gamma^g)*a\\
    \end{align*}
    Then by Cauchy-Schwartz inequality and the triangle inequality, we have
    \begin{align*}
        |(\nabla^h)^2 a|_h^2 = |\partial^2 a|_h^2 & \leq 3|(\nabla^g)^2 a|_h^2 + C(n) \max |\Gamma^g|^2 |\nabla^g a|_h^2 + C(n) (\max |\partial \Gamma^g| + \max |\Gamma^g|^2)^2 |a|_h^2\\
        & \leq (\lambda_{\max})^8 |(\nabla^g)^2 a|_g^2 + C(n) \max |\Gamma^g|^2 (\lambda_{\max})^6 |\nabla^g a|_g^2\\
        & + C(n)(\max |\partial \Gamma^g| + \max |\Gamma^g|^2)^2 (\lambda_{\max})^4 |a|_g^2
    \end{align*}
    Then by letting $C_1 (M,g,U,2) = \max \{ (\lambda_{\max})^8, (2 + C(n)\max |\Gamma^g|^2)(\lambda_{\max})^6, (2 + C(n) \max|\Gamma^g|^2 + C(n)(\max |\partial \Gamma^g| + \max |\Gamma^g|^2)^2 )(\lambda_{\max})^4 \}$ and from $k=0$ and $k=1$, we have
    \begin{align*}
        |(\nabla^h)^2 a|_h^2 + |\nabla^h a|_h^2 + |a|_h^2 & \leq (\lambda_{\max})^8 |(\nabla^g)^2 a|_g^2 + C(n) \max |\Gamma^g|^2 (\lambda_{\max})^6 |\nabla^g a|_g^2\\
        & + C(n)(\max |\partial \Gamma^g| + \max |\Gamma^g|^2)^2 (\lambda_{\max})^4 |a|_g^2\\
        & + 2 (\lambda_{\max})^6 |\nabla^g a|_g^2 + (1+C(n) \max |\Gamma^g|^2 )(\lambda_{\max})^4 |a|_g^2 + (\lambda_{\max})^4 |a|_g^2\\
        & = (\lambda_{\max})^8 |(\nabla^g)^2 a|_g^2 + (2 + C(n)\max |\Gamma^g|^2)(\lambda_{\max})^6 |\nabla^g a|_g^2\\
        & + (2 + C(n) \max|\Gamma^g|^2 + C(n)(\max |\partial \Gamma^g| + \max |\Gamma^g|^2)^2 )(\lambda_{\max})^4 |a|_g^2\\
        & \leq C_1 (M,g,U,2) (|(\nabla^g)^2 a|_g^2 + |\nabla^g a|_g^2 + |a|_g^2)
    \end{align*}
    Then, in the other direction, we have
    \begin{align*}
        (\lambda_{\min})^8 |(\nabla^g)^2a|_g^2 \leq |(\nabla^g)^2a|_h^2 & \leq 3 |(\nabla^h)^2a |_h^2 + C(n) \max|\Gamma^g|^2 |\nabla^ga|_h^2 + C(n) (\max |\partial \Gamma^g| + \max |\Gamma^g|^2)^2 |a|_h^2\\
        & \leq 3 |(\nabla^h)^2a |_h^2 + C(n) \max|\Gamma^g|^2 |\nabla^ha|_h^2\\
        & + C(n)((\max |\partial \Gamma^g| + \max |\Gamma^g|^2)^2 + \max|\Gamma^g|^2) |a|_h^2
    \end{align*}
    where the second inequality comes from case $k=1$. Then let $C = \max \{ \frac{3}{(\lambda_{\min})^8}, C(n) \big( \frac{\max|\Gamma^g|^2}{(\lambda_{\min})^8} + \frac{1}{(\lambda_{\min})^6} \big), C(n) \big( \frac{\max |\partial \Gamma^g| + \max |\Gamma^g|^2)^2 + \max|\Gamma^g|^2}{(\lambda_{\min})^8} + \frac{\max|\Gamma^g|^2}{(\lambda_{\min})^6} + \frac{1}{(\lambda_{\min})^4} \big) \}$, and let $C_2(M,g,U,2) = \frac{1}{C}$ and we have
    \begin{equation*}
        C_2 (M,g,U,2) (|(\nabla^g)^2 a|_g^2 + |\nabla^g a|_g^2 + |a|_g^2) \leq |(\nabla^h)^2 a|_h^2 + |\nabla^h a|_h^2 + |a|_h^2
    \end{equation*}
    
    So, we have
    \begin{equation*}
        \sqrt{ C_2 (M,g, U, 2)} \parallel a \parallel_{C^2 (U,g)} \leq  \parallel a \parallel_{C^2 (U,h)} \leq  \sqrt{  C_1 (M,g, U, 2) } \parallel a \parallel_{C^2 (U,g)}
    \end{equation*}    
\end{enumerate}
Then, for $k=0,1,2$
\begin{equation*}
    \parallel \cdot \parallel_{C^k (U,g)} \sim  \parallel \cdot \parallel_{C^k (U,h)}
\end{equation*}
as desired.
\end{proof}

\section{Proofs for Section~\ref{sec:metricaprox}}
In this section, we provide the proofs of the statements in Section~\ref{sec:metricaprox}.

\subsection{Proof of Theorem \ref{thrm:neat}}

\phantomsection\label{app:neat}
\begin{proof} [Proof of Theorem \ref{thrm:neat}]

     First, let's construct an orthonormal frame with respect to $g_1$ along $\gamma$. Without loss of generality, we can assume that $\gamma$ has unit speed on the interval $[0,1]$. Then let $e_1^1 (t) = \gamma^\prime (t)$. Then using the Gram-Schmidt process to create an orthogonal basis $\{e_1^1, \dots , e_n^1 \}$ with respect to $g_1$ in $T_pM$, and take the parallel transport of each basis vector along $\gamma$ to create an orthonormal frame $\{ e_1^1 (t), \dots , e_n^1 (t) \}$ with respect to $g_1$ along $\gamma (t)$. Since these $e_j^1 (t)$ are parallel transports, they're independent of the choice of coordinate charts. So, we can choose one. Then, $e_1^1 (t), \dots , e_n^1 (t)$ is a basis for all points along the curve $\gamma$. Then from proposition 8.11, p.178, ~\cite{lee2018introduction} there's an open set $V$ containing $\gamma ([0,1])$ which has a local frame $X_1, \dots , X_n$ such that $X_j (\gamma (t)) = e_j^1 (t)$.  From here, we want to compute $V_2^\prime \bigg|_{g_1} (t) = \frac{D_{g_1}}{dt} V_2 (t) = \lambda_2^i \nabla_{\gamma^\prime (t)}^{g_1} e_i^2 (t)$.
   So, let $e_1^2 (t), \dots , e_n^2 (t)$ be an orthonormal basis with respect to $g_2$ along $\gamma$. Then we have
   \begin{align*}
       \lambda_1^i e_i^1 (0)  & = \lambda_2^i e_i^2 (0) =v\\
       P_{g_1}^\gamma (v) & = \lambda_1^i e_i^1 (1)\\
       P_{g_2}^\gamma (v) & = \lambda_2^i e_i^2 (1)
   \end{align*}   
    with vector fields along $V_1 (t) = \lambda_{1}^i e_i^1 (t)$ and $V_2 (t) = \lambda_{2}^i e_i^2 (t)$. Then we see that $|\lambda_m^i| \leq \sqrt{ \sum_i (\lambda_m^i)^2 } = |v|_{g_m} \leq L$.
    Then let $A (t) = (a_{ij}(t))$ be the matrix function such that $e_i^1 (t) = a_{ij}(t) e_j^2 (t)$. This implies
    \begin{align*}
     V_1 (t) & = \lambda_1^i e_i^1 (t)\\
    & = \begin{bmatrix}
        \lambda_1^1 & \dots & \lambda_1^n
    \end{bmatrix} \cdot \begin{bmatrix}
        a_{11}(t) & \dots & a_{1n} (t)\\
        \vdots & & \vdots\\
        a_{n1}(t) & \dots & a_{nn}(t)
    \end{bmatrix} \cdot \begin{bmatrix}
        e_1^2 (t)\\
        \vdots\\
        e_n^2 (t)
    \end{bmatrix}\\
    & = \Vec{ \lambda_{1}} \cdot A \cdot \Vec{e}_1
    \end{align*}
    and it also implies $e_j^2 (t) = a^{ji}(t) e_i^1 (t)$ with
    \begin{equation*}
        V_2 (t) = \lambda_{2}^i e_i^2 (t) = \Vec{\lambda_{2}} \cdot A^{-1} \cdot \Vec{e}_1 (t)
    \end{equation*}
    Then we observe that
    \begin{align*}
        (A \cdot A^{-1})^\prime & = I^\prime\\
        A^\prime \cdot A^{-1} + A \cdot (A^{-1})^\prime & = 0\\
        (A^{-1})^\prime & = - A^{-1} \cdot A^\prime \cdot A^{-1}
    \end{align*}
    Then the formula for $D = A \cdot B \cdot C$ is 
    \begin{equation*}
        d_{ij}= \sum_{uv} a_{iu}b_{uv}c_{vj}
    \end{equation*}
    Then we have
    \begin{align*}
        \nabla_{e_1^1(t)}^{g_2} e_k^1(t) & = \nabla_{\gamma^\prime (t)}^{g_2} e_k^1 (t)\\
        & = \nabla_{\gamma^\prime (t)}^{g_2} a_{ku}(t) e_u^2 (t)\\
        & = \Dot{a}_{ku}(t) e_u^2 (t) + a_{ku}(t) \nabla_{\gamma^\prime (t)}^{g_2} e_u^2 (t)\\
        & = \Dot{a}_{ku}(t) e_u^2 (t)\\
        & = \Dot{a}_{ku}(t) a^{uj}(t) e_j^1 (t)\\
        a^{ik}(t) \nabla_{\gamma^\prime (t) }^{g_2} e_k^1 (t) & = a^{ik}(t) \Dot{a}_{ku}(t) a^{uj}(t) e_j^1 (t)\\
        & = - \Dot{a}^{ij}(t) e_j^1 (t)\\
    \end{align*}
    Then we must also have
    \begin{align*}
        \nabla_{\gamma^\prime (t)}^{g_1} e_i^2 (t) & = \nabla _{\gamma^\prime (t)}^{g_1} (a^{ij}(t)e_j^1 (t))\\
        & = \Dot{a}^{ij}(t)e_j^1 (t) + a^{ij}(t)  \nabla_{\gamma^\prime (t)}^{g_1} e_j^1 (t)\\
        & = \Dot{a}^{ij}(t)e_j^1 (t)
    \end{align*}
    Combining these two observations, we see
    \begin{align*}
        \nabla_{\gamma^\prime (t) }^{g_2} e_k^1 (t) & = - a_{ki}(t) \Dot{a}^{ij}(t) e_j^1 (t)\\
        & = - a_{ki}(t) ( \Dot{a}^{ij}(t) e_j^1 (t))\\
        & = - a_{ki}(t) \nabla_{\gamma^\prime (t)}^{g_1} e_i^2 (t)
    \end{align*}
    Which means that
    \begin{equation*}
        \nabla_{\gamma^\prime (t)}^{g_1} e_i^2 (t) = -  a^{ik}(t) \nabla_{ \gamma^\prime (t)}^{g_2} e_k^1 (t)
    \end{equation*}
    Then we see that
    \begin{align*}
        0 & = \nabla_{\gamma^\prime (t)}^{g_1} e_i^1 (t)\\
        & = \Dot{a}_{ij}(t) e_j^2 (t) - a_{ij}(t) a^{jk}(t) \nabla_{\gamma^\prime (t) }^{g_2} e_k^1 (t)\\
        & = \Dot{a}_{ij}(t) e_j^2 (t) - \delta_i^k \nabla_{\gamma^\prime (t) }^{g_2} e_k^1 (t)\\
        & = \Dot{a}_{ij}(t) e_j^2 (t) - \nabla_{\gamma^\prime (t) }^{g_2} e_i^1 (t) = \star
    \end{align*}
    Then because of the local frames $\{ X_j \}$, where $X_j (\gamma (t)) = e_j^1 (t)$, we have
    \begin{equation*}
        \nabla_{\gamma^\prime (t)}^{g_2} e_i^1 (t)= \nabla_{e_1^1 (t)}^{g_2} e_i^1 (t) = \nabla_{X_1}^{g_2} X_i (\gamma (t)) = \Gamma_{1i}^{k,g_2} (\gamma (t)) X_k (\gamma (t)) = \Gamma_{1i}^{k,g_2} (\gamma (t)) e_k^1 (t)
    \end{equation*}
    Then along $\gamma$, we have
    \begin{align*}
        \nabla_{\gamma^\prime (t) }^{g_2} e_i^1 (t) & = \Gamma_{1i}^{k,g_2} (\gamma (t)) e_k^1 (t)\\
        & = \Gamma_{1i}^{k,g_2} (\gamma (t)) \big(  a_{kj}(t) e_j^2 (t) \big)
    \end{align*}
    Then, $\star$ now gives us
    \begin{equation*}
        \star = \Dot{a}_{ij}(t) e_j^2 (t) - \Gamma_{1i}^{k,g_2} (\gamma (t)) \big(  a_{kj}(t)e_j^2 (t) \big)
    \end{equation*}
    This means that
    \begin{equation*}
        \Dot{a}_{ij}(t)= \Gamma_{1i}^{k,g_2} (\gamma (t)) a_{kj}(t)
    \end{equation*}
    Then going further, we have
    \begin{align*}
        \Dot{a}^{ij}(t) & = -  a^{iu}(t) \Dot{a}_{uv}(t)a^{vj}(t)\\
        & = -   a^{iu}(t) \big(  \Gamma_{1u}^{k,g_2} (\gamma (t)) a_{kv}(t) \big) a^{vj}(t)\\
        & = -   a^{iu}(t) \big(  \Gamma_{1u}^{k,g_2} (\gamma (t)) \delta_k^j \big)\\
        & = -   a^{iu}(t) \Gamma_{1u}^{j,g_2} (\gamma (t))
    \end{align*}
    Then we have
    \begin{equation*}
        \Dot{a}^{ij}(t) = -   a^{ik}(t) \Gamma_{1k}^{j,g_2} (\gamma (t))
    \end{equation*}
    Which gives us
    \begin{align*}
        V_2^\prime \bigg|_{g_1} (t) & = \sum_i \nabla_{\gamma^\prime (t)}^{g_1} (\lambda_{2}^i e_i^2 (t))\\
        & = \sum_i \nabla_{\gamma^\prime (t)}^{g_1} (\lambda_{2}^i a^{ij}(t) e_j^1 (t))\\
        & = \sum_i \lambda_{2}^i \Dot{a}^{ij}(t) e_j^1 (t) + \lambda_{2}^i a^{ij}(t) \nabla_{\gamma^\prime (t)}^{g_1}  e_j^1 (t)\\
        & = \sum_i \lambda_{2}^i \Dot{a}^{ij}(t) e_j^1 (t)\\
        & = - \sum_i \lambda_{2}^i \big( \sum_{k} a^{ik}(t) \Gamma_{1k}^{j,g_2} (\gamma (t)) \big) e_j^1 (t)
    \end{align*}
    Since this computation of $V_2^\prime \big|_{g_1}(t)$ is independent of any coordinate chart, fix $t_0 \in [0,1]$ and choose the normal coordinate system about the point $\gamma (t_0)$ with respect to $g_1$. Since $|e_i^2|_{g_1(\gamma (t))} = \sqrt{ \sum_j (a^{ij}(t))^2}$, from proposition~\ref{prop:neat}, we have a $\delta_0$ such that $\parallel g_1 - g_2 \parallel_{C^1 (M,g_1)}< \delta_0$ implies that along $\gamma$
    \begin{align*}
        \sqrt{1 - \delta_0} \ |e_i^2 (t)|_{g_1 (\gamma (t))} & < |e_i^2 (t)|_{g_2 (\gamma (t))} = 1\\
        |a^{ij}(t) e_j^1 (t)|_{g_1 (\gamma (t))} & < \frac{1}{\sqrt{1 - \delta_0}} < 2\\
        \sqrt{\sum_{j=1}^n (a^{ij}(t))^2} & < 2
    \end{align*}
    That means that
    \begin{equation*}
        |a^{ij}(t)| = \sqrt{(a^{ij}(t))^2} \leq \sqrt{ \sum_{j=1}^n (a^{ij}(t))^2} < 2
    \end{equation*}
    Then $|a^{ij}(t)| < 2$. Then, at $t_0$, we have
    \begin{align*}
     |V_2^\prime \bigg|_{g_1} (t_0)|_{g_1 (\gamma (t_0))}^2 & =  |- \sum_i \lambda_{2}^i \big( \sum_{k} a^{ik}(t_0) \Gamma_{1k}^{j,g_2} (\gamma (t_0)) \big) e_j^1 (t_0)  |_{g_1 (\gamma (t_0))}^2\\
     & = \sum_i (\lambda_2^i)^2 \big( \sum_{k} a^{ik}(t_0) \Gamma_{1k}^{j,g_2} (\gamma (t_0)) \big)^2\\
     & < \sum_i (\lambda_2^i)^2 \big( \sum_{k} 2 \Gamma_{1k}^{j,g_2} (\gamma (t_0)) \big)^2 = \star \star
    \end{align*}
    From here, let $\delta (L, \ \epsilon)= \min \{ \frac{1}{2}, \delta_0, \frac{\epsilon}{12n^{\frac{5}{2}}L} \}$. Then if $\parallel g_1 - g_2 \parallel_{C^1} < \delta (L, \epsilon)$, lemma~\ref{lemma:ineq0}  gives us $|\Gamma_{ij}^{k,g_2} (\gamma (t_0))|< 3n \delta (L, \epsilon) < \frac{\epsilon}{4n^{\frac{3}{2}}L}$ for all $1 \leq i,j,k \leq n$ along $\gamma (t)$, and 
    \begin{align*}
        \star \star & < \sum_i L^2 \big( \sum_k 2 \frac{\epsilon}{4n^{\frac{3}{2}}L} \big)^2\\
        & = \sum_i L^2 \big( \frac{\epsilon}{2n^{\frac{1}{2}}L} \big)^2 = \frac{\epsilon^2 }{4}
    \end{align*}
    Then, under our normal coordinate system at $t_0$, we have $|V_2^\prime \big|_{g_1} (t_0)|_{g_1 (\gamma (t_0))}^2 <  \frac{\epsilon^2}{4}$, in which the bound is independent of the coordinate chart. This means that under any coordinate chart, $|V_2^\prime \big|_{g_1} (t_0)|_{g_1 (\gamma (t_0))}^2 <  \frac{\epsilon^2}{4}$, and since $t_0 \in [0,1]$ was arbitrary, it holds for all $t \in [0,1]$.

Let $W(t):=V_1(t)-V_2(t)$ and $u(t):=|W(t)|_{g_1(\gamma(t))}^2$. Since $W(0)=v-v=0$, we have $u(0)=0$. By Cauchy-Schwarz,
\begin{equation*}
    u^\prime (t) = 2 g_1 (W^\prime (t), W(t))_{(\gamma(t))} \leq 2 |W^\prime (t)|_{g_1(\gamma(t))} |W(t)|_{g_1(\gamma(t))} = 2|W^\prime(t)|_{g_1(\gamma(t))} \sqrt{u(t)}
\end{equation*}
Since each $e_i^1(t)$ is $g_1$-parallel along $\gamma$, we have $V_1^\prime\big|_{g_1}(t)=0$, so $W^\prime(t) = -V_2^\prime\big|_{g_1}(t)$, and by our bound above, $|W^\prime(t)|_{g_1(\gamma(t))} < \frac{\epsilon}{2}$. Then
\begin{equation*}
    u^\prime (t) < \epsilon \sqrt{u(t)}
\end{equation*}
for all $t \in [0,1]$. Since $u$ is continuous on the compact interval $[0,1]$, let $u_{\max}:= \max_{t \in [0,1]} u(t)$, achieved at some $t_1 \in [0,1]$. For all $s \in [0,t_1]$, we have $u^\prime(s) < \epsilon \sqrt{u(s)} \leq \epsilon \sqrt{u_{\max}}$. Integrating from $0$ to $t_1$ and using $u(0)=0$, gives us
\begin{equation*}
    u_{\max} = u(t_1) = u(t_1) - u(0)= \int_0^{t_1} u^\prime (s) \ ds < \int_0^{t_1} \epsilon \sqrt{u_{\max}} \ ds = \epsilon \sqrt{u_{\max}} \ t_1 \leq \epsilon \sqrt{u_{\max}}
\end{equation*}
If $u_{\max} > 0$, dividing both sides by $\sqrt{u_{\max}}$ gives $\sqrt{u_{\max}} < \epsilon$, so $u_{\max} < \epsilon^2$; this also holds trivially if $u_{\max}=0$. Then
\begin{equation*}
    |u_1 - u_2|_{g_1 (q)}^2 = u(1) \leq u_{\max} < \epsilon^2
\end{equation*}
so $|u_1 - u_2|_{g_1(q)} < \epsilon$, as desired.

\end{proof}

\subsection{Proof of Proposition \ref{prop:prop1}}

\phantomsection\label{app:prop-prop1}
\begin{proof}[Proof of Proposition \ref{prop:prop1}]

Fix our neighborhood $U$ (recall from remark\ref{def:rmk1}, that there's a chart $V$ such that $\overline{U} \subset V$), and choose $0 < \epsilon < 1$.

\begin{equation*}
    \parallel \cdot \parallel_{C^0 (M, g_1)} \leq  \parallel \cdot \parallel_{C^1 (M, g_1)} \leq  \parallel \cdot \parallel_{C^2 (M, g_1)}
\end{equation*}
We'll use $\parallel \cdot \parallel_{C^k (M,g_1)}$ for $k = 0,1,2$ and specify which $k$ we're using (so that we can re-use it later). Also, we see that our $\delta (M,g_1, U, k, \epsilon) = C_\alpha (M, g_1 , U, k) \epsilon$. So, we'll be trying to identify $C_\alpha (M, g_1, U,k)$.

Since we're trying to obtain local control over $U$, we can compare the metric components directly via the Euclidean metric. So, for $k=0,1,2$, we have that
\begin{equation*}
    \parallel g_1 - g_2 \parallel_{C^k (U,h)}^2 = \sum_{ij} (g_{1ij} - g_{2 ij})^2 + \sum_{1 \leq |\alpha| \leq k} \sum_{ij} (\partial^\alpha g_{1 ij} - \partial^\alpha g_{2 ij})^2
\end{equation*}
where $h$ is the Euclidean metric from definition~\ref{def:eucl}. Then from lemma~\ref{lemma:ineq1}, we have
\begin{equation*}
    |g_{1ij} - g_{2 ij}| \leq \parallel g_1 - g_2 \parallel_{C^k (U,h)} \leq C_0 \parallel g_1 - g_2 \parallel_{C^k (M, g_1)}
\end{equation*}
We'll use this to bound $|g_2^{ij}| \leq C_1$ and show that
\begin{equation*}
    |g_1^{ij} - g_2^{ij}| \leq C_2 \parallel g_1 - g_2 \parallel_{C^k (M, g_1)} 
\end{equation*}
where $C_1$ and $C_2$ are based on the manifold $M$, the metric $g_1$, the neighborhood $U$, and $k$.

First, we'll think of $g_1$ and $g_2$ as matrices: $ g_1 = (g_{1ij})$ and $g_2 = (g_{2ij})$. Their inverse matrices are $g_1^{-1} = (g_1^{ij})$ and $g_2^{-1} = (g_2^{ij})$. Let $G_1^{-1} = \max_{p \in \overline{U}} \parallel g_1^{-1} (p) \parallel_F$ where $\parallel \cdot \parallel_F$ is the Frobenius norm (defined in observation~\ref{def:obs1}), so that  $\parallel g_1^{-1} (p) \parallel_F \leq G_1^{-1}$ for all $p \in \overline{U}$. Take the matrix $K= g_2 - g_1$, then $g_2 = g_1 +K$, and we have
\begin{equation*}
    \begin{split}
        g_2 & = g_1 (I +g_1^{-1}K)\\
        g_2^{-1} & = (I + g_1^{-1}K)^{-1} g_1^{-1}
    \end{split}
\end{equation*}
Now, let $C_\alpha = \min \{ \frac{1}{2 C_0 G_1^{-1}}, 1 \}$ (we'll add more constraints as we go), so that once $\parallel g_1 - g_2 \parallel_{C^k (M,g_1)}< \delta = C_\alpha \epsilon$, we have $\parallel g_1 - g_2 \parallel_{C^k (M,g_1)} < \frac{1}{2 C_0 G_1^{-1}}$. Then, from observation~\ref{def:obs1}, the Cauchy-Schwarz inequality, and lemma~\ref{lemma:ineq1}, we have
\begin{align*}
    \parallel g_1^{-1} K \parallel_F \leq \parallel g_1^{-1} \parallel_F \parallel K \parallel_F & \leq C_0 G_1^{-1} \parallel K \parallel_{C^k (M, g_1)}\\
    & = C_0 G_1^{-1} \parallel g_1 - g_2 \parallel_{C^k (M, g_1)}\\
    & < C_0 G_1^{-1} \frac{1}{2 C_0 G_1^{-1}} = \frac{1}{2} <1
\end{align*}
This means that from lemma 2.3.3 (p.58,~\cite{golub1996matrix}) we can use the geometric series expansion for $(I + g_1^{-1}K)^{-1}$. We have
\begin{align*}
    \parallel (I + g_1^{-1}K)^{-1} \parallel_F & = \parallel  \sum_{i=0}^\infty (-1)^i (g_1^{-1}K)^i \parallel_F\\
    & \leq  \sum_{i=0}^\infty \parallel (g_1^{-1}K) \parallel_F^i\\
    & < \sum_{i=0}^\infty \big( \frac{1}{2} \big)^i\\
    & = 2
\end{align*}
Then,
\begin{align*}
    |g_2^{ij}| = \sqrt{(g_2^{ij})^2} \leq \parallel g_2^{-1} \parallel_F & = \parallel   (I + g_1^{-1}K)^{-1} g_1^{-1} \parallel_F\\
    & \leq \parallel g_1^{-1} \parallel_F \parallel (I + g_1^{-1}K)^{-1} \parallel_F\\
    & \leq G_1^{-1} 2\\
    & = C_1
\end{align*}
where $C_1 = 2 G_1^{-1}$, which is based on the manifold $M$, the metric $g_1$, the neighborhood $U$, and $k$ for $k=0,1,2$.

Now, for the matrix $g_1^{-1} - g_2^{-1}$, used to bound $|g_1^{ij} - g_2^{ij}|$, we have
\begin{equation*}
    g_1^{-1} - g_2^{-1} = (g_1^{-1}g_2 g_2^{-1} - g_1^{-1}g_1 g_2^{-1}) = g_1^{-1} (g_2 - g_1) g_2^{-1}
\end{equation*}
Then, using this in our Frobenius norm, we have
\begin{equation*}
    \begin{split}
        |g_1^{ij} - g_2^{ij}| & \leq \parallel g_1^{-1} -g_2^{-1}\parallel_F\\
        & = \parallel  g_1^{-1} (g_2 - g_1) g_2^{-1} \parallel_F\\
        & \leq \parallel  g_1^{-1} \parallel_F \parallel g_2 - g_1 \parallel_F \parallel g_2^{-1} \parallel_F\\
        & \leq \parallel  g_1^{-1} \parallel_F \parallel g_2 - g_1 \parallel_F C_1\\
        & \leq C_0 C_1 G_1^{-1} \parallel g_2 - g_1 \parallel_{C^k (M, g_1)}\\
        & = C_2 \parallel g_2 - g_1 \parallel_{C^k (M, g_1)}
    \end{split}
\end{equation*}
where $C_2 = C_0 C_1 G_1^{-1}$, which is based on the manifold $M$, the metric $g_1$, the neighborhood $U$, and $k$ for $k=0,1,2$.

Now, we'll let $k=1,2$, and we see that from lemma~\ref{lemma:ineq1}, we have

\begin{equation*}
    |\partial_k g_{1ij} - \partial_k g_{2ij}|  \leq C_0 \parallel g_1 - g_2 \parallel_{C^k (M, g_1)}
\end{equation*}
Thus, we want to show that $|\Gamma_{ij}^{k, g_1} - \Gamma_{ij}^{k, g_2} | \leq C_3 \parallel g_1 - g_2 \parallel_{C^k (M, g_1)}$, where $C_3 >0$ is based on the manifold $M$, the metric $g_1$, the neighborhood $U$, and $k$ where $k=1,2$.

Then, we have

\begin{align*}
    |\Gamma_{ij}^{k, g_1} - \Gamma_{ij}^{k, g_2} | & = | \frac{1}{2} g_1^{lk} (\partial_i g_{1jl} + \partial_j g_{1li} - \partial_l g_{1ij}) -  \frac{1}{2} g_2^{lk} (\partial_i g_{2jl} + \partial_j g_{2li} - \partial_l g_{2ij})|\\
    & \leq | \frac{1}{2}  g_1^{lk} (\partial_i g_{1jl} + \partial_j g_{1li} - \partial_l g_{1ij}) - \frac{1}{2} g_2^{lk} (\partial_i g_{1jl} + \partial_j g_{1li} - \partial_l g_{1ij})|\\
    & + |\frac{1}{2} g_2^{lk} (\partial_i g_{1jl} + \partial_j g_{1li} - \partial_l g_{1ij}) - \frac{1}{2} g_2^{lk} (\partial_i g_{2jl} + \partial_j g_{2li} - \partial_l g_{2ij})|\\
    & \leq \frac{1}{2} |g_1^{lk} - g_2^{lk}| (|\partial_i g_{1jl}|+|\partial_j g_{1il}|+|\partial_l g_{1ij}|)\\
    & +\frac{1}{2}|g_2^{lk}|( |\partial_i g_{1jl} - \partial_i g_{2jl}| + |\partial_j g_{1il} - \partial_j g_{2il}| + |\partial_l g_{1ij} - \partial_l g_{2ij}| )\\
    & \leq \frac{3}{2} \max |\partial g_1| |g_1^{lk} - g_2^{lk}| + \frac{3}{2} C_1 |\partial_l g_{1ij} - \partial_l g_{2ij}|\\
    & \leq  \frac{3}{2} \max |\partial g_1| C_2 \parallel g_1 - g_2 \parallel_{C^k (M, g_1)} + \frac{3}{2} C_1 C_0 \parallel g_1 - g_2 \parallel_{C^k (M, g_1)}\\
    & = (\frac{3}{2} \max |\partial g_1| C_2 + \frac{3}{2} C_1 C_0) \parallel g_1 - g_2 \parallel_{C^k (M, g_1)}\\
    & = C_3 \parallel g_1 - g_2 \parallel_{C^k (M, g_1)}
\end{align*}
where $C_3 = \frac{3}{2} \max |\partial g_1| C_2 + \frac{3}{2} C_1 C_0$, which is based on the manifold $M$, the metric $g_1$, the neighborhood $U$, and $k$ where $k=1,2$. Now, we'll use this information to add to our constraints on $C_\alpha = \min \{ \frac{1}{2 C_0 G_1^{-1}}, \frac{1}{C_3}, 1  \} $, and our $\delta (M, g_1, U, k, \epsilon) = C_\alpha \epsilon$. Then,
\begin{equation*}
    \parallel g_1 - g_2 \parallel_{C^k (M, g_1)} <  \delta (M, g_1, U, k, \epsilon)
\end{equation*}
implies that
\begin{equation*}
    |\Gamma_{ij}^{k, g_1} - \Gamma_{ij}^{k, g_2} | < C_3 \parallel g_1 - g_2 \parallel_{C^k (M, g_1)} < C_3 C_\alpha \epsilon \leq \epsilon
\end{equation*}

Now, choose $\epsilon \geq 1$, then let $\delta (M, g_1,U,k, \epsilon) = \delta_1 (M,g_1,U,k, \epsilon)$ where $\parallel g_1 - g_2 \parallel_{C^k (M,g_1)} < \delta_1 (M,g_1,U,k, \epsilon)$ implies that on $U$, we have
\begin{equation*}
    |\Gamma_{ij}^{k, g_1} - \Gamma_{ij}^{k, g_2}| < \frac{1}{2}
\end{equation*}
From above, we know that such a $\delta_1 (M,g_1,U,k, \epsilon)$ exists. Then we see that $\parallel g_1 - g_2 \parallel_{C^k (M,g_1)} < \delta (M, g_1,U,k, \epsilon)$ implies that on $U$, we have
\begin{equation*}
    |\Gamma_{ij}^{k, g_1} - \Gamma_{ij}^{k, g_2}| < \frac{1}{2} < \epsilon
\end{equation*}
Thus, for all $\epsilon >0$, we have our desired result.    
\end{proof}

\subsection{Proof of Theorem \ref{def:thrm3}}

\phantomsection\label{app:thrm3}
\begin{proof}[Proof of Theorem \ref{def:thrm3}]
    Since geodesics are the solutions to the differential equation

    \begin{equation*}
    \frac{d^2}{dt^2}\gamma_m  + \sum_{ij} \Gamma_{ij}^{k, g_m} (\gamma_m (t)) \dot{\gamma}_m^i \dot{\gamma}_m^j =0
    \end{equation*}
    we can use lemma~\ref{def:example2} where $f_m (t) = \gamma_m (t)$, $\lambda_{mij}^k (q) = \Gamma_{ij}^{k,g_m} (q)$ for $q \in U$, and $y_m (t) = 
    \begin{bmatrix}
    \gamma_m (t)\\
    \gamma_m^\prime (t)
    \end{bmatrix}$.

    Then, we have that for all $p \in U$, and for $v \in T_pM$, there exists constants $A_{g_1},B_{g_1} >0$, such that

    \begin{equation}\label{def:eq11}
        A_{g_1} |v| \leq |v|_{g_1 (p)} \leq B_{g_1} |v|
    \end{equation}
    where $|\cdot|$ is the Euclidean norm from definition~\ref{def:eucl}. Then, from proposition~\ref{prop:neat} for $\parallel g_1 - g_2 \parallel_{C^k (M, g_1)} < \delta < \frac{1}{2}$, we have
    \begin{equation*}
        |v|_{g_1 (p)} \sqrt{1 - \delta} \leq |v|_{g_2 (p)} \leq \sqrt{1+\delta}|v|_{g_1 (p)}
    \end{equation*}
    which gives us
    \begin{equation*}
        |v| A_{g_1} \sqrt{1- \delta} \leq |v|_{g_2 (p)} \leq \sqrt{1+ \delta} B_{g_1} |v|
    \end{equation*}
    Since $\gamma_m : [0,1] \to M$ is a geodesic with respect to $g_m$, we have that $|\gamma_m^\prime (t)|_{g_m (\gamma_m (t))} =c_m$ from~\cite{do1992riemannian}, p.61. Since $L_{g_m} (\gamma_m) < L$, we have

    \begin{equation*}
        c_m = c_m (1-0) = \int_0^1 c_m dt = \int_0^1 |\gamma_m^\prime (t)|_{g_m (\gamma_m (t))} dt = L_{g_m}(\gamma_m) < L
    \end{equation*}

    Then $|\gamma_m^\prime (t)|_{g_m (\gamma_m (t))} < L $, and we have

    \begin{equation}\label{def:eq12}
        \begin{split}
            |\Dot{\gamma}_1^i (t)| & \leq |\gamma_1^\prime (t)| \leq \frac{1}{A_{g_1}} |\gamma_1^\prime (t)|_{g_1 (\gamma_1 (t))} \leq \frac{ L}{A_{g_1}}\\
            |\Dot{\gamma}_2^i (t)| & \leq |\gamma_2^\prime (t)| \leq \frac{2}{A_{g_1}} |\gamma_2^\prime (t)|_{g_2 (\gamma_2 (t))} \leq \frac{2 L}{A_{g_1}}\\
        \end{split}
    \end{equation}
     then $|\gamma_m^\prime (t)| < \frac{2L}{A_{g_1}}$, so let $C_2 = \frac{2L}{A_{g_1}}$. Then since $M$ is compact we can let $C_1 = \max |\Gamma^{g_1}|$ where $\max |\Gamma^{g_1}|$ is from remark~\ref{def:rmk1}. Additionally, since $\Gamma^{g_1}$ is smooth on $\overline{U}$, which is compact, it has bounded derivatives, and therefore has a Lipschitz constant $\max |\partial \Gamma^{g_1}|$ which was defined in remark~\ref{def:rmk1}. So, let $C_0 =   \max |\partial \Gamma^{g_1}|$. We have
    \begin{equation*}
        |\Gamma_{ij}^{k, g_1} (\gamma_1 (t)) - \Gamma_{ij}^{k, g_1} (\gamma_2 (t))| \leq C_0 |\gamma_1 (t) - \gamma_2 (t)|
    \end{equation*}
    From here, we'll suppose that
    \begin{equation*}
        |\gamma_1^\prime (0) - \gamma_2^\prime (0)|_{g_1 (p)} < \zeta_1 = \delta_2 (M,g_1, U,L, \epsilon)
    \end{equation*}
    Then we see that
    \begin{equation*}
        |\gamma_1^\prime (0) - \gamma_2^\prime (0)| \leq \frac{1}{A_{g_1}}|\gamma_1^\prime (0) - \gamma_2^\prime (0)|_{g_1 (p)} < \frac{\zeta_1}{A_{g_1}}
    \end{equation*}

    Then from proposition~\ref{prop:prop1}, we know that there exists $\delta_1 (M, g_1, U,k, \epsilon)>0$ such that $\parallel g_1 - g_2 \parallel_{C^k (M, g_1)} < \delta_1 (M, g_1, U,k, \epsilon) < \frac{1}{2}$ implies that
    \begin{equation*}
        |\Gamma_{ij}^{k,g_1} - \Gamma_{ij}^{k,g_2}| < \zeta_2
    \end{equation*}
    on $U$. Then from lemma~\ref{def:example2} (with $\zeta = \zeta_2$), we'll have that
    \begin{align*}
    |y_1 (t) - y_2 (t)| & \leq Be^A\\
    & = (|y_1 (0) - y_2 (0)| + n^3 C_2^2 \zeta_2)e^A = \star
    \end{align*}
    where $A = 1 + \frac{4n^3L^2C_0}{A_{g_1}^2} + \frac{4n^3 LC_1 }{A_{g_1}}$ and $B= |y_1 (0) - y_2 (0)| + \frac{4n^3 L^2}{A_{g_1}^2} \zeta_2$. We also note that $\gamma_1 (0) = p = \gamma_2 (0)$, so $|\gamma_1 (0) - \gamma_2 (0)|=0$. Then we have
    \begin{equation*}
        \star < (|\gamma_1^\prime (0) - \gamma_2^\prime (0)| + \frac{4n^3 L^2}{A_{g_1}^2} \zeta_2)e^A 
    \end{equation*}
    What this means for us that since $y_1 (t) - y_2 (t) = \begin{bmatrix}
        \gamma_1 (t) - \gamma_2 (t)\\
        \gamma_1^\prime (t) - \gamma_2^\prime (t)
    \end{bmatrix}$
    then we have
    \begin{equation}\label{eq:eq2}
         |\gamma_1 (t) - \gamma_2 (t)| \leq |y_1 (t) - y_2(t)| <  \frac{\zeta_1 e^A}{A_{g_1}} + \frac{4n^3L^2 e^A}{A_{g_1}^2} \zeta_2
    \end{equation}
    and we have
    \begin{equation}\label{def:eq3}
        |\gamma_1^\prime (t) - \gamma_2^\prime (t)| \leq |y_1 (t) - y_2(t)| <  \frac{\zeta_1 e^A}{A_{g_1}} + \frac{4n^3L^2 e^A}{A_{g_1}^2} \zeta_2        
    \end{equation}
    Then let $\zeta_1 = \min \{ \frac{ A_{g_1} \epsilon}{2B_{g_1}e^A}, \frac{ A_{g_1} \epsilon}{2e^A} \}$ and $\zeta_2 = \min \{ \frac{A_{g_1}^2  \epsilon}{8B_{g_1}n^3 L^2 e^A}, \frac{A_{g_1}^2  \epsilon}{8n^3 L^2 e^A} \}$, where $B_{g_1}$ comes from equation~\ref{def:eq11}. Since $\zeta_2$ is also based on $L$, our $\delta_1$ will reflect that in its criteria, i.e. there exists $\delta_1 (M,g_1,U,k,L, \epsilon)>0$, such that $\parallel g_1 - g_2 \parallel_{C^k (M, g_1)} < \delta_1 (M,g_1,U,k,L, \epsilon)$ implies that

    \begin{align*}
        d_{g_1} (\gamma_1 (t), \gamma_2 (t)) \leq B_{g_1} |\gamma_1 (t) - \gamma_2 (t)| < B_{g_1} \big( \frac{\epsilon}{2 B_{g_1}} + \frac{\epsilon}{2 B_{g_1}} \big) = \epsilon
    \end{align*}
    and we have
    \begin{align*}
        |\gamma_1^\prime (t) - \gamma_2^\prime (t)| & \leq |y_1 (t) - y_2(t)|\\
        & <  \frac{\zeta_1 e^A}{A_{g_1}} + \frac{4n^3L^2 e^A}{A_{g_1}^2} \zeta_2\\
        & \leq \frac{e^A}{A_{g_1}} \cdot \frac{ A_{g_1} \epsilon}{2e^A}  + \frac{4n^3L^2 e^A}{A_{g_1}^2} \cdot \frac{A_{g_1}^2  \epsilon}{8n^3 L^2 e^A}\\
        & = \frac{\epsilon}{2} + \frac{\epsilon}{2} = \epsilon
    \end{align*}
    as desired.
\end{proof}

\subsection{Proof of Theorem \ref{def:thrm5}}

\phantomsection\label{app:thrm5}
\begin{proof}[Proof of Theorem \ref{def:thrm5}]
    Choose $0< \epsilon$. Let $\{ X_l \}$ be the coordinate frame affiliated with $U$, and let $V_m (t)$ be the parallel transport of $v$ along $\gamma_m$ with respect to $g_m$ for $m=1,2$. Then we have
\begin{equation*}
    V_m (t) = \sum_k V_m^k X_k (\gamma_m (t))
\end{equation*}
where $V_m^k (t)$ for $k = 1, \dots , n$ is the solution to the system of homogeneous linear differential equations

\begin{equation*}
    0 = \dot{V}_m^k + \sum_{ij} V_m^j \dot{\gamma}_m^i \Gamma_{ij}^{k, g_m} (\gamma_m (t))
\end{equation*}
Then we see that

\begin{equation*}
    |P_{g_1}^{\gamma_1} (v) - P_{g_2}^{\gamma_2} (v)|_{g_1 (q)} = |V_1 (1) - V_2 (1)|_{g_1 (q)}
\end{equation*}

From here, we'll use lemma~\ref{def:example1} to establish our bounds, where $f_m (t) = V_m (t)$ and $\lambda_{mj}^k (t) = \sum_i \dot{\gamma}_m^i \Gamma_{ij}^{k,g_m}(\gamma_m (t))$.

Then for $\lambda_{1j}^k (t) = \sum_i \dot{\gamma}_1^i \Gamma_{ij}^{k, g_1} (\gamma_1 (t))$ and from equation~\ref{def:eq12}, we have
\begin{align*}
    |\dot{\gamma}_1^i| & \leq \frac{L}{A_{g_1}}\\
    |\Gamma_{ij}^{k,g_1}(\gamma_1 (t))| & \leq \max |\Gamma^{g_1}|
\end{align*}
where $\max |\Gamma^{g_1}|$ was defined in remark~\ref{def:rmk1}. So, let $C_0 = \frac{nL \max |\Gamma^{g_1}|}{A_{g_1}}$.

Next we observe that since parallel transports are isometries, we have
\begin{equation*}
    |V_1 (t)|_{g_1 (\gamma_1 (t))} = |V_1 (0)|_{g_1 (p)}= |v |_{g_1 (p)} <L
\end{equation*}
If we introduce the Euclidean norm from definition~\ref{def:eucl} and equation~\ref{def:eq12}, this means that
\begin{equation*}
    |V_1^k (t)| \leq |V_1 (t)| \leq \frac{1}{A_{g_1}} |V_1 (t)|_{g_1 (\gamma_1 (t))} < \frac{L}{A_{g_1}}
\end{equation*}
Then, let $C_1 = \frac{L}{A_{g_1}}$.

Next, by combining what we know from proposition~\ref{prop:prop1} and theorem~\ref{def:thrm3}, there exists $\delta_1 (M, g_1, U, k, L, \epsilon) >0$ and $\delta_2 (M, g_1, U, L, \epsilon)>0$ such that
\begin{align*}
    \parallel g_1 - g_2 \parallel_{C^k (M, g_1)} & < \delta_1 (M, g_1, U, k, L, \epsilon)\\
    |\gamma_1^\prime (0) - \gamma_2^\prime (0)|_{g_1 (p)} &< \delta_2 (M, g_1, U, L, \epsilon)
\end{align*}
implies that on $U$, we have the following bounds (which hold at $\gamma_1 (t)$ and $\gamma_2 (t)$ since both lie in $U$ for all $t \in [0,1]$):
\begin{align*}
    |\gamma_1^\prime (t) - \gamma_2^\prime (t)| & < \zeta_1\\
    |\gamma_1 (t) - \gamma_2 (t)| & < \zeta_2\\
    |\Gamma_{ij}^{k, g_1} - \Gamma_{ij}^{k, g_2}| & < \zeta_3
\end{align*}

Then we see that
\begin{align*}
    |\dot{\gamma}_1^i \Gamma_{ij}^{k, g_1} (\gamma_1 (t)) - \dot{\gamma}_2^i \Gamma_{ij}^{k, g_2} (\gamma_2 (t))| & \leq |\dot{\gamma}_1^i \Gamma_{ij}^{k, g_1} (\gamma_1 (t)) - \dot{\gamma}_1^i \Gamma_{ij}^{k, g_1} (\gamma_2 (t))| \\
    & + |\dot{\gamma}_1^i \Gamma_{ij}^{k, g_1} (\gamma_2 (t)) - \dot{\gamma}_2^i \Gamma_{ij}^{k, g_1} (\gamma_2 (t))|\\
    & + |\dot{\gamma}_2^i \Gamma_{ij}^{k, g_1} (\gamma_2 (t)) - \dot{\gamma}_2^i \Gamma_{ij}^{k, g_2} (\gamma_2 (t))|\\
    & = I + II + III
\end{align*}
    
From the proof of theorem~\ref{def:thrm3} we have that $|\dot{\gamma}_m^i| \leq |\gamma_m^\prime (t)| \leq \frac{2L}{A_{g_1}}$. Then for $I$, we have

\begin{align*}
    I & = |\dot{\gamma}_1^i \Gamma_{ij}^{k, g_1} (\gamma_1 (t)) - \dot{\gamma}_1^i \Gamma_{ij}^{k, g_1} (\gamma_2 (t))|\\
    & = |\dot{\gamma}_1^i| |\Gamma_{ij}^{k, g_1} (\gamma_1 (t)) - \Gamma_{ij}^{k, g_1} (\gamma_2 (t))|\\
    & \leq \frac{2L}{A_{g_1}} \max |\partial \Gamma^{g_1}| |\gamma_1 (t) - \gamma_2 (t)|\\
    & < \frac{2L \max |\partial \Gamma^{g_1}|}{A_{g_1}}\zeta_2
\end{align*}
where $\max |\partial \Gamma^{g_1}|$ was defined in remark~\ref{def:rmk1}.

Then for $II$, we have
\begin{align*}
    II & = |\dot{\gamma}_1^i \Gamma_{ij}^{k, g_1} (\gamma_2 (t)) - \dot{\gamma}_2^i \Gamma_{ij}^{k, g_1} (\gamma_2 (t))|\\
    & = |\Gamma_{ij}^{k, g_1}(\gamma_2 (t))||\dot{\gamma}_1^i - \dot{\gamma}_2^i|\\
    & \leq \max |\Gamma^{g_1}| |\gamma_1^\prime (t) - \gamma_2^\prime (t)|\\
    & < \max |\Gamma^{g_1}| \zeta_1
\end{align*}
where $\max |\Gamma^{g_1}|$ was defined in remark~\ref{def:rmk1}.

Then for $III$, we have
\begin{align*}
    III & = |\dot{\gamma}_2^i \Gamma_{ij}^{k, g_1} (\gamma_2 (t)) - \dot{\gamma}_2^i \Gamma_{ij}^{k, g_2} (\gamma_2 (t))|\\
    & = |\dot{\gamma}_2^i| |\Gamma_{ij}^{k, g_1} (\gamma_2 (t)) - \Gamma_{ij}^{k, g_2} (\gamma_2 (t))|\\
    & < \frac{2L}{A_{g_1}} \zeta_3
\end{align*}

Then, by putting everything back together, we have
\begin{align*}
    |\dot{\gamma}_1^i \Gamma_{ij}^{k, g_1} (\gamma_1 (t)) - \dot{\gamma}_2^i \Gamma_{ij}^{k, g_2} (\gamma_2 (t))| & \leq I + II + III\\
    & < \frac{2L \max |\partial \Gamma^{g_1}|}{A_{g_1}}\zeta_2 + \max |\Gamma^{g_1}| \zeta_1 + \frac{2L}{A_{g_1}} \zeta_3
\end{align*}

So, let $C_2 = \frac{2nL \max |\partial \Gamma^{g_1}|}{A_{g_1}}\zeta_2 + n \max |\Gamma^{g_1}| \zeta_1 + \frac{2nL}{A_{g_1}} \zeta_3$. 

Finally, from lemma~\ref{def:example1}, if we let $A= n^2 (C_0 + C_2)$, and $B = |V_1 (0) - V_2(0)| + \frac{2n^3L^2 \max |\partial \Gamma^{g_1}|}{A_{g_1}^2}\zeta_2 +  \frac{n^3L \max |\Gamma^{g_1}|}{A_{g_1}} \zeta_1 + \frac{2n^3L^2}{A_{g_1}^2} \zeta_3 $, we'll have
\begin{equation*}
    |V_1 (t) - V_2 (t)| \leq Be^A 
\end{equation*}
Noting $|V_1 (0) - V_2 (0)|= |v-v|=0$, and expanding out $B$, we have
\begin{align*}
    |V_1 (t) - V_2 (t)| & \leq Be^A\\
    & =  (|V_1 (0) - V_2(0)| + \frac{2n^3L^2 \max |\partial \Gamma^{g_1}|}{A_{g_1}^2}\zeta_2 +  \frac{n^3L \max |\Gamma^{g_1}|}{A_{g_1}} \zeta_1 + \frac{2n^3L^2}{A_{g_1}^2} \zeta_3 )e^A\\
    & = \big(  \frac{2n^3L^2 \max |\partial \Gamma^{g_1}|}{A_{g_1}^2}\zeta_2 +  \frac{n^3L \max |\Gamma^{g_1}|}{A_{g_1}} \zeta_1 + \frac{2n^3L^2}{A_{g_1}^2} \zeta_3 \big) e^A
\end{align*}
    Let $\zeta_1 = \frac{ A_{g_1} \epsilon}{3 B_{g_1}n^3 L \max |\Gamma^{g_1}|e^A}$, $\zeta_2 = \frac{ A_{g_1}^2 \epsilon}{6B_{g_1}n^3L^2 \max |\partial \Gamma^{g_1}| e^A}$, and $\zeta_3 = \frac{ A_{g_1}^2 \epsilon}{6B_{g_1}n^3 L^2 e^A}$. Then we see that
\begin{align*}
          |P_{g_1}^{\gamma_1} (v) - P_{g_2}^{\gamma_2} (v)|_{g_1 (q)} & = |V_1 (1) - V_2 (1)|_{g_1 (q)}\\
          & \leq B_{g_1} |V_1 (1) - V_2 (1)|\\
          & < B_{g_1} (\frac{\epsilon}{3B_{g_1}} + \frac{\epsilon}{3B_{g_1}} + \frac{\epsilon}{3B_{g_1}}) = \epsilon
\end{align*}

Now for the second part of the theorem, since $|v_1|_{g_1 (p)}, |v_2|_{g_1 (p)} <L$, we know that there exists $\delta_1 (M, g_1, U, k, L, \epsilon)>0$ and $\delta_2 (M, g_1, U, L, \epsilon)>0$, such that

    \begin{equation*}
    \begin{split}
      \parallel g_1 - g_2 \parallel_{C^k (M, g_1)} & < \delta_1 (M, g_1, U, k, L, \epsilon)\\
      |\gamma_1^\prime (0) - \gamma_2^\prime (0)|_{g_1 (p)} & < \delta_2 (M, g_1, U, L, \epsilon)
    \end{split}
    \end{equation*}
    implies that we have
    \begin{equation*}
        |P_{g_1}^{\gamma_1} (v_2) - P_{g_2}^{\gamma_2} (v_2) |_{g_1 (q)} <  \frac{\epsilon}{2}
    \end{equation*}
Let $\delta_3 = \frac{\epsilon}{2}$. Then $|v_1 - v_2|_{g_1 (p)} < \delta_3 = \frac{\epsilon}{2}$ implies that

\begin{align*}
    |P_{g_1}^{\gamma_1} (v_1) - P_{g_2}^{\gamma_2} (v_2) |_{g_1 (q)} & \leq |P_{g_1}^{\gamma_1} (v_1) - P_{g_1}^{\gamma_1} (v_2) |_{g_1 (q)} + |P_{g_1}^{\gamma_1} (v_2) - P_{g_2}^{\gamma_2} (v_2) |_{g_1 (q)}\\
    & = |v_1 - v_2 |_{g_1 (p)} + |P_{g_1}^{\gamma_1} (v_2) - P_{g_2}^{\gamma_2} (v_2) |_{g_1 (q)}\\
    & < \frac{\epsilon}{2} + \frac{\epsilon}{2} = \epsilon
\end{align*}

as desired.
 
\end{proof}

\section{Proofs for Section~\ref{sec:C2}}
In this section, we provide the proofs of the statements in section~\ref{sec:C2}.

\subsection{Proof of Proposition \ref{def:prop2}}

\phantomsection\label{app:prop2}
\begin{proof}[Proof of Proposition \ref{def:prop2}]
    First, fix $U$, because of how we set up the proposition~\ref{prop:prop1}, for $k=2$, we can bring in all of the infrastructure from that proposition where we'll omit $k$. We'll also assume that $\parallel g_1 - g_2 \parallel_{C^2 (M, g_1)} <  \delta (M, g_1, U, \epsilon) \leq 1$, and we'll add more constraints to our final $\delta (M, g_1, U, \epsilon)$. Then we'll start by showing
    \begin{equation*}
        |\partial_k g_1^{ij} - \partial_k g_2^{ij}| \leq C_4 \parallel g_1-g_2 \parallel_{C^2 (M, g_1)}
    \end{equation*}
    where $C_4$ is based on the manifold $M$, the metric $g_1$, and the neighborhood $U$. From lemma~\ref{lemma:ineq1}, we know that
    \begin{equation*}
        |\partial_k g_{1ij} - \partial_k g_{2ij}| \leq C_0 \parallel g_1 - g_2 \parallel_{C^2 (M, g_1)}
    \end{equation*}
    Then, we have for matrix functions
    \begin{equation*}
        \begin{split}
            A^{-1}A & = I\\
            (\partial A^{-1})A + A^{-1}(\partial A) & = 0\\
            \partial A^{-1} & = - A^{-1} \partial A A^{-1}
        \end{split}
    \end{equation*}
    This means that
    \begin{equation*}
        |\partial_k g_1^{ij} | \leq |g_1^{ij}| |\partial_k g_{1ij}| |g_1^{ij}| \leq C_0 (G_1^{-1})^2 \parallel g_1 \parallel_{C^2 (M,g_1)}
    \end{equation*}

    Then we have
    \begin{align*}
        |\partial_k g_1^{ij} - \partial_k g_2^{ij}| & \leq \parallel \partial_k g_1^{-1} - \partial_k g_2^{-1} \parallel_F = \parallel g_1^{-1} \partial_k g_1 g_1^{-1} -  g_2^{-1} \partial_k g_2 g_2^{-1}\parallel_F\\
        & = \parallel g_1^{-1} \partial_k g_1 g_1^{-1} - g_2^{-1} \partial_k g_1 g_2^{-1} + g_2^{-1} \partial_k g_1 g_2^{-1} - g_2^{-1} \partial_k g_2 g_2^{-1} \parallel_F\\
        & \leq \parallel g_1^{-1} \partial_k g_1 g_1^{-1} - g_2^{-1} \partial_k g_1 g_2^{-1} \parallel_F + \parallel g_2^{-1} \partial_k g_1 g_2^{-1} - g_2^{-1} \partial_k g_2 g_2^{-1} \parallel_F\\
        & =  \parallel g_2^{-1} (\partial_k g_1 - \partial_k g_2) g_2^{-1} \parallel_F\\
        & + \parallel g_2^{-1} \partial_k g_1 g_2^{-1} -g_2^{-1} \partial_k g_1 g_1^{-1} + g_2^{-1} \partial_k g_1 g_1^{-1} - g_1^{-1} \partial_k g_1 g_1^{-1} \parallel_F\\
        & \leq \parallel g_2^{-1} (\partial_k g_1 - \partial_k g_2) g_2^{-1} \parallel_F + \parallel (g_2^{-1} \partial_k g_1) (g_2^{-1} - g_1^{-1}) \parallel_F + \parallel (g_2^{-1} - g_1^{-1})(\partial_k g_1 g_1^{-1}) \parallel_F\\
        & \leq \parallel g_2^{-1} \parallel_F \parallel \partial_k g_2 - \partial_k g_1 \parallel_F \parallel g_2^{-1} \parallel_F + \parallel g_2^{-1} \parallel_F \parallel \partial_k g_1  \parallel_F \parallel g_2^{-1} - g_1^{-1} \parallel_F\\
        & + \parallel g_2^{-1} - g_1^{-1} \parallel_F  \parallel \partial_k g_1  \parallel_F \parallel g_1^{-1} \parallel_F\\
        & \leq C_1^2 \parallel \partial_k g_2 - \partial_k g_1 \parallel_F + C_1 \parallel \partial_k g_1 \parallel_F \parallel g_2^{-1} - g_1^{-1}
        \parallel_F
         + \parallel g_2^{-1} - g_1^{-1} \parallel_F \parallel  \partial_k g_1  \parallel_F \parallel g_1^{-1} \parallel_F\\
         & \leq C_1^2 \parallel \partial_k g_2 - \partial_k g_1 \parallel_F + C_1 \parallel \partial_k g_1 \parallel_F C_2 \parallel g_2 - g_1 \parallel_{C^2 (M, g_1)}\\
         &+ C_2 \parallel g_2 - g_1 \parallel_{C^2 (M,g_1)} \parallel  \partial_k g_1  \parallel_F \parallel g_1^{-1} \parallel_F\\
         & \leq C_1^2C_0 \parallel g_2 - g_1 \parallel_{C^2 (M,g_1)}  + C_1 C_2 C_0 \parallel g_1 \parallel_{C^2 (M,g_1)} \parallel g_2 - g_1 \parallel_{C^2 (M,g_1)}\\
         & + C_2C_0 \parallel g_1 \parallel_{C^2 (M,g_1)} G_1^{-1}  \parallel g_1 - g_2 \parallel_{C^2 (M,g_1)}\\
         & =  \big( C_1^2C_0  + C_1 C_2 C_0 \parallel g_1 \parallel_{C^2 (M,g_1)} + C_2C_0 \parallel g_1 \parallel_{C^2 (M,g_1)} G_1^{-1} \big) \parallel g_1 - g_2 \parallel_{C^2 (M,g_1)}\\
         & = C_4 \parallel g_1 - g_2 \parallel_{C^2 (M, g_1)}\\
    \end{align*}

    Where $C_4 =  C_1^2C_0  + C_1 C_2 C_0 \parallel g_1 \parallel_{C^2 (M,g_1)} + C_2C_0 \parallel g_1 \parallel_{C^2 (M,g_1)} G_1^{-1}$, which is based on the manifold $M$, the metric $g_1$, and the neighborhood $U$.

    Now, for an arbitrary metric on $M$, $a$, we get
    \begin{equation*}
        \begin{split}
            \partial_\eta \Gamma_{ij}^{k,a} & = \partial_\eta(\frac{1}{2} a^{lk}(\partial_i a_{jl} + \partial_j a_{il} - \partial_l a_{ij}))\\
            & = \frac{1}{2} \partial_\eta a^{lk} (\partial_i a_{jl} + \partial_j a_{il} - \partial_l a_{ij})\\
            & + \frac{1}{2} a^{lk} ( \partial_\eta \partial_i a_{jl} +  \partial_\eta \partial_j a_{il} -  \partial_\eta \partial_l a_{ij})
        \end{split}
    \end{equation*}

    Then, we want to show that $|\partial_\eta \Gamma_{ij}^{k, g_1} - \partial_\eta \Gamma_{ij}^{k, g_2}| < C_5 \parallel g_1 - g_2 \parallel_{C^2 (M, g_1)}$ where $C_5 >0$ is based on the manifold $M$, the metric $g_1$, and the neighborhood $U$. Then,
    \begin{align*}
        |\partial_\eta \Gamma_{ij}^{k, g_1} - \partial_\eta \Gamma_{ij}^{k, g_2}| & = |\frac{1}{2} \partial_\eta g_1^{lk} (\partial_i g_{1jl} + \partial_j g_{1il} - \partial_l g_{1ij}) + \frac{1}{2} g_1^{lk} ( \partial_\eta \partial_i g_{1jl} +  \partial_\eta \partial_j g_{1il} -  \partial_\eta \partial_l g_{1ij})\\
        & - \frac{1}{2} \partial_\eta g_2^{lk} (\partial_i g_{2jl} + \partial_j g_{2il} - \partial_l g_{2ij})
        + \frac{1}{2} g_2^{lk} ( \partial_\eta \partial_i g_{2jl} +  \partial_\eta \partial_j g_{2il} -  \partial_\eta \partial_l g_{2ij})|\\
        & \leq |\frac{1}{2} \partial_\eta g_1^{lk} (\partial_i g_{1jl} + \partial_j g_{1il} - \partial_l g_{1ij}) - \frac{1}{2} \partial_\eta g_2^{lk} (\partial_i g_{1jl} + \partial_j g_{1il} - \partial_l g_{1ij})|\\
        & + |\frac{1}{2} \partial_\eta g_2^{lk} (\partial_i g_{1jl} + \partial_j g_{1il} - \partial_l g_{1ij}) -\frac{1}{2} \partial_\eta g_2^{lk} (\partial_i g_{2jl} + \partial_j g_{2il} - \partial_l g_{2ij})  |\\
        & + |\frac{1}{2}  g_1^{lk} ( \partial_\eta \partial_i g_{1jl} + \partial_\eta \partial_j g_{1il} - \partial_\eta \partial_l g_{1ij}) -\frac{1}{2}  g_2^{lk} ( \partial_\eta \partial_i g_{1jl} + \partial_\eta \partial_j g_{1il} - \partial_\eta \partial_l g_{1ij}) |\\
        & + |\frac{1}{2}  g_2^{lk} ( \partial_\eta \partial_i g_{1jl} + \partial_\eta \partial_j g_{1il} - \partial_\eta \partial_l g_{1ij}) -\frac{1}{2}  g_2^{lk} ( \partial_\eta \partial_i g_{2jl} + \partial_\eta \partial_j g_{2il} - \partial_\eta \partial_l g_{2ij}) |\\
        & = I + II + III + IV = \star
    \end{align*}
    Then, breaking this equation into pieces, we see that
    \begin{align*}
        I & \leq \frac{1}{2} |\partial_\eta g_1^{lk} - \partial_\eta g_2^{lk}| (|\partial_i g_{1jl}| + |\partial_j g_{1il}| + |\partial_l g_{1ij} |)\\
        & \leq \frac{1}{2} |\partial_\eta g_1^{lk} - \partial_\eta g_2^{lk}|  3C_0 \parallel g_1 \parallel_{C^2 (M, g_1)}\\
        & \leq \frac{3}{2} C_0 C_4 \parallel g_1 \parallel_{C^2 (M, g_1)} \parallel g_1 - g_2 \parallel_{C^2 (M, g_1)}\\
        II & \leq \frac{1}{2} |\partial_\eta g_2^{lk}| ( |\partial_i g_{1jl} - \partial_i g_{2jl}|+ |\partial_j g_{1il} - \partial_j g_{2il}| + |\partial_l g_{1ij} - \partial_l g_{2ij}|)\\
        & \leq \frac{1}{2} (C_0 (G_1^{-1})^2 \parallel g_1 \parallel_{C^2 (M, g_1)} + C_4)3 C_0 \parallel g_1 - g_2 \parallel_{C^2 (M, g_1)}\\
        III & \leq \frac{1}{2} |g_1^{lk} - g_2^{lk}| (|\partial_\eta \partial_i g_{1jl}| + |\partial_\eta \partial_j g_{1il}| + |\partial_\eta \partial_l g_{1ij}|)\\
        & \leq \frac{3 C_0}{2} \parallel g_1 \parallel_{C^2 (M, g_1)} |g_1^{lk} - g_2^{lk}|\\
        & \leq \frac{3 C_0}{2} \parallel g_1 \parallel_{C^2 (M, g_1)}  C_2 \parallel g_1 - g_2 \parallel_{C^2 (M, g_1)}\\
        IV & \leq \frac{1}{2} |g_2^{lk}| (|\partial_\eta \partial_i g_{1jl} - \partial_\eta \partial_i g_{2jl}| + |\partial_\eta \partial_j g_{1il} - \partial_\eta \partial_j g_{2il}| + |\partial_\eta \partial_l g_{1ij} - \partial_\eta \partial_l g_{2ij}|)\\
        & \leq \frac{3}{2} C_1 C_0 \parallel g_1 -g_2 \parallel_{C^2 (M, g_1)}
    \end{align*}
    Then we can put this together,
    \begin{align*}
        \star & \leq \big( \frac{3 C_0 C_4}{2} \parallel g_1 \parallel_{C^2 (M, g_1)} + \frac{3 C_0}{2} (C_0 (G_1^{-1})^2 \parallel g_1 \parallel_{C^2 (M, g_1)} + C_4)\\
        & + \frac{3 C_0}{2} \parallel g_1 \parallel_{C^2 (M, g_1)}  C_2 + \frac{3}{2} C_1 C_0  \big) \parallel g_1 -g_2 \parallel_{C^2 (M, g_1)}\\
        & = C_5 \parallel g_1 -g_2 \parallel_{C^2 (M, g_1)}
    \end{align*}
    where $C_5 = \big( \frac{3 C_0 C_4}{2} \parallel g_1 \parallel_{C^2 (M, g_1)} + \frac{3 C_0}{2} (C_0 (G_1^{-1})^2 \parallel g_1 \parallel_{C^2 (M, g_1)} + C_4) + \frac{3 C_0}{2} \parallel g_1 \parallel_{C^2 (M, g_1)}  C_2 + \frac{3}{2} C_1 C_0  \big)$ which is based on the manifold $M$, the metric $g_1$, and the neighborhood $U$. 
    
    Now for the curvature, we have
    
    \begin{align*}
        |R_{ijk}^{l,g_1} - R_{ijk}^{l, g_2}| & = | \partial_i \Gamma_{jk}^{l, g_1}  -  \partial_j \Gamma_{ik}^{l, g_1}  + \Gamma_{jk}^{s, g_1} \Gamma_{is}^{l, g_1} - \Gamma_{ik}^{s, g_1} \Gamma_{js}^{l, g_1} -  \partial_i \Gamma_{jk}^{l, g_2}  -  \partial_j \Gamma_{ik}^{l, g_2}  + \Gamma_{jk}^{s, g_2} \Gamma_{is}^{l, g_2} - \Gamma_{ik}^{s, g_2} \Gamma_{js}^{l, g_2}|\\
        & \leq |\partial_i \Gamma_{jk}^{l, g_1} - \partial_i \Gamma_{jk}^{l, g_2}| + |\partial_j \Gamma_{ik}^{l, g_1}  - \partial_j \Gamma_{ik}^{l, g_2} |\\
        & + |\Gamma_{jk}^{s, g_1} \Gamma_{is}^{l, g_1} - \Gamma_{jk}^{s, g_2} \Gamma_{is}^{l, g_2} | + |\Gamma_{ik}^{s, g_1} \Gamma_{js}^{l, g_1} - \Gamma_{ik}^{s, g_2} \Gamma_{js}^{l, g_2} |\\
        & = \star \star
    \end{align*}
    Now we see that
    \begin{align*}
        |\Gamma_{jk}^{s, g_1} \Gamma_{is}^{l, g_1} - \Gamma_{jk}^{s, g_2} \Gamma_{is}^{l, g_2} | & \leq |\Gamma_{jk}^{s, g_1} \Gamma_{is}^{l, g_1} - \Gamma_{jk}^{s, g_1} \Gamma_{is}^{l, g_2} | + |\Gamma_{jk}^{s, g_1} \Gamma_{is}^{l, g_2} - \Gamma_{jk}^{s, g_2} \Gamma_{is}^{l, g_2} |\\
        & = |\Gamma_{jk}^{s,g_1}||\Gamma_{is}^{l, g_1} - \Gamma_{is}^{l, g_2}| + |\Gamma_{is}^{l,g_2}||\Gamma_{jk}^{s, g_1} - \Gamma_{jk}^{s, g_2}|\\
        & \leq \max |\Gamma^{g_1}| |\Gamma_{is}^{l, g_1} - \Gamma_{is}^{l, g_2}| + (\max |\Gamma^{g_1}| + C_3) |\Gamma_{jk}^{s, g_1} - \Gamma_{jk}^{s, g_2}|\\
        & \leq \max |\Gamma^{g_1}| C_3 \parallel g_1 - g_2 \parallel_{C^2 (M, g_1)} + (\max |\Gamma^{g_1}| + C_3) C_3 \parallel g_1 - g_2 \parallel_{C^2 (M, g_1)}\\
    \end{align*}
    Then,
    \begin{align*}
        \star \star & \leq 2 C_5 \parallel g_1 - g_2 \parallel_{C^2 (M, g_1)} + 2(\max |\Gamma^{g_1}| C_3  + (\max |\Gamma^{g_1}| + C_3) C_3) \parallel g_1 - g_2 \parallel_{C^2 (M, g_1)}\\
        & = (2 C_5  + 2 \max |\Gamma^{g_1}| C_3  + 2(\max |\Gamma^{g_1}| + C_3) C_3) \parallel g_1 - g_2 \parallel_{C^2 (M, g_1)}\\
        & = C_6 \parallel g_1 - g_2 \parallel_{C^2 (M, g_1)}
    \end{align*}
    where $C_6 = 2 C_5  + 2 \max |\Gamma^{g_1}| C_3  + 2(\max |\Gamma^{g_1}| + C_3) C_3$, which is based on the manifold $M$, the metric $g_1$, and the neighborhood $U$. Then, let our $C_\alpha$ from the proof of proposition~\ref{prop:prop1} be defined as
    \begin{equation*}
        C_\alpha = \min \{ \frac{1}{2 C_0 G_1^{-1}}, \frac{1}{C_3}, \frac{1}{C_6} , \frac{1}{\epsilon}  \}
    \end{equation*}
    Let $\delta (M, g_1, U, \epsilon) = C_\alpha \epsilon$. Then $\parallel g_1 - g_2 \parallel_{C^2 (M, g_1)} < \delta (M, g_1, U, \epsilon)$ implies that, on the coordinate neighborhood $U$, we have
    \begin{align*}
       |R_{ijk}^{l,g_1} - R_{ijk}^{l, g_2}| & \leq C_6 \parallel g_1 - g_2 \parallel_{C^2 (M, g_1)}\\
       & < C_6 C_\alpha \epsilon \leq \epsilon
    \end{align*}
    as desired.
    
\end{proof}

\subsection{Proof of Theorem \ref{thrm:neat1}}

\phantomsection\label{app:neat1}
\begin{proof}[Proof of Theorem \ref{thrm:neat1}]

To assist us, we'll use lemma~\ref{def:example3} with 
$y_m (t) = \begin{bmatrix}
    f_m (t)\\
    f_m^\prime (t)
\end{bmatrix}$ where $f_m (t)$ is the vector of components of $J_m$ from theorem~\ref{thrm:neat1}, which satisfies the system in equation~\ref{def:jaceq}, and $M(t, \lambda_m) = \begin{bmatrix}
    0 & I\\
    - (b_{kj}^{g_m}(t)) & - (a_{kj}^{g_m}(t)) 
\end{bmatrix}$.
Then, we see that
\begin{align*}
    |M(t, \lambda_1)| & \leq \sqrt{n} + \sum_{kj} |a_{kj}^{g_1}(t)|+|b_{kj}|\\
    & = \sqrt{n} + \frac{2n^3L \max|\Gamma^{g_1}|}{A_{g_1}}  +  \frac{n^4L^2 \max |\partial \Gamma^{g_1}|}{A_{g_1}^2} +  \frac{2n^5 L^2 \max |\Gamma^{g_1}|^2}{A_{g_1}^2} +  \frac{n^4L^2 \max |R^{g_1}|}{A_{g_1}^2}
\end{align*}
where $\max|\Gamma^{g_1}|$ and $\max |\partial \Gamma^{g_1}|$ were defined in remark~\ref{def:rmk1}, and $\max |R^{g_1}|$ was defined in remark~\ref{def:rmk2}. So, let $C_1 = \sqrt{n} +  \frac{ 2n^3L \max|\Gamma^{g_1}|}{A_{g_1}} +  \frac{ n^4L^2 \max |\partial \Gamma^{g_1}| }{A_{g_1}^2} +  \frac{ 2n^5 L^2 \max |\Gamma^{g_1}|^2}{A_{g_1}^2} + \frac{ n^4L^2 \max |R^{g_1}| }{A_{g_1}^2}$.

Then, we want to bound $y_1 (t)$. We've established that $|M(t, \lambda_1)|< C_1$. Since
\begin{equation*}
    y_1 (t) = y_1 (0) + \int_0^t y_1^\prime (s) \ ds
\end{equation*}
we have
\begin{align*}
    |y_1 (t)| & \leq |y_1 (0)| + \int_0^t |y_1^\prime (s)| \ ds\\
    & \leq |y_1 (0)| + \int_0^t C_1 |y_1 (s)| \ ds
\end{align*}
So, we've satisfied the conditions for the integral form of Gr\"onwall's inequality, lemma~\ref{gron:int}, with $A = C_1$ and $B = |y_1 (0)|$. Then, we have
\begin{equation*}
    |y_1 (t)| \leq |y_1 (0)|e^{C_1}
\end{equation*}
and, since $y_1 (0) = 
\begin{bmatrix}
    f_1 (0)\\
    f_1^\prime (0)
\end{bmatrix}
= 
\begin{bmatrix}
    0\\
    w
\end{bmatrix}$, 
\begin{equation*}
    |y_1 (t)| \leq |w|e^{C_1} \leq \frac{Le^{C_1}}{A_{g_1}}
\end{equation*}
Then let $C_0 = \frac{Le^{C_1}}{A_{g_1}}$.

Next, our goal is to show that on $U$, for each $\epsilon_0>0$ there exists $\delta_1 (M, g_1, U, L, \epsilon_0)>0$ such that $\parallel g_1 - g_2 \parallel_{C^2 (M, g_1)}< \delta_1 (M, g_1, U, L, \epsilon_0)$ implies that
\begin{equation*}
    |a_{kj}^{g_1}(t) - a_{kj}^{g_2}(t)| < \epsilon_0
\end{equation*}
where $a_{kj}^{g_m} (t) = 2 \Gamma_{ij}^{k,g_m} (\gamma_m (t)) \dot{\gamma}_m^i$. From proposition~\ref{prop:prop1} and theorem~\ref{def:thrm3}, we know that there exists a $\delta_0 (M, g_1, U, L, \min \{ \zeta_1, \zeta_2, \zeta_3 \})>0$, such that $\parallel g_1 - g_2 \parallel_{C^2 (M, g_1)}< \delta_0 (M, g_1, U, L, \min \{ \zeta_1, \zeta_2, \zeta_3 \})$ implies that on $U$, we have
\begin{equation*}
\begin{split}
|\gamma_1 (t) - \gamma_2 (t)| & < \zeta_1\\
|\gamma_1^\prime (t) - \gamma_2^\prime (t)| & < \zeta_2\\
|\Gamma_{ij}^{k, g_1} - \Gamma_{ij}^{k, g_2}| & < \zeta_3
\end{split}
\end{equation*}
where the first two inequalities come from equations~\ref{eq:eq2} and~\ref{def:eq3} respectively.

Then we have
\begin{align*}
    |a_{kj}^{g_1}(t) - a_{kj}^{g_2}(t)| & \leq |  2 \Gamma_{ij}^{k, g_1} (\gamma_1 (t)) \dot{\gamma}_1^i - 2 \Gamma_{ij}^{k, g_2} (\gamma_2 (t)) \dot{\gamma}_2^i |\\
    & \leq |2 \Gamma_{ij}^{k, g_1} (\gamma_1 (t)) \dot{\gamma}_1^i - 2 \Gamma_{ij}^{k, g_1} (\gamma_2 (t)) \dot{\gamma}_1^i|\\
    & + |2 \Gamma_{ij}^{k, g_1} (\gamma_2 (t)) \dot{\gamma}_1^i - 2 \Gamma_{ij}^{k, g_1} (\gamma_2 (t)) \dot{\gamma}_2^i|\\
    & + |2 \Gamma_{ij}^{k, g_1} (\gamma_2 (t)) \dot{\gamma}_2^i - 2 \Gamma_{ij}^{k, g_2} (\gamma_2 (t)) \dot{\gamma}_2^i|\\
    & = I+II+III
\end{align*}
Then, breaking this into pieces, we first see that by equation~\ref{def:eq12} we have

\begin{equation*}
\begin{split}
    |\dot{\gamma}_1^i| & \leq \frac{L}{A_{g_1}}\\
    |\dot{\gamma}_2^i| & \leq \frac{2L}{A_{g_1}}\\
\end{split}
\end{equation*}

\begin{align*}
    I & \leq 2 |\dot{\gamma}_1^i| |\Gamma_{ij}^{k, g_1} (\gamma_1 (t)) - \Gamma_{ij}^{k, g_1} (\gamma_2 (t))|\\
    & =  \frac{2nL \max |\partial \Gamma^{g_1}|}{A_{g_1}}  |\gamma_1 (t) - \gamma_2 (t)|\\
    & <  \frac{2nL \max |\partial \Gamma^{g_1}|}{A_{g_1}} \zeta_1
\end{align*}

Then we have

\begin{align*}
    II & \leq 2 |\Gamma_{ij}^{k, g_1}(\gamma_2 (t))| |\dot{\gamma}_1^i - \dot{\gamma}_2^i|\\
    & < 2 n \max |\Gamma^{g_1}| \zeta_2
\end{align*}
Finally, we have
\begin{align*}
   III & \leq 2 |\dot{\gamma}_2^i| |\Gamma_{ij}^{k, g_1} (\gamma_2 (t)) - \Gamma_{ij}^{k, g_2} (\gamma_2 (t))|\\
   & < \frac{4Ln}{A_{g_1}} \zeta_3
\end{align*}
Then putting this back together and letting $\zeta_1 = \frac{A_{g_1}\epsilon_0}{6 n L \max |\partial \Gamma^{g_1}|}$, $\zeta_2 = \frac{\epsilon_0}{6 n \max |\Gamma^{g_1}|}$, $\zeta_3 = \frac{A_{g_1} \epsilon_0}{12Ln}$, and by letting $\delta_0 (M,g_1,U,L, \min \{ \zeta_1, \zeta_2, \zeta_3 \}) = \delta_1 (M,g_1,U,L, \epsilon_0 )$, we have

\begin{equation*}
    I + II+ III < \frac{\epsilon_0}{3} + \frac{\epsilon_0}{3} + \frac{\epsilon_0}{3} = \epsilon_0
\end{equation*}

Now, our next goal is to show that on $U$, for all $\epsilon_1 >0$ there exists $\delta_2 (M, g_1, U, L, \epsilon_1)>0$  such that $\parallel g_1 - g_2 \parallel_{C^2 (M, g_1)} < \delta_2 (M, g_1, U, L, \epsilon_1)$ implies that
\begin{equation*}
    |b_{k j}^{g_1}(t) - b_{k j}^{g_2}(t)|< \epsilon_1
\end{equation*}
where $b_{k j}^{g_m}=  \dot{\gamma}_m^\mu \dot{\gamma}_m^s \frac{\partial \Gamma_{\mu j}^{k, g_m}}{\partial x^s} (\gamma_m (t)) - \dot{\gamma}_m^\nu \dot{\gamma}_m^c \Gamma_{c \nu}^{\mu, g_m} (\gamma_m (t)) \Gamma_{\mu j}^{k, g_m} (\gamma_m (t)) + \dot{\gamma}_m^\mu \dot{\gamma}_m^\nu \Gamma_{\mu j}^{\beta, g_m} (\gamma_m (t)) \Gamma_{\nu \beta}^{k, g_m} (\gamma_m (t)) + \dot{\gamma}_m^\mu \dot{\gamma}_m^\nu R_{j \nu \mu}^{k, g_m} (\gamma_m (t))$. Now from corollary~\ref{def:cor1}, theorem~\ref{def:thrm3}, and proposition~\ref{def:prop2}, we know that there exists $\delta_\alpha (M, g_1, U, L, \min \{ \sigma_1, \sigma_2, \sigma_3, \sigma_4 \})>0$, such that $\parallel g_1 - g_2 \parallel_{C^2 (M, g_1)} < \delta_\alpha (M, g_1, U, L, \min \{ \sigma_1, \sigma_2, \sigma_3, \sigma_4 \})$ implies that
\begin{equation*}
\begin{split}
|\gamma_1 (t) - \gamma_2 (t)| & < \sigma_1\\
|\gamma_1^\prime (t)- \gamma_2^\prime (t)| & < \sigma_2\\
|\frac{\partial \Gamma_{\mu j}^{k, g_1}}{\partial x^s} - \frac{\partial \Gamma_{\mu j}^{k, g_2}}{\partial x^s}| & < \sigma_3\\
|R_{j \nu \mu}^{k, g_1} - R_{j \nu \mu}^{k, g_2}| & < \sigma_4\\
\end{split}
\end{equation*}
From proposition~\ref{prop:prop1}, we know that $|\Gamma_{ij}^{k,g_1} - \Gamma_{ij}^{k,g_2}|$ can be made arbitrarily small, and from theorem~\ref{def:thrm3}, we know that $d_{g_1}(\gamma_1 (t), \gamma_2 (t))$ can also be made arbitrarily small. Since
\begin{align*}
    |\Gamma_{\mu j}^{k,g_1}(\gamma_1 (t)) - \Gamma_{\mu j}^{k,g_2}(\gamma_2 (t))| & \leq |\Gamma_{\mu j}^{k,g_1}(\gamma_1 (t)) - \Gamma_{\mu j}^{k,g_1}(\gamma_2 (t))| + |\Gamma_{\mu j}^{k,g_1}(\gamma_2 (t)) - \Gamma_{\mu j}^{k,g_2}(\gamma_2 (t))|\\
    & \leq \max|\partial \Gamma^{g_1}| |\gamma_1 (t) - \gamma_2 (t)| + |\Gamma_{\mu j}^{k,g_1}(\gamma_2 (t)) - \Gamma_{\mu j}^{k,g_2}(\gamma_2 (t))|
\end{align*}
So, we'll also suppose that $\parallel g_1 - g_2 \parallel_{C^2 (M, g_1)} < \delta_\alpha (M, g_1, U, L, \min \{ \sigma_1, \sigma_2, \sigma_3, \sigma_4, \sigma_5, 1 \})$ implies that
\begin{equation*}
    |\Gamma_{\mu j}^{k,g_1}(\gamma_1 (t)) - \Gamma_{\mu j}^{k,g_2}(\gamma_2 (t))|  < \sigma_5
\end{equation*}

Additionally, we'll suppose that $\parallel g_1 - g_2 \parallel_{C^2 (M, g_1)} < \delta_\alpha (M, g_1, U, L, \min \{ \sigma_1, \sigma_2, \sigma_3, \sigma_4, 1 \})$ implies that
\begin{equation*}
        |\Gamma_{ij}^{k,g_1} - \Gamma_{ij}^{k,g_2}| < 1
\end{equation*}
which means that
\begin{equation*}\label{def:eq22}
    | \Gamma_{ij}^{k,g_2} | < \max |\Gamma^{g_1}|+1
\end{equation*}
which we'll use later in this proof.

Then we have
\begin{align*}
    |b_{kj}^{g_1}(t) - b_{kj}^{g_2}(t)| & = | \dot{\gamma}_1^\mu \dot{\gamma}_1^s \frac{\partial \Gamma_{\mu j}^{k, g_1}}{\partial x^s} (\gamma_1 (t)) - \dot{\gamma}_1^\nu \dot{\gamma}_1^c \Gamma_{c \nu}^{\mu, g_1} (\gamma_1 (t)) \Gamma_{\mu j}^{k, g_1} (\gamma_1 (t))\\
    & + \dot{\gamma}_1^\mu \dot{\gamma}_1^\nu \Gamma_{\mu j}^{\beta, g_1} (\gamma_1 (t)) \Gamma_{\nu \beta}^{k, g_1} (\gamma_1 (t)) + \dot{\gamma}_1^\mu \dot{\gamma}_1^\nu R_{j \nu \mu}^{k, g_1} (\gamma_1 (t))\\
    &- ( \dot{\gamma}_2^\mu \dot{\gamma}_2^s \frac{\partial \Gamma_{\mu j}^{k, g_2}}{\partial x^s} (\gamma_2 (t)) - \dot{\gamma}_2^\nu \dot{\gamma}_2^c \Gamma_{c \nu}^{ \mu, g_2} (\gamma_2 (t)) \Gamma_{\mu j}^{k, g_2} (\gamma_2 (t))\\
    & + \dot{\gamma}_2^\mu \dot{\gamma}_2^\nu \Gamma_{\mu j}^{\beta, g_2} (\gamma_2 (t)) \Gamma_{\nu \beta}^{k, g_2} (\gamma_2 (t)) + \dot{\gamma}_2^\mu \dot{\gamma}_2^\nu R_{j \nu \mu}^{k, g_2} (\gamma_2 (t)) )|\\
    & \leq |\dot{\gamma}_1^\mu \dot{\gamma}_1^s \frac{\partial \Gamma_{\mu j}^{k, g_1}}{\partial x^s} (\gamma_1 (t)) - \dot{\gamma}_2^\mu \dot{\gamma}_2^s \frac{\partial \Gamma_{\mu j}^{k, g_2}}{\partial x^s} (\gamma_2 (t)) |\\
    & + |\dot{\gamma}_1^\nu \dot{\gamma}_1^c \Gamma_{c \nu}^{\mu, g_1} (\gamma_1 (t)) \Gamma_{\mu j}^{k, g_1} (\gamma_1 (t)) - \dot{\gamma}_2^\nu \dot{\gamma}_2^c \Gamma_{c \nu}^{\mu, g_2} (\gamma_2 (t)) \Gamma_{\mu j}^{k, g_2} (\gamma_2 (t)) |\\
    & + |\dot{\gamma}_1^\mu \dot{\gamma}_1^\nu \Gamma_{\mu j}^{\beta, g_1} (\gamma_1 (t)) \Gamma_{\nu \beta}^{k, g_1} (\gamma_1 (t)) - \dot{\gamma}_2^\mu \dot{\gamma}_2^\nu \Gamma_{\mu j}^{\beta, g_2} (\gamma_2 (t)) \Gamma_{\nu \beta}^{k, g_2} (\gamma_2 (t)) |\\
    & + |\dot{\gamma}_1^\mu \dot{\gamma}_1^\nu R_{j \nu \mu}^{k, g_1} (\gamma_1 (t)) - \dot{\gamma}_2^\mu \dot{\gamma}_2^\nu R_{j \nu \mu}^{k, g_2} (\gamma_2 (t)) |\\
    & = A + B + C + D
\end{align*}

For $A$, we have

\begin{align*}
    A & = |\dot{\gamma}_1^\mu \dot{\gamma}_1^s \frac{\partial \Gamma_{\mu j}^{k, g_1}}{\partial x^s} (\gamma_1 (t)) - \dot{\gamma}_2^\mu \dot{\gamma}_2^s \frac{\partial \Gamma_{\mu j}^{k, g_2}}{\partial x^s} (\gamma_2 (t)) |\\
    & \leq |\dot{\gamma}_1^\mu \dot{\gamma}_1^s \frac{\partial \Gamma_{\mu j}^{k, g_1}}{\partial x^s} (\gamma_1 (t)) - \dot{\gamma}_1^\mu \dot{\gamma}_1^s \frac{\partial \Gamma_{\mu j}^{k, g_1}}{\partial x^s} (\gamma_2 (t))|\\
    & + |\dot{\gamma}_1^\mu \dot{\gamma}_1^s \frac{\partial \Gamma_{\mu j}^{k, g_1}}{\partial x^s} (\gamma_2 (t)) - \dot{\gamma}_2^\mu \dot{\gamma}_2^s \frac{\partial \Gamma_{\mu j}^{k, g_1}}{\partial x^s} (\gamma_2 (t)) |\\
    & + |\dot{\gamma}_2^\mu \dot{\gamma}_2^s \frac{\partial \Gamma_{\mu j}^{k, g_1}}{\partial x^s} (\gamma_2 (t)) - \dot{\gamma}_2^\mu \dot{\gamma}_2^s \frac{\partial \Gamma_{\mu j}^{k, g_2}}{\partial x^s} (\gamma_2 (t)) |\\
    & = I + II + III
\end{align*}
Then, by breaking this into pieces, for $I$ we have
\begin{align*}
    I & \leq |\dot{\gamma}_1^\mu \dot{\gamma}_1^s \frac{\partial \Gamma_{\mu j}^{k, g_1}}{\partial x^s} (\gamma_1 (t)) - \dot{\gamma}_1^\mu \dot{\gamma}_1^s \frac{\partial \Gamma_{\mu j}^{k, g_1}}{\partial x^s} (\gamma_2 (t))|\\
    & \leq |\dot{\gamma}_1^\mu | | \dot{\gamma}_1^s | \max |\partial^2 \Gamma^{g_1}| \cdot |\gamma_1 (t) - \gamma_2 (t)|\\
    & \leq \frac{n^2L^2}{A_{g_1}^2} \max |\partial^2 \Gamma^{g_1}| \sigma_1
\end{align*}
where $\max |\partial^2 \Gamma^{g_1}|$ was defined in remark~\ref{def:rmk1}.

Then for $II$, we have

\begin{align*}
    II & = |\dot{\gamma}_1^\mu \dot{\gamma}_1^s \frac{\partial \Gamma_{\mu j}^{k, g_1}}{\partial x^s} (\gamma_2 (t)) - \dot{\gamma}_2^\mu \dot{\gamma}_2^s \frac{\partial \Gamma_{\mu j}^{k, g_1}}{\partial x^s} (\gamma_2 (t)) |\\
    & \leq \max |\partial \Gamma^{g_1}| \cdot |\dot{\gamma}_1^\mu \dot{\gamma}_1^s - \dot{\gamma}_2^\mu \dot{\gamma}_2^s|\\
    & \leq \max |\partial \Gamma^{g_1}| (| \dot{\gamma}_1^\mu ||\dot{\gamma}_1^s  - \dot{\gamma}_2^s | + |\dot{\gamma}_2^s ||\dot{\gamma}_1^\mu - \dot{\gamma}_2^\mu |)\\
    & \leq \max | \partial \Gamma^{g_1}| ( \frac{L}{A_{g_1}} |\dot{\gamma}_1^s  - \dot{\gamma}_2^s | + \frac{2L}{A_{g_1}} |\dot{\gamma}_1^\mu - \dot{\gamma}_2^\mu |)\\
    & \leq \frac{n^2 3 L}{A_{g_1}} \max | \partial \Gamma^{g_1}| \sigma_2
\end{align*}

Then, finally for $III$, we have

\begin{align*}
    III & \leq |\dot{\gamma}_2^\mu \dot{\gamma}_2^s \frac{\partial \Gamma_{\mu j}^{k, g_1}}{\partial x^s} (\gamma_2 (t)) -  \dot{\gamma}_2^\mu \dot{\gamma}_2^s  \frac{\partial \Gamma_{\mu j}^{k, g_2}}{\partial x^s} (\gamma_2 (t)) |\\
    & = |\dot{\gamma}_2^\mu | | \dot{\gamma}_2^s| |\frac{\partial \Gamma_{\mu j}^{k, g_1}}{\partial x^s} (\gamma_2 (t))  - \frac{\partial \Gamma_{\mu j}^{k, g_2}}{\partial x^s} (\gamma_2 (t))|\\
    & \leq \frac{4L^2}{A_{g_1}^2} |\frac{\partial \Gamma_{\mu j}^{k, g_1}}{\partial x^s} (\gamma_2 (t))  - \frac{\partial \Gamma_{\mu j}^{k, g_2}}{\partial x^s} (\gamma_2 (t))|\\
    & \leq \frac{4n^2L^2}{A_{g_1}^2} \sigma_3
\end{align*}

We'll add some more constraints to $\sigma_1$, $\sigma_2$, and $\sigma_3$, but for the moment let $\sigma_1 = \frac{A_{g_1}^2 \epsilon_1}{12 n^2 L^2 \max |\partial^2 \Gamma^{g_1}|}$, $\sigma_2 = \frac{A_{g_1} \epsilon_1}{36 n^2 L \max |\partial \Gamma^{g_1}|}$, $\sigma_3 = \frac{A_{g_1}^2 \epsilon_1}{48 n^2 L^2}$, then putting everything back together, we have
\begin{equation*}
    A \leq I + II + III < \frac{\epsilon_1}{4}
\end{equation*}

We'll handle $B$ and $C$ at the same time, by breaking them into pieces,

\begin{align*}
    B, C & = |\dot{\gamma}_1^\nu \dot{\gamma}_1^c \Gamma_{c \nu}^{\mu, g_1} (\gamma_1 (t)) \Gamma_{\mu j}^{k, g_1} (\gamma_1 (t)) - \dot{\gamma}_2^\nu \dot{\gamma}_2^c \Gamma_{c \nu}^{\mu , g_2} (\gamma_2 (t)) \Gamma_{\mu j}^{k, g_2} (\gamma_2 (t)) |\\
    & \leq |\dot{\gamma}_1^\nu \dot{\gamma}_1^c \Gamma_{c \nu}^{\mu, g_1} (\gamma_1 (t)) \Gamma_{\mu j}^{k, g_1} (\gamma_1 (t)) - \dot{\gamma}_2^\nu \dot{\gamma}_2^c \Gamma_{c \nu}^{\mu , g_1} (\gamma_1 (t)) \Gamma_{\mu j}^{k, g_1} (\gamma_1 (t))|\\
    & + |\dot{\gamma}_2^\nu \dot{\gamma}_2^c \Gamma_{c \nu}^{\mu , g_1} (\gamma_1 (t)) \Gamma_{\mu j}^{k, g_1} (\gamma_1 (t)) - \dot{\gamma}_2^\nu \dot{\gamma}_2^c \Gamma_{c \nu}^{\mu , g_2} (\gamma_2 (t)) \Gamma_{\mu j}^{k, g_2} (\gamma_2 (t)) |\\
    & = I + II
\end{align*}
Then for $I$, we'll have

\begin{align*}
    I & = |\dot{\gamma}_1^\nu \dot{\gamma}_1^c \Gamma_{c \nu}^{\mu , g_1} (\gamma_1 (t)) \Gamma_{\mu j}^{k, g_1} (\gamma_1 (t)) - \dot{\gamma}_2^\nu \dot{\gamma}_2^c \Gamma_{c \nu}^{\mu, g_1} (\gamma_1 (t)) \Gamma_{\mu j}^{k, g_1} (\gamma_1 (t))|\\
    & \leq \max |\Gamma^{g_1}|^2  |\dot{\gamma}_1^\nu \dot{\gamma}_1^c  - \dot{\gamma}_2^\nu \dot{\gamma}_2^c |\\
    & \leq \max |\Gamma^{g_1}|^2 (| \dot{\gamma}_1^\nu ||\dot{\gamma}_1^c  - \dot{\gamma}_2^c | + |\dot{\gamma}_2^c ||\dot{\gamma}_1^\nu - \dot{\gamma}_2^\nu |)\\
    & \leq \max |\Gamma^{g_1}|^2 ( \frac{L}{A_{g_1}} |\dot{\gamma}_1^c  - \dot{\gamma}_2^c | + \frac{2L}{A_{g_1}} |\dot{\gamma}_1^\nu - \dot{\gamma}_2^\nu |)\\
    & \leq \frac{n^3 3 L}{A_{g_1}} \max |\Gamma^{g_1}|^2 \sigma_2
\end{align*}

Then for $II$, we'll have

\begin{align*}
    II & = |\dot{\gamma}_2^\nu \dot{\gamma}_2^c \Gamma_{c \nu}^{\mu , g_1} (\gamma_1 (t)) \Gamma_{\mu j}^{k, g_1} (\gamma_1 (t)) - \dot{\gamma}_2^\nu \dot{\gamma}_2^c \Gamma_{c \nu}^{\mu , g_2} (\gamma_2 (t)) \Gamma_{\mu j}^{k, g_2} (\gamma_2 (t)) |\\
    & = |\dot{\gamma}_2^\nu| |\dot{\gamma}_2^c| |\Gamma_{c \nu}^{\mu , g_1} (\gamma_1 (t)) \Gamma_{\mu j}^{k, g_1} (\gamma_1 (t)) -  \Gamma_{c \nu}^{\mu, g_2} (\gamma_2 (t)) \Gamma_{\mu j}^{k, g_2} (\gamma_2 (t)) |\\
    & \leq \frac{4 L^2}{A_{g_1}^2} \bigg( |\Gamma_{c \nu}^{\mu , g_1} (\gamma_1 (t)) \Gamma_{\mu j}^{k, g_1} (\gamma_1 (t)) - \Gamma_{c \nu}^{\mu , g_1} (\gamma_1 (t)) \Gamma_{\mu j}^{k, g_2} (\gamma_2 (t))|\\
    & + |\Gamma_{c \nu}^{\mu , g_1} (\gamma_1 (t)) \Gamma_{\mu j}^{k, g_2} (\gamma_2 (t)) - \Gamma_{c \nu}^{\mu , g_2} (\gamma_2 (t)) \Gamma_{\mu j}^{k, g_2} (\gamma_2 (t))| \bigg)\\
     & \leq \frac{4 L^2}{A_{g_1}^2} \bigg(  |\Gamma_{c \nu}^{\mu , g_1} (\gamma_1 (t))| | \Gamma_{\mu j}^{k, g_1} (\gamma_1 (t)) -  \Gamma_{\mu j}^{k, g_2} (\gamma_2 (t))|\\
    & + |\Gamma_{\mu j}^{k, g_2} (\gamma_2 (t)) | |\Gamma_{c \nu}^{\mu , g_1} (\gamma_1 (t))  - \Gamma_{c \nu}^{\mu, g_2} (\gamma_2 (t))| \bigg)\\
    & \leq \frac{4 n^3 L^2}{A_{g_1}^2} (2\max |\Gamma^{g_1}| +1) \sigma_5
\end{align*}
which comes from equation~\ref{def:eq22}.
We'll add some more constraints to $\sigma_2$, but for the moment let $\sigma_2 = \min \{  \frac{A_{g_1} \epsilon_1}{36 n^2 L \max |\partial \Gamma^{g_1}|} , \frac{A_{g_1} \epsilon_1}{24 n^3 L \max |\Gamma^{g_1}|^2} \}$ and let $\sigma_5 = \frac{A_{g_1}^2 \epsilon_1}{32 n^3 L^2 (2 \max |\Gamma^{g_1}| + 1)}$ then putting everything back together, we have

\begin{equation*}
    B, C \leq I + II < \frac{\epsilon_1}{4}
\end{equation*}

Then for $D$, we have

\begin{align*}
    D & = |\dot{\gamma}_1^\mu \dot{\gamma}_1^\nu  R_{j \nu \mu }^{k, g_1} (\gamma_1 (t)) - \dot{\gamma}_2^\mu \dot{\gamma}_2^\nu  R_{j \nu \mu}^{k, g_2} (\gamma_2 (t)) |\\
    & \leq |\dot{\gamma}_1^\mu \dot{\gamma}_1^\nu R_{j \nu \mu}^{k, g_1} (\gamma_1 (t)) - \dot{\gamma}_1^\mu \dot{\gamma}_1^\nu R_{j \nu \mu}^{k, g_1} (\gamma_2 (t))|\\
    & + | \dot{\gamma}_1^\mu \dot{\gamma}_1^\nu R_{j \nu \mu}^{k, g_1} (\gamma_2 (t)) - \dot{\gamma}_2^\mu \dot{\gamma}_2^\nu R_{j \nu \mu}^{k, g_1} (\gamma_2 (t))|\\
    & + | \dot{\gamma}_2^\mu \dot{\gamma}_2^\nu R_{j \nu \mu}^{k, g_1} (\gamma_2 (t)) - \dot{\gamma}_2^\mu \dot{\gamma}_2^\nu R_{j \nu \mu}^{k, g_2} (\gamma_2 (t))|\\
    & = I+II+III = \star
\end{align*}

Then by breaking this into pieces, first we have

\begin{align*}
    I & \leq |\dot{\gamma}_1^\mu \dot{\gamma}_1^\nu  R_{j \nu \mu}^{k, g_1} (\gamma_1 (t)) - \dot{\gamma}_1^\mu \dot{\gamma}_1^\nu R_{j \nu \mu}^{k, g_1} (\gamma_2 (t))|\\
    & \leq | \dot{\gamma}_1^\mu | |\dot{\gamma}_1^\nu| \max|\partial R^{g_1}| |\gamma_1 (t) - \gamma_2 (t)|\\
    & \leq \frac{n^2L^2}{A_{g_1}^2} \max|\partial R^{g_1}| |\gamma_1 (t) - \gamma_2 (t)|\\
    & < \frac{n^2L^2}{A_{g_1}^2} \max |\partial R^{g_1}| \sigma_1
\end{align*}
where $\max |\partial R^{g_1}|$ was defined in remark~\ref{def:rmk2}.

Next, we have

\begin{align*}
    II & \leq |\dot{\gamma}_1^\mu \dot{\gamma}_1^\nu  R_{j \nu \mu}^{k, g_1}(\gamma_2 (t)) -  \dot{\gamma}_2^\mu \dot{\gamma}_2^\nu   R_{j \nu \mu}^{k, g_1}(\gamma_2 (t))|\\
    & \leq \max |R^{g_1}| |\dot{\gamma}_1^\mu \dot{\gamma}_1^\nu  - \dot{\gamma}_2^\mu \dot{\gamma}_2^\nu |\\
    & \leq \max |R^{g_1}| (| \dot{\gamma}_1^\mu ||\dot{\gamma}_1^\nu  - \dot{\gamma}_2^\nu | + |\dot{\gamma}_2^\nu ||\dot{\gamma}_1^\mu - \dot{\gamma}_2^\mu |)\\
    & \leq \max |R^{g_1}| ( \frac{L}{A_{g_1}} |\dot{\gamma}_1^\nu  - \dot{\gamma}_2^\nu | + \frac{2L}{A_{g_1}} |\dot{\gamma}_1^\mu - \dot{\gamma}_2^\mu |)\\
    & \leq \frac{n^2 3 L}{A_{g_1}} \max |R^{g_1}| \sigma_2
\end{align*}
where $\max |R^{g_1}|$ is defined in remark~\ref{def:rmk2}.

Then, finally

\begin{align*}
    III & \leq |\dot{\gamma}_2^\mu \dot{\gamma}_2^\nu  R_{j \nu \mu}^{k, g_1}(\gamma_2 (t)) -  \dot{\gamma}_2^\mu \dot{\gamma}_2^\nu  R_{j \nu \mu }^{k, g_2}(\gamma_2 (t))|\\
    & = |\dot{\gamma}_2^\mu | | \dot{\gamma}_2^\nu| |R_{j \nu \mu}^{k, g_1}(\gamma_2 (t)) - R_{j \nu \mu}^{k, g_2}(\gamma_2 (t))|\\
    & \leq \frac{4L^2}{A_{g_1}^2} |R_{j \nu \mu}^{k, g_1}(\gamma_2 (t)) - R_{j \nu \mu}^{k, g_2}(\gamma_2 (t))|\\
    & \leq \frac{4n^2L^2}{A_{g_1}^2} \sigma_4
\end{align*}

Then, by putting these pieces back together, we let $\sigma_1 = \min \{\frac{A_{g_1}^2 \epsilon_1}{24 n^2 L^2 \max |\partial^2 \Gamma^{g_1}|},  \frac{A_{g_1}^2 \epsilon_1}{12n^2L^2 \max |\partial R^{g_1}|} \}$, $\sigma_2 = \min \{  \frac{A_{g_1} \epsilon_1}{36 n^2 L \max |\partial \Gamma^{g_1}|} , \frac{A_{g_1} \epsilon_1}{24 n^3 L \max |\Gamma^{g_1}|^2}, \frac{A_{g_1} \epsilon_1}{36n^2L \max |R^{g_1}|} \}$, and $\sigma_4 = \frac{A_{g_1}^2 \epsilon_1}{48 n^2 L^2}$, then we have
\begin{equation*}
    D \leq I + II+ III < \frac{\epsilon_1}{4} 
\end{equation*}
Then we can put everything back together and we have that
\begin{equation*}
    A + B + C + D < \frac{\epsilon_1}{4} + \frac{\epsilon_1}{4} + \frac{\epsilon_1}{4} + \frac{\epsilon_1}{4} = \epsilon_1
\end{equation*}
So, by letting $\delta_\alpha (M,g_1,U,L, \min \{ \sigma_1, \sigma_2, \sigma_3, \sigma_4 \}) = \delta_2 (M,g_1,U,L, \epsilon_1)$, and $\parallel g_1 - g_2 \parallel_{C^2 (M,g_1)} < \delta_2 (M,g_1,U,L, \epsilon_1)$ implies that on $U$, we have
\begin{equation*}
    |b_{kj}^{g_1}(t) - b_{kj}^{g_2}(t)| < \epsilon_1
\end{equation*}

Then, we know that $\delta (M, g_1, U, L, \min \{ \chi_1, \chi_2 \}) = \min \{ \delta_1, \delta_2 \}>0$, is such that $\parallel g_1 - g_2 \parallel_{C^2 (M,g_1)} < \delta (M, g_1, U, L, \min \{ \chi_1, \chi_2 \})$ implies that

\begin{align*}
    |a_{kj}^{g_1}(t) - a_{kj}^{g_2}(t)| & < \chi_1\\
    |b_{kj}^{g_1}(t) - b_{kj}^{g_2}(t)| & < \chi_2
\end{align*}

Then

\begin{align*}
    |A_1 (t) - A_2 (t)|+ |B_1 (t) - B_2 (t)| & \leq |a_{kj}^{g_1}(t) - a_{kj}^{g_2}(t)| + |b_{kj}^{g_1}(t) - b_{kj}^{g_2}(t)|\\
    & < n^2 (\chi_1 + \chi_2)
\end{align*}

So, let $\zeta = n^2 (\chi_1 + \chi_2)$, and we'll assume that $\chi_1, \chi_2 \leq 1$ (we'll add more constraints to $\chi_1$ and $\chi_2$). Then from lemma~\ref{def:example3}, we'll have
\begin{align*}
    |y_1 (t) - y_2 (t)| & \leq (|y_1 (0) - y_2 (0)| + C_0 \zeta)e^{C_1+ 2n^2}\\
    & \leq (|f_1 (0) - f_2 (0)| + |f_1^\prime (0) - f_2^\prime (0)|+C_0 \zeta)e^{C_1+2n^2}\\
    & = C_0 \zeta e^{C_1 +2n^2}\\
    & = C_0 n^2 (\chi_1 + \chi_2)e^{C_1+2n^2}
\end{align*}
Then let $\chi_1 = \chi_2 = \min \{ \frac{\epsilon}{2n^2 C_0 e^{C_1 +2n^2}} , 1\}$, we have

\begin{align*}
        |f_1 (t) - f_2 (t)| \leq |y_1 (t) - y_2 (t)| & < C_0 n^2 (\chi_1 + \chi_2)e^{C_1+2n^2}\\
        & \leq C_0 n^2 (\frac{\epsilon}{2n^2 C_0 e^{C_1 +2n^2}} + \frac{\epsilon}{2n^2 C_0 e^{C_1 +2n^2}}) e^{C_1+2n^2} \\
        & = \frac{\epsilon}{2} + \frac{\epsilon}{2}= \epsilon
\end{align*}

as desired.
 
\end{proof}

\subsection{Proof of Lemma \ref{lemma:lemma3}}

\phantomsection\label{app:lemma3}
\begin{proof}[Proof of Lemma \ref{lemma:lemma3}]

\begin{enumerate}

\item  First, we know that $exp_p^{g_1}: ID^{g_1}(p) \to M \setminus Cut_{g_1}(p)$ is a diffeomorphism, and $\overline{U}_1 \subset \overline{U}_2 \subset M \setminus Cut_{g_1}(p)$. Let $exp_p^{g_1,-1}(U_i)=V_i$ and we see that $\overline{V}_1 \subset \overline{V}_2 \subset ID^{g_1}(p)$ and $\overline{V}_i$ are compact.

More specifically, $\overline{V}_2$ is compact in $T_pM$, and there's a bound $L > 0$ such that for all $w \in \overline{V}_2$, we have $|w|_{g_1 (p)}<L$. From proposition~\ref{prop:neat} there's a $\delta_0 > 0$, such that $\parallel g_1 - g_2 \parallel_{C^2 (M,g_1)} < \delta_0$ implies that $|v|_{g_2 (p)} \leq 2|v|_{g_1(p)}$, then for all $v \in \overline{V}_2$, let $L_0 = 2L +1$, and we have $|v|_{g_m(p)} < L_0$. 

Since $\overline{U}_1$ is $g_1$-geodesically star-shaped with respect to $p$, then from remark~\ref{rmk:star}, we see that for all $w \in \overline{V}_1$, on the interval $[0,1]$, where $\eta_{1w}(t) = exp_p^{g_1}(tw)$, then $\eta_{1w} ([0,1]) \subset \overline{U}_1 \subset U_2$. So, we want to show that there exists a $\delta_1 (M,g_1,U_2, L_0)>0$ such that $\parallel g_1 - g_2 \parallel_{C^2 (M,g_1)} < \delta_1 (M,g_1,U_2, L_0)$ implies that for $\eta_{2w}(t) = exp_p^{g_2}(tw)$, then $\eta_{2w}([0,1]) \subset U_2$.

Since $\overline{U}_1$ is compact and $\overline{U}_1 \subset U_2$, then let $\Omega = d_{g_1} (\overline{U}_1, \partial U_2)$. So, for all $q \in \overline{U}_1$, then $B_{g_1}(q, \Omega) \subset U_2$. Then from corollary~\ref{def:cor1}, there exists $\delta_1 (M,g_1,U_2,L_0)>0$ such that $\parallel g_1 - g_2 \parallel_{C^2 (M,g_1)} < \delta_1 (M,g_1,U_2, L_0)$ implies that for all $t \in (0,1]$ such that $\eta_{2w}([0,t]) \subset U_2$, we have
\begin{equation*}
    d_{g_1} (\eta_{1w}(t), \eta_{2w}(t)) < \frac{\Omega}{2}
\end{equation*}
Then let $S = \{ t \in [0,1] \ : \ \eta_{2w}([0,t]) \subset U_2 \}$. Since $\eta_{2w}(0)=p$, by its continuity we know that there exists $t_0 \in (0,1]$ such that $\eta_{2w}([0,t_0]) \subset U_2$. So, $S$ is nonempty. Let $t^* = \sup S$, then there exists $t_n \in S$ such that $t_n \to t^*$. For each $t_n$, we have $\eta_{2w}([0,t_n]) \subset U_2$, then 
\begin{equation*}
    d_{g_1} (\eta_{1w} (t_n), \eta_{2w} (t_n)) < \frac{\Omega}{2}
\end{equation*}
Since both $\eta_{1w}$ and $\eta_{2w}$ are geodesics, hence continuous on $[0,1]$, and $d_{g_1}: M \times M \to \mathbb{R}$ is continuous, then by taking $n \to \infty$, we have
\begin{equation*}
    d_{g_1} (\eta_{1w} (t_n), \eta_{2w} (t_n)) \to d_{g_1} (\eta_{1w} (t^*), \eta_{2w} (t^*))
\end{equation*}
Then we have
\begin{equation*}
    d_{g_1} (\eta_{1w} (t^*), \eta_{2w} (t^*)) \leq \frac{\Omega}{2} < \Omega
\end{equation*}
Since $\eta_{1w}(t^*) \in \overline{U}_1$ and $B_{g_1}(\eta_{1w}(t^*), \Omega) \subset U_2$, this means that $\eta_{2w}(t^*) \in U_2$. Combining this with $\eta_{2w}([0,t^*)) \subset U_2$, we get $\eta_{2w}([0,t^*]) \subset U_2$, i.e. $t^* \in S$.  Now suppose that $t^* <1$. Since we just showed that $\eta_{2w}(t^*) \in U_2$ and $U_2$ is open, there exists $r >0$ with $B_{g_1}(\eta_{2w}(t^*),r) \subset U_2$. Then by the continuity of $\eta_{2w}$, there exists $\epsilon_0 >0$ such that $d_{g_1}(\eta_{2w}(t), \eta_{2w}(t^*)) < r$ for all $t \in [t^*, t^* + \epsilon_0 )$, hence $\eta_{2w}(t) \in U_2$ for those $t$ too. That extends containment to $[0, t^*+ \epsilon_0] \subset S$, contradicting $t^* = \sup S$. So, for all $w \in \overline{V}_1$, then $\parallel g_1 - g_2 \parallel_{C^2 (M,g_1)} < \delta_1 (M,g_1,U_2,L_0)$ implies that $\eta_{2w}([0,1]) \subset U_2$.

Let $dist_{g_1 (p)} (\overline{V}_1, \partial V_2) = \mathcal{R}_0$, then for all $v \in \overline{V}_1$, we have $B_{\frac{\mathcal{R}_0}{2}}^{g_1}(v) \subset V_2$. Since $\frac{\mathcal{R}_0}{2} < \mathcal{R}_0 \leq dist_{g_1 (p)} (v, \partial V_2)$ for every $v \in \overline{V}_1$, then $\overline{B_{\frac{\mathcal{R}_0}{2}}^{g_1}(v)} \subset V_2$ as well. Let $V_3 = \cup_{v \in \overline{V}_1} B_{\frac{\mathcal{R}_0}{2}}^{g_1}(v)$; then $\overline{V}_3 = \cup_{v \in \overline{V}_1} \overline{B_{\frac{\mathcal{R}_0}{2}}^{g_1}(v)} \subset V_2$. 

Now we want to show that $\overline{V}_3$ is star-shaped with respect to $0$. Let $u \in \overline{V}_3$, by how we defined $\overline{V}_3$, there exists a $v \in \overline{V}_1$ such that $|u-v|_{g_1 (p)} \leq \frac{\mathcal{R}_0}{2}$. Fix $t \in [0,1]$. Since $\overline{U}_1$ is $g_1$-geodesically star-shaped with respect to $p$, by remark~\ref{rmk:star} we know that $\overline{V}_1$ is star-shaped with respect to $0$, so $tv \in \overline{V}_1$. Then we have
\begin{equation*}
    |tu - tv|_{g_1 (p)} = t|u - v|_{g_1 (p)} \leq |u - v|_{g_1 (p)} \leq \frac{\mathcal{R}_0}{2}
\end{equation*}
for $t \in [0,1]$. Hence $tu \in \overline{B_{\frac{\mathcal{R}_0}{2}}^{g_1}(tv)} \subset \overline{V}_3$, since $tv \in \overline{V}_1$. Thus, $\overline{V}_3$ is star-shaped with respect to $0$. Then, by remark~\ref{rmk:star}, for $\overline{U}_3 : = exp_p^{g_1}(\overline{V}_3)$, $\overline{U}_3$ is $g_1$-geodesically star-shaped with respect to $p$. Then from corollary~\ref{def:cor1}, and what we just showed for $\overline{U}_1$, there exists $\delta_2 (M,g_1,U_2, L_0)>0$ such that $\parallel g_1 - g_2 \parallel_{C^2 (M,g_1)}< \delta_2 (M,g_1,U_2, L_0)$ implies that for all $w \in \overline{V}_3$, then $\eta_{2w}([0,1]) \subset U_2$.

Then, we want to find $R_0 (U_1,U_2)>0$ such that $exp_p^{g_1} \big|_{B_{R_0 }^{g_1}(v)}$ is diffeomorphic to its image, $ exp_p^{g_m} (B_{R_0 }^{g_1}(v)) \subset U_2$, and for all $w \in B_{R_0}^{g_1}(v)$ then $|w|_{g_m (p)} <L_0$. Then let $R_0 = \frac{\mathcal{R}_0}{2}$. Since $B_{R_0}^{g_1}(v) \subset V_2 \subset ID^{g_1}(p)$ for all $v\in \overline{V}_1$, and $exp_p^{g_1}$ is a diffeomorphism on $ID^{g_1}(p)$, its restriction to $B_{R_0}^{g_1}(v)$ is a diffeomorphism to its image. Then for all $v \in \overline{V}_1$, $exp_p^{g_1} ( B_{R_0}^{g_1}(v)) \subset U_2$, and for all $w \in B_{R_0}^{g_1}(v)$, $|w|_{g_m(p)} < L_0$.

Then we're interested in the geodesics $\eta_{2w} (t) = exp_p^{g_2} (tw)$ for $ w \in V_3$. From what we showed above, we have $\eta_{2w}([0,1]) \subset U_2$. Since $\overline{V}_3 \subset V_2$, then for all $w \in \overline{V}_3$, we have $|w|_{g_1 (p)} < L_0$. Then from theorem~\ref{thrm:neat3}, we know that there exists $\delta_3 (M, g_1, U_2)>0$ such that $\parallel g_1 - g_2 \parallel_{C^2 (M, g_1)}< \delta_3 (M, g_1, U_2)$ implies that for all $w \in V_3$, the geodesic $\eta_{2w} (t)$ will have no conjugate points for $t \leq \frac{\pi}{2 L_0 \sqrt{C(g_1)+1}}$. Then let $a = \min \{ \frac{1}{2}, \frac{\pi}{2L_0 \sqrt{C(g_1)+1}} \}$. This means that $\parallel g_1 - g_2 \parallel_{C^2 (M, g_1)}< \delta_3 (M, g_1, U_2)$ implies that for $w \in V_3$ and $x \in T_pM$ the Jacobi field $J_{2wx} (t)$ will be non-vanishing on $(0,a]$.

Then for all $v \in \overline{V}_1 \subset ID^{g_1}(p)$ and all nonzero $w \in T_pM$, $d(exp_p^{g_1})_v(w)$ is continuous with respect to $v$, and the Jacobi field $J_{1vw}(t)$ is non-vanishing for $t \in (0,1]$. Now, let $S^{g_1}(0) = \{x \in T_pM \ | \ |x|_{g_1 (p)}=1 \}$, then let

\begin{equation*}
    \lambda := \min_{v \in \overline{V}_1, u \in S^{g_1}(0), t \in [a,1]} |f_{1vu}(t)| >0
\end{equation*}

Then for all $v,w \in V_2$, we know that $|v|_{g_m(p)}, |w|_{g_m(p)} < L_0$, and we see that $|v-w|_{g_1 (p)} < 2L < L_0$. So, from lemma~\ref{lemma:ineq2}, for all $u \in S^{g_1}(0)$, the Jacobi fields $J_{1vu}(t)$ and $J_{1wu}(t)$ are such that
\begin{equation*}
    |f_{1vu}(t) - f_{1wu}(t)| < K_0 |v-w|_{g_1 (p)}
\end{equation*}

Then let $\mathcal{R}_1 = \frac{\lambda}{K_0 +1}$, and let $R_1 = \min \{ R_0, \mathcal{R}_1 \}$. Then for all $v \in \overline{V}_1$, for all $w \in B_{R_1}^{g_1}(v)$, and for all $u \in S^{g_1}(0)$, we have
\begin{equation*}
    |f_{1vu}(t) - f_{1wu}(t)| < \frac{K_0 \lambda}{K_0+1}
\end{equation*}
Then by letting $\epsilon = \lambda - \frac{K_0 \lambda}{K_0+1} >0$ then from theorem~\ref{thrm:neat1}, there exists $\delta_4 (M, g_1, U_2, L_0)>0$ such that $\parallel g_1 - g_2 \parallel_{C^2 (M,g_1)} < \delta_4 (M, g_1, U_2, L_0)$ implies that for all $v \in \overline{V}_1$, for all $w \in B_{R_1}^{g_1}(v)$, and for all $u \in S^{g_1}(0)$, we have
\begin{align*}
    |f_{1vu}(t) - f_{2wu}(t)| & \leq |f_{1vu}(t) - f_{1wu}(t)| + |f_{1wu}(t) - f_{2wu}(t)|\\
    & < \frac{K_0 \lambda}{K_0 + 1} + \lambda - \frac{K_0 \lambda}{K_0 + 1} = \lambda
\end{align*}

Then, $\parallel g_1 - g_2 \parallel_{C^2 (M,g_1)} < \delta_4 (M, g_1, U_2, L_0)$ implies that on the interval $[a,1]$, we have
\begin{align*}
    |f_{2wu} (t)| & \geq |f_{1vu}(t)| - |f_{1vu}(t) - f_{2wu}(t)|\\
    & \geq \lambda - |f_{1vu}(t) - f_{2wu}(t)|\\
    & > \lambda - \lambda =0
\end{align*}
So, $|f_{2wu}(t)|>0$, on $[a,1]$.

Now choose nonzero $x \in T_pM$, and let $y = \frac{x}{|x|_{g_1 (p)}}$, then $y \in S^{g_1}(0)$. Then, for all $v \in \overline{V}_1$ and for all $w \in B_{R_1}^{g_1}(v)$, the Jacobi field $J_{2wy}(t)$ is non-vanishing on $[a,1]$. Then we see that on $[a,1]$
\begin{align*}
    J_{2wx} (t) & = d (exp_p^{g_2})_{tw} (tx)\\
    & = |x|_{g_1 (p)} d (exp_p^{g_2})_{tw} (t \frac{x}{|x|_{g_1 (p)} })\\
    & = |x|_{g_1 (p)} d (exp_p^{g_2})_{tw} (t y )\\
    & = |x|_{g_1 (p)} J_{2wy} (t)
\end{align*}
Then $J_{2wx}(t)$ must be non-vanishing on $[a,1]$. Now let $\delta_5 = \min \{ \delta_0, \delta_1, \delta_2, \delta_3, \delta_4 \}$ and $R_2 = \min \{ R_0, R_1 \}$, then $\parallel g_1 - g_2 \parallel_{C^2 (M, g_1)} < \delta_5 (M,g_1,U_2,L_0)$ implies that for all $v \in \overline{V}_1$, for all $w \in B_{R_2}^{g_1}(v)$, which are contained in the injectivity domain, for all non-zero $x \in T_pM$ the Jacobi fields $J_{mwx}(t)$ are non-vanishing. This allows us to conclude that $\parallel g_1 - g_2 \parallel_{C^2 (M,g_1)}< \delta_5 (M,g_1,U_2,L_0)$ implies that for all $v \in \overline{V}_1$ and for all $w \in B_{R_2}^{g_1}(v)$, the linear maps $d(exp_p^{g_m})_w$ are invertible.

From here, we'll modify the proof of the inverse function theorem, (theorem C.34, p. 657,~\cite{lee2003smooth}). Since $\overline{V}_1 \subset \overline{V}_3 \subset V_2 \subset ID^{g_1} (p)$, then $d (exp_p^{g_1})$ is continuous on $V_2$ and therefore uniformly continuous on $\overline{V}_3$. So, for all $\epsilon >0$ there exists $\mathcal{R}_2>0$ such that for all $v \in \overline{V}_1$, $w_1, w_2 \in B_{\mathcal{R}_2}^{g_1}(v)$ implies that

\begin{equation*}
    \parallel d (exp_p^{g_1})_{w_1} - d (exp_p^{g_1})_{w_2} \parallel < \epsilon
\end{equation*}

Additionally, since $\overline{V}_3$ is compact, then

\begin{equation*}
    \sup_{v \in \overline{V}_3} \parallel d (exp_p^{g_1})_{v}  \parallel = C_1 < \infty
\end{equation*}

Now since $T_pM \cong \mathbb{R}^n$, we'll let the isomorphism be $d \phi_p$ (for the chart $(V, \phi)$ with $\overline{U}_2 \subset V$ from the hypothesis) then let $g^* = (d \phi_p)_* g_1 (p)$ be the pushforward of $g_1 (p)$ via $d \phi_p$, and then $B_{R_0}^{g^*} (v_0) = d \phi_p ( B_{R_0}^{g_1}(v) )$, where $v_0 = d \phi_p (v)$.
Then, define $F_{m}: (\mathbb{R}^n, g^*) \to \mathbb{R}^n$ such that $F_{m} = \phi \circ exp_p^{g_m} \circ d \phi_p^{-1}$. We see that $\parallel g_1 - g_2 \parallel_{C^2 (M,g_1)}< \delta_5 (M,g_1,U_2,L_0)$ implies that for all $v \in \overline{V}_1$, and for all $w \in B_{R_2}^{g_1}(v)$, $DF_{m}(w_0)$ is invertible.

Now for the differentials, or any linear map we're dealing with, for $A \in M(n, \mathbb{R})$, we'll define
\begin{equation*}
    \parallel A \parallel = \sup_{x \in \mathbb{R}^n, x \neq 0} \frac{|Ax|}{|x|_{g^*}}
\end{equation*}
where $|\cdot|_{g^*} = \sqrt{ g^* (\cdot , \cdot) }$ is the norm on $\mathbb{R}^n$ induced by $g^*$ (defined above), and $|\cdot|$ is the Euclidean norm from definition~\ref{def:eucl}.

Then, accounting for this change by the smoothness and therefore the continuity of $\phi$, for all $\epsilon >0$ there exists $\mathcal{R}_3$ such that for all $v \in \overline{V}_1$, with $x,y \in B_{\mathcal{R}_3}^{g^*}(v_0)$ implies that

\begin{equation*}
    \parallel DF_{1}(x) - DF_{1}(y) \parallel < \epsilon
\end{equation*}
Additionally, for all $w_0 \in B^{g^*}_{R_0}(v_0)$, we have $d \phi_p^{-1}(w_0)=w \in \overline{V}_3$, since $B_{R_0}^{g^*}(v_0) = d\phi_p(B_{R_0}^{g_1}(v)) \subset d \phi_p(\overline{V}_3)$. Then we see that
\begin{equation*}
    \parallel D F_{1}(w_0) \parallel \leq  C_1 < \infty
\end{equation*}
where $\parallel \cdot \parallel$ is the norm defined above.

By combining what we've just shown with corollary~\ref{cor:neat1} we know that for arbitrary $\zeta_1 >0$, there exists $\delta_6 (M, g_1, U_2, L_0) >0$ such that $\parallel g_1 - g_2 \parallel_{C^2 (M, g_1)} < \delta_6 (M, g_1, U_2, L_0)$ implies that for all $v \in \overline{V}_1$ and for all $w_0 \in B_{R_1}^{g^*} (v_0)$, we have
\begin{equation*}
    \parallel DF_{1}(w_0) - DF_{2} (w_0) \parallel < \zeta_1
\end{equation*}

Fix $v \in \overline{V}_1$. Next, let's define $G_{m}: B_{R_0}^{g^*} (0) \to \mathbb{R}^n$ such that $G_{m}(x) = F_{m}(x+v_0) - F_{m}(v_0)$. We see that $G_{m}(0)=0$, $G_{m}$ is injective if and only if $F_{m}$ is injective if and only if $exp_p^{g_m}$ is injective. Since
\begin{equation*}
    DG_{m} (x) = DF_{m}(x+v_0)
\end{equation*}
then for all $v \in \overline{V}_1$ and $x \in B_{R_0}^{g^*}(0)$, $DG_{m} (x)$ is invertible, and
\begin{equation*}
    \parallel DG_{1}(x) - DG_{2}(x) \parallel = \parallel DF_{1}(x+v_0) - DF_{2}(x+v_0) \parallel < \zeta_1
\end{equation*}

From here we want to use this inequality to show that for all $\epsilon_0 >0$, there exists $\delta_6 (M, g_1, U_2, L_0) > 0$ such that $\parallel g_1 - g_2 \parallel_{C^2 (M, g_1)} < \delta_6 (M, g_1, U_2, L_0)$ implies 
\begin{equation*}
    \parallel (DG_{1}(0))^{-1} - (DG_{2}(0))^{-1} \parallel < \epsilon_0
\end{equation*}
for all $v \in \overline{V}_1$. 

Let $\epsilon_0>0$. By proposition B.36 (p. 634, \cite{lee2003smooth}) matrix inversion is smooth and is therefore continuous. So, for all $v \in \overline{V}_1$, we have
\begin{equation*}
     \parallel (DG_1(0))^{-1} \parallel = \parallel (DF_{1}(v_0))^{-1} \parallel \leq  \sup_{v \in \overline{V}_1} \parallel (DF_{1}(v_0))^{-1} \parallel = C_2 < \infty
\end{equation*}
we also have that for all $v \in \overline{V}_1$, and all $x \in B_{R_1}^{g^*}(0)$, 

\begin{equation*}
    \parallel DG_{1}(x) - DG_{2} (x) \parallel < \zeta_1
\end{equation*}

So, let $k (x) = DG_{2}(x) - DG_{1}(x)$, and let $0 < \zeta_1 \leq \frac{1}{2C_2}$ (we'll put more constraints on $\zeta_1$ as we go). Then we see that

\begin{align*}
    \parallel (DG_{1}(0))^{-1}  k (0) \parallel & \leq \parallel (DG_{1}(0))^{-1} \parallel \cdot \parallel k(0) \parallel\\
    & < \frac{\parallel (DG_{1}(0))^{-1} \parallel}{2 C_2} \leq \frac{1}{2} < 1
\end{align*}

Then by the geometric series expansion for $(I_n + (DG_{1}(0))^{-1} k (0))^{-1}$, (lemma 2.3.3, p.58,~\cite{golub1996matrix}) we have
\begin{equation*}
    (I_n + (DG_{1}(0))^{-1} k (0))^{-1} = \sum_{i=0}^\infty (-1)^i ( (DG_{1}(0))^{-1} k (0))^i
\end{equation*}
which converges since $\parallel (DG_{1}(0))^{-1} k(0) \parallel <1$ and $I_n + (DG_{1}(0))^{-1} k (0)$ is invertible. Then we have
\begin{align*}
    \parallel (DG_{2}(0))^{-1} \parallel & = \parallel   (I_n + (DG_{1}(0))^{-1} k (0))^{-1} \cdot (DG_{1}(0))^{-1} \parallel\\
    & \leq  \parallel (I_n + (DG_{1}(0))^{-1} k (0))^{-1}  \parallel \cdot  \parallel  (DG_{1}(0))^{-1} \parallel\\
    & =  \parallel \sum_{i=0}^\infty (-1)^i ((DG_{1}(0))^{-1} k(0))^i \parallel \cdot \parallel  (DG_{1}(0))^{-1} \parallel\\
    & \leq  \bigg( \sum_{i=0}^\infty \parallel (DG_{1}(0))^{-1} k(0) \parallel^i\bigg) \cdot \parallel  (DG_{1}(0))^{-1} \parallel\\
    & \leq  \bigg( \sum_{i=0}^\infty \big( \frac{1}{2} \big)^i\bigg) \cdot \parallel  (DG_{1}(0))^{-1} \parallel\\
    & = 2 \cdot \parallel  (DG_{1}(0))^{-1} \parallel \leq 2C_2
\end{align*}
Now, let $\zeta_1 = \min \{ \frac{1}{2C_2}, \frac{\epsilon_0}{2 C_2^2} \}$, and we have
\begin{align*}
    \parallel (DG_{1}(0))^{-1} -  (DG_{2}(0))^{-1} \parallel & = \parallel (DG_{1}(0))^{-1}(DG_{2} (0) - DG_{1} (0)) (DG_{2}(0))^{-1} \parallel\\
    & \leq \parallel (DG_{1}(0))^{-1} \parallel \cdot \parallel ( DG_{2} (0) - DG_{1} (0)) \parallel \cdot \parallel (DG_{2}(0))^{-1} \parallel\\
    & \leq 2 C_2^2 \cdot \parallel ( DG_{2} (0) - DG_{1} (0) \parallel\\
    & < 2 C_2^2 \zeta_1 \leq 2 C_2^2 \frac{\epsilon_0}{2 C_2^2}= \epsilon_0
\end{align*}

Then define $P_{m}: B_{R_2}^{g^*} (0) \to \mathbb{R}^n$ such that $P_{m}(x) = (DG_{m}(0))^{-1} G_{m}(x)$, and we see that $DP_{m}(0) = I_n$ and $P_{m}(0)=0$. We want to show that for all $\epsilon_1 >0$, there exists $\delta_7 (M, g_1, U_2, L_0)>0$ such that $\parallel g_1 - g_2 \parallel_{C^2 (M, g_1)} < \delta_7 (M, g_1, U_2, L_0)$ implies that for all $v \in \overline{V}_1$ and all $x \in B_{R_1}^{g^*}(0)$, we have
\begin{equation*}
    \parallel DP_1 (x) - DP_2 (x) \parallel < \epsilon_1
\end{equation*}

From what we've shown so far, we know that for arbitrary $\zeta_2 >0$ and $\zeta_3 >0$, there's a $\delta_7 (M, g_1, U_2, L_0)>0$ such that $\parallel g_1 - g_2 \parallel_{C^2 (M, g_1)} < \delta_7 (M, g_1, U_2, L_0)$ implies that for all $v \in \overline{V}_1$, we have
\begin{equation*}
        \parallel DG_{1}(x) -  DG_{2} (x) \parallel < \zeta_2
\end{equation*}
for all $x \in B_{R_1}^{g^*} (0)$, and
\begin{equation*}
    \parallel (DG_{1}(0))^{-1} -  (DG_{2} (0))^{-1} \parallel < \zeta_3
\end{equation*}
Then we have 
\begin{align*}
    \parallel DP_{1}(x) -  DP_{2} (x) \parallel & = \parallel (DG_{1}(0))^{-1} DG_{1}(x) - (DG_{2}(0))^{-1} DG_{2}(x) \parallel\\
    & \leq \parallel DG_{1} (x) \parallel \cdot \parallel (DG_{1}(0))^{-1} -  (DG_{2} (0))^{-1}\parallel\\
    & + \parallel (DG_{2}(0))^{-1} \parallel \cdot \parallel DG_{1}(x) - DG_{2} (x) \parallel\\
    & < C_1 \zeta_3 + 2C_2\zeta_2
\end{align*}

So, let $\zeta_3 =  \frac{\epsilon_1}{2C_1}$, and $\zeta_2 = \frac{\epsilon_1}{4 C_2}$. Now, we know that there exists $\delta_8 (M,g_1,U_2, L_0)>0$ such that $\parallel g_1 - g_2 \parallel_{C^2 (M, g_1)} < \delta_8 (M, g_1, U_2, L_0)$ implies that for all $v \in \overline{V}_1$ and all $x \in B_{R_1}^{g^*}(0)$
\begin{equation*}
    \parallel DP_{1}(x) -  DP_{2} (x) \parallel < \zeta_4
\end{equation*}
where $\epsilon_1 = \zeta_4$.

Then define $H_{m} : B_{R_2}^{g^*}(0) \to \mathbb{R}^n$, such that $H_{m} (x) = x - P_{m}(x)$. Then we see that $DH_{m} (0) = I_n - DP_{m}(0) = I_n - I_n =0$. Additionally, we see that for all $v \in \overline{V}_1$ and all $x \in B_{R_2}^{g^*}(0)$, we have
\begin{align*}
    \parallel DH_{1} (x) - DH_{2} (x) \parallel & = \parallel  I_n - DP_{1} (x) - (I_n - DP_{2}(x))\parallel\\
    & = \parallel DP_{1}(x) -  DP_{2} (x) \parallel < \zeta_4
\end{align*}

Then from above if we let $\epsilon = \frac{1}{4 C_2}$, then there exists $\mathcal{R}_2 >0$ such that for all $v \in \overline{V}_1$, for all $u,w \in B_{\mathcal{R}_2}^{g_1}(v)$, we have $\parallel d(exp_p^{g_1})_w  - d (exp_p^{g_1})_u \parallel < \epsilon$. Then, by how we defined $F$ for our fixed $v$, there exists $R_3 >0$, such that $x,y \in B_{R_3}^{g^*}(v_0)$ implies that
\begin{equation*}
    \parallel DF_{1} (x) - DF_{1} (y) \parallel < \frac{1}{4 C_2}
\end{equation*}
Then for all $v \in \overline{V}_1$, $x,y \in B_{R_3}^{g^*}(0) $ implies that

\begin{align*}
    \parallel DP_{1}(x) - DP_{1}(y) \parallel & \leq \parallel (DG_{1} (0))^{-1} \parallel \cdot \parallel DG_{1}(x) - DG_{1}(y) \parallel\\
    & \leq C_2 \parallel DF_{1}(x+v_0) - DF_{1}(y+v_0) \parallel\\
    & < C_2 \frac{1}{4 C_2} = \frac{1}{4}
\end{align*}

Then for all $v \in \overline{V}_1$, $x,y \in B_{R_3}^{g^*}(0)$, we have

\begin{align*}
    \parallel DH_{1}(x) - DH_{1}(y) \parallel & = \parallel I_n-DP_{1}(x) - (I_n - DP_{1}(y)) \parallel\\
    & =  \parallel DP_{1}(x) - DP_{1}(y) \parallel\\
    & <   \frac{1}{4}
\end{align*}

Then for all $v \in \overline{V}_1$ and all $x \in B_{R_3}^{g^*}(0)$, we have
\begin{equation*}
    \parallel DH_{1}(x) \parallel = \parallel DH_{1}(x) - DH_{1}(0) \parallel < \frac{1}{4}
\end{equation*}

Let $R = \min \{ R_1, R_2, R_3 \}$, then for all $v \in \overline{V}_1$ and all $x \in B_R^{g^*}(0)$, we have

\begin{align*}
    \parallel DH_{2} (x) \parallel & \leq \parallel DH_{2} (x) - DH_{1} (x) \parallel + \parallel DH_{1} (x) \parallel\\
    & < \zeta_4 + \frac{1}{4}
\end{align*}
So, let $\zeta_4 = \frac{1}{4}$, and for all $v \in \overline{V}_1$ and all $x \in B_R^{g^*} (0)$, we have $\parallel DH_{m}(x) \parallel < \frac{1}{2}$. This means that for 

$\delta_9 (M,g_1, U_2, L_0) = \min \{ \delta_5, \delta_6, \delta_7, \delta_8 \}$, $\parallel g_1 - g_2 \parallel_{C^2 (M,g_1)} < \delta_9 (M,g_1, U_2, L_0)$ implies that  $\frac{1}{2}$ is a Lipschitz constant for $H_{m}$ for all $v \in \overline{V}_1$ on $B_R^{g^*}(0)$, i.e. for all $x,y \in B_R^{g^*}(0)$, we have
\begin{equation*}
    \parallel H_{m} (x) - H_{m} (y) \parallel_{g^*} < \frac{1}{2} \parallel x - y \parallel_{g^*}
\end{equation*}
Finally, we observe that
\begin{equation*}
    x - y = P_{m}(x) - P_{m}(y) + H_{m} (x)- H_{m}(y)
\end{equation*}
This means that
\begin{equation*}
    \begin{split}
        \parallel x - y \parallel_{g^*} & \leq \parallel P_{m}(x) - P_{m}(y) \parallel_{g^*} + \parallel H_{m} (x)- H_{m}(y) \parallel_{g^*}\\
        & < \parallel P_{m}(x) - P_{m}(y) \parallel_{g^*} + \frac{1}{2} \parallel x - y \parallel_{g^*}
    \end{split}
\end{equation*}
for which we can conclude that
\begin{equation*}
    \parallel x - y \parallel_{g^*}  \leq 2 \parallel P_{m}(x) - P_{m}(y) \parallel_{g^*} 
\end{equation*}
This means that on $B_R^{g^*}(0)$, $P_{m}$ is injective.  Which means that on $B_R^{g^*}(0)$, $G_{m}$ is injective. Which means that on $B_R^{g_1}(v)$, $exp_p^{g_m}$ must be injective. 

Then for all $v \in \overline{V}_1$ and $w \in B_{R}^{g_1}(v)$, we have the linear map $d(exp_p^{g_m})_w : T_w (T_pM) \to T_{exp_p^{g_m}(w)}M$, where $\dim (T_w(T_pM)) = \dim (T_pM)=n$ and by the Rank-Nullity Law (Corollary B.21, p. 627, ~\cite{lee2003smooth}), we have

\begin{equation*}
    \begin{split}
        \dim (\ker ( d(exp_p^{g_m})_w)) + \dim (Im( d(exp_p^{g_m})_w)) & = \dim (T_pM)\\
        0 + \dim (Im( d(exp_p^{g_m})_w)) & = n
    \end{split}
\end{equation*}

Since $w \in B_R^{g_1}(v)$ was arbitrary on $B_R^{g_1}(v)$, the map $exp_p^{g_m}$ has constant rank equal to $n$. Then from the global rank theorem (theorem 4.14, p. 83, ~\cite{lee2003smooth}), we know that $exp_p^{g_m} \bigg|_{B_R^{g_1} (v)}$ is a diffeomorphism.

Since $R_1$, $R_2$, and $R_3$ are fixed for all $v \in \overline{V}_1$ then $R$ is also fixed for $v \in \overline{V}_1$. Then for all $v \in \overline{V}_1$, then $exp_p^{g_m} \bigg|_{B_R^{g_1} (v)}$ is a diffeomorphism.

\item Let $R>0$ be the radius from our hypothesis. Then let $R_1 = \frac{R}{2}$, then $\overline{B_{R_1}^{g_1}(v)} \subset B_R^{g_1}(v)$ and $exp_p^{g_m} \big|_{\overline{B_{R_1}^{g_1} (v)}}$ is diffeomorphic to its image. From here, we want to show that there exists $\delta >0$ such that $\parallel g_1 - g_2 \parallel_{C^2(M,g_1)}< \delta$ implies that $exp_p^{g_1}(v) \in exp_p^{g_2} (B_{R_1}^{g_1}(v))$, for all $v \in W$. 

So assume, for contradiction, that for all $\delta >0$ there exists a Riemannian metric $g$ on $M$ with $\parallel g_1 -g \parallel_{C^2 (M,g_1)}< \delta$ such that for all $w \in W$, $exp_p^g \bigg|_{B_R^{g_1}(w)}$ is a diffeomorphism, and $exp_p^g ( B_R^{g_1}(w)) \subset U$, and such that there exists a vector $v \in W$  with $exp_p^{g_1}(v) \notin exp_p^{g}(B_{R_1}^{g_1}(v))$. Then by letting $\delta=\frac{1}{i}$, we obtain a sequence of Riemannian metrics $\{ g^{(i)} \}$ and a sequence of vectors $v_i \in W$, such that $\parallel g^{(i)} - g_1 \parallel_{C^2 (M, g_1)}< \frac{1}{i}$ and $q_i := exp_p^{g_1}(v_i) \notin exp_p^{g^{(i)}}(B_{R_1}^{g_1}(v_i))$. 

Let $U_i = exp_p^{g^{(i)}}(B_{R_1}^{g_1}(v_i))$. Since $exp_p^{g^{(i)}} \bigg|_{B_{R}^{g_1}(v_i)}$ is a diffeomorphism, and $exp_p^{g^{(i)}} (B_{R}^{g_1}(v_i)) \subset U$ then $\overline{U}_i = exp_p^{g^{(i)}}(\overline{B_{R_1}^{g_1}(v_i)})$.

Let $L:= 2 (\max_{v \in W} |v|_{g_1 (p)}+R)+1$ and for large $i$, $|w|_{g^{(i)}} \leq 2|w|_{g_1 (p)}<L$ for all $w \in B_R^{g_1}(v)$ and all $v \in W$. Let $a_i:= exp_p^{g^{(i)}}(v_i)$ and let $\rho_i := d_{g_1}(q_i, a_i)$. By corollary~\ref{def:cor1}, $\rho_i \to 0$ as $g^{(i)} \to g_1$ in $C^2 (M,g_1)$ and the convergence depends on $(M,g_1,U,L)$.

Now $a_i \in U_i$, and $q_i \notin U_i$, by our assumption. Since $\rho_i \to 0$, there exists $i_0$ such that for all $i \geq i_0$, $\rho_i < inj(M, g_1)$ (the injectivity radius of $(M,g_1)$). 

So, for all $i \geq i_0 $, let $\eta_{1,(i)}$ be the unique minimizing geodesic with respect to $g_1$ from $a_i$ to $q_i$. Since $\eta_{1,(i)}$ starts at $a_i$ in $U_i$ and ends at $q_i$ outside of $U_i$, there exists $t_i$ with $x_i := \eta_{1,(i)}(t_i) \in \partial U_i$, and $d_{g_1}(q_i, x_i) \leq d_{g_1}(q_i,a_i) = \rho_i \to 0$.
Then we have
\begin{equation*}
    \partial U_i = exp_p^{g^{(i)}} (\partial B_{R_1}^{g_1}(v_i))
\end{equation*}
So, $x_i = exp_p^{g^{(i)}}(w_i)$ for some $w_i \in \partial B_{R_1}^{g_1}(v_i)$. 

Since $W$ is compact, and for all $v_i \in W$, $w_i \in \overline{B_R^{g_1}(v_i)}$, $|w_i|_{g_1 (p)} < L$, then we can pass to a subsequence $v_{i_k} \to v_\infty \in W$ and $w_{i_k} \to w_\infty$. Then by continuity, $|w_\infty - v_\infty|_{g_1 (p)} = R_1 > 0$, and therefore $w_\infty \neq v_\infty$. Since $q_i = exp_p^{g_1}(v_i) \to q_\infty : = exp_p^{g_1}(v_\infty)$ and $d_{g_1}(q_i, x_i) \to 0$, we get $x_{i_k} \to q_\infty$. Then by theorem~\ref{def:thrm3}, $\parallel g^{(i)} - g_1\parallel_{C^2 (M,g_1)} \to 0$ and $|w_{i_k} - w_\infty|_{g_1 (p)} \to 0$ implies that, 
\begin{equation*}
    d_{g_1} (x_{i_k}, exp_p^{g_1}(w_\infty)) \to 0
\end{equation*}
Then, since $M$ is Hausdorff, we have $exp_p^{g_1}(w_\infty)=q_\infty = exp_p^{g_1}(v_\infty)$. Then, $w_\infty, v_\infty \in \overline{B_{R_1}^{g_1}(v_\infty)}$, where $exp_p^{g_1}$ is injective. Then,  $w_\infty = v_\infty$, which is a contradiction because $|w_\infty - v_\infty|_{g_1 (p)} = R_1 > 0$. Therefore, there does exist $\delta >0$ such that $\parallel g_1 - g_2 \parallel_{C^2 (M,g_1)} < \delta$ implies that for all $v \in W$, $exp_p^{g_1}(v) \in exp_p^{g_2}(B_{R_1}^{g_1}(v)) \subset exp_p^{g_2}(B_{R}^{g_1}(v))$.

\end{enumerate}
\end{proof}

\section{Proofs for Section~\ref{sec:app}}
In this section, we provide the proofs of the statements in section~\ref{sec:app}.

\subsection{Proof of Theorem \ref{thrm:neat5}}

\phantomsection\label{app:neat5}
\begin{proof}[Proof of Theorem \ref{thrm:neat5}]

     Assume, for contradiction, that for all $\delta_1 >0$, there exists a metric $g_{\delta_1}$ such that $\parallel g_1 - g_{\delta_1} \parallel_{C^2 (M,g_1)} < \delta_1$ and $inj (M,g_{\delta_1}) < \frac{inj(M,g_1)}{2}$. Let $\delta_{n} = \frac{1}{n}$, and let $g_n = g_{\delta_n}$. Then we have a sequence of metrics $g_n$ such that $\parallel g_1 - g_n \parallel_{C^2 (M,g_1)} < \frac{1}{n}$ (i.e. $g_n \to g_1$ in $C^2 (M,g_1)$), with $inj(M,g_n) < \frac{inj(M,g_1)}{2}$ for all $n$. Since $M$ is compact, for each $g_n$, there exists $p_n$ such that $d_{g_n}(p_n,Cut_{g_n}(p_n)) = \inf_{x \in M} d_{g_n}(x, Cut_{g_n}(x))$ and since $Cut_{g_n}(p_n)$ is compact, there exists $q_n$ such that $d_{g_n}(q_n,p_n) = \inf_{x \in Cut_{g_n}(p_n)} d_{g_n}(x,p_n)$. Then for each $n$, we have
    \begin{equation*}
        d_{g_n}(p_n,q_n) = inj(M,g_n) < \frac{inj(M,g_1)}{2}
    \end{equation*}
    Since $(M,g_1)$ is a compact metric space, there exists a subsequence $p_{n_1} \to p_0$ in $d_{g_1}$. Then there's another subsequence of $n_1$, indexed by $n_2$, such that $q_{n_2} \to q_0$ in $d_{g_1}$, with $p_{n_2} \to p_0$ in $d_{g_1}$ because $M$ is Hausdorff. From here, we'd like to show that
    \begin{equation*}
        \lim_{n_2 \to \infty} d_{g_{n_2}}(p_{n_2}, q_{n_2})= d_{g_1}(p_0,q_0)
    \end{equation*}

    First, we'll show that $d_{g_{n_2}} \to d_{g_1}$ uniformly as continuous functions on $M \times M$. Let $\epsilon_{n_2} = \frac{1}{n_2}$, then from proposition~\ref{prop:neat}, for all $x,y \in M$, we have
    \begin{align*}
        ( \sqrt{1 - \epsilon_{n_2}} - 1) \ d_{g_1}(x,y) & = \sqrt{1-\epsilon_{n_2}} \ d_{g_1}(x,y) - d_{g_1}(x,y)\\
        & \leq  d_{g_{n_2}}(x,y) - d_{g_1}(x,y)\\
        & \leq ( \sqrt{1 + \epsilon_{n_2}} - 1) \ d_{g_1}(x,y)
    \end{align*}
    then we see that
    \begin{align*}
        (1 - \sqrt{1 + \epsilon_{n_2}}) \ d_{g_1}(x,y) & =  \ d_{g_1}(x,y) - \sqrt{1 + \epsilon_{n_2}}d_{g_1}(x,y)\\
        & \leq    d_{g_1}(x,y) - d_{g_{n_2}}(x,y)\\
        & \leq (1 - \sqrt{1 - \epsilon_{n_2}} ) \ d_{g_1}(x,y)
    \end{align*}    
    which means that
    \begin{align*}
        |d_{g_{n_2}}(x,y) - d_{g_1}(x,y)| & \leq (1 - \sqrt{1 - \epsilon_{n_2}}) \ d_{g_1}(x,y)\\
        & \leq (1 - \sqrt{1 - \epsilon_{n_2}}) \ \max_{x,y \in M} \ d_{g_1}(x,y) \to 0
    \end{align*}
    as $n_2 \to \infty$, since $M$ is compact. Thus, $d_{g_{n_2}} \to d_{g_1}$ uniformly, as desired.
    
    Next, take
    \begin{align*}
        |d_{g_{n_2}}(p_{n_2},q_{n_2}) - d_{g_1}(p_0,q_0)| & \leq |d_{g_{n_2}}(p_{n_2},q_{n_2}) - d_{g_{n_2}}(p_0,q_0)| + |d_{g_{n_2}}(p_0,q_0) - d_{g_1}(p_0,q_0)| = I + II
    \end{align*}
    For II, we have
    \begin{equation*}
        II \leq \sup_{x,y \in M} |d_{g_{n_2}}(x,y) - d_{g_1}(x,y)| \to 0
    \end{equation*}
    For I, we observe
    \begin{align*}
        d_{g_{n_2}}(p_{n_2},q_{n_2}) & \leq d_{g_{n_2}} (p_{n_2}, p_0) + d_{g_{n_2}} (p_0, q_{n_2})\\
        & \leq d_{g_{n_2}} (p_{n_2}, p_0) + d_{g_{n_2}} (p_0, q_0) + d_{g_{n_2}} (q_0, q_{n_2})
    \end{align*}
    Similarly, we have
    \begin{equation*}
        d_{g_{n_2}} (p_0, q_0) \leq  d_{g_{n_2}} (p_{n_2},p_0) + d_{g_{n_2}} (p_{n_2}, q_{n_2}) + d_{g_{n_2}} (q_{n_2}, q_0)
    \end{equation*}
    and we have
    \begin{align*}
        |d_{g_{n_2}}(p_{n_2},q_{n_2}) - d_{g_{n_2}} (p_0, q_0)| & \leq d_{g_{n_2}} (p_{n_2},p_0) + d_{g_{n_2}} (q_{n_2}, q_0)\\
        & \leq \sqrt{1 + \epsilon_{n_2}} (d_{g_{1}} (p_{n_2},p_0) + d_{g_{1}} (q_{n_2}, q_0) ) \to 0
    \end{align*}
    Then putting this together we have
    \begin{equation*}
        |d_{g_{n_2}}(p_{n_2},q_{n_2}) - d_{g_1}(p_0,q_0)| \leq I + II \to 0
    \end{equation*}
    which means that $d_{g_{n_2}}(p_{n_2},q_{n_2}) \to d_{g_1}(p_0,q_0)$ as $n_2 \to \infty$. Then we see that
    \begin{align*}
        d_{g_1}(p_0, q_0) & = \lim_{n_2 \to \infty} d_{g_{n_2}} (p_{n_2}, q_{n_2})\\
        & \leq \lim_{n_2 \to \infty} \frac{inj(M,g_1)}{2}\\
        & = \frac{inj(M,g_1)}{2}
    \end{align*}
    Since $q_{n_2}$ is a point of $Cut_{g_{n_2}}(p_{n_2})$ closest to $p_{n_2}$, proposition 2.12 (p.274,~\cite{do1992riemannian}) gives, for each $n_2$, either:
    \begin{enumerate}
        \item $q_{n_2}$ is the first conjugate point of $p_{n_2}$ along the unique minimizing geodesic with respect to $g_{n_2}$ from $p_{n_2}$ to $q_{n_2}$, or
        \item There exist exactly two distinct minimizing geodesics with respect to $g_{n_2}$ from $p_{n_2}$ to $q_{n_2}$, with $\eta_{1 n_2}^\prime (1) = - \eta_{2 n_2}^\prime (1)$, and $| \eta_{1 n_2}^\prime (0) |_{g_{n_2}(p_{n_2})} = | \eta_{2 n_2}^\prime (0) |_{g_{n_2}(p_{n_2})} = d_{g_{n_2}}(p_{n_2},q_{n_2})$.
    \end{enumerate}
    Since one of the two cases holds for infinitely many $n_2$, after passing to a subsequence we may assume that either case A holds for every $n_2$ or case B holds for every $n_2$.
    
    We'll first assume case A holds for every $n_2$, and derive a contradiction. Since, $q_{n_2}$ is the first conjugate point of $p_{n_2}$ along the unique minimizing geodesic with respect to $g_{n_2}$ from $p_{n_2}$ to $q_{n_2}$, then there exists a unique $v_{n_2} \in T_{p_{n_2}}M$ such that for $\eta_{n_2} (t) = exp_{p_{n_2}}^{g_{n_2}}(t v_{n_2})$, $\eta_{n_2} (1) = q_{n_2}$, and
    \begin{equation*}
       |v_{n_2}|_{g_{n_2}(p_{n_2})} = d_{g_{n_2}} (p_{n_2}, q_{n_2})
    \end{equation*}
    Let $v_0 \in T_{p_0}M$ be such that $\eta_0 (t) = exp_{p_0}^{g_1}(tv_0)$ and $\eta_0 (1)=q_0$. Then we have
    \begin{equation*}
        |v_0|_{g_1 (p_0)} = d_{g_1}(p_0,q_0) = \lim_{n_2 \to \infty} d_{g_{n_2}} (p_{n_2}, q_{n_2}) = \lim_{n_2 \to \infty} |v_{n_2}|_{g_{n_2}(p_{n_2})}
    \end{equation*}
    Since $q_{n_2}$ is a conjugate point of $p_{n_2}$, there exists a unit vector $u_{n_2} \in T_{p_{n_2}}M$ with respect to $g_{n_2}(p_{n_2})$, such that we have a Jacobi field $f_{n_2}(t)$ along $\eta_{n_2}$, with $f_{n_2}(0)=0$, $f_{n_2}^\prime (0)= u_{n_2}$ and $f_{n_2}(1) = 0$. Now let's construct a neighborhood centered at $p_0$. Since $d_{g_1}(p_0, q_0) \leq \frac{inj(M, g_1)}{2}$, we'll let $U= B_{g_1}(p_0, \mathcal{R})$ and now we'll identify $\mathcal{R}$.
    For the geodesic $\eta_{n_2}(t)$, we see that
    \begin{equation*}
        d_{g_1}(\eta_{n_2}(t), p_0) \leq d_{g_1}(\eta_{n_2}(t), p_{n_2}) + d_{g_1}(p_{n_2}, p_0)
    \end{equation*}
    Then we see that
    \begin{align*}
        d_{g_1}(\eta_{n_2}(t), p_{n_2}) & \leq \frac{1}{\sqrt{1 - \epsilon_{n_2}}} d_{g_{n_2}}(\eta_{n_2}(t), p_{n_2})\\
        & \leq \frac{1}{\sqrt{1 - \epsilon_{n_2}}} d_{g_{n_2}}(q_{n_2}, p_{n_2})\\
        & < \frac{1}{\sqrt{1 - \epsilon_{n_2}}} \frac{inj(M, g_1)}{2}\\
        & < \frac{inj(M, g_1)}{2 (1 - \epsilon_{n_2})} = \star
    \end{align*}
    Then there exists $N_0$ such that $n_2 \geq N_0$ implies that $(1- \epsilon_{n_2}) > \frac{4}{5}$, then for $n_2 \geq N_0$ we have
    \begin{equation*}
        \star < \frac{inj(M,g_1)}{2 (1 - \epsilon_{n_2})} < \frac{5 inj(M,g_1)}{8}
    \end{equation*}
    Then because $d_{g_1}(p_{n_2}, p_0) \to 0$, there exists $N_1$ such that $n_2 \geq N_1$ implies that $d_{g_1}(p_{n_2}, p_0) < \frac{inj(M,g_1)}{8}$. Then for $n_2 \geq \max \{ N_0,N_1 \}$, we have
    \begin{equation*}
         d_{g_1}(\eta_{n_2}(t), p_0) < \frac{5 inj(M,g_1)}{8} + \frac{ inj(M,g_1)}{8} = \frac{3 \ inj(M,g_1)}{4}
    \end{equation*}
    So, let $\mathcal{R} = \frac{3 \ inj(M,g_1)}{4}$, and $U= B_{g_1}(p_0, \frac{3 \ inj(M,g_1)}{4} )$. Then $n_2 \geq \max \{ N_0, N_1 \}$ implies that $\eta_{n_2}([0,1]) \subset U$.
    Now, we have a fixed $U$, and we can fix a chart $(V, \phi)$ that contains $\overline{U}$, and we'll have a matrix representation for $g_1 = (g_{1ij})$ with smooth and therefore continuous entries. Then $p_{n_2} \to p_0$ implies that $(g_{1ij} (p_{n_2})) \to (g_{1ij}(p_0))$. Since $T_{p_{n_2}}M \cong \mathbb{R}^n$ via $d \phi_{p_{n_2}}$, then let $v_{n_2}^\prime = d \phi_{p_{n_2}} (v_{n_2}) \in \mathbb{R}^n$. Recall:
    \begin{align*}
        |v_0|_{g_1 (p_0)} = d_{g_1}(p_0,q_0) = \lim_{n_2 \to \infty} d_{g_{n_2}} (p_{n_2}, q_{n_2}) = \lim_{n_2 \to \infty} |v_{n_2}|_{g_{n_2}(p_{n_2})}
    \end{align*}
    Then $\{ |v_{n_2}|_{g_{n_2}(p_{n_2})} \}$ is a convergent sequence in $\mathbb{R}$, and is therefore bounded for sufficiently large $n_2$. Thus, there exists $C_0 > 0$ and $N_2$ such that $n_2 \geq N_2$ implies
    \begin{equation*}
        |v_{n_2}|_{g_{n_2}(p_{n_2})} \leq C_0
    \end{equation*}
    Then for $v_{n_2} \in T_{p_{n_2}}M$, we have
    \begin{equation*}
        |v_{n_2}|_{g_1(p_{n_2})}  \leq \frac{|v_{n_2}|_{g_{n_2}(p_{n_2})}}{\sqrt{1 - \epsilon_{n_2}}}
    \end{equation*}
    Then let $N_3$ be such that $n_2 \geq N_3$ implies that $\sqrt{1 - \epsilon_{n_2}} \geq \frac{1}{2}$ then $n_2 \geq \max \{ N_2, N_3 \}$ implies that
    \begin{equation*}
        |v_{n_2}|_{g_1(p_{n_2})} \leq 2 |v_{n_2}|_{g_{n_2}(p_{n_2})} \leq 2C_0
    \end{equation*}
    Again since $(g_{1ij}(p_{n_2})) \to (g_{1ij}(p_0))$ there exists $N_4$ such that $n_2 \geq N_4$ implies that for all $v \in \mathbb{R}^n$, we have 
    \begin{equation*}
        \frac{1}{2} \ v^T (g_{1ij}(p_0))v \leq v^T (g_{1ij}(p_{n_2}))v \leq 2 \ v^T (g_{1ij}(p_0))v
    \end{equation*}
    Then for $v_{n_2}^\prime \in \mathbb{R}^n$, we'll have $n_2 \geq \max \{ N_2,N_3,N_4 \}$ implies that
    \begin{equation*}
        | v_{n_2}^\prime|_{g_1 (p_0)} \leq | v_{n_2}^\prime|_{g_1 (p_{n_2})} < \sqrt{2} \cdot 2 C_0 < 4C_0
    \end{equation*}
    Now since $g_1 (p_0)$ is a fixed symmetric positive-definite matrix, thus for $\lambda_1 = \sqrt{\lambda_{\min}(g_1 (p_0))}$ and $\lambda_2 = \sqrt{\lambda_{\max}(g_1 (p_0))}$ we have
    \begin{equation*}
        \lambda_1 |v_{n_2}^\prime| \leq |v_{n_2}^\prime|_{g_1 (p_0)} \leq \lambda_2 |v_{n_2}^\prime|
    \end{equation*}
    which means that for $|\cdot|$ being the Euclidean norm from definition~\ref{def:eucl}, we have
    \begin{equation*}
        |v_{n_2}^\prime| \leq \frac{|v_{n_2}^\prime|_{g_1 (p_0)}}{\lambda_1} \leq \frac{4C_0}{\lambda_1} = C_1
    \end{equation*}
    and for $n_2 \geq \max \{ N_2 , N_3, N_4 \}$, we have that $v_{n_2}^\prime \in B_{C_1}^{|\cdot|} (0)$ in $\mathbb{R}^n$, and we have a subsequence $v_{n_3}^\prime \to v_0^\prime$ in $(\mathbb{R}^n, |\cdot|)$. 
    Now for $u_{n_3}^\prime = d \phi_{p_{n_3}}(u_{n_3})$, we know that $|u_{n_3}^\prime|_{g_{n_3} (p_{n_3})} = 1$. Then using the same logic as before, $n_3 \geq \max \{ N_2, N_3, N_4 \}$ we have
    \begin{equation*}
        |u_{n_3}^\prime| \leq \frac{|u_{n_3}^\prime|_{g_1 (p_0)}}{\lambda_1} \leq \frac{2|u_{n_3}^\prime|_{g_1 (p_{n_3})}}{\lambda_1} \leq \frac{4|u_{n_3}^\prime|_{g_{n_3} (p_{n_3})}}{\lambda_1} = \frac{4}{\lambda_1} = C_2
    \end{equation*}
    where the first inequality comes from what we computed above, by how we defined $\lambda_1$. Then $\{ u_{n_3}^\prime \} \in B_{C_2}^{| \cdot |} (0)$, and thus we have a subsequence $\{ u_{n_4}^\prime \}$ such that $u_{n_4}^\prime \to u_0^\prime$. 
    
    To apply lemma~\ref{def:example2}, we'll use initial points $p_{n_4}$ and initial velocity $v_{n_4}^\prime$. Then we'll have $x_{n_4} (t) = \begin{bmatrix}
        \eta_{n_4} (t)\\  \eta_{n_4}^\prime (t)
    \end{bmatrix}$ with $x_{n_4} (0) = \begin{bmatrix}
        p_{n_4} \\  v_{n_4}^\prime
    \end{bmatrix}$, and then we'll have
    \begin{equation*}
        x_{n_4} (0) - x_0 (0) = \begin{bmatrix}
            p_{n_4} - p_0\\
            v_{n_4}^\prime - v_0^\prime
        \end{bmatrix}
    \end{equation*}
    which converges to $0$ by how we set up $p_{n_4}$ and $v_{n_4}^\prime$. Then from proposition~\ref{prop:prop1} on $U$, we have $|\Gamma_{ij}^{k,g_{n_4}} - \Gamma_{ij}^{k,g_1}| < \zeta_{n_4} \to 0$ as $n_4 \to \infty$. Let $Z_0 = \max |\partial \Gamma^{g_1}|$ and $Z_1 = \max |\Gamma^{g_1}|$, which were defined in remark~\ref{def:rmk1}. Since $|v_{n_4}^\prime| < C_1$, let $Z_2 = C_1$. Then let $A = 1 + n^3 Z_2^2Z_0 + 2n^3 Z_1 Z_2$ and $B_{n_4}= |x_{n_4} (0) - x_0 (0)| + n^3 Z_2^2 \zeta_{n_4} \to 0$ as $n_4 \to \infty$. Then
    \begin{equation*}
        |x_{n_4} (t) - x_0 (t)| \leq B_{n_4}e^A \to 0
    \end{equation*}
    Then
    \begin{equation*}
        |\eta_{n_4}(t) - \eta_0 (t)|, |\eta_{n_4}^\prime (t) - \eta_0^\prime (t)| \to 0
    \end{equation*}
   
    Now, we want to deal with the Jacobi field $f_{n_4}(t)$ along $\eta_{n_4}(t)$. Let $y_{n_4}(t) = \begin{bmatrix}
        f_{n_4}(t)\\ f_{n_4}^\prime (t)
    \end{bmatrix}$, and we'll use lemma~\ref{def:example3}, where $f_{n_4}(0)=0$ and $f_{n_4}^\prime (0) = u_{n_4}^\prime$, we see that
    \begin{equation*}
        y_{n_4} (0) - y_0 (0) = \begin{bmatrix}
            f_{n_4}(0) - f_0 (0)\\ f_{n_4}^\prime (0) - f_0^\prime (0)
        \end{bmatrix} = \begin{bmatrix}
            0\\
            u_{n_4}^\prime - u_0^\prime
        \end{bmatrix}
    \end{equation*}
    Then $ |y_{n_4} (0) - y_0 (0)| = |u_{n_4}^\prime - u_0^\prime| \to 0$ by how we set up $u_{n_4}^\prime$. We observe that $|y_0 (t)| < \Omega_0$ and $|M(t, \gamma^{g_1})| < \Omega_1$. Then for $a_{kj}^{n_4}(t) = 2 \sum_{i} \dot{\eta}_{n_4}^i \Gamma_{ij}^{k,g_{n_4}}(\eta_{n_4}(t))$, from the proof of theorem~\ref{thrm:neat1} in Appendix~\ref{app:neat1}, we have $|a_{kj}^{n_4}(t) - a_{kj}^{0}(t) | < A_{g_{n_4}} \to 0$. For $b_{kj}^{n_4}(t)$ as defined in equation~\ref{def:jaceq}, we have $|b_{kj}^{n_4}(t) - b_{kj}^{0}(t)| < B_{g_{n_4}} \to 0$. Then
    \begin{equation*}
        |A_{n_4}(t) - A_0 (t)| + |B_{n_4}(t) - B_0 (t)| < (A_{g_{n_4}} + B_{g_{n_4}})
    \end{equation*}
    Then we have
    \begin{equation*}
        | y_{n_4} (t) - y_0 (t) | \leq (|u_{n_4}^\prime - u_0^\prime| + \Omega_0 (A_{g_{n_4}} + B_{g_{n_4}} ))e^{\Omega_1+ A_{g_{n_4}} + B_{g_{n_4}}} \to 0
    \end{equation*}
    as $n_4 \to \infty$. This means that for all $t \in [0,1]$, we have
    \begin{equation*}
        |f_{n_4}(t) - f_0 (t)| \to 0
    \end{equation*}
    Then we see that
    \begin{align*}
        f_0 (1) & = \lim_{n_4 \to \infty} f_{n_4}(1)\\
        & = \lim_{n_4 \to \infty} 0 = 0
    \end{align*}
    but this is a contradiction because $d_{g_1}(p_0,q_0) < inj(M,g_1)$ which means that $f_0 (t) \neq 0$ on $(0,1]$.
    
    Now, we'll assume that case B holds for all $n_2$ and derive a contradiction. There exists exactly two minimizing geodesics with respect to $g_{n_2}$, $\eta_{1 n_2}$ and $\eta_{2n_2}$ from $p_{n_2}$ to $q_{n_2}$, where $\eta_{1n_2}^\prime (1) = - \eta_{2n_2}^\prime (1)$, with $d_{g_{n_2}}(p_{n_2},q_{n_2}) = L_{g_{n_2}}(\eta_{1n_2}) = L_{g_{n_2}}(\eta_{2n_2})$. 

    Let $v_{n_2}$ denote the initial velocity of $\eta_{1n_2}$, so $|v_{n_2}|_{g_{n_2}(p_{n_2})} = d_{g_{n_2}}(p_{n_2}, q_{n_2})$. Then the construction of $U$, the constants $N_0$ to $N_4$, $C_0$ and $C_1$, and the subsequence above uses only that $\eta_{1n_2}$ is a minimizing geodesic of this length, so it applies verbatim. Hence there's a subsequence $n_3$ with $v_{n_3}^\prime \to v_0^\prime$ and, as in the Gr\"onwall estimate for Case A, $\eta_{1n_3} \to \eta_{10}$.

    Then for each $\eta_{1n_3}(t)$, there exists $\eta_{2n_3} (t)$ and $\eta_{2n_3}^\prime (0)= \alpha_{n_3} \in T_{p_{n_3}}M$. Additionally, we have $|\alpha_{n_3}|_{g_{n_3}(p_{n_3})} = d_{g_{n_3}}(p_{n_3},q_{n_3}) = |v_{n_3}|_{g_{n_3}(p_{n_3})}$. Then for sufficiently large $n_3$, $\eta_{2n_3}([0,1]) \subset U$, then $\alpha_{n_3}^\prime = d \phi_{p_{n_3}}(\alpha_{n_3})$ is also bounded in $\mathbb{R}^n$. This allows us to pass to a subsequence that $\alpha_{n_4}^\prime \to \alpha_0^\prime$ in $(\mathbb{R}^n, | \cdot |)$. Then with the same constants as with $\eta_{1n_4} (t)$, lemma~\ref{def:example2} tells us that $\eta_{2n_4}(t) \to \eta_{20}(t)$.
    Then since $\eta_{1n_4}^\prime (1) = - \eta_{2n_4}^\prime (1)$, if we combine this with the fact that $|\eta_{2n_4}^\prime (t) - \eta_{20}^\prime (t)| \to 0$, then 
    \begin{equation*}
        \eta_{10}^\prime (1) = \lim_{n_4 \to \infty} \eta_{1 n_4}^\prime (1) = \lim_{n_4 \to \infty} - \eta_{2 n_4}^\prime (1) = - \eta_{20}^\prime (1)        
    \end{equation*}
    Then $d_{g_{n_4}}(q_{n_4},Cut_{g_{n_4}}(p_{n_4})) = d_{g_{n_4}}(q_{n_4}, p_{n_4})$, so let $\gamma_{in_4}(t) = \eta_{i n_4} (1-t)$, which are geodesics from $q_{n_4}$ to $p_{n_4}$, and $L_{g_{n_4}}(\gamma_{1n_4}) = L_{g_{n_4}}(\gamma_{2n_4}) = d_{g_{n_4}}(q_{n_4}, p_{n_4})$. These are exactly the two geodesics from case B, so $\gamma_{1n_4}^\prime (1) = - \gamma_{2n_4}^\prime (1)$. We see that
    \begin{equation*}
        \eta_{1n_4}^\prime (0) = - \gamma_{1n_4}^\prime (1) = - ( - \gamma_{2n_4}^\prime (1)) = \gamma_{2n_4}^\prime (1) = - \eta_{2n_4}^\prime (0)
    \end{equation*}
    Then we have
    \begin{equation*}
        \eta_{10}^\prime (0) = \lim_{n_4 \to \infty} \eta_{1n_4}^\prime (0) = \lim_{n_4 \to \infty} - \eta_{2n_4}^\prime (0) = - \eta_{20}^\prime (0)
    \end{equation*}
    Now, let
    \begin{equation*}
        \sigma_0 (t) = \begin{cases}
            \eta_{10} (t) & if \  t\in [0, 1]\\
            \eta_{20} (2-t) & if \  t\in [1,2]
        \end{cases}
    \end{equation*}
    Then we see that
    \begin{equation*}
        \lim_{t \to 1^-} \sigma_0^\prime (t) =  \lim_{t \to 1^-} \eta_{10}^\prime (t) = \eta_{10}^\prime (1)
    \end{equation*}
    then from the other direction we have
    \begin{align*}
        \lim_{t \to 1^+} \sigma_0^\prime (t) = \lim_{t \to 1^+} \frac{d}{dt} \eta_{20}(2-t) = \lim_{t \to 1^+} - \eta_{20}^\prime (2-t) = - \eta_{20}^\prime (1) = \eta_{10}^\prime (1) =  \lim_{t \to 1^-} \sigma_0^\prime (t)
    \end{align*}
    Then $\sigma_0^\prime (1)$ exists and is continuous. Next, we have
    \begin{equation*}
        \sigma_0^\prime (2) = \frac{d}{dt} \big|_{t=2} \eta_{20}(2-t) = - \eta_{20}^\prime (0) = \eta_{10}^\prime (0) = \sigma_0^\prime (0)
    \end{equation*}
    Then $\sigma_0(t)$ is a closed geodesic, and we have
    \begin{align*}
        L_{g_1}(\sigma_0) & = \lim_{n_4 \to \infty} L_{g_{n_4}} (\sigma_{n_4}) =  \lim_{n_4 \to \infty} L_{g_{n_4}} (\eta_{1 n_4}) + L_{g_{n_4}} (\eta_{2 n_4})\\
        & = \lim_{n_4 \to \infty} 2 d_{g_{n_4}} (p_{n_4},q_{n_4}) = 2 d_{g_1}(p_0,q_0)\\
        & \leq 2 \cdot \frac{inj(M, g_1)}{2} = inj(M,g_1)
    \end{align*}
    Which is a contradiction because proposition 2.13 (p.275,~\cite{do1992riemannian}) says that all closed geodesics have length strictly larger than $inj(M,g_1)$.
\end{proof}

\subsection{Proof of Corollary \ref{cor:neat5}}

\phantomsection\label{app:cor-neat5}
\begin{proof}[Proof of Corollary \ref{cor:neat5}]

    Assume for contradiction, that there exists $\epsilon >0$ such that for every $\delta(r) \leq \min \{ \delta_0, \delta_1 \}$, there exists a metric $g_{\delta (r)}$ with $\parallel g_1 - g_{\delta (r)} \parallel_{C^2 (M,g_1)} < \delta (r)$, there exists $x \in M$, and a $y \in B_{g_1}(x,r)$, such that $|exp_x^{g_1,-1}(y) - exp_x^{g_{\delta (r)},-1}(y)|_{g_1 (x)} \geq \epsilon$. Letting $\delta (r)=\frac{1}{n}$ for all $n \in \mathbb{N}$ (taking $n$ large enough that $\frac{1}{n} \leq \min \{ \delta_0, \delta_1 \}$ which guarantees that $exp_x^{g_n, -1}$ is well-defined by theorem~\ref{thrm:neat5} and by $\delta_0$), we obtain a sequence of metrics $g_n$ with $\parallel g_1 - g_n \parallel_{C^2 (M,g_1)} < \frac{1}{n} \leq \min \{ \delta_0, \delta_1 \}$, and the points $x_n \in M$, and $y_n \in B_{g_1}(x_n,r)$, such that $|exp_{x_n}^{g_1,-1}(y_n) - exp_{x_n}^{g_n,-1}(y_n)|_{g_1 (x_n)} \geq \epsilon$ for all $n$.
    
    Then because $(M,g_1)$ is a compact metric space, for $\{ x_n \}$ we have a subsequence $x_{n_1} \to x_0$ in $d_{g_1}$ and another subsequence $y_{n_2} \to y_0$ in $d_{g_1}$. 
    Let's define
    \begin{equation*}
        \begin{split}
            w_{1n_2} & := exp_{x_{n_2}}^{g_1,-1} (y_{n_2})\\
            w_{2n_2} & := exp_{x_{n_2}}^{g_{n_2},-1} (y_{n_2})
        \end{split}
    \end{equation*}
    which are uniquely defined since $y_{n_2} \in B_{g_1}(x_{n_2},r) \subset B_{g_{n_2}}(x_{n_2}, \frac{inj(M, g_1)}{2}) $. Then for the geodesics, we have
    \begin{equation*}
        \begin{split}
            \eta_{1n_2} (t) & : = exp_{x_{n_2}}^{g_1}(t w_{1n_2})\\
            \eta_{2n_2}(t) & : = exp_{x_{n_2}}^{g_{n_2}}(t w_{2n_2})
        \end{split}
    \end{equation*}
    for $t \in [0,1]$, with $w_{i 0} \in T_{x_0}M$ such that $\eta_{i0}(t) = exp_{x_0}^{g_1}(tw_{i0})$ and $\eta_{i0}(1)=y_0$. Then by letting $\epsilon_{n_2} = \frac{1}{n_2}$
    \begin{align*}
        d_{g_{n_2}}(x_{n_2},y_{n_2}) & \leq \sqrt{1 + \epsilon_{n_2}} \ d_{g_1}(x_{n_2}, y_{n_2})\\
        & < \sqrt{1 + \epsilon_{n_2}} \ r
    \end{align*}
    then there exists $N_0$ such that $n_2 \geq N_0$ implies that $\sqrt{1 + \epsilon_{n_2}} < \frac{5r}{4}$. This means that  
    \begin{equation*}
        d_{g_{n_2}} (x_{n_2}, y_{n_2}) < \frac{5r}{4}
    \end{equation*}
    then
    \begin{equation*}
        d_{g_{n_2}}(x_{n_2}, \eta_{2n_2}(t)) \leq |w_{2n_2}|_{g_{n_2}(x_{n_2})} =  d_{g_{n_2}}(x_{n_2}, y_{n_2}) < \frac{5r}{4}
    \end{equation*}
    for all $t \in [0,1]$. Then
    \begin{align*}
        d_{g_1}(x_{n_2}, \eta_{2n_2}(t)) & \leq \frac{1}{\sqrt{1 - \epsilon_{n_2}}} d_{g_{n_2}} (x_{n_2}, \eta_{2n_2}(t))\\
        & < \frac{1}{\sqrt{1 - \epsilon_{n_2}}} \frac{5r}{4}
    \end{align*}
    Let $N_1$ be such that $n_2 \geq N_1$ implies that $\frac{1}{\sqrt{1 - \epsilon_{n_2}}} < \frac{6}{5}$, giving us
    \begin{equation*}
        d_{g_1}(x_{n_2}, \eta_{2n_2}(t)) < \frac{6}{5} \cdot \frac{5}{4} \cdot r = \frac{3r}{2}
    \end{equation*}
    Then for $i=1,2$, we have
    \begin{align*}
        d_{g_1}(\eta_{in_2}(t), x_0) & \leq d_{g_1}(\eta_{in_2}(t), x_{n_2}) + d_{g_1}(x_{n_2}, x_0)\\
        & < \frac{3r}{2} + \frac{inj(M,g_1)}{8} = \star
    \end{align*}
    where $\frac{inj(M,g_1)}{8}$ is because $x_{n_2} \to x_0$, and $n_2 \geq N_2$ gives us our desired inequality. Then since $r < \frac{inj(M,g_1)}{2}$, we have
    \begin{equation*}
        \star < \frac{3 \ inj(M,g_1)}{4} + \frac{inj(M,g_1)}{8} = \frac{7 \ inj(M,g_1)}{8}
    \end{equation*}
    Let $U : = B_{g_1}(x_0, \frac{7 \ inj(M,g_1)}{8})$. Then $n_2 \geq \max \{ N_0, N_1, N_2 \}$, implies that 
    \begin{equation*}
        \eta_{i n_2} ([0,1]) \subset U
    \end{equation*}
    This gives us a fixed $U$. Let $\phi$ be the $g_1$-normal coordinates centered at $x_0$. They are defined on $B_{g_1}(x_0, inj(M,g_1)) \supset U$, so $\phi$ is a chart on all of $U$, and we use it throughout (in particular for proposition~\ref{prop:prop1} and lemma~\ref{def:example2}). Then with the chart, let $w_{1n_2}^\prime = d \phi_{x_{n_2}} ( w_{1n_2})$ and $w_{2n_2}^\prime = d \phi_{x_{n_2}} ( w_{2n_2})$. Since we have
    \begin{align*}
        \lim_{n_2 \to \infty} |w_{1n_2}|_{g_1 (x_{n_2})} & = \lim_{n_2 \to \infty} d_{g_1}(x_{n_2}, y_{n_2})\\
        & \leq \lim_{n_2 \to \infty} r = r
    \end{align*}
    the sequence $\{ |w_{1n_2}|_{g_1 (x_{n_2})} \}$ is bounded. Then by the continuity of the matrix functions of $(g_{1ij})$, there exists $N_3$ such that $n_2 \geq N_3$ implies that for all $v \in \mathbb{R}^n$, we have
    \begin{equation*}
        \frac{1}{2} v^T (g_{1ij}(x_0)) v \leq v^T (g_{1ij}(x_{n_2})) v \leq 2 v^T (g_{1ij}(x_0)) v
    \end{equation*}
    Let $\lambda_1 = \sqrt{\lambda_{\min}(g_1 (x_0))}$, then we have
    \begin{equation*}
        |w_{1n_2}^\prime| \leq \frac{|w_{1n_2}^\prime|_{g_1 (x_0)} }{\lambda_1} \leq \frac{2|w_{1n_2}^\prime|_{g_1 (x_{n_2})} }{\lambda_1} \leq \frac{2r}{\lambda_1} = C_0
    \end{equation*}
    Then the sequence $\{ w_{1n_2}^\prime \}$ is bounded, which means that there exists a convergent subsequence $w_{1n_3}^\prime \to w_{10}^\prime$ in $(\mathbb{R}^n, | \cdot |)$. 
    Since
    \begin{equation*}
        \lim_{n_3 \to \infty}|w_{2n_3}|_{g_{n_3}(x_{n_3})} = |w_{20}|_{g_1 (x_0)}
    \end{equation*}
    there exists $N_4$ and a $C_1 >0$ such that $n_3 \geq N_4$ implies that $|w_{2n_3}|_{g_{n_3}(x_{n_3})} \leq C_1$. Then because there exists $N_5$ such that $n_3 \geq N_5$ implies that $\sqrt{1 - \epsilon_{n_3}} \geq \frac{1}{2}$. For $n_3 \geq \max \{ N_3,N_4,N_5 \}$, we have
    \begin{align*}
        |w_{2n_3}^\prime| & \leq \frac{|w_{2n_3}^\prime|_{g_1 (x_0)} }{\lambda_1} \leq \frac{2|w_{2n_3}^\prime|_{g_1 (x_{n_3})} }{\lambda_1}\\
        & \leq \frac{4|w_{2n_3}^\prime|_{g_{n_3} (x_{n_3})} }{\lambda_1} \leq \frac{4 C_1}{\lambda_1} = C_2
    \end{align*}
    Then we have another bounded sequence $\{ w_{2n_3}^\prime \}$, which means that we have a convergent subsequence $w_{2n_3}^\prime \to w_{20}^\prime$ in $(\mathbb{R}^n, | \cdot |)$.
    By employing lemma~\ref{def:example2}, for $\eta_{1n_4}(t)$, we have initial position $x_{n_4}$ and initial velocity $w_{1n_4}^\prime$. For $y_{in_4} (t) = \begin{bmatrix}
        \eta_{in_4}(t)\\
        \eta_{in_4}^\prime (t)
    \end{bmatrix}$ we have
    \begin{equation*}
        y_{1n_4}(0) - y_{10}(0) = \begin{bmatrix}
            x_{n_4} - x_0\\
            w_{1n_4}^\prime - w_{10}^\prime
        \end{bmatrix}
    \end{equation*}
    In order to apply lemma~\ref{def:example2}, let $K_0 = \max | \partial \Gamma^{g_1}|$, $K_1 = \max |  \Gamma^{g_1}|$, $|\eta_{2n_4}^\prime (t)|< K_2 = C_2$. By proposition~\ref{prop:prop1} $|\Gamma_{ij}^{k,g_1} - \Gamma_{ij}^{k,g_{n_4}}| < \zeta_{n_4} \to 0$ as $n_4 \to \infty$. Then let $A= 1 + n^3 K_2^2 K_0 + 2n^3 K_1 K_2$ and $B_{n_4} = |y_{1n_4}(0) - y_{10}(0)| +  n^3 K_2^2 \zeta_{n_4} \to 0$ as $n_4 \to \infty$. This gives us
    \begin{equation*}
        |y_{1n_4}(t) - y_{10}(t)|  \leq B_{n_4}e^A \to 0
    \end{equation*}
    This means that
    \begin{equation*}
        y_0 = \lim_{n_4 \to \infty} y_{n_4} = \lim_{n_4 \to \infty} \eta_{1n_4}(1) = \eta_{10}(1) = exp_{x_0}^{g_1}(w_{10})
    \end{equation*}
    Then for $\eta_{2n_4}(t)$ and $\eta_{20}(t)$, we have
    \begin{equation*}
        y_{2n_4} (0) - y_{20}(0) = \begin{bmatrix}
            x_{n_4} - x_0\\
            w_{2n_4}^\prime - w_{20}^\prime
        \end{bmatrix}
    \end{equation*}
    Then the same $A$ and $B_{n_4}$ will also apply. So, we have
    \begin{equation*}
        |y_{2n_4}(t) - y_{20}(t)|  \leq B_{n_4}e^A \to 0
    \end{equation*}
    Then, $d_{g_1}(\eta_{2n_4}(t), \eta_{20}(t)) \to 0$ as $n_4 \to \infty$, so we have
    \begin{equation*}
        d_{g_1}(y_{n_4}, \eta_{20}(1)) = d_{g_1}(\eta_{2n_4}(1), \eta_{20}(1)) \to 0
    \end{equation*}
    this means that $y_0 = \eta_{20}(1) = exp_{x_0}^{g_1} (w_{20})$, and we have $ exp_{x_0}^{g_1} (w_{20}) = y_0 = exp_{x_0}^{g_1}(w_{10})$, and since $|w_{10}|_{g_1 (x_0)}, |w_{20}|_{g_1 (x_0)} \leq r < \frac{inj(M,g_1)}{2} < inj(M,g_1)$, then $w_{10}=w_{20}$. Since $|w_{10} - w_{20}|_{g_1(x_0)}=0$, we have
    \begin{equation*}
        |w_{10} - w_{20}|_{g_1(x_0)} = \lim_{n_4 \to \infty} |w_{1n_4} - w_{2n_4}|_{g_1(x_{n_4})} \geq \epsilon > 0
    \end{equation*}
    which is a contradiction.
\end{proof}

\subsection{Proof of Lemma \ref{lemma:fin}}

\phantomsection\label{app:lemma-fin}
\begin{proof}[Proof of Lemma \ref{lemma:fin}]

We'll first show that for $\parallel g_1 -g_2 \parallel_{C^2 (M, g_1)}< \zeta$, where  $0 < \zeta < 1$ for the moment. Define the function $z: M \to \mathbb{R}$ such that $z(p) =1 - \frac{Vol_{g_2} (p)}{Vol_{g_1}(p)}$, with
\begin{equation*}
    |z(p)| < C_1 \zeta
\end{equation*}
where $C_1$ is a constant independent of $g_1$ and $g_2$. We'll fix $p \in M$, and choose a normal coordinate chart around $p$, with a local coordinate system $x^1, \dots , x^n$, such that
\begin{align*}
          g_{1ij} (p) = \delta_{ij} & & \frac{\partial g_{1ij}}{\partial x^k} (p)=0
\end{align*}
Then, since $g_2 = (g_{2ij})$ is a positive real symmetric matrix, it's diagonalizable by an orthogonal matrix. Let such a matrix be $O$. Also, let $(y^j)$ be the normal coordinate chart related to our original chart, by means of $O$, i.e. $y^j = O_i^j x^i$ from proposition 5.23, p.132~\cite{lee2018introduction}, and then we have
\begin{equation*}
    O^T (g_{2ij}) O = \begin{bmatrix}
        \lambda_1 &   & 0\\
         & \ddots &\\
        0 &   & \lambda_n
    \end{bmatrix}
\end{equation*}
and
\begin{equation*}
    O^T (g_{1ij})O = I
\end{equation*}
Then at $p$, we have
\begin{align*}
        \bigg|  \begin{bmatrix}
        1 &   & 0\\
         & \ddots &\\
        0 &   & \ 1
    \end{bmatrix}  -  \begin{bmatrix}
        \lambda_1 &   & 0\\
         & \ddots &\\
        0 &   & \lambda_n
    \end{bmatrix} \bigg|_{\mathbb{R}^{n \times n}} & = \sqrt{\sum_{k=1}^n (\lambda_k -1)^2}\\
    & \leq \parallel g_1 - g_2 \parallel_{C^2} < \zeta
\end{align*}
Furthermore, we see that
\begin{align*}
    |\lambda_1 -1| & < \zeta\\
    \vdots\\
    |\lambda_n -1| & < \zeta
\end{align*}
Then given this coordinate chart representation at $p$, we have
\begin{align*}
    \bigg( \frac{Vol_{g_2}(p)}{Vol_{g_1}(p)} \bigg)^2 & = \bigg( \frac{\sqrt{\det (g_{2ij})} dy^1 \wedge \dots \wedge dy^n}{\sqrt{\det (g_{1ij})} dy^1 \wedge \dots \wedge dy^n} \bigg)^2\\
    & = \lambda_1 \cdots \lambda_n < (1 + \zeta)^n\\
    & = 1 + \sum_{k=1}^n \binom{n}{k}  \zeta^k\\
    & = 1 + \big( \sum_{k=1}^n \binom{n}{k}  \zeta^{k-1} \big) \zeta\\
    & < 1 + \big( \sum_{k=1}^n \binom{n}{k} \big) \zeta
\end{align*}
Additionally, we see that
\begin{align*}
    \bigg( \frac{Vol_{g_2}(p)}{Vol_{g_1}(p)} \bigg)^2 & = \bigg( \frac{\sqrt{\det (g_{2ij})} dy^1 \wedge \dots \wedge dy^n}{\sqrt{\det (g_{1ij})} dy^1 \wedge \dots \wedge dy^n} \bigg)^2\\
    & = \lambda_1 \cdots \lambda_n > (1 - \zeta)^n\\
    & = 1 + \sum_{k=1}^n \binom{n}{k} (-\zeta)^k\\
    & \geq 1 - \big( \sum_{k=1}^n \binom{n}{k} \zeta^k \big)\\
    & = 1 - \big( \sum_{k=1}^n \binom{n}{k}  \zeta^{k-1} \big) \zeta\\
    & > 1 - \big( \sum_{k=1}^n \binom{n}{k} \big) \zeta
\end{align*}
Let $C_1 = \sum_{k=1}^n \binom{n}{k}$, then at $p$, under this coordinate representation we have
\begin{equation*}
    | \bigg( \frac{Vol_{g_2}(p)}{Vol_{g_1}(p)} \bigg)^2-1| < C_1 \zeta
\end{equation*}
Since  $(g_{2ij})$ is a symmetric positive definite matrix, $\lambda_i >0$ for $i=1,2, \dots , n$. Then we have
\begin{align*}
    |  \frac{Vol_{g_2}(p)}{Vol_{g_1}(p)} -1| & = |  \frac{Vol_{g_2}(p)}{Vol_{g_1}(p)} -1| \cdot (1)\\
    & \leq |  \frac{Vol_{g_2}(p)}{Vol_{g_1}(p)} -1| \cdot \bigg( \bigg(\frac{Vol_{g_2}(p)}{Vol_{g_1}(p)} \bigg) + 1 \bigg)\\
    & = |  \frac{Vol_{g_2}(p)}{Vol_{g_1}(p)} -1| \cdot |  \frac{Vol_{g_2}(p)}{Vol_{g_1}(p)} + 1 |\\
    & = | \big( \frac{Vol_{g_2}(p)}{Vol_{g_1}(p)} -1 \big) \big( \frac{Vol_{g_2}(p)}{Vol_{g_1}(p)} + 1 \big) |\\
    & = | \bigg( \frac{Vol_{g_2}(p)}{Vol_{g_1}(p)} \bigg)^2-1 | < C_1 \zeta
\end{align*}

Then since $p$ was arbitrary, and since $Vol_{g_i}$ for $i=1,2$ are independent of a change of coordinate charts, we have that for all $p \in M$
\begin{equation*}
    |z(p)| = |1 - \frac{Vol_{g_2}(p)}{Vol_{g_1}(p)}| < C_1 \zeta
\end{equation*}
as desired. Then, in our local coordinates, we have
\begin{equation*}
    d \mu_m (y) = \sqrt{ \det (g_{mij})} dy^1 \wedge \dots \wedge dy^n = Vol_{g_m} (y) dy^1 \wedge \dots \wedge dy^n
\end{equation*}
which means that
\begin{align*}
       d \mu_1 (y) - d \mu_2 (y) & = Vol_{g_1} (y) dy^1 \wedge \dots \wedge dy^n - Vol_{g_2} (y) dy^1 \wedge \dots \wedge dy^n\\
       & =\bigg( 1 - \frac{Vol_{g_2}(y)}{Vol_{g_1}(y)}  \bigg) Vol_{g_1}(y) dy^1 \wedge \dots \wedge dy^n\\
       & = \bigg( 1 - \frac{Vol_{g_2}(y)}{Vol_{g_1}(y)}  \bigg) d \mu_1 (y)  = z(y) d \mu_1 (y)
\end{align*}
Then from above, we have
\begin{equation*}
    \begin{split}
        \big| \int_M f (y)K_2(x,y) \ d \mu_1 (y)  - \int_M f(y)K_2 (x,y) \ d \mu_2 (y) \big| & = \big| \int_M f (y) K_2(x,y) \bigg( 1 - \frac{Vol_{g_2}(y)}{Vol_{g_1}(y)}  \bigg) d \mu_1 (y)\\
        & = \big| \int_M f(y) K_2(x,y) \ z(y) \ d \mu_1 (y) \big| \\
        & \leq \int_M | f (y)| \ |K_2 (x,y)| \ |z(y)| \ d \mu_1 (y)\\
        & < C_1 \zeta \int_M |f (y)| \ |K_2 (x,y)| \ d \mu_1 (y)\\
        & < C_1\cdot F \cdot K_0 \cdot Vol(M, g_1) \zeta
    \end{split}
\end{equation*}
where $F = \max_{y \in M} |f (y)|$, and $|K_m (x,y)| \leq K_0 = \max_{z \in M } |K(x_0,z)|$. Then $C_2 = C_1 \cdot F \cdot K_0 \cdot Vol (M, g_1)$, and let $\zeta = \min \{ \frac{\epsilon}{C_2}, \frac{1}{2} \}$, then we have
\begin{equation*}
   \big| \int_M f (y) \ K_2 (x,y) \ d \mu_1 (y) - \int_M f(y) \ K_2 (x,y) \ d \mu_2 (y) \big| < C_2 \zeta \leq C_2 \frac{\epsilon}{C_2} = \epsilon
\end{equation*}
as desired.
\end{proof}

\section{Derivation of the Jacobi equation in coordinates}\label{app:jaceq}

In this section we derive the coordinate form of the Jacobi equation used in section~\ref{sec:jac}, with the notation set up there. Recall that the Jacobi field $J_m$ along $\gamma_m$ has components $f_{mj}$ in the frame $\{X_j\}$ and solves $(\nabla^{g_m}_{\gamma_m'})^2J_m + R^{g_m}(J_m,\gamma_m')\gamma_m'=0$. We compute the two terms of this equation separately, starting with the curvature term.

\begin{align*}
    R^{g_m}(J_m(t), \gamma_m^\prime (t)) \gamma_m^\prime (t) & = R^{g_m} ( f_{mj}(t)X_j , \dot{\gamma}_m^\nu X_\nu ) \dot{\gamma}_m^\mu X_\mu\\
    & = f_{mj}(t) \dot{\gamma}_m^\nu \dot{\gamma}_m^\mu R^{g_m} ( X_j ,  X_\nu )  X_\mu\\
    & = f_{mj}(t) \dot{\gamma}_m^\nu \dot{\gamma}_m^\mu R_{j \nu \mu}^{k,g_m}X_k
\end{align*}

For the covariant derivative, we have

\begin{align*}
    \nabla_{\gamma_m^\prime (t)}^{g_m} J_m (t) & = f_{m \beta}^\prime (t) X_\beta (\gamma_m (t)) + f_{mj}(t) \dot{\gamma}_m^i \nabla_{X_i (\gamma_m (t))}^{g_m} X_j (\gamma_m (t))\\
    & = f_{m \beta}^\prime (t) X_\beta (\gamma_m (t)) + \Gamma_{ij}^{\beta , g_m} (\gamma_m (t)) f_{mj}(t) \dot{\gamma}_m^i X_\beta (\gamma_m (t))\\
    & = (f_{m \beta}^\prime (t) + \Gamma_{ij}^{\beta , g_m} (\gamma_m (t)) f_{mj}(t) \dot{\gamma}_m^i ) X_\beta (\gamma_m (t))
\end{align*}
With $u_{m \beta}(t): = f_{m \beta}^\prime (t) + \Gamma_{ij}^{\beta , g_m} (\gamma_m (t)) f_{mj}(t) \dot{\gamma}_m^i$, we have

\begin{align*}
    (\nabla_{\gamma_m^\prime (t)}^{g_m})^2 J_m (t) & = \nabla_{\gamma_m^\prime (t)}^{g_m} (u_{m \beta}(t) X_\beta (\gamma_m (t)))\\
    & = (u_{m k}^\prime (t) + \Gamma_{i \beta}^{k , g_m} (\gamma_m (t)) u_{m\beta}(t) \dot{\gamma}_m^i ) X_k (\gamma_m (t))
\end{align*}
Differentiating $u_{mk}$ and using the geodesic equation $\ddot{\gamma}_m^i = - \Gamma_{pq}^{i,g_m} \dot{\gamma}_m^p \dot{\gamma}_m^q$, we obtain
\begin{align*}
    u_{mk}^\prime (t) & = f_{mk}^{\prime \prime} (t) + f_{mj}(t) \dot{\gamma}_m^i  \frac{d}{dt} \bigg[ \Gamma_{ij}^{k,g_m} (\gamma_m (t)) \bigg] + f_{mj}^\prime (t) \dot{\gamma}_m^i \Gamma_{ij}^{k,g_m} (\gamma_m (t)) + f_{mj}(t) \Gamma_{ij}^{k,g_m} (\gamma_m (t)) \ddot{\gamma}_m^i\\
    & =f_{mk}^{\prime \prime} (t) + f_{mj}(t) \dot{\gamma}_m^i  \frac{\partial \Gamma_{ij}^{k,g_m}}{\partial x^s} (\gamma_m (t)) \dot{\gamma}_m^s + f_{mj}^\prime (t) \dot{\gamma}_m^i \Gamma_{ij}^{k,g_m} (\gamma_m (t)) + f_{mj}(t) \Gamma_{ij}^{k,g_m} (\gamma_m (t)) \ddot{\gamma}_m^i\\
    & = f_{mk}^{\prime \prime} (t) + f_{mj}(t) \dot{\gamma}_m^i \dot{\gamma}_m^s  \frac{\partial \Gamma_{ij}^{k,g_m}}{\partial x^s} (\gamma_m (t)) + f_{mj}^\prime (t) \dot{\gamma}_m^i \Gamma_{ij}^{k,g_m} (\gamma_m (t))\\
    &- f_{mj}(t) \Gamma_{ij}^{k,g_m} (\gamma_m (t)) \Gamma_{pq}^{i, g_m}(\gamma_m (t)) \dot{\gamma}_m^p \dot{\gamma}_m^q
\end{align*}
Substituting into the Jacobi equation, we have
\begin{align*}
    0 & = (\nabla_{\gamma_m^\prime(t)}^{g_m})^2J_m(t)+ R^{g_m}(J_m (t), \gamma_m^\prime (t)) \gamma_m^\prime (t)\\
    & = \bigg( f_{mk}^{\prime \prime}(t) + \big( 2 \Gamma_{ij}^{k,g_m}(\gamma_m (t)) \dot{\gamma}_m^i \big) f_{mj}^\prime (t)\\
    & + \big( \dot{\gamma}_m^\mu \dot{\gamma}_m^s \frac{\partial \Gamma_{ \mu j}^{k,g_m}}{\partial x^s} (\gamma_m (t))  - \dot{\gamma}_m^\nu \dot{\gamma}_m^c \Gamma_{c \nu}^{\mu,g_m}(\gamma_m (t)) \Gamma_{\mu j}^{k, g_m}(\gamma_m (t))\\
    & + \dot{\gamma}_m^\mu \dot{\gamma}_m^\nu \Gamma_{\mu j}^{\beta, g_m}(\gamma_m(t)) \Gamma_{\nu \beta}^{k,g_m}(\gamma_m (t)) + \dot{\gamma}_m^\mu \dot{\gamma}_m^\nu R_{j \nu \mu}^{k,g_m}(\gamma_m(t))  \big) f_{mj}(t) \bigg) X_k (\gamma_m (t))
\end{align*}

Since the vectors $X_k$ are linearly independent, the coefficient of each $X_k$ must vanish, so the components satisfy the linear homogeneous system
\begin{equation*}
    f_{mk}^{\prime \prime}(t) + a_{kj}^{g_m}(t) f_{mj}^\prime (t) + b_{k j}^{g_m}(t) f_{m j}(t) =0    
\end{equation*}
where, in Einstein's summation convention, $a_{kj}^{g_m}(t) = 2  \Gamma_{ij}^{k, g_m}(\gamma_m (t)) \dot{\gamma}_m^i$ and $b_{k j}^{g_m}(t) =   \dot{\gamma}_m^\mu \dot{\gamma}_m^s \frac{\partial \Gamma_{\mu j}^{k, g_m}}{\partial x^s} (\gamma_m (t)) - \dot{\gamma}_m^\nu \dot{\gamma}_m^c \Gamma_{c \nu}^{\mu, g_m} (\gamma_m (t)) \Gamma_{\mu j}^{k, g_m} (\gamma_m (t)) + \dot{\gamma}_m^\mu \dot{\gamma}_m^\nu \Gamma_{\mu j}^{\beta, g_m} (\gamma_m (t)) \Gamma_{\nu \beta}^{k, g_m} (\gamma_m (t)) + \dot{\gamma}_m^\mu \dot{\gamma}_m^\nu R_{j \nu \mu }^{k, g_m} (\gamma_m (t)) $, which are smooth along $\gamma_m (t)$. This is the system~\eqref{def:jaceq} of section~\ref{sec:jac}.

\section{A Jacobi field estimate for a fixed metric}

\begin{lemma}\label{lemma:ineq2}
    Fix a coordinate neighborhood $U$, and fix $L>0$. Then choose $p \in U$, distinct $v,w \in T_pM$ with $|v|_{g (p)}, |w|_{g (p)}, |v-w|_{g (p)} < L$. Then, for $z=v,w$, suppose that the geodesic $\eta_z (t) = exp_p (tz)$ on $[0,1]$ satisfies $\eta_z ([0,1]) \subset U$. Then let $J_z (t)$ be a Jacobi field along $\eta_z (t)$, with $J_z (0)=0$ and $J_v^\prime (0) = J_w^\prime (0)$ with $|J_z^\prime (0)|_{g(p)} < L$, then
    \begin{equation*}
        J_z (t) = f_{zi}(t) X_i (\eta_z (t))
    \end{equation*}
    where $\{ X_i \}$ is the coordinate frame with respect to $U$. Then let 
    \begin{equation*}
        f_z (t) = (f_{z1}(t), f_{z2}(t), \dots , f_{zn}(t))
    \end{equation*}
    Then there exists a constant $C >0$ such that for all $t \in [0,1]$, we have
    \begin{equation*}
        |f_v (t) - f_w (t)| < C |v-w|_{g (p)}
    \end{equation*}  
    where $| \cdot |$ is the Euclidean norm from definition~\ref{def:eucl}.
\end{lemma}

\begin{proof}

First, we'll use lemma~\ref{def:example2}, to show that there's a constant $K_0$ such that
\begin{align*}
    |\eta_v (t) - \eta_w (t)| & \leq K_0 |v-w|_{g (p)}\\
    |\eta_v^\prime (t) - \eta_w^\prime (t)| & \leq K_0 |v-w|_{g (p)}
\end{align*}

To avoid confusion, for this part of the proof, we'll be using constants $Z_0$, $Z_1$, and $Z_2$ instead of $C_0$, $C_1$, and $C_2$ and we'll use $\zeta_0 \geq 0$. We'll also use $x_z (t)$ instead of $y_z (t)$ for our geodesic system. Then, we'll have, $x_z (t) = \begin{bmatrix}
    \eta_z (t)\\
    \eta_z^\prime (t)
\end{bmatrix}$,
and $\lambda_{1ij}^k = \lambda_{2ij}^k = \Gamma_{ij}^k$. Then, we see that
\begin{equation*}
    |\Gamma_{ij}^k (\eta_v (t)) - \Gamma_{ij}^k (\eta_w (t))| \leq \max |\partial \Gamma| |\eta_v (t) - \eta_w (t)|
\end{equation*}
So, let $Z_0 = \max |\partial \Gamma|$ where the metric is $g$, which was defined in remark~\ref{def:rmk1}.

Then we see that
\begin{equation*}
    |\Gamma_{ij}^k| \leq \max |\Gamma|
\end{equation*}
So, let $Z_1 = \max |\Gamma|$ where the metric is $g$, which was defined in remark~\ref{def:rmk1}.
Then we have
\begin{equation*}
    |\dot{\eta}_z^i| \leq |\eta_z^\prime (t)| \leq \frac{1}{A_g} |\eta_z^\prime (t)|_{g(p)} < \frac{L}{A_g}
\end{equation*}
So, let $Z_2 = \frac{L}{A_g}$. Since $\lambda_{1ij}^k = \lambda_{2ij}^k = \Gamma_{ij}^k$, then for $x \in \eta_w ([0,1])$, we have $|\lambda_{1ij}^k (x) - \lambda_{2ij}^k (x)|=0$. Let $\zeta_0 = 0$.

Then from lemma~\ref{def:example2}, let $A=1+n^3 Z_2^2 Z_0 + 2n^3 Z_1 Z_2$ and $B= |x_v (0) - x_w (0)|$, then we have

\begin{align*}
    |\eta_v^\prime (t) - \eta_w^\prime (t)|,  |\eta_v (t) - \eta_w (t)| & \leq |x_v (t) - x_w (t)| \\ 
    & \leq  |\eta_v^\prime (0) - \eta_w^\prime (0)|e^A\\
    & \leq  \frac{e^A}{A_g} |v-w|_{g (p)}\\
    & = K_0 |v-w|_{g (p)}
\end{align*}

Next, from equation~\ref{def:jaceq}, for $z=v,w$, with $\eta_z (t) = exp_p (tz)$ and $\{ X_i \}$ being the coordinate frame associated with $U$, we have the Jacobi field
\begin{equation*}
    J_z (t) = f_{zi}(t) X_i (\eta_z (t))
\end{equation*}
for which $f_z (t)$ is the solution to the second order linear homogeneous system of differential equations
\begin{equation*}
    f_{zk}^{\prime \prime}(t) + a_{kj}^z (t)f_{zj}^\prime (t) + b_{kj}^z (t) f_{zj}(t) =0
\end{equation*}
where 
\begin{align*}
    a_{kj}^z (t) & = 2 \dot{\eta}_z^i \Gamma_{ij}^k (\eta_z (t))\\
    b_{kj}^z (t) & = \dot{\eta}_z^\mu \dot{\eta}_z^s \frac{\partial \Gamma_{\mu j}^k}{\partial x^s}(\eta_z (t)) -  \dot{\eta}_z^\nu \dot{\eta}_z^c \Gamma_{c \nu}^\mu (\eta_z (t)) \Gamma_{\mu j}^k (\eta_z (t))\\
    & +  \dot{\eta}_z^\mu \dot{\eta}_z^\nu \Gamma_{\mu j}^\beta (\eta_z (t)) \Gamma_{\nu \beta}^k (\eta_z (t)) + \dot{\eta}_z^\mu \dot{\eta}_z^\nu R_{j \nu \mu}^k (\eta_z (t))
\end{align*}
Then we'll use lemma~\ref{def:example3}. In this part of the proof, we'll use constants $C_0$ and $C_1$, variable $\zeta_1 >0$, the system $y_z (t) = \begin{bmatrix}
    f_z (t)\\
    f_z^\prime (t)
\end{bmatrix}$,
and the parameter matrix 
$M (t, \lambda_z) = \begin{bmatrix}
    0 & I\\
   - (b_{kj}^z (t)) & - (a_{kj}^z (t)) 
\end{bmatrix}$.
We observe that
\begin{align*}
    |M(t, \lambda_v)| & \leq \sqrt{n} + |a_{kj}^v (t)| + |b_{kj}^v (t)|\\
    & \leq \sqrt{n} + \frac{ 2n^3L \max|\Gamma|}{A_g} + \frac{n^4 L^2 \max |\partial \Gamma|}{A_g^2} +  \frac{2n^5 L^2 \max |\Gamma|^2}{A_g^2} + \frac{n^4 L^2 \max |R|}{A_g^2}
\end{align*}
where $g$ is the metric, $\max |\Gamma|$ and $\max |\partial \Gamma|$ are defined in remark~\ref{def:rmk1} and $\max |R|$ is defined in remark~\ref{def:rmk2}. So, let $C_1 = \sqrt{n} + \frac{ 2n^3L \max|\Gamma|}{A_g} + \frac{n^4 L^2 \max |\partial \Gamma|}{A_g^2} +  \frac{2n^5 L^2 \max |\Gamma|^2}{A_g^2} + \frac{n^4 L^2 \max |R|}{A_g^2}$.
Then we have
\begin{equation*}
    y_v (t) = y_v (0) + \int_0^t y_v^\prime (s) ds
\end{equation*}
This gives us
\begin{equation*}
     |y_v (t) | \leq |y_v (0)| + \int_0^t C_1|y_v (s)| ds
\end{equation*}
Then from lemma~\ref{gron:int}, for $A_0 = C_1$ and $B_0 = |y_v (0)|$, for all $t \in [0,1]$, we have

\begin{align*}
    |y_v (t) | & \leq B_0e^{A_0}\\
    & = |y_v (0)|e^{C_1}\\
    & = |f_v^\prime (0)|e^{C_1}\\
    & \leq \frac{L e^{C_1}}{A_g}
\end{align*}
Then let $C_0 = \frac{Le^{C_1}}{A_g}$.

So far, we have a constant $K_0$ such that
\begin{equation*}
    |\eta_v^\prime (t) - \eta_w^\prime (t)|,  |\eta_v (t) - \eta_w (t)| \leq K_0 |v-w|_{g (p)}
\end{equation*}
Then, we see that

\begin{align*}
    |a_{kj}^w (t) - a_{kj}^v (t)| & \leq |2 \Gamma_{ij}^k (\eta_w (t)) \dot{\eta}_w^i - 2 \Gamma_{ij}^k (\eta_v (t)) \dot{\eta}_v^i |\\
    & \leq |2 \Gamma_{ij}^k (\eta_w (t)) \dot{\eta}_w^i - 2 \Gamma_{ij}^k (\eta_v (t)) \dot{\eta}_w^i |\\
    & + |2 \Gamma_{ij}^k (\eta_v (t)) \dot{\eta}_w^i - 2 \Gamma_{ij}^k (\eta_v (t)) \dot{\eta}_v^i |\\
    & = I + II
\end{align*}
Then by breaking this into pieces, we have

\begin{align*}
    I & = 2 |\dot{\eta}_w^i| |\Gamma_{ij}^k (\eta_w (t)) - \Gamma_{ij}^k (\eta_v (t))|\\
    & \leq \frac{2n L}{A_g} \max |\partial \Gamma| |\eta_w (t) - \eta_v (t)|\\
    & \leq \frac{2n L}{A_g} \max |\partial \Gamma| K_0 |v-w|_{g (p)}
\end{align*}
Then we have
\begin{align*}
    II & = 2 |\Gamma_{ij}^k (\eta_v (t))| |\dot{\eta}_w^i - \dot{\eta}_v^i|\\
    & \leq 2 n \max |\Gamma| |\eta_w^\prime (t) - \eta_v^\prime (t)|\\
    & \leq 2n \max |\Gamma| K_0 |v-w|_{g (p)}
\end{align*}
Then by putting this back together, we have

\begin{equation*}
    |a_{kj}^w (t) - a_{kj}^v (t)| < ( \frac{2n L }{A_g} \max |\partial \Gamma| + 2n  \max |\Gamma| ) K_0 |v-w|_{g (p)} = K_1 |v-w|_{g (p)}
\end{equation*}

Next, we have

\begin{align*}
    |b_{k j}^w (t) - b_{k j}^v(t)| & = | \dot{\eta}_w^\mu \dot{\eta}_w^s \frac{\partial \Gamma_{\mu j}^{k}}{\partial x^s} (\eta_w (t)) -  \dot{\eta}_w^\nu \dot{\eta}_w^c \Gamma_{c \nu}^{\mu} (\eta_w (t)) \Gamma_{\mu j}^{k} (\eta_w (t))\\
    & + \dot{\eta}_w^\mu \dot{\eta}_w^\nu \Gamma_{\mu j}^{\beta} (\eta_w (t)) \Gamma_{\nu \beta}^{k} (\eta_w (t)) + \dot{\eta}_w^\mu \dot{\eta}_w^\nu R_{j \nu \mu}^{k} (\eta_w (t))\\
    & - ( \dot{\eta}_v^\mu \dot{\eta}_v^s \frac{\partial \Gamma_{\mu j}^{k}}{\partial x^s} (\eta_v (t)) -  \dot{\eta}_v^\nu \dot{\eta}_v^c \Gamma_{c \nu}^{\mu} (\eta_v (t)) \Gamma_{\mu j}^{k} (\eta_v (t))\\
    & +  \dot{\eta}_v^\mu \dot{\eta}_v^\nu \Gamma_{\mu j}^{\beta} (\eta_v (t)) \Gamma_{\nu \beta}^{k} (\eta_v (t)) +  \dot{\eta}_v^\mu \dot{\eta}_v^\nu R_{j \nu \mu}^{k} (\eta_v (t))|\\
    & \leq  |\dot{\eta}_w^\mu \dot{\eta}_w^s \frac{\partial \Gamma_{\mu j}^{k}}{\partial x^s} (\eta_w (t)) - \dot{\eta}_v^\mu \dot{\eta}_v^s \frac{\partial \Gamma_{\mu j}^{k}}{\partial x^s} (\eta_v (t))|\\
    & +  |\dot{\eta}_w^\nu \dot{\eta}_w^c \Gamma_{c \nu}^{\mu} (\eta_w (t)) \Gamma_{\mu j}^{k} (\eta_w (t)) - \dot{\eta}_v^\nu \dot{\eta}_v^c \Gamma_{c \nu}^{\mu} (\eta_v (t)) \Gamma_{\mu j}^{k} (\eta_v (t)) |\\
    & + |\dot{\eta}_w^\mu \dot{\eta}_w^\nu \Gamma_{\mu j}^{\beta} (\eta_w (t)) \Gamma_{\nu \beta}^{k} (\eta_w (t)) - \dot{\eta}_v^\mu \dot{\eta}_v^\nu \Gamma_{\mu j}^{\beta} (\eta_v (t)) \Gamma_{\nu \beta}^{k} (\eta_v (t))|\\
    & + |\dot{\eta}_w^\mu \dot{\eta}_w^\nu R_{j \nu \mu}^{k} (\eta_w (t)) - \dot{\eta}_v^\mu \dot{\eta}_v^\nu R_{j \nu \mu}^{k} (\eta_v (t))|\\
    & = A + B + C+ D
\end{align*}

For $A$, we have

\begin{align*}
    A & = |\dot{\eta}_w^\mu \dot{\eta}_w^s \frac{\partial \Gamma_{\mu j}^{k}}{\partial x^s} (\eta_w (t)) - \dot{\eta}_v^\mu \dot{\eta}_v^s \frac{\partial \Gamma_{\mu j}^{k}}{\partial x^s} (\eta_v (t)) |\\
    & \leq |\dot{\eta}_w^\mu \dot{\eta}_w^s \frac{\partial \Gamma_{\mu j}^{k}}{\partial x^s} (\eta_w (t)) - \dot{\eta}_w^\mu \dot{\eta}_w^s \frac{\partial \Gamma_{\mu j}^{k}}{\partial x^s} (\eta_v (t))|\\
    & + |\dot{\eta}_w^\mu \dot{\eta}_w^s \frac{\partial \Gamma_{\mu j}^{k}}{\partial x^s} (\eta_v (t)) - \dot{\eta}_v^\mu \dot{\eta}_v^s \frac{\partial \Gamma_{\mu j}^{k}}{\partial x^s} (\eta_v (t)) |\\
    & = I + II 
\end{align*}
Then, by breaking this into pieces, for $I$ we have
\begin{align*}
    I & \leq |\dot{\eta}_w^\mu \dot{\eta}_w^s \frac{\partial \Gamma_{\mu j}^{k}}{\partial x^s} (\eta_w (t)) - \dot{\eta}_w^\mu \dot{\eta}_w^s \frac{\partial \Gamma_{\mu j}^{k}}{\partial x^s} (\eta_v (t))|\\
    & \leq |\dot{\eta}_w^\mu| | \dot{\eta}_w^s | \max |\partial^2 \Gamma| \cdot |\eta_w (t) - \eta_v (t)|\\
    & \leq \frac{n^2L^2}{A_{g}^2} \max |\partial^2 \Gamma| K_0 |v-w|_{g(p)}\\
    & = \frac{n^2L^2 \max |\partial^2 \Gamma|}{A_{g}^2}  K_0 |v-w|_{g(p)}
\end{align*}
where $\max |\partial^2 \Gamma|$ was defined in remark~\ref{def:rmk1}.

Then for $II$, we have

\begin{align*}
    II & = |\dot{\eta}_w^\mu \dot{\eta}_w^s \frac{\partial \Gamma_{\mu j}^{k}}{\partial x^s} (\eta_v (t)) - \dot{\eta}_v^\mu \dot{\eta}_v^s \frac{\partial \Gamma_{\mu j}^{k}}{\partial x^s} (\eta_v (t)) |\\
    & \leq \max |\partial \Gamma| \cdot |\dot{\eta}_w^\mu \dot{\eta}_w^s - \dot{\eta}_v^\mu \dot{\eta}_v^s|\\
    & \leq \max |\partial \Gamma| (| \dot{\eta}_w^\mu ||\dot{\eta}_w^s  - \dot{\eta}_v^s | + |\dot{\eta}_v^s ||\dot{\eta}_w^\mu - \dot{\eta}_v^\mu |)\\
    & \leq \max | \partial \Gamma| ( \frac{L}{A_{g}} |\dot{\eta}_w^s  - \dot{\eta}_v^s | + \frac{L}{A_{g}} |\dot{\eta}_w^\mu - \dot{\eta}_v^\mu |)\\
    & \leq \frac{2n^2  L}{A_{g}} \max | \partial \Gamma| K_0 |v-w|_{g (p)}\\
    & = \frac{2n^2  L \max | \partial \Gamma|}{A_{g}}  K_0 |v-w|_{g (p)}\\
\end{align*}

Then putting everything back together, we have
\begin{equation*}
    A \leq I + II \leq  (\frac{n^2 L^2 \max |\partial^2 \Gamma|}{A_g^2} + \frac{2n^2L \max |\partial \Gamma|} {A_g}) K_0 |v-w|_{g (p)}
\end{equation*}

We'll handle $B$ and $C$ at the same time, by breaking them into pieces,

\begin{align*}
    B, C & = |\dot{\eta}_w^\nu \dot{\eta}_w^c \Gamma_{c \nu}^{\mu} (\eta_w (t)) \Gamma_{\mu j}^{k} (\eta_w (t)) - \dot{\eta}_v^\nu \dot{\eta}_v^c \Gamma_{c \nu}^{\mu} (\eta_v (t)) \Gamma_{\mu j}^{k} (\eta_v (t)) |\\
    & \leq |\dot{\eta}_w^\nu \dot{\eta}_w^c \Gamma_{c \nu}^{\mu} (\eta_w (t)) \Gamma_{\mu j}^{k} (\eta_w (t)) - \dot{\eta}_v^\nu \dot{\eta}_v^c \Gamma_{c \nu}^{\mu} (\eta_w (t)) \Gamma_{\mu j}^{k} (\eta_w (t))|\\
    & + |\dot{\eta}_v^\nu \dot{\eta}_v^c \Gamma_{c \nu}^{\mu} (\eta_w (t)) \Gamma_{\mu j}^{k} (\eta_w (t)) - \dot{\eta}_v^\nu \dot{\eta}_v^c \Gamma_{c \nu}^{\mu} (\eta_v (t)) \Gamma_{\mu j}^{k} (\eta_v (t)) |\\
    & = I + II
\end{align*}
Then for $I$, we'll have

\begin{align*}
    I & = |\dot{\eta}_w^\nu \dot{\eta}_w^c \Gamma_{c \nu}^{\mu} (\eta_w (t)) \Gamma_{\mu j}^{k} (\eta_w (t)) - \dot{\eta}_v^\nu \dot{\eta}_v^c \Gamma_{c \nu}^{\mu} (\eta_w (t)) \Gamma_{\mu j}^{k} (\eta_w (t))|\\
    & \leq \max |\Gamma|^2  |\dot{\eta}_w^\nu \dot{\eta}_w^c  - \dot{\eta}_v^\nu \dot{\eta}_v^c |\\
    & \leq \max |\Gamma|^2 (| \dot{\eta}_w^\nu ||\dot{\eta}_w^c  - \dot{\eta}_v^c | + |\dot{\eta}_v^c ||\dot{\eta}_w^\nu - \dot{\eta}_v^\nu |)\\
    & \leq \max |\Gamma|^2 ( \frac{L}{A_{g}} |\dot{\eta}_w^c  - \dot{\eta}_v^c | + \frac{L}{A_{g}} |\dot{\eta}_w^\nu - \dot{\eta}_v^\nu |)\\
    & \leq \frac{2n^3 L}{A_{g}} \max |\Gamma|^2 K_0 |v-w|_{g (p)}\\
    & = \frac{2n^3  L\max |\Gamma|^2}{A_{g}}  K_0 |v-w|_{g (p)}
\end{align*}

Then for $II$, we'll have

\begin{align*}
    II & = |\dot{\eta}_v^\nu \dot{\eta}_v^c \Gamma_{c \nu}^{\mu} (\eta_w (t)) \Gamma_{\mu j}^{k} (\eta_w (t)) - \dot{\eta}_v^\nu \dot{\eta}_v^c \Gamma_{c \nu}^{\mu} (\eta_v (t)) \Gamma_{\mu j}^{k} (\eta_v (t)) |\\
    & = |\dot{\eta}_v^\nu| |\dot{\eta}_v^c| |\Gamma_{c \nu}^{\mu} (\eta_w (t)) \Gamma_{\mu j}^{k} (\eta_w (t)) -  \Gamma_{c \nu}^{\mu} (\eta_v (t)) \Gamma_{\mu j}^{k} (\eta_v (t)) |\\
    & \leq \frac{ L^2}{A_{g}^2} \bigg(  |\Gamma_{c \nu}^{\mu} (\eta_w (t)) \Gamma_{\mu j}^{k} (\eta_w (t)) - \Gamma_{c \nu}^{\mu} (\eta_w (t)) \Gamma_{\mu j}^{k} (\eta_v (t))|\\
    & +  |\Gamma_{c \nu}^{\mu} (\eta_w (t)) \Gamma_{\mu j}^{k} (\eta_v (t)) - \Gamma_{c \nu}^{\mu} (\eta_v (t)) \Gamma_{\mu j}^{k} (\eta_v (t))| \bigg)\\
     & \leq \frac{ L^2}{A_{g}^2} \bigg( |\Gamma_{c \nu}^{\mu} (\eta_w (t))| | \Gamma_{\mu j}^{k} (\eta_w (t)) -  \Gamma_{\mu j}^{k} (\eta_v (t))|\\
    & + |\Gamma_{\mu j}^{k} (\eta_v (t)) | |\Gamma_{c \nu}^{\mu} (\eta_w (t))  - \Gamma_{c \nu}^{\mu} (\eta_v (t))| \bigg)\\
    & \leq  \frac{n^3 L^2}{A_{g}^2} 2 \max |\Gamma| \max|\partial \Gamma||\eta_w (t) - \eta_v (t)|\\
    & \leq \frac{2n^3 L^2 \max |\Gamma| \max|\partial \Gamma|}{A_{g}^2} K_0 |v-w|_{g (p)}
\end{align*}
Then putting everything back together, for $B$ and $C$ individually, we have

\begin{equation*}
    B , C \leq  I + II  \leq (\frac{2n^3 L  \max |\Gamma|^2}{A_g} + \frac{2n^3 L^2  \max |\Gamma| \max|\partial \Gamma|}{A_{g}^2} ) K_0 |v-w|_{g (p)}
\end{equation*}

Finally, for $D$, we have

\begin{align*}
    D & = |\dot{\eta}_w^\mu \dot{\eta}_w^\nu R_{j \nu \mu}^k (\eta_w (t)) - \dot{\eta}_v^\mu \dot{\eta}_v^\nu R_{j \nu \mu}^k (\eta_v (t)) |\\
    & \leq |\dot{\eta}_w^\mu \dot{\eta}_w^\nu R_{j \nu \mu}^k (\eta_w (t)) - \dot{\eta}_w^\mu \dot{\eta}_w^\nu R_{j \nu \mu}^k (\eta_v (t)) |\\
    & + |\dot{\eta}_w^\mu \dot{\eta}_w^\nu R_{j \nu \mu}^k (\eta_v (t)) - \dot{\eta}_v^\mu \dot{\eta}_v^\nu R_{j \nu \mu}^k (\eta_v (t)) |\\
    & = I + II
\end{align*}
Then by breaking this into pieces, we have

\begin{align*}
    I & \leq |\dot{\eta}_w^\mu| |\dot{\eta}_w^\nu| |R_{j \nu \mu}^k (\eta_w (t)) - R_{j \nu \mu}^k (\eta_v (t))|\\
    & \leq \frac{n^2 L^2}{A_g^2} \max |\partial R| |\eta_w (t) - \eta_v (t)|\\
    & \leq \frac{n^2 L^2 }{A_g^2} \max |\partial R| K_0 |v-w|_{g (p)}
\end{align*}
where $g$ is the metric, and $\max |\partial R|$ was defined in remark~\ref{def:rmk2}.

Next we have

\begin{align*}
    II & \leq |R_{j\nu \mu}^k (\eta_v (t))| |\dot{\eta}_w^\mu \dot{\eta}_w^\nu - \dot{\eta}_v^\mu \dot{\eta}_v^\nu|\\
    & \leq \max |R| (|\dot{\eta}_w^\mu| |\dot{\eta}_v^\nu - \dot{\eta}_w^\nu| + |\dot{\eta}_v^\nu| |\dot{\eta}_v^\mu - \dot{\eta}_w^\mu|)\\
    & \leq \frac{2n^2L}{A_g} \max |R| |\dot{\eta}_w (t) - \dot{\eta}_v (t)|\\
    & \leq \frac{2n^2L }{A_g} \max |R| K_0 |v-w|_{g (p)}\\
    & = \frac{2n^2L \max |R| }{A_g}  K_0 |v-w|_{g (p)}\\
\end{align*}
Then by putting this back together, we have

\begin{equation*}
    D \leq I + II < ( \frac{n^2 L^2 }{A_g^2} \max |\partial R| + \frac{2n^2L }{A_g} \max |R|) K_0 |v-w|_{g (p)}
\end{equation*}
Then by putting these four components together, we have
\begin{equation*}
\begin{split}
        |b_{kj}^w (t) - b_{kj}^v (t)| & \leq A + B + C + D\\
        & \leq \bigg(\frac{n^2 L^2 \max |\partial^2 \Gamma|}{A_g^2} + \frac{2n^2L \max |\partial \Gamma| }{A_g}\\
        & + 2 \big( \frac{2n^3 L  \max |\Gamma|^2}{A_g} + \frac{2n^3 L^2  \max |\Gamma| \max|\partial \Gamma|}{A_{g}^2} \big) + \frac{n^2 L^2 }{A_g^2} \max |\partial R|\\
        &+ \frac{2n^2L }{A_g} \max |R| \bigg) K_0 |v-w|_{g (p)}\\
        & = K_2 |v-w|_{g (p)}
\end{split}
\end{equation*}

Then we have
\begin{align*}
    |A_v (t) - A_w (t)| + |B_v (t) - B_w (t)| & \leq \sum_{kj} |a_{kj}^v (t) - a_{kj}^w (t)| + |b_{kj}^v (t) - b_{kj}^w (t)|\\
    & < n^2 (K_1 + K_2) |v-w|_{g (p)} < n^2 (K_1 + K_2)L
\end{align*}
Then let $\zeta_1 = n^2 (K_1 + K_2) |v-w|_{g (p)} $. Then from lemma~\ref{def:example3}, we have

\begin{equation*}
       |f_v (t) - f_w (t)| \leq n^2 (K_1 + K_2)C_0 e^{C_1 + n^2 (K_1 + K_2)|v-w|_{g (p)}}|v-w|_{g (p)} < n^2 (K_1 + K_2)C_0 e^{C_1 + n^2 (K_1 + K_2)L}|v-w|_{g (p)}
\end{equation*}

Then $C = n^2 (K_1 + K_2)C_0 e^{C_1 + n^2 (K_1 + K_2)L}$ is our desired constant.

\end{proof}

\end{document}